\documentclass[review]{elsarticle}

\usepackage{amssymb}
\usepackage[cmex10]{amsmath}
\usepackage[caption=false,font=normalsize,labelfont=sf,textfont=sf]{subfig}
\usepackage{stmaryrd}
\usepackage{multirow}
\usepackage{tabularx}
\usepackage{booktabs}
\usepackage{algorithm}
\usepackage{algorithmic}
\usepackage{dsfont}
\usepackage[latin1]{inputenc}
\usepackage{array}
\usepackage{longtable}
\usepackage{tabu}
\usepackage{lscape}
\usepackage[pass]{geometry}
\usepackage{mathtools}
\usepackage[title]{appendix}
\usepackage{etoolbox}
\usepackage{adjustbox}
\usepackage{graphicx}
\graphicspath{{../pdf/}{../jpeg/}{../eps/}{../eps/cmp/}}
\DeclareGraphicsExtensions{.pdf,.jpeg,.png,.eps}

\makeatletter
\def\ps@pprintTitle{%
  \let\@oddhead\@empty
  \let\@evenhead\@empty
  \def\@oddfoot{\reset@font\hfil\thepage\hfil}
  \let\@evenfoot\@oddfoot
}
\makeatother

\newcommand{\E}{\mathrm{E}}
\newcommand{\Prob}{P}
\newcommand{\Var}{\mathrm{Var}}
\DeclareMathSymbol{\Delta}{\mathalpha}{operators}{1}

\newcommand{\titlePartII}{Untangling~AdaBoost-based Cost-Sensitive~Classification\protect\\Part II: Empirical Analysis}

\journal{Pattern Recognition}

\begin{document}

%%%%%%%%%%%%
%% PART 2 %%
%%%%%%%%%%%%
%\part*{ } \pdfbookmark{Part II: Empirical Analysis}{part:II}

\begin{frontmatter}

\title{\titlePartII}

\author{Iago Landesa-V\'azquez, Jos\'e Luis Alba-Castro}
\address{Signal Theory and Communications Department, University of Vigo, Maxwell Street, 36310, Vigo, Spain}
\ead{iagolv@gts.uvigo.es, jalba@gts.uvigo.es}

\begin{abstract}
A lot of approaches, each following a different strategy, have been proposed in the literature to provide AdaBoost with cost-sensitive properties. In the first part of this series of two papers, we have presented these algorithms in a homogeneous notational framework, proposed a clustering scheme for them and performed a thorough theoretical analysis of those approaches with a fully theoretical foundation. The present paper, in order to complete our analysis, is focused on the empirical study of all the algorithms previously presented over a wide range of heterogeneous classification problems. The results of our experiments, confirming the theoretical conclusions, seem to reveal that the simplest approach, just based on cost-sensitive weight initialization, is the one showing the best and soundest results, despite having been recurrently overlooked in the literature.
\end{abstract}


\end{frontmatter}

%\linenumbers
\mathtoolsset{showonlyrefs}

\section{Introduction}
\label{sec:Intro2}

AdaBoost \citep{FreundSchapire97}, the quintessential boosting \citep{Schapire90} algorithm and one of the main representatives of the \emph{Ensemble Classifiers} \citep{Polikar06} paradigm, has been subject of extensive research in the fields of machine learning, pattern recognition and computer vision during the last few years (e.g. \citep{Schapire98, SchapireSinger99,Opitz99,Friedman00,MeaseWyner08a,ViolaJones04,MasnadiVasconcelos11, LandesaAlba12}), attracting an attention ``rarely matched in computational intelligence'' \citep{Polikar06}.

In practice, among the different scenarios for classification, those involving \emph{cost-sensitive} or \emph{asymmetric} conditions (e.g. disaster prediction, fraud detection, medical diagnosis, object detection, etc.) hold a noteworthy position. In those cases, different decisions may have different associated costs (depending on the nature of the decision or on class priors) so that obtained classifiers must focus their attention in the rare/most valuable class \citep{Elkan01, Provost97, Weiss03}. 

The intersection between these two worlds, AdaBoost and Cost-Sensitive learning, is represented by a significant set of works in the literature devoted to provide AdaBoost with asymmetric properties (e.g. \citep{Fan99, Ting00, ViolaJones04, ViolaJones02, Sun07, MasnadiVasconcelos07, MasnadiVasconcelos11, LandesaAlba12, LandesaAlba13}), and the practical relevance of the problem is evidenced in the role of AdaBoost as learning algorithm in the widespread Viola-Jones object detector framework \citep{ViolaJones04} that inherently deals with a clearly asymmetric scenario. 

Nonetheless, the different cost-sensitive AdaBoost methods proposed in the literature are very heterogeneous, and they have been presented to the researcher as a succession of algorithms with no clear properties to rule their use and behavior in practice. 

In the first part of this series of two papers \citep{LandesaAlba??a} we have presented, in a homogeneous notational framework, the different asymmetric AdaBoost approaches in the literature, proposing a clustering scheme for them based on the way asymmetry is inserted in the learning process (\emph{a posteriori}, \emph{heuristically} or \emph{theoretically}):

\begin{itemize}
\item \emph{A posteriori}
\begin{itemize}
\item  AdaBoost with threshold modification \citep{ViolaJones04}
\end{itemize}
\item \emph{Heuristic}
\begin{itemize}
\item AsymBoost \citep{ViolaJones02}
\item AdaCost \citep{Fan99}
\item CSB0, CSB1 and CSB2 \citep{Ting98,Ting00}
\item AdaC1, AdaC2 and AdaC3 \citep{Sun05, Sun07}
\end{itemize}
\item \emph{Theoretical}
\begin{itemize}
\item Cost-Sensitive AdaBoost \citep{MasnadiVasconcelos07, MasnadiVasconcelos11}
\item AdaBoostDB \citep{LandesaAlba13}
\item Cost-Generalized AdaBoost \citep{LandesaAlba12}
\end{itemize}
\end{itemize}

Then, for those algorithms with a fully theoretical derivation, we performed a thorough theoretical analysis and discussion adopting the two major perspectives used to derive and explain AdaBoost: the Error Bound Minimization perspective \citep{SchapireSinger99} and the Statistical View of Boosting \citep{Friedman00}.

Such analysis demonstrates, from whatever line of reasoning we may follow, that the simple asymmetric weight initialization strategy followed by Cost-Generalized AdaBoost, despite having been recurrently overlooked (and even rejected) in the literature \citep{Fan99, Ting00, ViolaJones02, MasnadiVasconcelos07, MasnadiVasconcelos11}, is a completely valid mechanism to build theoretically sound cost-sensitive boosted classifiers. Moreover, the error bound defined by Cost-Generalized AdaBoost, able to preserve the class-dependent loss ratio regardless of the training round, seems to be more consistent than that used by the other theoretical alternatives, Cost-Sensitive AdaBoost and AdaBoostDB, showing a tendency to increasingly emphasize the least costly class (\emph{asymmetry swapping}).

The present paper, in order to complete our study, covers the empirical analysis of all the methods previously enumerated, including the non-fully-theoretical approaches (a posteriori and heuristic) as well as the fully theoretical ones. 

The article is organized as follows: next section describes the thorough experimental framework we used for our experiments, in Section \ref{sec:results} the obtained results are detailed, and in Section \ref{sec:discussion} we analyze and discuss the results connecting them to the theoretical insights from the accompanying paper. At last, in Section \ref{sec:Conclusions2}, final conclusions are highlighted.

\section{Experimental Framework}
\label{sec:experimental_framework} 

Aiming to evaluate all the algorithms under the same conditions (training/test sets, costs) and over a broad range of heterogeneous classification problems, we designed the following experimental framework:

\subsection{Datasets}
\label{subsec:datasets} 

We have used both \emph{synthetic} and \emph{real} datasets for our tests. The synthetic group is composed of two different (though similar) sets, both conceived to allow an easy visual interpretation of the classification task.

\begin{figure}[!htb]
\centering
\includegraphics[width=.95\columnwidth]{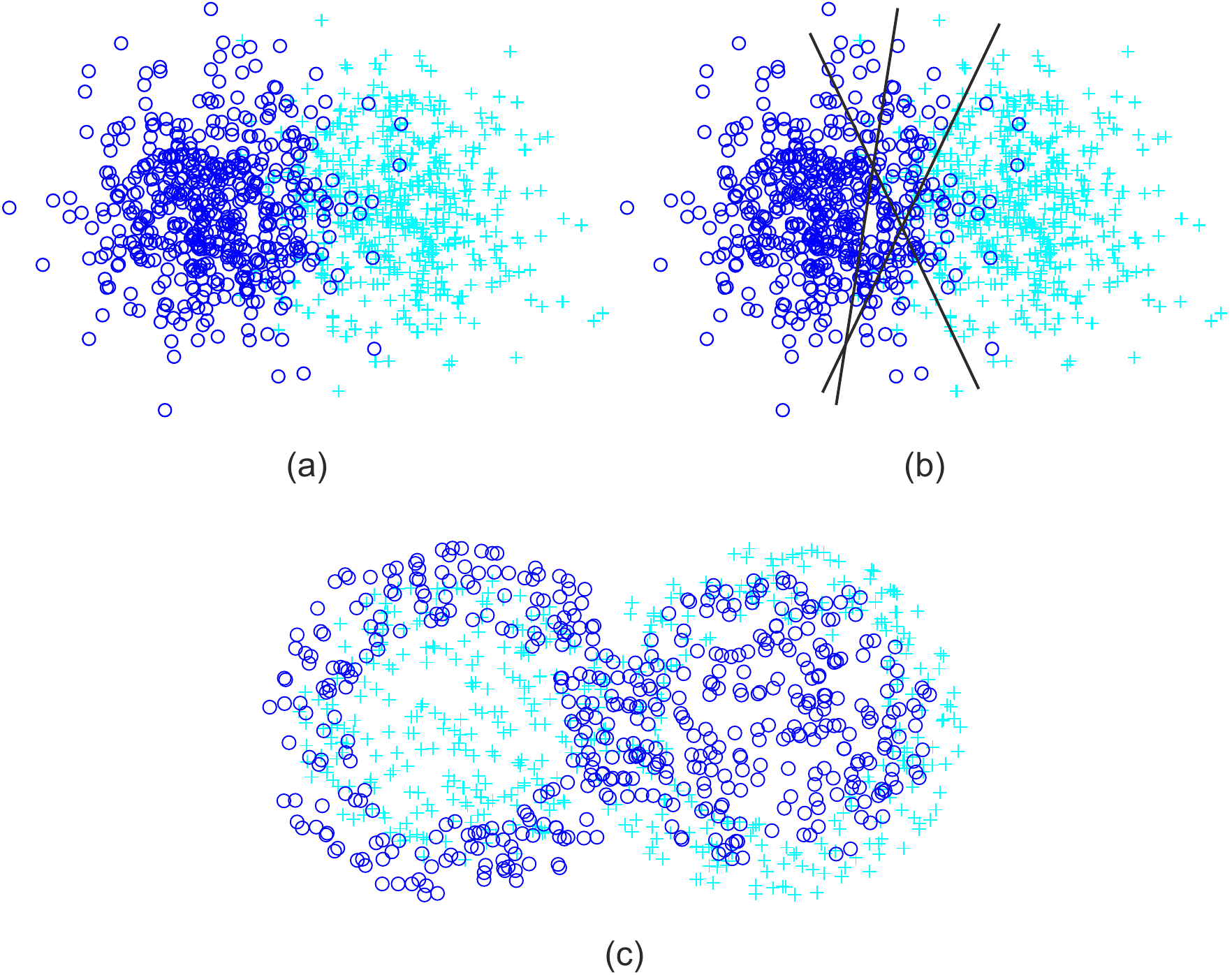}
\caption{Synthetic Datasets: (a) Bayes Dataset; (b) weak classifiers on the Bayes Dataset; (c) Two Clouds Dataset.  Positive examples are marked as `$+$', while negatives are `$\circ$'.}
\label{synthetic_datasets} % caption for the whole figure
\end{figure}

\begin{itemize}

\item \emph{Bayes Dataset}: Positives and negatives are modeled by bivariate normal distributions, both with the same priors and covariance matrices, but different means (Figure \ref{synthetic_datasets}a). Features for classification are the projections of each sample point on a discrete collection of angles in the 2D space (see Figure \ref{synthetic_datasets}b). In this scenario the \emph{optimal} classifier can be easily computed for every cost requirement, according to the Bayes risk rule.

%\begin{figure}[!htb]
%\centering
%\includegraphics[width=\columnwidth]{BayesClouds.eps}
%\caption{\emph{Bayes Dataset} examples. Positive examples are marked as `$+$', while negatives are `$\circ$'. In figure b examples of weak classifiers are shown.}
%\label{bayes_clouds} % caption for the whole figure
%\end{figure}

\item \emph{Two Clouds Dataset}: Inspired by the example in the work by Viola and Jones \citep{ViolaJones02}, this dataset can be seen as a more complex version of the Bayes Dataset. Positives and negatives are uniformly distributed into two clouds (one circular and one annular), with different centers and overlapping each other (Figure \ref{synthetic_datasets}c). Features are, again, the projections of each example on a discrete range of angles in the 2D space.

%\begin{figure}[!htb]
%\centering
%\includegraphics[width=.8\columnwidth]{TwoClouds.eps}
%\caption{\emph{Two Clouds Dataset} example. Positive examples are marked as `$
%+$', while `$\circ$' are the negative ones (note that positive and negative 
%classes are overlapped in both cases).}
%\label{twoclouds_dataset}
%\label{two_clouds} % caption for the whole figure
%\end{figure}

\end{itemize}

On the other hand, we have used six different datasets extracted from real problems. They can be grouped into two classes:

\begin{itemize}

\item \emph{UCI Datasets}: We have selected five datasets from the UCI Machine Learning Repository \citep{UCIRepository10} that are characterized by having an intrinsic asymmetric nature: Breast Cancer, Credit, Diabetes, Ionosphere and Spam. Examples belonging to the most valuable class of each dataset, as defined in the original problem, are considered as positives.

\item \emph{CBCL Face Database}: As an example of a real-world asymmetric problem in which boosting is used in practice, we have taken 1000 faces and 1000 non-faces (19x19 pixels resolution) from the CBCL face database \citep{heisele2000}. Following the proposal by Viola and Jones \citep{ViolaJones04}, we use their same dictionary of Haar-like features to build the classifiers.

\end{itemize}

%(giving a total of 63960 features per image for 19x19 pixel images).

For a more homogeneous cost-sensitive benchmark across the different databases, we have imposed that the number of positives and the number of negatives should match each other in every used dataset. As a result, in UCI datasets, we have discarded some examples of the most populated class to match its cardinality with that of the less populated one. With this condition, we can ensure that, regardless the specific problem being tested, the priors of each class are always the same (0.5). A summarizing listing of the datasets used for our experiments is shown in Table \ref{tab:datasets}.

\begin{table}[htbp]
  \centering
	 \scriptsize
{
\renewcommand{\arraystretch}{1.5}
\vspace{8pt}
    \begin{tabular}{|l|c|c|c|p{5.5cm}|}
    \hline
    \textbf{Name}  & \textbf{Pos.} & \textbf{Neg.} & \textbf{Feat.} & \textbf{Short Description} \\
    \hline
    \textbf{Bayes} & 250   & 250   & 31    & Two bivariate normal distributions (Figure \ref{synthetic_datasets}a)\\
    \textbf{Two Clouds} & 500   & 500   & 31    & Two bivariate clouds with more complex distributions (Figure \ref{synthetic_datasets}c) \\
    \textbf{UCI Breast} & 239   & 239   & 10    & Classification of benign or malignant breast tumors (Original Wisconsin Breast Cancer Database) \\
    \textbf{UCI Credit} & 300   & 300   & 24    & Set of attributes as good or bad credit risks (German Credit Data) \\
    \textbf{UCI Diabetes} & 268   & 268   & 8     & Set of attributes as tested positive or not for diabetes (Pima Indians Diabetes Database) \\
    \textbf{UCI Ionosphere} & 126   & 126   & 34    & Classification of radar returns from the ionosphere as good or bad (Johns Hopkins University Ionosphere Database) \\
    \textbf{UCI Spam} & 1813  & 1813  & 57    & Classification of  email as spam or non-spam \\
    \textbf{CBCL}  & 1000  & 1000  & 63960 & Classification of face and no-face images using Haar-like features \\
    \hline
    \end{tabular}%
		}
		  \caption{Listing of the used datasets, showing the number of positive and negative samples, the number of features and a short semantic description of each set.}
  \label{tab:datasets}%
\end{table}%

\subsection{Costs}
\label{subsec:costs}

In order to sweep a wide range of asymmetries, we have defined nineteen different cost combinations to evaluate:

\begin{equation}
\label{cost_comb}
\begin{split}
(C_P, C_N)\in \{ & (1,100), (1,50), (1,25), (1,10), (1,7), (1,5),\\
                 & (1,3), (1,2), (2,3), (1,1), (3,2), (2,1), (3,1),\\
                 & (5,1), (7,1), (10,1), (25,1), (50,1), (100,1)\}
\end{split}
\end{equation}

\subsection{Algorithms}
\label{subsec:algorithms}

For each defined dataset and cost combination, we have trained classifiers with all the different algorithms analyzed in this work: AdaBoost with threshold modification, AsymBoost, AdaCost, CSB0, CSB1, CSB2, AdaC1, AdaC2, AdaC3, Cost-Sensitive AdaBoost/AdaBoostDB and Cost-Generalized AdaBoost. In addition, for the Bayes Dataset case, we have the optimal Bayes classifier as ground truth reference.

\subsection{Performance Metrics}
\label{subsec:metric}

The analysis of ROC curves \citep{Provost97,Fawcett06} has been the traditional way to evaluate and compare the behavior of different classifiers across different working points. Nevertheless, C. Drummond and R.C. Holte \citep{Drummond00} proposed an alternative representation, based on expected costs and dual respect to traditional ROC curves, that has been shown to be more appropriate for cost-sensitive classification problems. Since cost is explicitly shown, this kind of representations allow direct visual cost-sensitive interpretations and comparisons, and they are based on two magnitudes: the \emph{Probability Cost Function} ($PCF$) and the \emph{Normalized Expected Cost} ($NEC$), that are defined in Equations (\refeq{PCF}) and (\refeq{NEC}). In these expressions, $\Prob(y=1)$ and $\Prob(y=-1)$ are the prior probabilities of an example to be positive or negative, while $FNR$ and $FPR$ are, respectively, the false negative and false positive rates obtained by the classifier.

\begin{gather}
\label{PCF}
PCF=\frac{\Prob(y=1)C_{P}}{\Prob(y=1)C_{P}+\Prob(y=-1)C_{N}}\\
\label{NEC}
NEC=FNR \cdot PCF+FPR \cdot (1-PCF)
\end{gather}

Our experimental analysis is based on these cost-oriented representations \citep{Drummond00} and their related magnitudes.

\subsection{Training and Testing Schemes}
\label{subsec:schemes}

As customary in many boosting works (e.g. \citep{SchapireSinger99,ViolaJones04,MasnadiVasconcelos11}) weak classifiers used in our experiments are the simplest ones, stumps, to further underscore the role of ``boosting'' in getting classification strength by combination. 

Instead of defining static training and test sets for each database, to further improve robustness we have implemented a \emph{3-fold cross-validation strategy}: every dataset is split into three parts, so the role of test set is iteratively assigned to one of the subsets (folds) while the other two define the respective training set. From the three possible training scenarios arising from this scheme, three different classifiers are obtained, and the global performance will be defined as the average of the individual performances obtained over the respective test sets. 

Whichever fold, cost requirement, database or learning algorithm involved, every training process has been run for as many boosting rounds as the total number of examples (positives and negatives) of the respective dataset. In any case, to evaluate the performance of the different algorithms in a more uniform way throughout the different databases and costs, we have also performed an ``a posteriori'' convergence test. A classifier initially trained for $K$ rounds is considered to converge at round $k<K$ if the next two conditions are met:

\begin{itemize}
\item The deviation about the mean of the Normalized Expected Cost (NEC) over the training set is less than $10^{-3}$, for \emph{all} the subsequent rounds ($k+1, k+2,\cdots K$).
\item The subsequent rounds are, at least 10\% of the total ($K-k \geq 0.1K$).
\end{itemize}

To extract our experimental results, the earliest round meeting these two conditions is used as cutoff of the classifier. Otherwise, in case no converging round has been found, the entire classifier is taken into account. This procedure is aimed to protect the results from overfitting artifacts that could degrade some comparative experiments.

Depending on the specific scenario, there are several exceptions to this framework: 

\begin{itemize}

\item Bayes theoretical classifier (only for the Bayes dataset) is optimal by definition, thus it is ``only one'' and requires no training. In this case cross-validation makes no sense and the classifier is directly tested over the whole database.

\item For AdaBoost with threshold modification we have used the whole training dataset both to train the strong classifier and to adjust its threshold, instead of having, as originally proposed by Viola and Jones \citep{ViolaJones04}, independent sets for each role. From our point of view this is the most appropriate and homogeneous way to compare this method with the other ones, since the ``threshold adjustment'' dataset is no more than an additional training set. It is not clear for us how to properly establish a boundary to distribute, with the right sizing, the training examples into two independent and smaller subsets, as well as how this split affects the overall performance.

\item On AsymBoost classifiers, the convergence test is not performed. The definition of the algorithm (see the previous paper of the series \citep{LandesaAlba??a}) states that the number of training rounds determines how the asymmetry is introduced and distributed in the classifier. As a consequence, pruning the number of rounds of the final classifier would violate the asymmetric premises on which training was defined, changing the error metric that has actually been minimized.

\end{itemize}

\section{Results}
\label{sec:results}
Following the guidelines presented in the previous subsection, we have trained all the combinations of algorithm, database, costs and fold to obtain a broad collection of classifiers. The individual performance of each classifier has been evaluated over its respective test dataset, and finally averaged across the cross-validation folds for each case. As a result of this process we obtained a large corpus of performance data that can be consulted in detail in Appendix.

Table \ref{tab:abbreviations} summarizes the abbreviations we have used to refer to the different algorithms in the forthcoming tables and figures of this section and Appendix.

\begin{table}[htbp]
  \centering
  \footnotesize
	{
    \begin{tabular}{|c|c|}
    \hline
    \textbf{Algorithm} & \textbf{Abbreviation} \\
		\hline
    AdaBoost with threshold modification & ABT \\
		%\hline
		AsymBoost & ASB \\
    %\hline
		AdaCost & ADC \\
    %\hline
		CSB0  & CB0 \\
		%\hline
    CSB1  & CB1 \\
		%\hline
    CSB2  & CB2 \\
		%\hline
    AdaC1 & AC1 \\
		%\hline
    AdaC2 & AC2 \\
		%\hline
    AdaC3 & AC3 \\
		%\hline
    Cost-Sensitive AdaBoost / AdaBoostDB & CSA \\
		%\hline
    Cost-Generalized Adaboost & CGA \\
		%\hline
    Optimal Bayes (for Bayes Dataset) & BAY \\
		\hline
    \end{tabular}%
		}
	\caption{Abbreviations used to refer to the tested algorithms in forthcoming figures and tables.}
	\label{tab:abbreviations}%
\end{table}%

\subsection{Global Analysis}
\label{subsec:global_analysis}

To make a global behavioral analysis of the classifiers obtained by each algorithm, we have defined a comparative performance measure based on Normalized Expected Costs ($NEC$) \citep{Drummond00}. For each combination of dataset and cost requirement we found which of the trained classifiers had a lowest $NEC$, and then computed the deviations between the $NEC$ values obtained by all classifiers and that minimum one. These deviation values, which we denote as $\Delta NEC$, measure the distance of each classifier to the best solution we have achieved for the same scenario. Repeating this process across databases and costs to gather all the $\Delta NEC$ values corresponding to the whole experimental framework, we will obtain a wide sample of a random variable ranking the performance of the trained classifiers.

\begin{equation}
\label{deltanec_def}
\begin{split}
\Delta NEC(alg, cost, set)=NEC(alg, cost, set)-\underset{alg}{\operatorname{arg min}}\left(NEC(alg, cost, set) \right)
\end{split}
\end{equation}

\begin{figure}[!htb]

\centerline{
\subfloat[]{
\centering
\includegraphics[width=5.9cm]{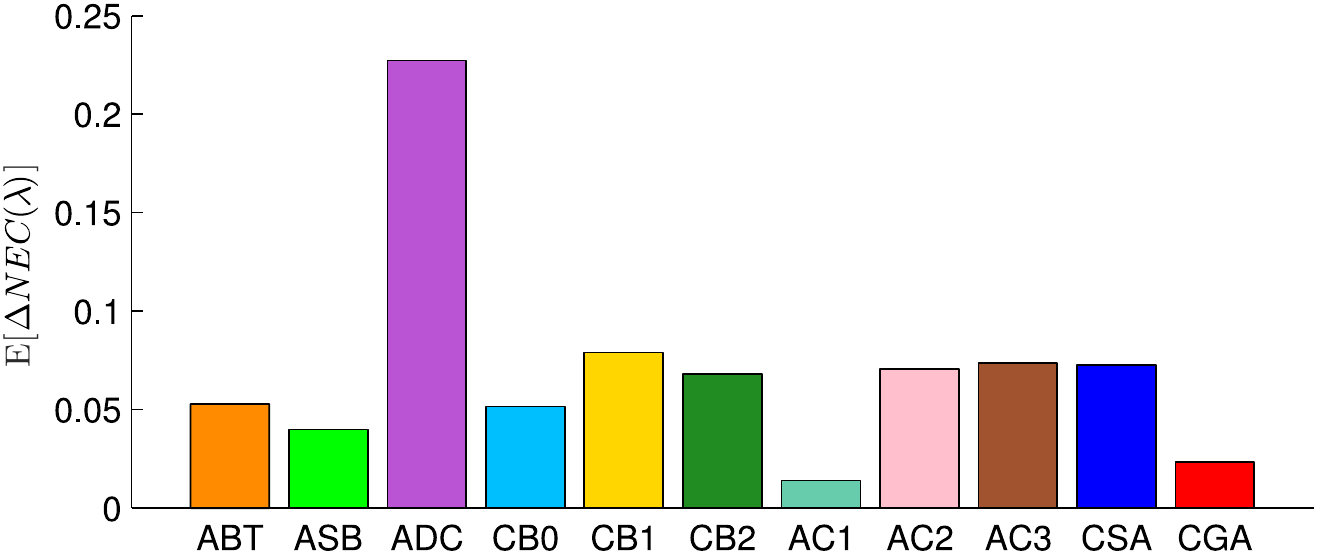}
\label{mean_delta_nec_comp}
}
\hfil
\subfloat[]{
\centering
\includegraphics[width=5.9cm]{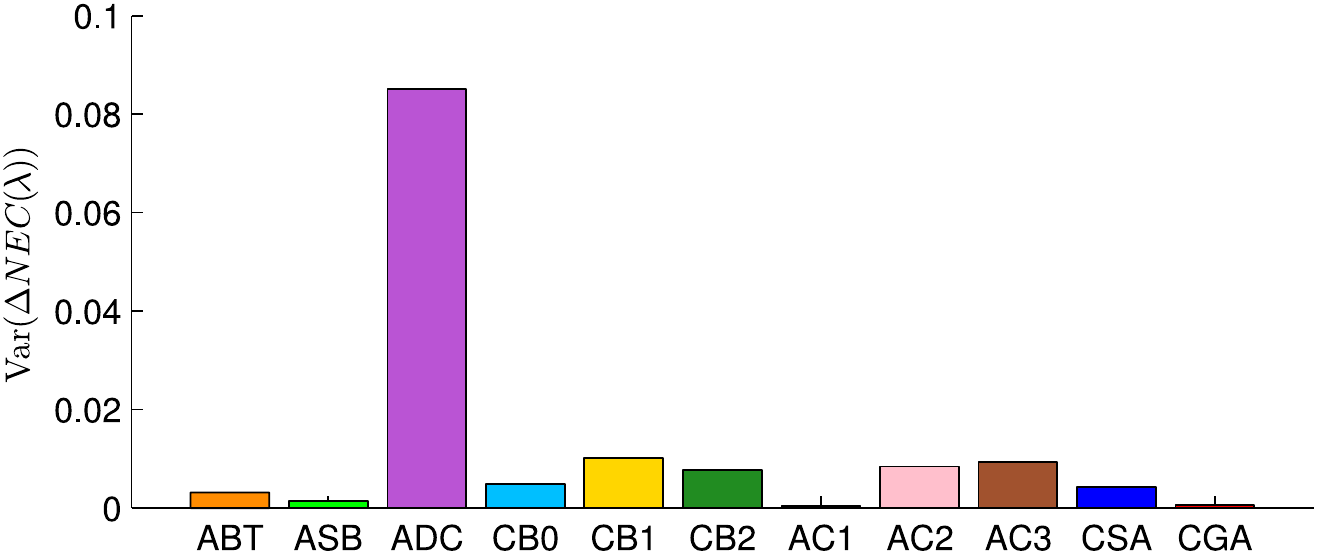}
\label{var_delta_nec_comp}
}
}
\caption{Global Conditional Expectations (a) and Variances (b) of $\Delta NEC$ for each tested algorithm.}
\label{fig_delta_nec_comp}
\end{figure}

As can be seen in Equation (\refeq{deltanec_def}), $\Delta NEC$ is a function of three variables: learning algorithm ($alg$), cost requirement ($cost$) and dataset ($set$)\footnote{Note that fold averaging was done in a previous step.}. Thus, if we compute the conditional expectation of $\Delta NEC$ for a given algorithm $\lambda$ (\refeq{deltanec_exp_a}), we will obtain the overall ranking score of all the classifiers trained by that specific algorithm throughout the whole experimental framework. Figure \ref{mean_delta_nec_comp} and Table \ref{tab:necglobalbehavior} depict the obtained values of $\E[\Delta NEC(\lambda)]$ for all the algorithms we are studying.

\begin{equation}
\label{deltanec_exp_a}
\begin{split}
\E[\Delta NEC(\lambda)]=\E[\Delta NEC(alg, cost, set) | alg=\lambda]
\end{split}
\end{equation}

\begin{table}[htbp]
  \centering
  \footnotesize
    \begin{tabular}{|c|c|c|}
    \hline
     & $\E[\Delta NEC(\lambda)]$ & $\Var(\Delta NEC(\lambda))$ \\ 		\hline
    \textbf{ABT}   & 5.273$\cdot10^{-2}$ & 3.136$\cdot10^{-3}$\\ 
    \textbf{ASB}   & 3.993$\cdot10^{-2}$ & 1.387$\cdot10^{-3}$\\ 
    \textbf{ADC}   & 2.272$\cdot10^{-1}$ & 8.510$\cdot10^{-2}$\\ 
    \textbf{CB0}   & 5.142$\cdot10^{-2}$ & 4.845$\cdot10^{-3}$\\ 
    \textbf{CB1}   & 7.891$\cdot10^{-2}$ & 1.012$\cdot10^{-2}$\\ 
    \textbf{CB2}   & 6.810$\cdot10^{-2}$ & 7.782$\cdot10^{-3}$\\ 
    \textbf{AC1}   & 1.399$\cdot10^{-2}$ & 3.681$\cdot10^{-4}$\\ 
    \textbf{AC2}   & 7.071$\cdot10^{-2}$ & 8.403$\cdot10^{-3}$\\ 
    \textbf{AC3}   & 7.350$\cdot10^{-2}$ & 9.399$\cdot10^{-3}$\\ 
    \textbf{CSA}   & 7.247$\cdot10^{-2}$ & 4.226$\cdot10^{-3}$\\ 
    \textbf{CGA}   & 2.343$\cdot10^{-2}$ & 5.935$\cdot10^{-4}$\\ 
		\hline
\end{tabular}%
	\caption{Global Conditional Expectation and Variance values of $\Delta NEC$ for each tested algorithm.}
	  \label{tab:necglobalbehavior}%
\end{table}%

Bearing in mind that the lower $\E[\Delta NEC]$, the better the performance, AdaC1 (one of the heuristic alternatives) is the algorithm showing best global results, followed by Cost-Generalized AdaBoost (one of the theoretical variants). After them, in a second tier, we find AsymBoost (heuristic), CSB0 (heuristic) and AdaBoost with threshold modification (a posteriori). On the opposite side, AdaCost (heuristic) is, by far, the algorithm showing the worst performance results. 

To analyze the stability (defined as the statistical precision) of these ranking scores across databases and costs, we have also computed the conditional variance of $\Delta NEC$ for each algorithm (\refeq{deltanec_var_a}), obtaining the results shown in Figure \ref{var_delta_nec_comp} and second column of Table \ref{tab:necglobalbehavior}. As can be seen, AdaC1 and Cost-Generalized AdaBoost are not only the algorithms giving the best average performance, they also are the most stable ones, with a difference of about one order of magnitude to the next.

\begin{equation}
\label{deltanec_var_a}
\begin{split}
\Var(\Delta NEC(\lambda))=\Var(\Delta NEC(alg, cost, set) | alg=\lambda)
\end{split}
\end{equation}

Going into a little more detail, if we inspect figures in Appendix, we will see that, for increasing asymmetries, several algorithms have a tendency to ``saturate'' and build ``all-positives'' or ``all-negatives'' solutions, instead of classifiers that, though biased to the most costly class, can still distinguish between two different labels. In order to globally evaluate and quantify this effect, we have defined the parameter $\Delta CE$ (\refeq{deltace_def}), analogous to $\Delta NEC$ but based on cost-insensitive Classification Error (CE, the ratio of correctly classified instances in the whole database), as a measure of the discriminative power of each obtained classifier with respect to the best distinguishing one for its same scenario. By definition, algorithms with a lower discriminative power (higher $\Delta CE$)  are more prone to saturation.

\begin{equation}
\label{deltace_def}
\begin{split}
\Delta CE(alg, cost, set)=CE(alg, cost, set)-\underset{alg}{\operatorname{arg min}}\left(CE(alg, cost, set) \right)
\end{split}
\end{equation}

\begin{figure}[!htb]

\centerline{
\subfloat[]{
\centering
\includegraphics[width=5.9cm]{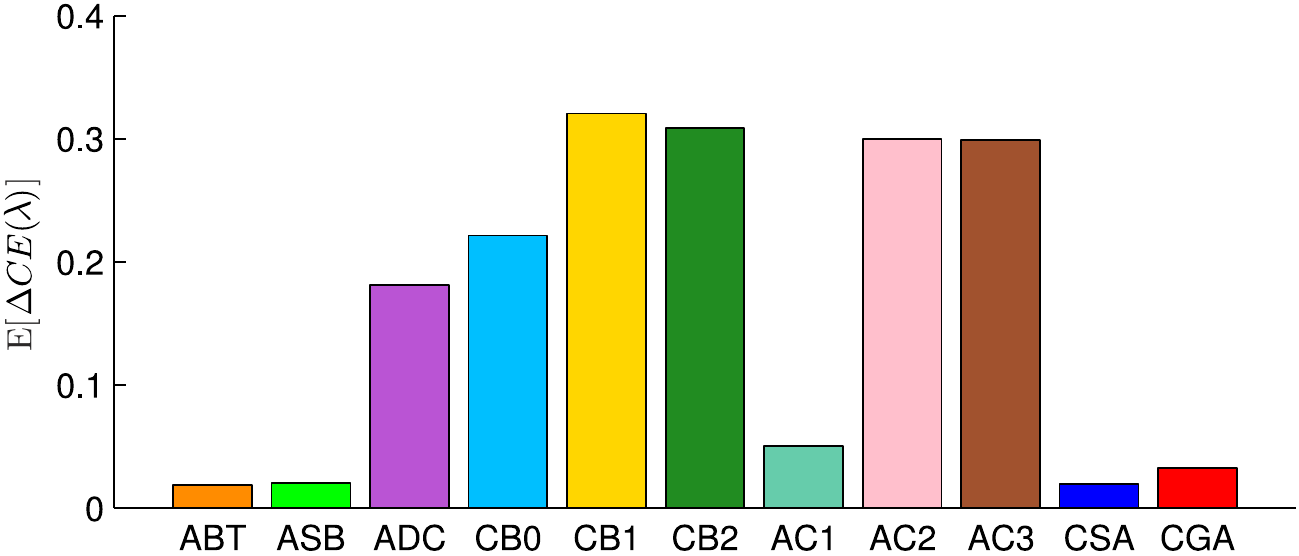}
\label{mean_delta_ce_comp}
}
\hfil
\subfloat[]{
\includegraphics[width=5.9cm]{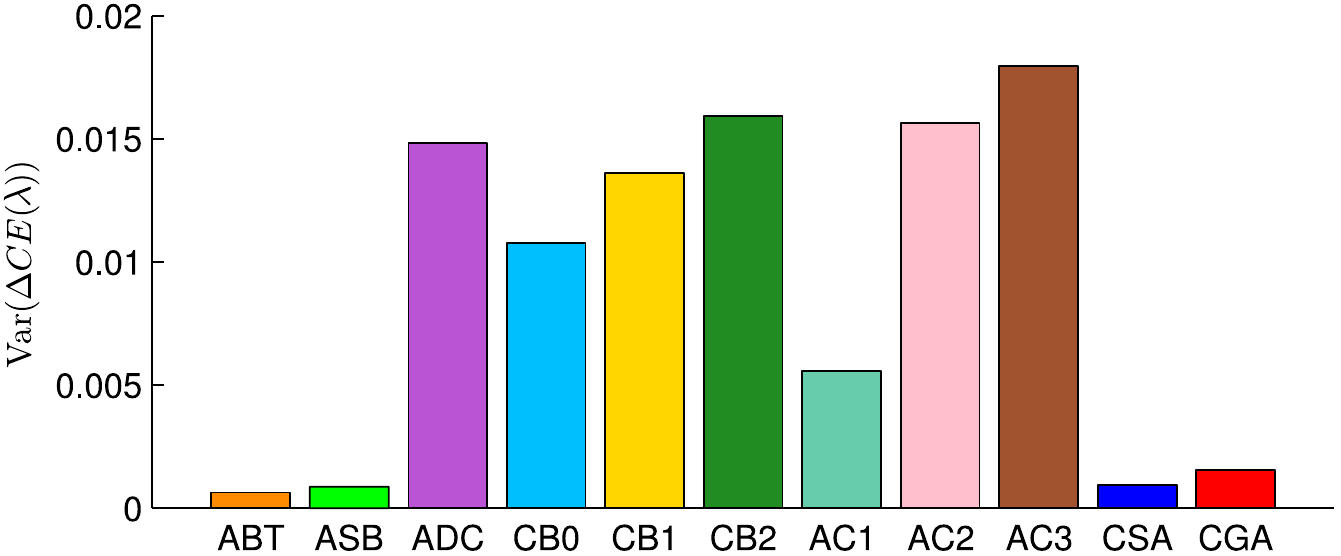}
\label{var_delta_ce_comp}
}
}
\caption{Global Conditional Expectations (a) and Variances (b) of $\Delta CE$ for each tested algorithm.}
\label{fig_delta_ce_comp}
\end{figure}

Values of conditional expectation and variance of $\Delta CE$ for each tested algorithm are shown in Figure \ref{fig_delta_ce_comp} and Table \ref{tab:ceglobalbehavior}. As can be seen, AdaBoost with threshold modification (a posteriori), Cost-Sensitive AdaBoost (theoretical) and AsymBoost (heuristic) are, in that order, the best discriminating algorithms, and also the most stable ones in this regard. After them, the two algorithms that showed better general performance are just the next in the ranking, but this time in reverse order: Cost-Generalized AdaBoost is more discriminative than AdaC1 and also more stable. The remaining algorithms present much poorer discrimination results.

\begin{table}[htbp]
  \centering
  \footnotesize
    \begin{tabular}{|c|c|c|}
    \hline
     & $\E[\Delta CE(\lambda)]$ & $\Var(\Delta CE(\lambda))$ \\ 		\hline
    \textbf{ABT}   & 1.861$\cdot10^{-2}$ & 6.292$\cdot10^{-4}$ \\ 
    \textbf{ASB}   & 2.002$\cdot10^{-2}$ & 8.651$\cdot10^{-4}$ \\ 
    \textbf{ADC}   & 1.813$\cdot10^{-1}$ & 1.482$\cdot10^{-2}$ \\ 
    \textbf{CB0}   & 2.212$\cdot10^{-1}$ & 1.076$\cdot10^{-2}$ \\ 
    \textbf{CB1}   & 3.203$\cdot10^{-1}$ & 1.361$\cdot10^{-2}$ \\ 
    \textbf{CB2}   & 3.087$\cdot10^{-1}$ & 1.592$\cdot10^{-2}$ \\ 
    \textbf{AC1}   & 5.056$\cdot10^{-2}$ & 5.576$\cdot10^{-3}$ \\ 
    \textbf{AC2}   & 2.998$\cdot10^{-1}$ & 1.564$\cdot10^{-2}$ \\ 
    \textbf{AC3}   & 2.990$\cdot10^{-1}$ & 1.797$\cdot10^{-2}$ \\ 
    \textbf{CSA}   & 1.965$\cdot10^{-2}$ & 9.236$\cdot10^{-4}$ \\ 
    \textbf{CGA}   & 3.247$\cdot10^{-2}$ & 1.554$\cdot10^{-3}$ \\ 
		\hline
\end{tabular}%
	\caption{Global Conditional Expectation and Variance values of $\Delta CE$ for each tested algorithm.}
	 \label{tab:ceglobalbehavior}%
\end{table}%

The analysis of the discriminant power is relevant because saturation may prevent a proper boosted evolution during learning, and cause a detrimental effect to the global performance of the final classifier. It is important to notice that the best classifier that a learning algorithm can build, no matter the specific cost scenario, is a classifier with null error. Such an ideal classifier must achieve null error in positives and in negatives, being, in fact, symmetric. Hence, an ideal cost-sensitive boosting algorithm should be aimed to approach that perfect (and symmetric) classifier as much as possible,\emph{but} following a consistent cost-sensitive iterative pathway. Thus, during learning, cost-sensitive boosting algorithms must keep a balance between the global asymmetry reached after each round, and the ability to evolve and further approximate the ideal solution in the forthcoming ones. In this scenario, the saturation effect may act as an anchor of the asymmetry and preclude a proper boosted evolution towards a classifier closest to the ideal one. We will discuss this in depth in Section \ref{subsec:algorithms_saturation}.

\subsection{Cost Analysis}
\label{subsec:cost_analysis}

Once we have made a general analysis on the global behavior of all the asymmetric AdaBoost variants we are testing, now it is time for a more detailed study able to asses the specific behavior of the algorithms through the cost spectrum. For a clearer analysis, we will prune the total set of algorithms to focus our attention on those yielding most interesting overall results according to the data presented in the previous subsection: on the one hand, AdaC1, Cost-Generalized AdaBoost and AsymBoost (the top three algorithms with lowest $\Delta NEC(\lambda)$); and, on the other hand, AdaBoost with threshold modification and Cost-Sensitive AdaBoost (among the remaining, those with less tendency to saturation). As can be seen, this selection includes algorithms from all the three types in which we clustered all the cost-sensitive AdaBoost variants in the previous paper of the series \citep{LandesaAlba??a}: \emph{A posteriori} (AdaBoost with threshold modification), \emph{Heuristic} (AsymBoost and AdaC1) and \emph{Theoretical} (Cost-Sensitive AdaBoost and Cost-Generalized AdaBoost).

\begin{figure}[!htb]
\centerline{
\subfloat[]{
\centering
\includegraphics[width=5.9cm]{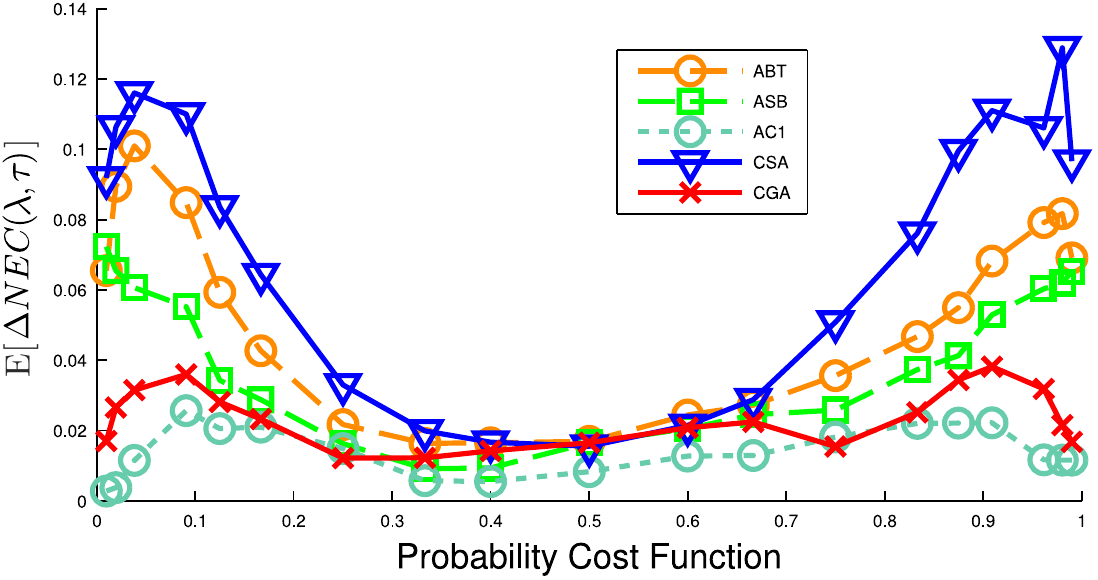}
}
\hfil
\subfloat[]{
\includegraphics[width=5.9cm]{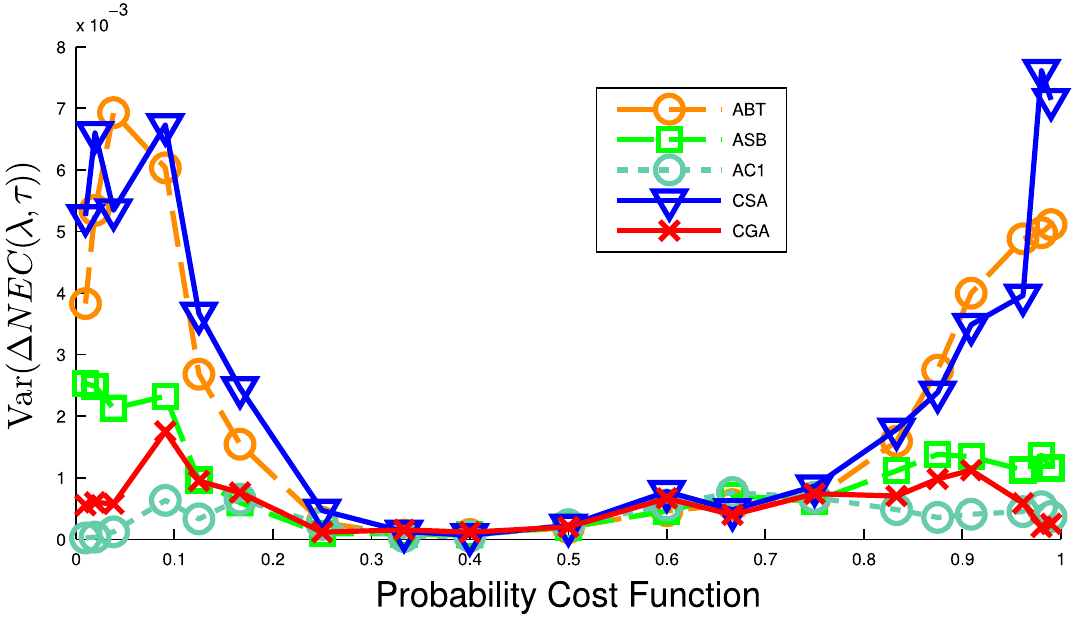}
}
}
\caption{Cost-Dependent Conditional Expectations (a) and Variances (b) of $\Delta NEC$ for each tested algorithm.}
\label{fig_delta_nec_cost}
\end{figure}

Now we are interested in computing the conditional expectation and variance of $\Delta NEC$ for each particular algorithm $\lambda$ and cost combination $\tau$, as shown in Equations (\refeq{deltanec_exp_a_c}) and (\refeq{deltanec_var_a_c}). Thus, we can obtain the cost-dependent ranking scores for all the algorithms we are studying, as shown in Figure \ref{fig_delta_nec_cost}. As can be seen, AdaC1 outperforms (both in performance and stability) the other alternatives in virtually all the tested cost scenarios, being closely followed by Cost-Generalized AdaBoost, while the results yielded by the other tested variants are clearly worse. It is remarkable that differences among the different algorithms tend to be more significant for increasing asymmetries (high values of $|PCF|$), while for moderate ones differences are more negligible.

\begin{gather}
\label{deltanec_exp_a_c}
\E[\Delta NEC(\lambda,\tau)]=\E[\Delta NEC(alg, cost, set) | alg=\lambda, cost=\tau]\\
\label{deltanec_var_a_c}
\Var(\Delta NEC(\lambda,\tau))=\Var(\Delta NEC(alg, cost, set) | alg=\lambda, cost=\tau)
\end{gather}

\begin{figure}[!htb]
\centerline{
\subfloat[]{
\centering
\includegraphics[width=5.9cm]{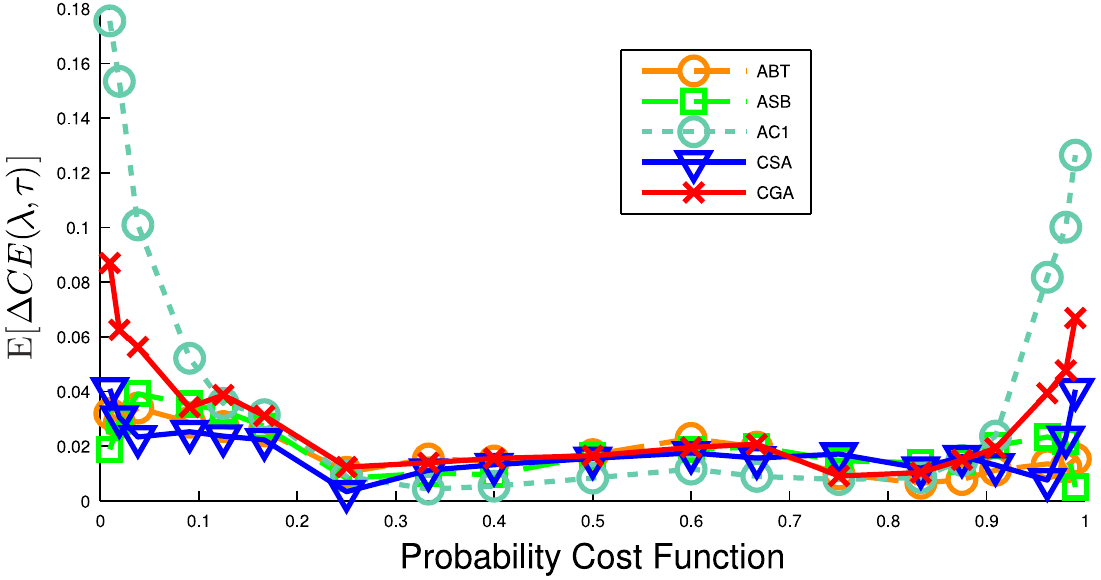}
}
\hfil
\subfloat[]{
\includegraphics[width=5.9cm]{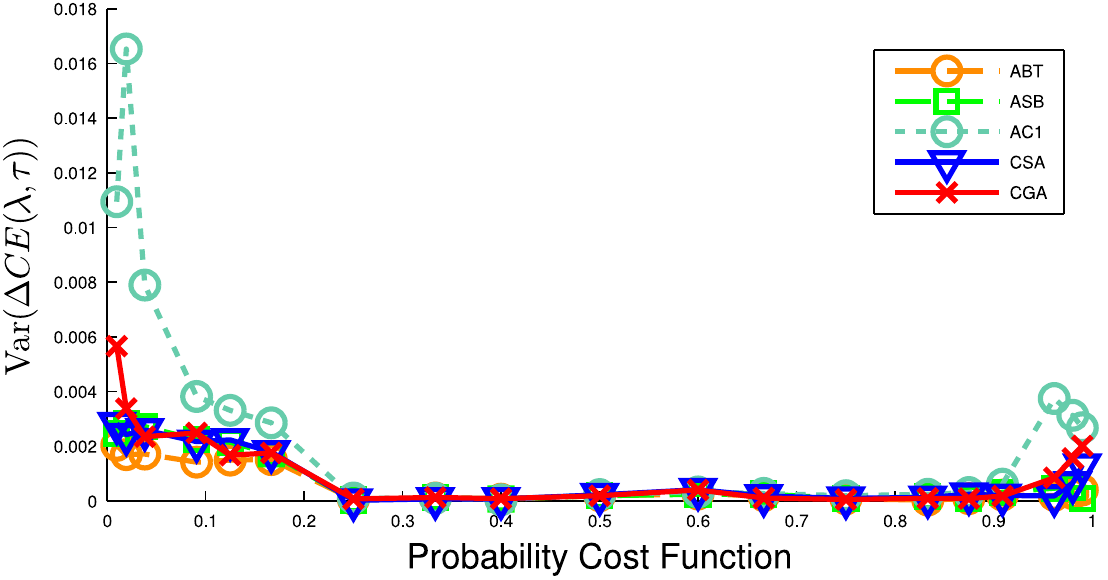}
}
}
\caption{Cost-Dependent Conditional Expectations (a) and Variances (b) of $\Delta CE$ for each tested algorithm.}
\label{fig_delta_ce_cost}
\end{figure}

An analogous procedure is followed to compute the cost-dependent discriminative power ($\Delta CE(\lambda,\tau)$), whose expectation and variance are depicted in Figure \ref{fig_delta_ce_cost}. Differences are again concentrated in high asymmetries, but this time with flipped roles: the two algorithms with better general performance are now the ones showing less discriminative power, which is especially noticeable for AdaC1.

\subsection{Training Time}
\label{subsec:train_time}

Another important aspect to consider when analyzing our experimental results is the computational burden associated to the training phase, especially bearing in mind that this kind of algorithms is widely used in learning systems dealing with a huge number of training samples and an even greater number of features (object detection in images, as the algorithm proposed by Viola and Jones \citep{ViolaJones04}, is a paradigmatic example), in which training can be extremely long. Training times obtained for each algorithm and dataset, averaged over the three folds, are shown in Table \ref{tab:trainingtime}.

\begin{table}[htbp]
  \centering
  \footnotesize
    \begin{tabular}{|c|c|c|c|c|}
    \hline
    & \textbf{~~~Bayes~~~} & \textbf{Two Clouds} & \textbf{~~~Breast~~~} & \textbf{~~~Credit~~~}\\
    \hline
    
    \textbf{ABT}   & 1.413 & 4.966 & 9.100$\cdot10^{-1}$ & 1.404  \\
    \textbf{ASB}   & 6.485$\cdot10^{-1}$ & 2.552 & 2.451$\cdot10^{-1}$ & 5.912$\cdot10^{-1}$  \\
    \textbf{ADC}   & 6.192$\cdot10^{-1}$ & 2.076 & 2.688$\cdot10^{-1}$ & 5.473$\cdot10^{-1}$ \\
    \textbf{CB0}   & 6.216$\cdot10^{-1}$ & 2.636 & 2.569$\cdot10^{-1}$ & 5.806$\cdot10^{-1}$ \\
    \textbf{CB1}   & 6.507$\cdot10^{-1}$ & 2.666 & 2.613$\cdot10^{-1}$ & 5.715$\cdot10^{-1}$ \\
    \textbf{CB2}   & 6.663$\cdot10^{-1}$ & 2.576 & 2.584$\cdot10^{-1}$ & 5.863$\cdot10^{-1}$  \\
    \textbf{AC1}   & 7.065$\cdot10^{-1}$ & 2.716 & 2.807$\cdot10^{-1}$ & 6.304$\cdot10^{-1}$ \\
    \textbf{AC2}   & 6.407$\cdot10^{-1}$ & 2.636 & 2.583$\cdot10^{-1}$ & 5.835$\cdot10^{-1}$ \\
    \textbf{AC3}   & 6.425$\cdot10^{-1}$ & 2.699 & 2.625$\cdot10^{-1}$ & 6.029$\cdot10^{-1}$ \\
    \textbf{CSA}   & 1.612$\cdot10^{1}$ & 4.708$\cdot10^{1}$ & 3.178 & 2.305 \\
    \textbf{CGA}   & 6.700$\cdot10^{-1}$ & 2.695 & 2.527$\cdot10^{-1}$ & 5.840$\cdot10^{-1}$ \\
\hline
\hline
    & \textbf{Diabetes} & \textbf{Ionosphere} & \textbf{Spam} & \textbf{CBCL}\\
    \hline
 \textbf{ABT}   & 1.014 & 5.868$\cdot10^{-1}$ & 3.551$\cdot10^{1}$ & 1.063$\cdot10^{4}$ \\
    \textbf{ASB}   & 3.032$\cdot10^{-1}$ & 2.716$\cdot10^{-1}$ & 2.528$\cdot10^{1}$ & 1.063$\cdot10^{4}$ \\
    \textbf{ADC}   & 2.715$\cdot10^{-1}$ & 2.637$\cdot10^{-1}$ & 2.040$\cdot10^{1}$ & 1.084$\cdot10^{4}$ \\
    \textbf{CB0}   & 2.844$\cdot10^{-1}$ & 2.645$\cdot10^{-1}$ & 2.543$\cdot10^{1}$ & 1.090$\cdot10^{4}$ \\
    \textbf{CB1}   & 2.982$\cdot10^{-1}$ & 2.788$\cdot10^{-1}$ & 2.571$\cdot10^{1}$ & 1.065$\cdot10^{4}$ \\
    \textbf{CB2}   & 2.972$\cdot10^{-1}$ & 2.798$\cdot10^{-1}$ & 2.569$\cdot10^{1}$ & 1.068$\cdot10^{4}$ \\
    \textbf{AC1}   & 3.082$\cdot10^{-1}$ & 2.828$\cdot10^{-1}$ & 2.782$\cdot10^{1}$ & 1.162$\cdot10^{4}$ \\
    \textbf{AC2}   & 2.731$\cdot10^{-1}$ & 2.737$\cdot10^{-1}$ & 2.715$\cdot10^{1}$ & 1.172$\cdot10^{4}$ \\
    \textbf{AC3}   & 2.741$\cdot10^{-1}$ & 2.687$\cdot10^{-1}$ & 2.822$\cdot10^{1}$ & 1.282$\cdot10^{4}$ \\
    \textbf{CSA}   & 2.692 & 2.395 & 1.115$\cdot10^{2}$ & 6.737$\cdot10^{5}$ \\
    \textbf{CGA}   & 2.928$\cdot10^{-1}$ & 2.699$\cdot10^{-1}$ & 2.523$\cdot10^{1}$ & 1.065$\cdot10^{4}$ \\
\hline
    \end{tabular}%
	\caption{Training time (s)}
		  \label{tab:trainingtime}%
\end{table}%

For an easier interpretation of these results, we have taken Cost-Generalized AdaBoost (the fully-theoretical alternative with a lower training time and also the reference used in Figures in Appendix) as a basis. Thus, we computed, for every dataset, the time consumed by each algorithm in relation to that needed by Cost-Generalized AdaBoost, and finally averaged the obtained ratios across datasets. Results are depicted in Table \ref{tab:extratime}.

\begin{table}[htbp]
  \centering
	\footnotesize
    \begin{tabular}{|c|c|}
		\hline
    & \textbf{Time Ratio} \\
		\hline
    \textbf{ABT}   & 2.250\\
    \textbf{ASB}   & 0.992\\
    \textbf{ADC}   & 0.928\\
    \textbf{CB0}   & 0.987\\
    \textbf{CB1}   & 1.005\\
    \textbf{CB2}   & 1.006\\
    \textbf{AC1}   & 1.068\\
    \textbf{AC2}   & 1.010\\
    \textbf{AC3}   & 1.036\\
    \textbf{CSA}   & 17.973\\
    \hline
		\end{tabular}%
	\caption{Average training time ratio needed by the tested algorithms compared to that consumed by Cost-Generalized AdaBoost.}
		\label{tab:extratime}%
\end{table}%

As can be seen, most of the algorithms are in a $\pm7\%$ range with regard to Cost-Generalized AdaBoost. However, there are two algorithms whose training times are much greater than the rest: AdaBoost with threshold modification (2.25 times slower than Cost-Generalized AdaBoost), due to the extra time needed to adjust the threshold after each iteration; and, above all, Cost-Sensitive AdaBoost (18 times slower than Cost-Generalized AdaBoost) that is, by far, the most complex algorithm we are testing. At this point we must remember that, though Cost-Sensitive AdaBoost and AdaBoostDB have been shown to reach the same solutions and their classification performance can be analyzed as if they were only one variant, their training times are markedly different. For feasibility, in our experimental framework, we have used AdaBoostDB, which, as reported in \citep{LandesaAlba13}, is more than 200 times faster than Cost-Sensitive AdaBoost. Hence, AdaBoostDB is the algorithm that actually needs 18 times the training time consumed by Cost-Generalized AdaBoost, while Cost-Sensitive AdaBoost would be even far slower.

\section{Discussion}
\label{sec:discussion}

We will now discuss the obtained results we have just shown to gain insights on the empirical behavior of the different algorithms.

\subsection{Saturation and Poor Performance}
\label{subsec:algorithms_saturation}

In Section \ref{subsec:global_analysis} we coined the concept ``saturation'' as the tendency of a learning algorithm to get ``all-positives'' or ``all-negatives'' classifiers when trained for increasing asymmetries, and defined an empirical measure ($\Delta CE$) of the tendency to saturation in terms of the discriminative power of the classifiers learned by a given algorithm. We also argued that, since the ideal classifier for any cost scenario is that obtaining null error in positives and in negatives (thus, a symmetric decision), saturation may be a detrimental effect precluding a proper approximation to that ideal classifier during the boosted learning process.

As we can see in Figure \ref{fig_delta_ce_comp} and Table \ref{tab:ceglobalbehavior}, five of the tested algorithms have an average discrimination power markedly lower than the remaining ones. These algorithms are CSB0, CSB1, CSB2, AdaC2, AdaC3 and AdaCost, and they are among the eight algorithms yielding poorest overall results in our experiments (Figure \ref{fig_delta_nec_comp} and Table \ref{tab:necglobalbehavior}). By inspecting figures in Appendix the tendency to saturation of these six algorithms becomes evident, empirically revealing the seeming correlation between tendency to saturation and bad performance results.

The listed algorithms have something in common, all of them are heuristic approaches based on direct manipulations of the weight update rule. Moreover CSB0, CSB1, CSB2, AdaC2 and AdaC3 have an unique feature that none of the other tested algorithms present: they are the only alternatives including costs as a \emph{multiplicative factor} in their respective weight update rules. This multiplicative mechanism is one of the most aggressive forms of inducing asymmetry in the boosting process, to the extent that, in view of the obtained results, it leads to a selection of weak classifiers that is systematically anchored to the costly class, thus avoiding a normal boosted evolution and building saturated strong classifiers.

Though also heuristic, AdaCost is a different case. Curiously (see Appendix), while for the Bayes dataset and almost all cost combinations in UCI Breast Cancer dataset AdaCost is able to obtain the best results, for the remaining datasets its performance is so poor that it even yields error rates greater than 0.5! As a consequence, AdaCost presents the worst overall performance, by far, of all the algorithms we have tested (see Figure \ref{fig_delta_nec_comp} and Table \ref{tab:necglobalbehavior}).

Revising AdaCost formulation, we can see that while weak classifier selection is made accordingly to the weight distribution $D(i)$, the goodness parameter $\alpha_t$ related to that selection depends on the weight distribution \emph{but also} on the cost-adjustment function $\beta(i)$. Thus, weak classifiers selection is guided by a criterion different to that of goodness computation and weight update, giving rise to some degree of decoupling in the process. In an extreme case, this decoupling may cause the $\alpha_t$ of the selected classifier to be negative, so the contribution of this classifier to the ensemble will also be negative (its reversed version, with swapped labels, will be, in fact, ``better'' than the selected one). This kind of situation was already pointed out by the authors of the algorithm (\citep{Fan99}, suggesting sign reversal as an a posteriori possible solution), and is responsible for the error rates exceeding 0.5.

\subsection{AdaBoost with Threshold Modification and AsymBoost}
\label{subsec:algorithms_interm}

AdaBoost with threshold modification and AsymBoost, the two cost-sensitive heuristic approaches proposed by Viola and Jones \citep{ViolaJones04, ViolaJones02}, take a step further in performance and also seem to be immune to saturation. Their results are significantly worse than those obtained by AdaC1 and Cost-Generalized AdaBoost, but despite of being heuristic modifications, AsymBoost and AdaBoost with threshold modification yield better results than Cost-Sensitive AdaBoost, one of the theoretically based approaches.

In addition to its lower performance compared to other tested alternatives, these two algorithms have other practical drawbacks: on the one hand, AdaBoost with threshold modification requires the definition of two training sets, one for symmetric learning and another one for threshold adjustment; on the other hand Asymboost needs to predefine a fixed number of training rounds to keep the desired asymmetric goal, resulting in a significant lack of flexibility (on-the-fly performance tests to stop training must be disabled, and boosted classifiers already learned can not be trimmed).

\subsection{The Theoretical Approaches under an Empirical Perspective}
\label{subsec:algorithms_theoretical}

In the previous paper of the series \citep{LandesaAlba??a} we performed a thorough analysis on the theoretical cost-sensitive AdaBoost variants proposed in the literature, showing that Cost-Sensitive Adaboost/AdaBoostDB and Cost-Generalized AdaBoost drive to different solutions. Now, our empirical results not only corroborate that the obtained classifiers for those algorithms are different, but also that Cost-Generalized AdaBoost outperforms Cost-Sensitive AdaBoost in all the scenarios (Figure \ref{fig_delta_nec_comp}, Table \ref{tab:necglobalbehavior} and Appendix), with a difference that is more marked the greater the asymmetry (Figure \ref{fig_delta_nec_cost}). Our experiments also show that both proposals seem to be immune to saturation (Figure \ref{fig_delta_ce_comp} and Table \ref{tab:ceglobalbehavior}), and that differences in training time (complexity of the algorithms) are vast: Cost-Generalized AdaBoost is 18 times faster than AdaBoostDB, which, in turn, has been shown to be more than 200 times faster than Cost-Sensitive AdaBoost \citep{LandesaAlba13}. 

Another result drawn from our theoretical analysis is that, as Cost-Sensitive Boosting training progresses, the ratio between positive and negative losses tends to decrease, to the extent that class prevalence may end up being inverted (see the accompanying paper \citep{LandesaAlba??a}). Meanwhile, the loss ratio for Cost-Generalized AdaBoost remains constant under the same circumstances throughout the whole training process. To visualize and assess this behavior in practice, we have defined the \emph{Classification Asymmetry} ($CA$) of a given classifier (over its respective test set) as the ratio of correct decisions on positives (true positive rate, $TPR$) to all the correct decisions made by the classifier\footnote{This definition of Classification Asymmetry is only valid for balanced test sets, as those in our experimental framework} (true positive and true negative rates, $TPR+TNR$).

\begin{equation}
\label{classif_asymmetry}
\begin{split}
CA=\frac{TPR}{TPR+TNR}=\frac{1-FPR}{(1-FPR)+(1-FNR)}\\
\end{split}
\end{equation}

Classification asymmetry is, in fact, a measure on how well the classifier is performing in one class with regard to the other one: A value of 0.5 means a balanced behavior, a value greater than 0.5 means that positives tend to be better classified than negatives, and a value lower than 0.5 means that negatives are better classified than positives. At this point is important to remember that, as commented in Section \ref{sec:results}, for whatever cost scenario, the ideal classifier is that making no mistakes in any of the two classes, and such a classifier has a classification asymmetry of 0.5 (it is symmetric). Thus, from an iterative perspective, the classification asymmetry obtained by any cost-sensitive boosting learning scheme should favor the costly class from the beginning, but vanish as the process progresses and the strong classifier evolves to approach the ideal one.

Hence, on the one hand, we have a parameter (the classification asymmetry) that, though biased towards the costly class, should evolve to a more balanced value during boosting learning. On the other hand, we have one algorithm, Cost-Sensitive Boosting, whose loss ratio changes during the learning process to the extent that the role of costly class may be swapped (see the theoretical analysis in the companion paper of the series \citep{LandesaAlba??a}). How are these effects reflected in practice?

\begin{figure}[!htb]
\centerline{
\centering
\includegraphics[width=0.8\columnwidth]{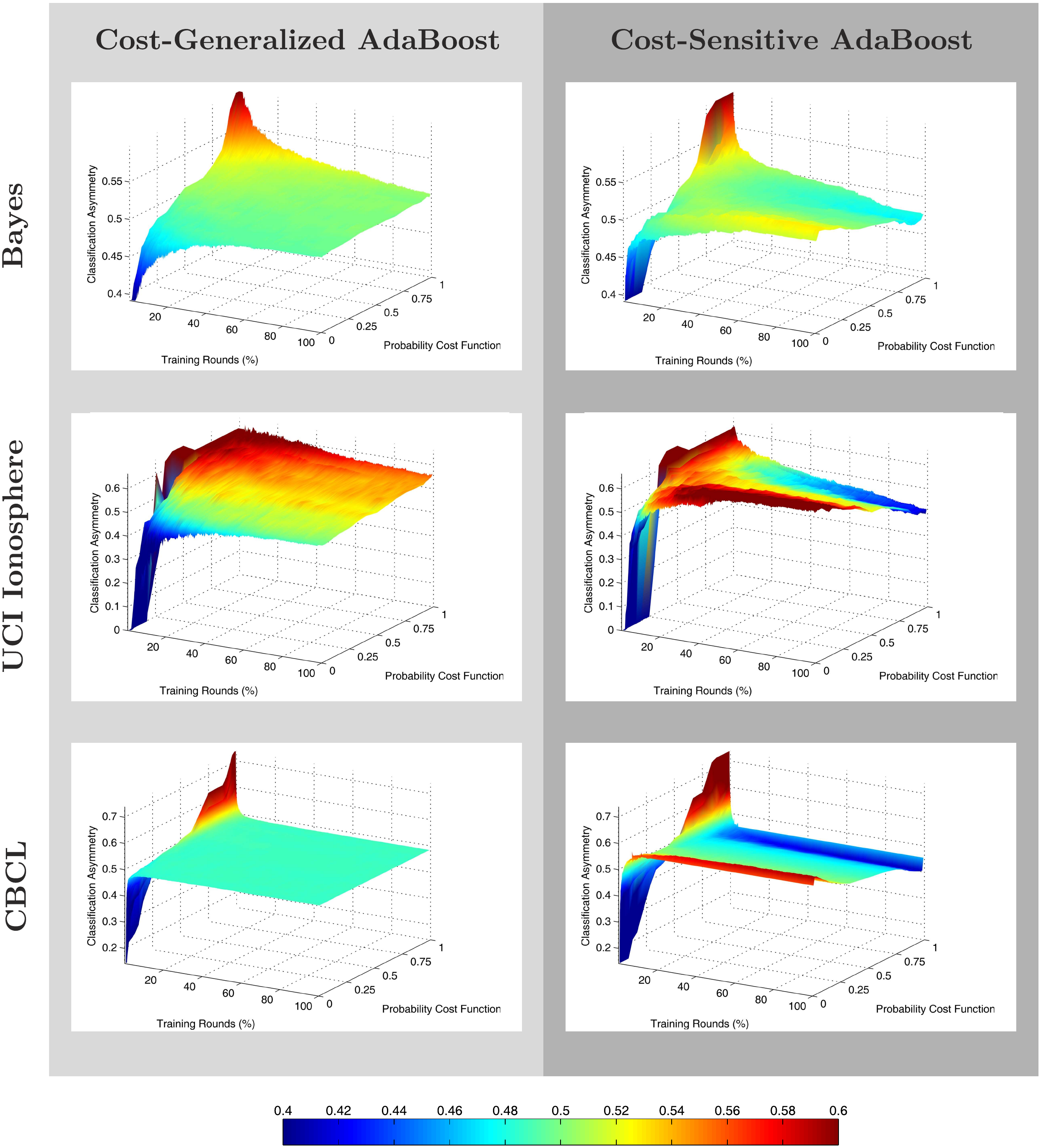}
}
\caption{Classification asymmetry comparative between Cost-Generalized AdaBoost and Cost-Sensitive AdaBoost, across training rounds and costs, for Bayes, UCI Ionosphere and CBCL datasets.}
\label{cga_csa_asym_evol_3d}
\end{figure}

Following our experimental framework, we have gathered the values of Classification Asymmetry obtained by Cost-Sensitive AdaBoost and Cost-Generalized AdaBoost after every training round of every dataset and fold, and for every cost combination. For an easier interpretation, we plotted 3D graphs condensing the obtained data for each dataset and algorithm at a glance, as those shown in Figure \ref{cga_csa_asym_evol_3d}. In these examples we can see, as expected, that classification asymmetry starts biased towards the costly class, and then progressively tends to vanish. But Figure \ref{cga_csa_asym_evol_3d} also makes empirically explicit the other effect we talked about: for Cost-Sensitive Boosting, classification asymmetry ends up being swapped, reaching final classifiers with an asymmetry opposite to that originally intended.

While asymmetry vanishing is a normal trend, it is straightforward to understand that swapping is an undesirable effect for a cost-sensitive boosted classifier, and it may be responsible for the performance decrease shown by Cost-Sensitive AdaBoost.

To build Figure \ref{cga_csa_asym_evol_3d} we selected those scenarios from our experimental framework in which asymmetry swapping is more evident. In the remaining examples swapping is more subtle, though it is still probably a detrimental factor on the overall performance yielded by Cost-Sensitive AdaBoost. It is also important to bear in mind that asymmetry swapping should be stronger the closer to overfitting the classifier is and, as explained in Section \ref{sec:experimental_framework}, our experimental data was gathered following a framework in which the risk of overfitting is highly controlled.

\subsection{Comparison of the Two Best Solutions}
\label{subsec:algorithms_best}

As a result of our empirical analysis, two algorithms stand above the rest: AdaC1 and Cost-Generalized AdaBoost, with performance figures that are consistently better than those shown by the other alternatives in virtually all the tested scenarios. Of these two algorithms, AdaC1 presents an overall performance slightly higher than Cost-Generalized AdaBoost (Figure \ref{fig_delta_nec_comp} and Table \ref{tab:necglobalbehavior}), that is kept across all the tested costs (Figure \ref{fig_delta_nec_cost}).

From a theoretical point of view, bearing in mind that AcaC1 is an algorithm of a heuristic nature while Cost-Generalized AdaBoost has an entirely theoretical derivation behind, we could expect that the latter would provide better results than the former. What is happening? Where is the supposed ``theoretical advantage'' of Cost-Generalized AdaBoost?

Inspecting Figure \ref{fig_delta_ce_comp} and Table \ref{tab:ceglobalbehavior} we can appreciate that both algorithms are far from those experimenting saturation problems, but, in comparative terms, Cost-Generalized AdaBoost has a higher and much more stable discriminative power than AdaC1. In fact, if we analyze $\Delta CE$ across costs (see Figure \ref{fig_delta_ce_cost}) we will appreciate that, though for moderate asymmetries AdaC1 is slightly more discriminative than Cost-Generalized AdaBoost, for high asymmetries the discriminative power of AdaC1 is significantly lower than that of Cost-Generalized AdaBoost. This behavior seems to suggest that AdaC1 applies asymmetry more strongly.

If we return to the algorithmic definition of AdaC1 (see the accompanying paper \citep{LandesaAlba??a}), we can see that the error measurement used to select the weak classifier at each iteration incorporates both the current weight distribution \emph{and} the cost associated to each example of the training dataset. The goodness parameter $\alpha_t$ related to the selected classifier is computed from that same error measurement, and weights are then updated by exponential factors depending on $\alpha_t$ and on the costs associated to each training example. This scheme implies that costs are actually included twice in the weight update rule of AdaC1: once through $\alpha_t$ (that depends on the error measurement which, in turn, depends on costs) and again through the direct incorporation of the costs to the exponential. Instead, the weight update rule in Cost-Generalized AdaBoost embeds asymmetry only through the $\alpha_t$ parameter. This ``two-way'' asymmetry embedding in the weight update of AdaC1 is the reason why it has a stronger cost-sensitive behavior than Cost-Generalized AdaBoost, and the core of the difference between the two algorithms.

\begin{figure}[!htb]
\centerline{
\centering
\includegraphics[width=\columnwidth]{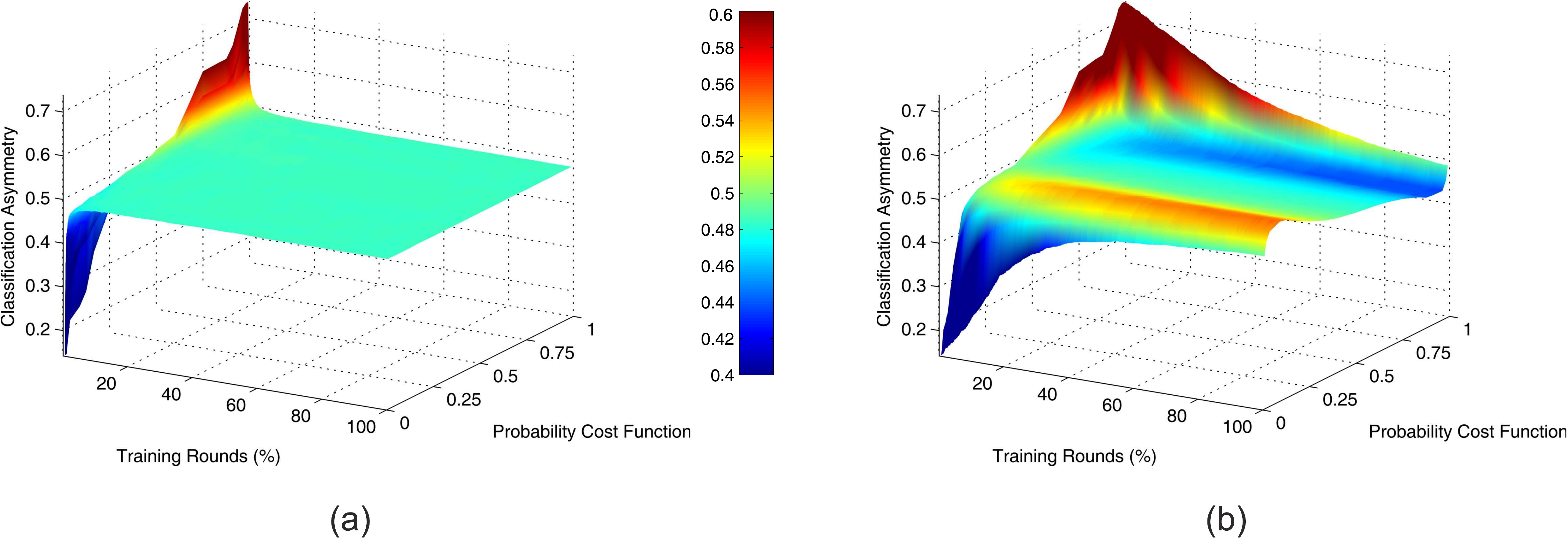}
}
\caption{Classification asymmetry comparative between Cost-Generalized AdaBoost and AdaC1,  across training rounds and costs, for the CBCL dataset.}
\label{cga_ac1_asym_evol_3d}
\end{figure}

After the empirical results we are showing, it seems difficult to find the supposed advantage of using Cost-Generalized AdaBoost with regard to AdaC1, due to the formal guarantees (intact compared to classical AdaBoost) of the former. As commented before, both algorithms are, undoubtedly, the two algorithms showing best performance, but AdaC1 gets results slightly better than Cost-Generalized AdaBoost after a broad empirical analysis. However, there is one scenario of our experimental framework in which Cost-Generalized AdaBoost seems to outperform AdaC1: the CBCL database (see Appendix, particularly, Figure A-8). Performing the ``Classification Asymmetry'' analysis proposed in the previous subsection we obtained the 3D graphs depicted in Figure \ref{cga_ac1_asym_evol_3d}. As can be seen, similarly to Cost-Sensitive AdaBoost, AdaC1 seems to suffer from some kind of asymmetry swapping at medium to high costs, obtaining classifiers with an asymmetry reversed to that desired.

The key to this problem may be in the loss functions somehow defined by the heuristic modifications inside AdaC1. The ratio between the positive and negative losses may be changing throughout the iterative learning process, leading, at some point, to a class prevalence change, similarly to Cost-Sensitive AdaBoost (but, probably, in a more moderate way). It is important to remember again that asymmetry swapping tends to be more detrimental the closer the obtained classifier is to overfitting, which will depend on several parameters, like the complexity of the problem, the complexity of the weak hypothesis, the number of training rounds or the number of training samples. In fact, CBCL is, by far, the most complex problem in our experimental framework: the one with most training samples, the one with a largest pool of weak classifiers and the one needing more boosting training rounds to reach its goals. In such conditions, Cost-Generalized AdaBoost remains totally immune to asymmetry swapping and other detrimental effects, showing us that its stability and versatility, are the real advantage of its full theoretical derivation and the associated formal guarantees, inherited from standard classical AdaBoost.

\section{Conclusions}
\label{sec:Conclusions2}

In this series of two papers we have performed, both theoretically (Part I) and practically (Part II), a thorough analysis on the different cost-sensitive variants of AdaBoost proposed in the literature, in order to provide a unifying framework for their definition, classification, comparison and assessment.

In Part I \citep{LandesaAlba??a}, after presenting, in a homogeneous framework, the bunch of algorithms of interest (AdaBoost with threshold modification \citep{ViolaJones04}; AsymBoost \citep{ViolaJones02}; AdaCost \citep{Fan99}; CSB0, CSB1 and CSB2 \citep{Ting98,Ting00}; AdaC1, AdaC2 and AdaC3 \citep{Sun05, Sun07}; Cost-Sensitive AdaBoost \citep{MasnadiVasconcelos07, MasnadiVasconcelos11}; AdaBoostDB \citep{LandesaAlba13}; and Cost-Generalized AdaBoost \citep{LandesaAlba12}) we proposed a clustering scheme to group them depending on the way asymmetry is inserted in the learning process (\emph{theoretically}, \emph{heuristically} or \emph{a posteriori}), and performed a deep theoretical analysis of those methods with a full theoretical derivation (Cost-Sensitive AdaBoost, AdaBoostDB and Cost-Generalized AdaBoost). Such analysis revealed that the asymmetric weight initialization giving rise to Cost-Generalized AdaBoost, the simplest mechanism of all (disregarded in many previous works), was shown to provide the most consistent error bound definition, being able to preserve the class-dependent loss ratio regardless of the training round, whereas Cost-Sensitive AdaBoost/AdaBoostDB seem to emphasize the least costly class as training progresses (\emph{asymmetry swapping} effect).

The present paper has covered the experimental part of our work, showing a detailed analysis of the behavior of all the algorithms under study (fully-theoretical or not) over a large range of classification problems. The obtained results confirm that Cost-Sensitive AdaBoost and AdaBoostDB suffer, in practice, from the asymmetry swapping effect disclosed on Part I of our work and are clearly outperformed by Cost-Generalized AdaBoost in all the scenarios. On the other hand, most of the heuristic algorithms have very low performance and are prone to obtain saturated classifiers (``all-positives'' or ``all-negatives'').

There are two algorithms that stand above the rest after our empirical analysis: AdaC1 (heuristic) and Cost-Generalized AdaBoost (theoretical). However, despite AdaC1's overall performance is slightly better than that of Cost-Generalized Adaboost, our tests suggest that the former potentially suffers from asymmetry swapping, which can be a side effect of the heuristic definition of the algorithm.

Hence, from a theoretical, practical or even algorithmic perspective, our unifying analysis points to Cost-Generalized AdaBoost as the best and also simplest option to provide AdaBoost with cost-sensitive sound capabilities, preserving all the formal guarantees from the classical (cost-insensitive) version of AdaBoost and exactly its same computational burden.

% Acknowledgements should go at the end, before appendices and references
\section{Acknowledgements}
\label{sec:acknowledgements}{This work has been supported by the Galician Government through the Research Contract GRC2014/024 (Modalidade: Grupos de Referencia Competitiva 2014), and also by the Spanish Government and the European Regional Development Fund (ERDF) under project TACTICA.}

%%%%%%%%%%%%%%%%%%%%%%%
%% PART 2 APPENDICES %%
%%%%%%%%%%%%%%%%%%%%%%%
\clearpage
%%%%%%%%%%%%%%%%%%%%%%%%%%%%%%%%%%%
%% PART 2 APPENDICES
%%%%%%%%%%%%%%%%%%%%%%%%%%%%%%%%%%%

% Manual newpage inserted to improve layout of sample file - not
% needed in general before appendices/bibliography.

\newpage

\begin{appendices}

\phantomsection
\addcontentsline{toc}{section}{\protect\numberline{}Appendix}
%\pdfbookmark{Appendix}{Appendix}

%\renewcommand{\thesection}{\Alph{section}}
%
%\newcounter{appcount2}
%\renewcommand{\theappcount2}{\alph{appcount2}}
%
\setcounter{section}{0}
\setcounter{figure}{0}
\setcounter{table}{0}
\makeatletter 
\renewcommand{\thefigure}{A-\@arabic\c@figure}
\makeatother

\makeatletter 
\renewcommand{\thetable}{A-\@arabic\c@table}
\makeatother

\newpage

\section{Detailed Experimental Results} 
\label{app:results}

Detailed results obtained by our experimental framework are shown in the following tables and figures.

%%%%%%%%%%%%%
%%% BAYES %%%
%%%%%%%%%%%%%

% TABLE

{
\tiny
\begin{longtabu} to \textwidth {llX[$l]@{}@{}X[$l]@{}@{}X[$l]@{}@{}X[$l]@{}@{}l@{}llX[$l]@{}@{}X[$l]@{}@{}X[$l]@{}@{}X[$l]@{}}

%{llX[$l]@{}@{}X[$l]@{}@{}X[$l]@{}@{}X[$l]@{}@{}l@{}llX[$l]@{}@{}X[$l]@{}@{}X[$l]@{}@{}X[$l]@{}}

\toprule
{\textbf{Cost}}  & {\textbf{Alg}}   & {\textbf{FNR}}    & {\textbf{FPR}}    & {\textbf{CE}}    & {\textbf{NEC}}   && {\textbf{Cost}}  & {\textbf{Alg}}   & {\textbf{FNR}}    & {\textbf{FPR}}    & {\textbf{CE}} & {\textbf{NEC}}\\* 
\cmidrule(r){1-6} \cmidrule(r){8-13}
\endfirsthead

%\multicolumn{13}{l}% {\tiny{Continued from previous page}} \\*
\toprule
{\textbf{Cost}}  & {\textbf{Alg}}   & {\textbf{FNR}}    & {\textbf{FPR}}    & {\textbf{CE}}    & {\textbf{NEC}}   && {\textbf{Cost}}  & {\textbf{Alg}}   & {\textbf{FNR}}    & {\textbf{FPR}}    & {\textbf{CE}} & {\textbf{NEC}}\\* 
\cmidrule(r){1-6} \cmidrule(r){8-13}
\endhead

\multicolumn{13}{r}{{\tiny{\tablename\ \thetable{} - Continued on next page $\rightarrow$}}} \\* 
\endfoot

\bottomrule
\caption{Results obtained for the Bayes Dataset.}
\label{tab:bayes_perform}%
\endlastfoot

    $[1, 100]$ & BAY   & 6.20\cdot10^{-1} & 0.000    & 3.10\cdot10^{-1} & 6.14\cdot10^{-3} &       & $[1, 50]$ & BAY   & 5.28\cdot10^{-1} & 0.000    & 2.64\cdot10^{-1} & 1.04\cdot10^{-2} \\*
          & ABT   & 1.29\cdot10^{-1} & 5.22\cdot10^{-2} & 9.04\cdot10^{-2} & 5.30\cdot10^{-2} &       &       & ABT   & 8.84\cdot10^{-2} & 7.23\cdot10^{-2} & 8.03\cdot10^{-2} & 7.26\cdot10^{-2} \\*
          & ASB   & 8.84\cdot10^{-2} & 6.02\cdot10^{-2} & 7.43\cdot10^{-2} & 6.05\cdot10^{-2} &       &       & ASB   & 8.43\cdot10^{-2} & 6.02\cdot10^{-2} & 7.23\cdot10^{-2} & 6.07\cdot10^{-2} \\*
          & ADC   & 3.61\cdot10^{-1} & 8.03\cdot10^{-3} & 1.85\cdot10^{-1} & 1.15\cdot10^{-2} &       &       & ADC   & 3.61\cdot10^{-1} & 8.03\cdot10^{-3} & 1.85\cdot10^{-1} & 1.50\cdot10^{-2} \\*
          & CB0   & 3.61\cdot10^{-1} & 8.03\cdot10^{-3} & 1.85\cdot10^{-1} & 1.15\cdot10^{-2} &       &       & CB0   & 3.61\cdot10^{-1} & 8.03\cdot10^{-3} & 1.85\cdot10^{-1} & 1.50\cdot10^{-2} \\*
          & CB1   & 1.000    & 0.000    & 5.00\cdot10^{-1} & 9.90\cdot10^{-3} &       &       & CB1   & 1.000    & 0.000    & 5.00\cdot10^{-1} & 1.96\cdot10^{-2} \\*
          & CB2   & 9.44\cdot10^{-1} & 0.000    & 4.72\cdot10^{-1} & 9.34\cdot10^{-3} &       &       & CB2   & 9.60\cdot10^{-1} & 0.000    & 4.80\cdot10^{-1} & 1.88\cdot10^{-2} \\*
          & AC1   & 3.61\cdot10^{-1} & 8.03\cdot10^{-3} & 1.85\cdot10^{-1} & 1.15\cdot10^{-2} &       &       & AC1   & 3.41\cdot10^{-1} & 1.20\cdot10^{-2} & 1.77\cdot10^{-1} & 1.85\cdot10^{-2} \\*
          & AC2   & 6.63\cdot10^{-1} & 4.02\cdot10^{-3} & 3.33\cdot10^{-1} & 1.05\cdot10^{-2} &       &       & AC2   & 9.28\cdot10^{-1} & 0.000    & 4.64\cdot10^{-1} & 1.82\cdot10^{-2} \\*
          & AC3   & 3.61\cdot10^{-1} & 8.03\cdot10^{-3} & 1.85\cdot10^{-1} & 1.15\cdot10^{-2} &       &       & AC3   & 3.65\cdot10^{-1} & 8.03\cdot10^{-3} & 1.87\cdot10^{-1} & 1.50\cdot10^{-2} \\*
          & CSA   & 1.69\cdot10^{-1} & 4.82\cdot10^{-2} & 1.08\cdot10^{-1} & 4.94\cdot10^{-2} &       &       & CSA   & 1.20\cdot10^{-1} & 4.82\cdot10^{-2} & 8.43\cdot10^{-2} & 4.96\cdot10^{-2} \\*
          & CGA   & 1.77\cdot10^{-1} & 2.81\cdot10^{-2} & 1.02\cdot10^{-1} & 2.96\cdot10^{-2} &       &       & CGA   & 1.45\cdot10^{-1} & 4.82\cdot10^{-2} & 9.64\cdot10^{-2} & 5.01\cdot10^{-2} \\*
					\cmidrule(r){1-6} \cmidrule(r){8-13}
    $[1, 25]$ & BAY   & 4.24\cdot10^{-1} & 4.00\cdot10^{-3} & 2.14\cdot10^{-1} & 2.02\cdot10^{-2} &       & $[1, 10]$ & BAY   & 2.76\cdot10^{-1} & 1.20\cdot10^{-2} & 1.44\cdot10^{-1} & 3.60\cdot10^{-2} \\*
          & ABT   & 7.23\cdot10^{-2} & 7.23\cdot10^{-2} & 7.23\cdot10^{-2} & 7.23\cdot10^{-2} &       &       & ABT   & 8.43\cdot10^{-2} & 7.63\cdot10^{-2} & 8.03\cdot10^{-2} & 7.70\cdot10^{-2} \\*
          & ASB   & 9.24\cdot10^{-2} & 6.43\cdot10^{-2} & 7.83\cdot10^{-2} & 6.53\cdot10^{-2} &       &       & ASB   & 8.43\cdot10^{-2} & 6.43\cdot10^{-2} & 7.43\cdot10^{-2} & 6.61\cdot10^{-2} \\*
          & ADC   & 3.09\cdot10^{-1} & 1.20\cdot10^{-2} & 1.61\cdot10^{-1} & 2.35\cdot10^{-2} &       &       & ADC   & 2.33\cdot10^{-1} & 2.01\cdot10^{-2} & 1.27\cdot10^{-1} & 3.94\cdot10^{-2} \\*
          & CB0   & 3.09\cdot10^{-1} & 1.20\cdot10^{-2} & 1.61\cdot10^{-1} & 2.35\cdot10^{-2} &       &       & CB0   & 3.61\cdot10^{-1} & 8.03\cdot10^{-3} & 1.85\cdot10^{-1} & 4.02\cdot10^{-2} \\*
          & CB1   & 1.000    & 0.000    & 5.00\cdot10^{-1} & 3.85\cdot10^{-2} &       &       & CB1   & 1.000    & 0.000    & 5.00\cdot10^{-1} & 9.09\cdot10^{-2} \\*
          & CB2   & 9.88\cdot10^{-1} & 0.000    & 4.94\cdot10^{-1} & 3.80\cdot10^{-2} &       &       & CB2   & 1.000    & 0.000    & 5.00\cdot10^{-1} & 9.09\cdot10^{-2} \\*
          & AC1   & 2.69\cdot10^{-1} & 2.01\cdot10^{-2} & 1.45\cdot10^{-1} & 2.97\cdot10^{-2} &       &       & AC1   & 1.57\cdot10^{-1} & 3.21\cdot10^{-2} & 9.44\cdot10^{-2} & 4.34\cdot10^{-2} \\*
          & AC2   & 6.10\cdot10^{-1} & 0.000    & 3.05\cdot10^{-1} & 2.35\cdot10^{-2} &       &       & AC2   & 6.14\cdot10^{-1} & 4.02\cdot10^{-3} & 3.09\cdot10^{-1} & 5.95\cdot10^{-2} \\*
          & AC3   & 4.18\cdot10^{-1} & 4.02\cdot10^{-3} & 2.11\cdot10^{-1} & 1.99\cdot10^{-2} &       &       & AC3   & 4.22\cdot10^{-1} & 4.02\cdot10^{-3} & 2.13\cdot10^{-1} & 4.20\cdot10^{-2} \\*
          & CSA   & 7.63\cdot10^{-2} & 8.84\cdot10^{-2} & 8.23\cdot10^{-2} & 8.79\cdot10^{-2} &       &       & CSA   & 6.43\cdot10^{-2} & 8.84\cdot10^{-2} & 7.63\cdot10^{-2} & 8.62\cdot10^{-2} \\*
          & CGA   & 1.24\cdot10^{-1} & 3.61\cdot10^{-2} & 8.03\cdot10^{-2} & 3.95\cdot10^{-2} &       &       & CGA   & 1.00\cdot10^{-1} & 4.82\cdot10^{-2} & 7.43\cdot10^{-2} & 5.29\cdot10^{-2} \\*
					\cmidrule(r){1-6} \cmidrule(r){8-13}
    $[1, 7]$ & BAY   & 2.36\cdot10^{-1} & 1.60\cdot10^{-2} & 1.26\cdot10^{-1} & 4.35\cdot10^{-2} &       & $[1, 5]$ & BAY   & 2.08\cdot10^{-1} & 1.60\cdot10^{-2} & 1.12\cdot10^{-1} & 4.80\cdot10^{-2} \\*
          & ABT   & 7.23\cdot10^{-2} & 7.63\cdot10^{-2} & 7.43\cdot10^{-2} & 7.58\cdot10^{-2} &       &       & ABT   & 6.83\cdot10^{-2} & 8.43\cdot10^{-2} & 7.63\cdot10^{-2} & 8.17\cdot10^{-2} \\*
          & ASB   & 8.84\cdot10^{-2} & 6.02\cdot10^{-2} & 7.43\cdot10^{-2} & 6.38\cdot10^{-2} &       &       & ASB   & 8.43\cdot10^{-2} & 6.43\cdot10^{-2} & 7.43\cdot10^{-2} & 6.76\cdot10^{-2} \\*
          & ADC   & 2.93\cdot10^{-1} & 1.61\cdot10^{-2} & 1.55\cdot10^{-1} & 5.07\cdot10^{-2} &       &       & ADC   & 2.57\cdot10^{-1} & 2.01\cdot10^{-2} & 1.39\cdot10^{-1} & 5.96\cdot10^{-2} \\*
          & CB0   & 3.61\cdot10^{-1} & 8.03\cdot10^{-3} & 1.85\cdot10^{-1} & 5.22\cdot10^{-2} &       &       & CB0   & 3.61\cdot10^{-1} & 8.03\cdot10^{-3} & 1.85\cdot10^{-1} & 6.69\cdot10^{-2} \\*
          & CB1   & 1.000    & 0.000    & 5.00\cdot10^{-1} & 1.25\cdot10^{-1} &       &       & CB1   & 1.000    & 0.000    & 5.00\cdot10^{-1} & 1.67\cdot10^{-1} \\*
          & CB2   & 1.000    & 0.000    & 5.00\cdot10^{-1} & 1.25\cdot10^{-1} &       &       & CB2   & 1.000    & 0.000    & 5.00\cdot10^{-1} & 1.67\cdot10^{-1} \\*
          & AC1   & 1.12\cdot10^{-1} & 3.61\cdot10^{-2} & 7.43\cdot10^{-2} & 4.57\cdot10^{-2} &       &       & AC1   & 8.43\cdot10^{-2} & 6.02\cdot10^{-2} & 7.23\cdot10^{-2} & 6.43\cdot10^{-2} \\*
          & AC2   & 7.91\cdot10^{-1} & 4.02\cdot10^{-3} & 3.98\cdot10^{-1} & 1.02\cdot10^{-1} &       &       & AC2   & 5.98\cdot10^{-1} & 4.02\cdot10^{-3} & 3.01\cdot10^{-1} & 1.03\cdot10^{-1} \\*
          & AC3   & 4.22\cdot10^{-1} & 4.02\cdot10^{-3} & 2.13\cdot10^{-1} & 5.62\cdot10^{-2} &       &       & AC3   & 4.22\cdot10^{-1} & 4.02\cdot10^{-3} & 2.13\cdot10^{-1} & 7.36\cdot10^{-2} \\*
          & CSA   & 6.02\cdot10^{-2} & 7.23\cdot10^{-2} & 6.63\cdot10^{-2} & 7.08\cdot10^{-2} &       &       & CSA   & 6.83\cdot10^{-2} & 6.43\cdot10^{-2} & 6.63\cdot10^{-2} & 6.49\cdot10^{-2} \\*
          & CGA   & 1.08\cdot10^{-1} & 5.22\cdot10^{-2} & 8.03\cdot10^{-2} & 5.92\cdot10^{-2} &       &       & CGA   & 1.00\cdot10^{-1} & 5.62\cdot10^{-2} & 7.83\cdot10^{-2} & 6.36\cdot10^{-2} \\*
					\cmidrule(r){1-6} \cmidrule(r){8-13}
		      
    $[1, 3]$ & BAY   & 1.56\cdot10^{-1} & 2.00\cdot10^{-2} & 8.80\cdot10^{-2} & 5.40\cdot10^{-2} &       & $[1, 2]$ & BAY   & 1.24\cdot10^{-1} & 2.80\cdot10^{-2} & 7.60\cdot10^{-2} & 6.00\cdot10^{-2} \\*
          & ABT   & 8.43\cdot10^{-2} & 7.63\cdot10^{-2} & 8.03\cdot10^{-2} & 7.83\cdot10^{-2} &       &       & ABT   & 8.43\cdot10^{-2} & 8.03\cdot10^{-2} & 8.23\cdot10^{-2} & 8.17\cdot10^{-2} \\*
          & ASB   & 8.43\cdot10^{-2} & 6.43\cdot10^{-2} & 7.43\cdot10^{-2} & 6.93\cdot10^{-2} &       &       & ASB   & 8.03\cdot10^{-2} & 6.43\cdot10^{-2} & 7.23\cdot10^{-2} & 6.96\cdot10^{-2} \\*
          & ADC   & 1.49\cdot10^{-1} & 2.81\cdot10^{-2} & 8.84\cdot10^{-2} & 5.82\cdot10^{-2} &       &       & ADC   & 1.04\cdot10^{-1} & 4.82\cdot10^{-2} & 7.63\cdot10^{-2} & 6.69\cdot10^{-2} \\*
          & CB0   & 3.61\cdot10^{-1} & 8.03\cdot10^{-3} & 1.85\cdot10^{-1} & 9.64\cdot10^{-2} &       &       & CB0   & 3.61\cdot10^{-1} & 8.03\cdot10^{-3} & 1.85\cdot10^{-1} & 1.26\cdot10^{-1} \\*
          & CB1   & 1.000    & 0.000    & 5.00\cdot10^{-1} & 2.50\cdot10^{-1} &       &       & CB1   & 1.000    & 0.000    & 5.00\cdot10^{-1} & 3.33\cdot10^{-1} \\*
          & CB2   & 1.000    & 0.000    & 5.00\cdot10^{-1} & 2.50\cdot10^{-1} &       &       & CB2   & 9.76\cdot10^{-1} & 0.000    & 4.88\cdot10^{-1} & 3.25\cdot10^{-1} \\*
          & AC1   & 7.23\cdot10^{-2} & 6.02\cdot10^{-2} & 6.63\cdot10^{-2} & 6.33\cdot10^{-2} &       &       & AC1   & 7.63\cdot10^{-2} & 6.83\cdot10^{-2} & 7.23\cdot10^{-2} & 7.10\cdot10^{-2} \\*
          & AC2   & 7.91\cdot10^{-1} & 8.03\cdot10^{-3} & 4.00\cdot10^{-1} & 2.04\cdot10^{-1} &       &       & AC2   & 4.66\cdot10^{-1} & 4.02\cdot10^{-3} & 2.35\cdot10^{-1} & 1.58\cdot10^{-1} \\*
          & AC3   & 4.22\cdot10^{-1} & 4.02\cdot10^{-3} & 2.13\cdot10^{-1} & 1.08\cdot10^{-1} &       &       & AC3   & 6.39\cdot10^{-1} & 4.02\cdot10^{-3} & 3.21\cdot10^{-1} & 2.16\cdot10^{-1} \\*
          & CSA   & 6.43\cdot10^{-2} & 7.23\cdot10^{-2} & 6.83\cdot10^{-2} & 7.03\cdot10^{-2} &       &       & CSA   & 7.23\cdot10^{-2} & 7.63\cdot10^{-2} & 7.43\cdot10^{-2} & 7.50\cdot10^{-2} \\*
          & CGA   & 9.64\cdot10^{-2} & 5.62\cdot10^{-2} & 7.63\cdot10^{-2} & 6.63\cdot10^{-2} &       &       & CGA   & 8.03\cdot10^{-2} & 6.83\cdot10^{-2} & 7.43\cdot10^{-2} & 7.23\cdot10^{-2} \\*
					\cmidrule(r){1-6} \cmidrule(r){8-13}
    $[2, 3]$ & BAY   & 9.20\cdot10^{-2} & 3.60\cdot10^{-2} & 6.40\cdot10^{-2} & 5.84\cdot10^{-2} &       & $[1, 1]$ & BAY   & 6.80\cdot10^{-2} & 5.60\cdot10^{-2} & 6.20\cdot10^{-2} & 6.20\cdot10^{-2} \\*
          & ABT   & 8.43\cdot10^{-2} & 8.03\cdot10^{-2} & 8.23\cdot10^{-2} & 8.19\cdot10^{-2} &       &       & ABT   & 8.43\cdot10^{-2} & 7.63\cdot10^{-2} & 8.03\cdot10^{-2} & 8.03\cdot10^{-2} \\*
          & ASB   & 8.43\cdot10^{-2} & 7.23\cdot10^{-2} & 7.83\cdot10^{-2} & 7.71\cdot10^{-2} &       &       & ASB   & 8.43\cdot10^{-2} & 7.63\cdot10^{-2} & 8.03\cdot10^{-2} & 8.03\cdot10^{-2} \\*
          & ADC   & 9.24\cdot10^{-2} & 4.82\cdot10^{-2} & 7.03\cdot10^{-2} & 6.59\cdot10^{-2} &       &       & ADC   & 5.62\cdot10^{-2} & 7.63\cdot10^{-2} & 6.63\cdot10^{-2} & 6.63\cdot10^{-2} \\*
          & CB0   & 3.61\cdot10^{-1} & 8.03\cdot10^{-3} & 1.85\cdot10^{-1} & 1.49\cdot10^{-1} &       &       & CB0   & 5.62\cdot10^{-2} & 7.63\cdot10^{-2} & 6.63\cdot10^{-2} & 6.63\cdot10^{-2} \\*
          & CB1   & 9.56\cdot10^{-1} & 0.000    & 4.78\cdot10^{-1} & 3.82\cdot10^{-1} &       &       & CB1   & 8.03\cdot10^{-2} & 5.22\cdot10^{-2} & 6.63\cdot10^{-2} & 6.63\cdot10^{-2} \\*
          & CB2   & 8.35\cdot10^{-1} & 0.000    & 4.18\cdot10^{-1} & 3.34\cdot10^{-1} &       &       & CB2   & 8.43\cdot10^{-2} & 7.63\cdot10^{-2} & 8.03\cdot10^{-2} & 8.03\cdot10^{-2} \\*
          & AC1   & 8.03\cdot10^{-2} & 6.83\cdot10^{-2} & 7.43\cdot10^{-2} & 7.31\cdot10^{-2} &       &       & AC1   & 8.43\cdot10^{-2} & 6.02\cdot10^{-2} & 7.23\cdot10^{-2} & 7.23\cdot10^{-2} \\*
          & AC2   & 6.43\cdot10^{-1} & 4.02\cdot10^{-3} & 3.23\cdot10^{-1} & 2.59\cdot10^{-1} &       &       & AC2   & 8.43\cdot10^{-2} & 7.63\cdot10^{-2} & 8.03\cdot10^{-2} & 8.03\cdot10^{-2} \\*
          & AC3   & 7.19\cdot10^{-1} & 4.02\cdot10^{-3} & 3.61\cdot10^{-1} & 2.90\cdot10^{-1} &       &       & AC3   & 8.43\cdot10^{-2} & 6.02\cdot10^{-2} & 7.23\cdot10^{-2} & 7.23\cdot10^{-2} \\*
          & CSA   & 6.83\cdot10^{-2} & 6.83\cdot10^{-2} & 6.83\cdot10^{-2} & 6.83\cdot10^{-2} &       &       & CSA   & 8.43\cdot10^{-2} & 7.63\cdot10^{-2} & 8.03\cdot10^{-2} & 8.03\cdot10^{-2} \\*
          & CGA   & 8.03\cdot10^{-2} & 7.23\cdot10^{-2} & 7.63\cdot10^{-2} & 7.55\cdot10^{-2} &       &       & CGA   & 8.43\cdot10^{-2} & 7.63\cdot10^{-2} & 8.03\cdot10^{-2} & 8.03\cdot10^{-2} \\*
					\cmidrule(r){1-6} \cmidrule(r){8-13}
    $[3, 2]$ & BAY   & 4.40\cdot10^{-2} & 8.80\cdot10^{-2} & 6.60\cdot10^{-2} & 6.16\cdot10^{-2} &       & $[2, 1]$ & BAY   & 3.60\cdot10^{-2} & 1.12\cdot10^{-1} & 7.40\cdot10^{-2} & 6.13\cdot10^{-2} \\*
          & ABT   & 8.03\cdot10^{-2} & 8.03\cdot10^{-2} & 8.03\cdot10^{-2} & 8.03\cdot10^{-2} &       &       & ABT   & 8.03\cdot10^{-2} & 8.03\cdot10^{-2} & 8.03\cdot10^{-2} & 8.03\cdot10^{-2} \\*
          & ASB   & 8.43\cdot10^{-2} & 8.03\cdot10^{-2} & 8.23\cdot10^{-2} & 8.27\cdot10^{-2} &       &       & ASB   & 7.63\cdot10^{-2} & 8.43\cdot10^{-2} & 8.03\cdot10^{-2} & 7.90\cdot10^{-2} \\*
          & ADC   & 5.22\cdot10^{-2} & 8.03\cdot10^{-2} & 6.63\cdot10^{-2} & 6.35\cdot10^{-2} &       &       & ADC   & 4.82\cdot10^{-2} & 8.84\cdot10^{-2} & 6.83\cdot10^{-2} & 6.16\cdot10^{-2} \\*
          & CB0   & 4.02\cdot10^{-3} & 3.29\cdot10^{-1} & 1.67\cdot10^{-1} & 1.34\cdot10^{-1} &       &       & CB0   & 4.02\cdot10^{-3} & 3.29\cdot10^{-1} & 1.67\cdot10^{-1} & 1.12\cdot10^{-1} \\*
          & CB1   & 0.000    & 9.20\cdot10^{-1} & 4.60\cdot10^{-1} & 3.68\cdot10^{-1} &       &       & CB1   & 0.000    & 9.84\cdot10^{-1} & 4.92\cdot10^{-1} & 3.28\cdot10^{-1} \\*
          & CB2   & 0.000    & 6.22\cdot10^{-1} & 3.11\cdot10^{-1} & 2.49\cdot10^{-1} &       &       & CB2   & 0.000    & 9.92\cdot10^{-1} & 4.96\cdot10^{-1} & 3.31\cdot10^{-1} \\*
          & AC1   & 8.03\cdot10^{-2} & 6.43\cdot10^{-2} & 7.23\cdot10^{-2} & 7.39\cdot10^{-2} &       &       & AC1   & 8.03\cdot10^{-2} & 6.02\cdot10^{-2} & 7.03\cdot10^{-2} & 7.36\cdot10^{-2} \\*
          & AC2   & 0.000    & 9.12\cdot10^{-1} & 4.56\cdot10^{-1} & 3.65\cdot10^{-1} &       &       & AC2   & 4.02\cdot10^{-3} & 8.51\cdot10^{-1} & 4.28\cdot10^{-1} & 2.86\cdot10^{-1} \\*
          & AC3   & 4.02\cdot10^{-3} & 8.03\cdot10^{-1} & 4.04\cdot10^{-1} & 3.24\cdot10^{-1} &       &       & AC3   & 0.000    & 9.92\cdot10^{-1} & 4.96\cdot10^{-1} & 3.31\cdot10^{-1} \\*
          & CSA   & 7.63\cdot10^{-2} & 7.23\cdot10^{-2} & 7.43\cdot10^{-2} & 7.47\cdot10^{-2} &       &       & CSA   & 7.63\cdot10^{-2} & 6.02\cdot10^{-2} & 6.83\cdot10^{-2} & 7.10\cdot10^{-2} \\*
          & CGA   & 6.83\cdot10^{-2} & 8.43\cdot10^{-2} & 7.63\cdot10^{-2} & 7.47\cdot10^{-2} &       &       & CGA   & 6.83\cdot10^{-2} & 9.24\cdot10^{-2} & 8.03\cdot10^{-2} & 7.63\cdot10^{-2} \\*
					\cmidrule(r){1-6} \cmidrule(r){8-13}
    $[3, 1]$ & BAY   & 2.80\cdot10^{-2} & 1.64\cdot10^{-1} & 9.60\cdot10^{-2} & 6.20\cdot10^{-2} &       & $[5, 1]$ & BAY   & 2.00\cdot10^{-2} & 2.00\cdot10^{-1} & 1.10\cdot10^{-1} & 5.00\cdot10^{-2} \\*
          & ABT   & 7.23\cdot10^{-2} & 7.63\cdot10^{-2} & 7.43\cdot10^{-2} & 7.33\cdot10^{-2} &       &       & ABT   & 6.83\cdot10^{-2} & 8.43\cdot10^{-2} & 7.63\cdot10^{-2} & 7.10\cdot10^{-2} \\*
          & ASB   & 8.03\cdot10^{-2} & 8.03\cdot10^{-2} & 8.03\cdot10^{-2} & 8.03\cdot10^{-2} &       &       & ASB   & 8.03\cdot10^{-2} & 7.23\cdot10^{-2} & 7.63\cdot10^{-2} & 7.90\cdot10^{-2} \\*
          & ADC   & 3.21\cdot10^{-2} & 1.41\cdot10^{-1} & 8.63\cdot10^{-2} & 5.92\cdot10^{-2} &       &       & ADC   & 1.61\cdot10^{-2} & 2.77\cdot10^{-1} & 1.47\cdot10^{-1} & 5.96\cdot10^{-2} \\*
          & CB0   & 4.02\cdot10^{-3} & 3.41\cdot10^{-1} & 1.73\cdot10^{-1} & 8.84\cdot10^{-2} &       &       & CB0   & 4.02\cdot10^{-3} & 3.29\cdot10^{-1} & 1.67\cdot10^{-1} & 5.82\cdot10^{-2} \\*
          & CB1   & 0.000    & 1.000    & 5.00\cdot10^{-1} & 2.50\cdot10^{-1} &       &       & CB1   & 0.000    & 1.000    & 5.00\cdot10^{-1} & 1.67\cdot10^{-1} \\*
          & CB2   & 0.000    & 1.000    & 5.00\cdot10^{-1} & 2.50\cdot10^{-1} &       &       & CB2   & 0.000    & 1.000    & 5.00\cdot10^{-1} & 1.67\cdot10^{-1} \\*
          & AC1   & 7.63\cdot10^{-2} & 6.43\cdot10^{-2} & 7.03\cdot10^{-2} & 7.33\cdot10^{-2} &       &       & AC1   & 4.82\cdot10^{-2} & 8.03\cdot10^{-2} & 6.43\cdot10^{-2} & 5.35\cdot10^{-2} \\*
          & AC2   & 4.02\cdot10^{-3} & 9.20\cdot10^{-1} & 4.62\cdot10^{-1} & 2.33\cdot10^{-1} &       &       & AC2   & 0.000    & 9.96\cdot10^{-1} & 4.98\cdot10^{-1} & 1.66\cdot10^{-1} \\*
          & AC3   & 0.000    & 1.000    & 5.00\cdot10^{-1} & 2.50\cdot10^{-1} &       &       & AC3   & 0.000    & 1.000    & 5.00\cdot10^{-1} & 1.67\cdot10^{-1} \\*
          & CSA   & 8.43\cdot10^{-2} & 6.43\cdot10^{-2} & 7.43\cdot10^{-2} & 7.93\cdot10^{-2} &       &       & CSA   & 8.84\cdot10^{-2} & 5.62\cdot10^{-2} & 7.23\cdot10^{-2} & 8.30\cdot10^{-2} \\*
          & CGA   & 4.82\cdot10^{-2} & 8.43\cdot10^{-2} & 6.63\cdot10^{-2} & 5.72\cdot10^{-2} &       &       & CGA   & 6.02\cdot10^{-2} & 8.84\cdot10^{-2} & 7.43\cdot10^{-2} & 6.49\cdot10^{-2} \\*
					\cmidrule(r){1-6} \cmidrule(r){8-13}
					
    $[7, 1]$ & BAY   & 2.00\cdot10^{-2} & 2.36\cdot10^{-1} & 1.28\cdot10^{-1} & 4.70\cdot10^{-2} &       & $[10, 1]$ & BAY   & 1.60\cdot10^{-2} & 2.72\cdot10^{-1} & 1.44\cdot10^{-1} & 3.93\cdot10^{-2} \\*
          & ABT   & 6.83\cdot10^{-2} & 8.43\cdot10^{-2} & 7.63\cdot10^{-2} & 7.03\cdot10^{-2} &       &       & ABT   & 6.83\cdot10^{-2} & 8.43\cdot10^{-2} & 7.63\cdot10^{-2} & 6.97\cdot10^{-2} \\*
          & ASB   & 7.63\cdot10^{-2} & 8.03\cdot10^{-2} & 7.83\cdot10^{-2} & 7.68\cdot10^{-2} &       &       & ASB   & 7.23\cdot10^{-2} & 8.03\cdot10^{-2} & 7.63\cdot10^{-2} & 7.30\cdot10^{-2} \\*
          & ADC   & 1.61\cdot10^{-2} & 2.89\cdot10^{-1} & 1.53\cdot10^{-1} & 5.02\cdot10^{-2} &       &       & ADC   & 1.20\cdot10^{-2} & 2.97\cdot10^{-1} & 1.55\cdot10^{-1} & 3.80\cdot10^{-2} \\*
          & CB0   & 4.02\cdot10^{-3} & 3.29\cdot10^{-1} & 1.67\cdot10^{-1} & 4.47\cdot10^{-2} &       &       & CB0   & 4.02\cdot10^{-3} & 3.41\cdot10^{-1} & 1.73\cdot10^{-1} & 3.47\cdot10^{-2} \\*
          & CB1   & 0.000    & 1.000    & 5.00\cdot10^{-1} & 1.25\cdot10^{-1} &       &       & CB1   & 0.000    & 1.000    & 5.00\cdot10^{-1} & 9.09\cdot10^{-2} \\*
          & CB2   & 0.000    & 1.000    & 5.00\cdot10^{-1} & 1.25\cdot10^{-1} &       &       & CB2   & 0.000    & 1.000    & 5.00\cdot10^{-1} & 9.09\cdot10^{-2} \\*
          & AC1   & 4.42\cdot10^{-2} & 1.04\cdot10^{-1} & 7.43\cdot10^{-2} & 5.17\cdot10^{-2} &       &       & AC1   & 4.42\cdot10^{-2} & 1.08\cdot10^{-1} & 7.63\cdot10^{-2} & 5.00\cdot10^{-2} \\*
          & AC2   & 0.000    & 9.56\cdot10^{-1} & 4.78\cdot10^{-1} & 1.19\cdot10^{-1} &       &       & AC2   & 0.000    & 9.92\cdot10^{-1} & 4.96\cdot10^{-1} & 9.02\cdot10^{-2} \\*
          & AC3   & 0.000    & 1.000    & 5.00\cdot10^{-1} & 1.25\cdot10^{-1} &       &       & AC3   & 0.000    & 1.000    & 5.00\cdot10^{-1} & 9.09\cdot10^{-2} \\*
          & CSA   & 9.24\cdot10^{-2} & 6.02\cdot10^{-2} & 7.63\cdot10^{-2} & 8.84\cdot10^{-2} &       &       & CSA   & 8.43\cdot10^{-2} & 5.22\cdot10^{-2} & 6.83\cdot10^{-2} & 8.14\cdot10^{-2} \\*
          & CGA   & 5.22\cdot10^{-2} & 8.84\cdot10^{-2} & 7.03\cdot10^{-2} & 5.67\cdot10^{-2} &       &       & CGA   & 5.22\cdot10^{-2} & 1.00\cdot10^{-1} & 7.63\cdot10^{-2} & 5.66\cdot10^{-2} \\*
					\cmidrule(r){1-6} \cmidrule(r){8-13}
    $[25, 1]$ & BAY   & 4.00\cdot10^{-3} & 4.08\cdot10^{-1} & 2.06\cdot10^{-1} & 1.95\cdot10^{-2} &       & $[50, 1]$ & BAY   & 0.000    & 4.84\cdot10^{-1} & 2.42\cdot10^{-1} & 9.49\cdot10^{-3} \\*
          & ABT   & 6.02\cdot10^{-2} & 8.43\cdot10^{-2} & 7.23\cdot10^{-2} & 6.12\cdot10^{-2} &       &       & ABT   & 6.43\cdot10^{-2} & 8.03\cdot10^{-2} & 7.23\cdot10^{-2} & 6.46\cdot10^{-2} \\*
          & ASB   & 7.63\cdot10^{-2} & 7.23\cdot10^{-2} & 7.43\cdot10^{-2} & 7.62\cdot10^{-2} &       &       & ASB   & 6.43\cdot10^{-2} & 8.43\cdot10^{-2} & 7.43\cdot10^{-2} & 6.47\cdot10^{-2} \\*
          & ADC   & 4.02\cdot10^{-3} & 3.29\cdot10^{-1} & 1.67\cdot10^{-1} & 1.65\cdot10^{-2} &       &       & ADC   & 4.02\cdot10^{-3} & 3.29\cdot10^{-1} & 1.67\cdot10^{-1} & 1.04\cdot10^{-2} \\*
          & CB0   & 4.02\cdot10^{-3} & 3.29\cdot10^{-1} & 1.67\cdot10^{-1} & 1.65\cdot10^{-2} &       &       & CB0   & 4.02\cdot10^{-3} & 3.29\cdot10^{-1} & 1.67\cdot10^{-1} & 1.04\cdot10^{-2} \\*
          & CB1   & 0.000    & 1.000    & 5.00\cdot10^{-1} & 3.85\cdot10^{-2} &       &       & CB1   & 0.000    & 1.000    & 5.00\cdot10^{-1} & 1.96\cdot10^{-2} \\*
          & CB2   & 0.000    & 1.000    & 5.00\cdot10^{-1} & 3.85\cdot10^{-2} &       &       & CB2   & 0.000    & 1.000    & 5.00\cdot10^{-1} & 1.96\cdot10^{-2} \\*
          & AC1   & 8.03\cdot10^{-3} & 3.25\cdot10^{-1} & 1.67\cdot10^{-1} & 2.02\cdot10^{-2} &       &       & AC1   & 4.02\cdot10^{-3} & 3.29\cdot10^{-1} & 1.67\cdot10^{-1} & 1.04\cdot10^{-2} \\*
          & AC2   & 0.000    & 9.96\cdot10^{-1} & 4.98\cdot10^{-1} & 3.83\cdot10^{-2} &       &       & AC2   & 0.000    & 9.96\cdot10^{-1} & 4.98\cdot10^{-1} & 1.95\cdot10^{-2} \\*
          & AC3   & 0.000    & 1.000    & 5.00\cdot10^{-1} & 3.85\cdot10^{-2} &       &       & AC3   & 0.000    & 1.000    & 5.00\cdot10^{-1} & 1.96\cdot10^{-2} \\*
          & CSA   & 6.83\cdot10^{-2} & 6.83\cdot10^{-2} & 6.83\cdot10^{-2} & 6.83\cdot10^{-2} &       &       & CSA   & 5.62\cdot10^{-2} & 9.24\cdot10^{-2} & 7.43\cdot10^{-2} & 5.69\cdot10^{-2} \\*
          & CGA   & 5.62\cdot10^{-2} & 1.08\cdot10^{-1} & 8.23\cdot10^{-2} & 5.82\cdot10^{-2} &       &       & CGA   & 3.61\cdot10^{-2} & 1.49\cdot10^{-1} & 9.24\cdot10^{-2} & 3.83\cdot10^{-2} \\*
					\cmidrule(r){1-6} \cmidrule(r){8-13}
    $[100, 1]$ & BAY   & 0.000    & 5.80\cdot10^{-1} & 2.90\cdot10^{-1} & 5.74\cdot10^{-3} &       &       &       &       &       &       &  \\*
          & ABT   & 3.21\cdot10^{-2} & 1.29\cdot10^{-1} & 8.03\cdot10^{-2} & 3.31\cdot10^{-2} &       &       &       &       &       &       &  \\*
          & ASB   & 6.02\cdot10^{-2} & 8.43\cdot10^{-2} & 7.23\cdot10^{-2} & 6.05\cdot10^{-2} &       &       &       &       &       &       &  \\*
          & ADC   & 4.02\cdot10^{-3} & 3.29\cdot10^{-1} & 1.67\cdot10^{-1} & 7.24\cdot10^{-3} &       &       &       &       &       &       &  \\*
          & CB0   & 4.02\cdot10^{-3} & 3.29\cdot10^{-1} & 1.67\cdot10^{-1} & 7.24\cdot10^{-3} &       &       &       &       &       &       &  \\*
          & CB1   & 0.000    & 8.76\cdot10^{-1} & 4.38\cdot10^{-1} & 8.67\cdot10^{-3} &       &       &       &       &       &       &  \\*
          & CB2   & 0.000    & 1.000    & 5.00\cdot10^{-1} & 9.90\cdot10^{-3} &       &       &       &       &       &       &  \\*
          & AC1   & 4.02\cdot10^{-3} & 3.29\cdot10^{-1} & 1.67\cdot10^{-1} & 7.24\cdot10^{-3} &       &       &       &       &       &       &  \\*
          & AC2   & 0.000    & 9.92\cdot10^{-1} & 4.96\cdot10^{-1} & 9.82\cdot10^{-3} &       &       &       &       &       &       &  \\*
          & AC3   & 0.000    & 9.60\cdot10^{-1} & 4.80\cdot10^{-1} & 9.50\cdot10^{-3} &       &       &       &       &       &       &  \\*
          & CSA   & 4.02\cdot10^{-2} & 1.61\cdot10^{-1} & 1.00\cdot10^{-1} & 4.14\cdot10^{-2} &       &       &       &       &       &       &  \\*
          & CGA   & 2.81\cdot10^{-2} & 1.61\cdot10^{-1} & 9.44\cdot10^{-2} & 2.94\cdot10^{-2} &       &       &       &       &       &       &  \\*

\end{longtabu}
}
\newpage

%%%%%%%%%%%%%
%%% BAYES %%%
%%%%%%%%%%%%%

\begin{landscape}
% GRAPHICS
\begin{figure}[p]
\centering
\includegraphics[width=.6\paperheight]{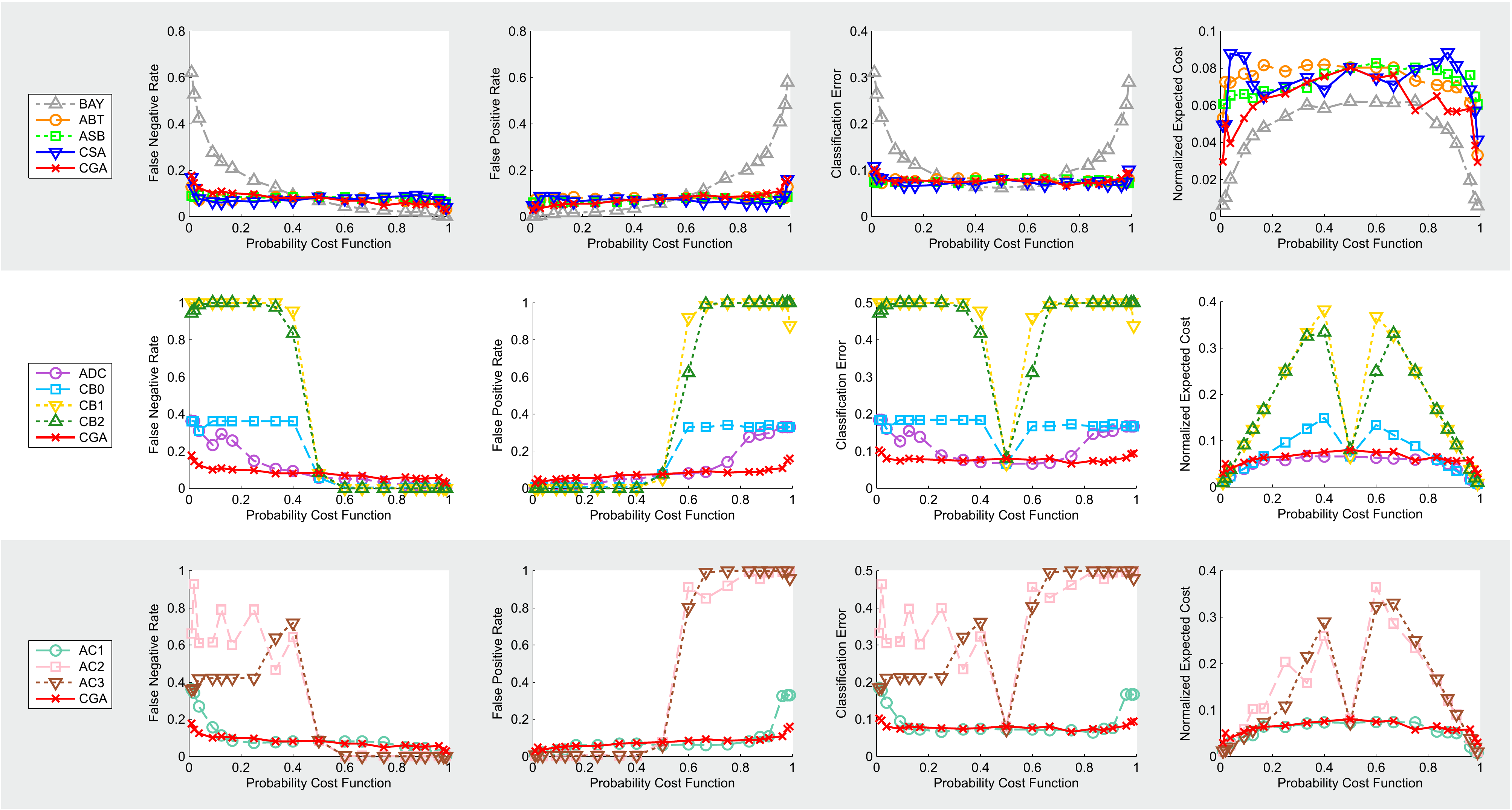}
\caption[Results obtained for the Bayes Dataset.]{Results obtained for the Bayes Dataset. First column of the illustration corresponds to False Negative Rate, second column to False Positive Rate, third column to Classification Error and the fourth one corresponds to Normalized Expected Cost. For a clearer visualization, algorithms have been divided into three groups so each row of the illustration corresponds to a different group. Cost Generalized AdaBoost is plotted in all the graphs to have a common reference across the representations.}
\label{fig:bayes_perform} % caption for the whole figure
\end{figure}

\end{landscape}
\newpage

%%%%%%%%%%%%%%%%%%
%%% TWO CLOUDS %%%
%%%%%%%%%%%%%%%%%%

% TABLE

{
\tiny
\begin{longtabu} to \textwidth {llX[$l]@{}@{}X[$l]@{}@{}X[$l]@{}@{}X[$l]@{}@{}l@{}llX[$l]@{}@{}X[$l]@{}@{}X[$l]@{}@{}X[$l]@{}}

\toprule
{\textbf{Cost}}  & {\textbf{Alg}}   & {\textbf{FNR}}    & {\textbf{FPR}}    & {\textbf{CE}}    & {\textbf{NEC}}   && {\textbf{Cost}}  & {\textbf{Alg}}   & {\textbf{FNR}}    & {\textbf{FPR}}    & {\textbf{CE}} & {\textbf{NEC}}\\* 
\cmidrule(r){1-6} \cmidrule(r){8-13}
\endfirsthead

%\multicolumn{13}{l}% {\tiny{Continued from previous page}} \\*
\toprule
{\textbf{Cost}}  & {\textbf{Alg}}   & {\textbf{FNR}}    & {\textbf{FPR}}    & {\textbf{CE}}    & {\textbf{NEC}}   && {\textbf{Cost}}  & {\textbf{Alg}}   & {\textbf{FNR}}    & {\textbf{FPR}}    & {\textbf{CE}} & {\textbf{NEC}}\\* 
\cmidrule(r){1-6} \cmidrule(r){8-13}
\endhead

\multicolumn{13}{r}{{\tiny{\tablename\ \thetable{} - Continued on next page $\rightarrow$}}} \\* 
\endfoot

\bottomrule
   \caption{Results obtained for the Two Clouds Dataset.}
	 \label{tab:twoclouds_perform}%
\endlastfoot

    $[1, 100]$ & ABT   & 4.14\cdot10^{-1} & 1.95\cdot10^{-1} & 3.04\cdot10^{-1} & 1.97\cdot10^{-1} &       & $[1, 50]$ & ABT   & 3.51\cdot10^{-1} & 2.39\cdot10^{-1} & 2.95\cdot10^{-1} & 2.41\cdot10^{-1} \\*
          & ASB   & 5.30\cdot10^{-1} & 8.63\cdot10^{-2} & 3.08\cdot10^{-1} & 9.07\cdot10^{-2} &       &       & ASB   & 5.66\cdot10^{-1} & 7.43\cdot10^{-2} & 3.20\cdot10^{-1} & 8.39\cdot10^{-2} \\*
          & ADC   & 3.33\cdot10^{-1} & 0.000    & 1.67\cdot10^{-1} & 3.30\cdot10^{-3} &       &       & ADC   & 3.33\cdot10^{-1} & 0.000    & 1.67\cdot10^{-1} & 6.54\cdot10^{-3} \\*
          & CB0   & 9.66\cdot10^{-1} & 0.000    & 4.83\cdot10^{-1} & 9.56\cdot10^{-3} &       &       & CB0   & 9.66\cdot10^{-1} & 0.000    & 4.83\cdot10^{-1} & 1.89\cdot10^{-2} \\*
          & CB1   & 9.66\cdot10^{-1} & 0.000    & 4.83\cdot10^{-1} & 9.56\cdot10^{-3} &       &       & CB1   & 1.000    & 0.000    & 5.00\cdot10^{-1} & 1.96\cdot10^{-2} \\*
          & CB2   & 9.66\cdot10^{-1} & 0.000    & 4.83\cdot10^{-1} & 9.56\cdot10^{-3} &       &       & CB2   & 1.000    & 0.000    & 5.00\cdot10^{-1} & 1.96\cdot10^{-2} \\*
          & AC1   & 9.12\cdot10^{-1} & 6.02\cdot10^{-3} & 4.59\cdot10^{-1} & 1.50\cdot10^{-2} &       &       & AC1   & 9.66\cdot10^{-1} & 0.000    & 4.83\cdot10^{-1} & 1.89\cdot10^{-2} \\*
          & AC2   & 9.12\cdot10^{-1} & 6.02\cdot10^{-3} & 4.59\cdot10^{-1} & 1.50\cdot10^{-2} &       &       & AC2   & 9.92\cdot10^{-1} & 0.000    & 4.96\cdot10^{-1} & 1.95\cdot10^{-2} \\*
          & AC3   & 9.66\cdot10^{-1} & 0.000    & 4.83\cdot10^{-1} & 9.56\cdot10^{-3} &       &       & AC3   & 9.66\cdot10^{-1} & 0.000    & 4.83\cdot10^{-1} & 1.89\cdot10^{-2} \\*
          & CSA   & 5.80\cdot10^{-1} & 8.03\cdot10^{-2} & 3.30\cdot10^{-1} & 8.53\cdot10^{-2} &       &       & CSA   & 4.70\cdot10^{-1} & 1.55\cdot10^{-1} & 3.12\cdot10^{-1} & 1.61\cdot10^{-1} \\*
          & CGA   & 8.45\cdot10^{-1} & 0.000    & 4.23\cdot10^{-1} & 8.37\cdot10^{-3} &       &       & CGA   & 7.01\cdot10^{-1} & 1.61\cdot10^{-2} & 3.58\cdot10^{-1} & 2.95\cdot10^{-2} \\*
  \cmidrule(r){1-6} \cmidrule(r){8-13}
		$[1, 25]$ & ABT   & 3.31\cdot10^{-1} & 2.51\cdot10^{-1} & 2.91\cdot10^{-1} & 2.54\cdot10^{-1} &       & $[1, 10]$ & ABT   & 3.05\cdot10^{-1} & 2.65\cdot10^{-1} & 2.85\cdot10^{-1} & 2.69\cdot10^{-1} \\*
          & ASB   & 5.64\cdot10^{-1} & 8.03\cdot10^{-2} & 3.22\cdot10^{-1} & 9.89\cdot10^{-2} &       &       & ASB   & 5.10\cdot10^{-1} & 1.10\cdot10^{-1} & 3.10\cdot10^{-1} & 1.47\cdot10^{-1} \\*
          & ADC   & 3.33\cdot10^{-1} & 0.000    & 1.67\cdot10^{-1} & 1.28\cdot10^{-2} &       &       & ADC   & 3.33\cdot10^{-1} & 0.000    & 1.67\cdot10^{-1} & 3.03\cdot10^{-2} \\*
          & CB0   & 9.66\cdot10^{-1} & 0.000    & 4.83\cdot10^{-1} & 3.71\cdot10^{-2} &       &       & CB0   & 9.12\cdot10^{-1} & 6.02\cdot10^{-3} & 4.59\cdot10^{-1} & 8.84\cdot10^{-2} \\*
          & CB1   & 1.000    & 0.000    & 5.00\cdot10^{-1} & 3.85\cdot10^{-2} &       &       & CB1   & 1.000    & 0.000    & 5.00\cdot10^{-1} & 9.09\cdot10^{-2} \\*
          & CB2   & 1.000    & 0.000    & 5.00\cdot10^{-1} & 3.85\cdot10^{-2} &       &       & CB2   & 1.000    & 0.000    & 5.00\cdot10^{-1} & 9.09\cdot10^{-2} \\*
          & AC1   & 9.04\cdot10^{-1} & 1.41\cdot10^{-2} & 4.59\cdot10^{-1} & 4.83\cdot10^{-2} &       &       & AC1   & 6.85\cdot10^{-1} & 1.81\cdot10^{-2} & 3.51\cdot10^{-1} & 7.87\cdot10^{-2} \\*
          & AC2   & 9.92\cdot10^{-1} & 0.000    & 4.96\cdot10^{-1} & 3.82\cdot10^{-2} &       &       & AC2   & 9.70\cdot10^{-1} & 0.000    & 4.85\cdot10^{-1} & 8.82\cdot10^{-2} \\*
          & AC3   & 9.90\cdot10^{-1} & 0.000    & 4.95\cdot10^{-1} & 3.81\cdot10^{-2} &       &       & AC3   & 9.80\cdot10^{-1} & 0.000    & 4.90\cdot10^{-1} & 8.91\cdot10^{-2} \\*
          & CSA   & 4.00\cdot10^{-1} & 2.29\cdot10^{-1} & 3.14\cdot10^{-1} & 2.35\cdot10^{-1} &       &       & CSA   & 3.49\cdot10^{-1} & 2.51\cdot10^{-1} & 3.00\cdot10^{-1} & 2.60\cdot10^{-1} \\*
          & CGA   & 6.04\cdot10^{-1} & 4.42\cdot10^{-2} & 3.24\cdot10^{-1} & 6.57\cdot10^{-2} &       &       & CGA   & 5.26\cdot10^{-1} & 1.06\cdot10^{-1} & 3.16\cdot10^{-1} & 1.45\cdot10^{-1} \\*
      \cmidrule(r){1-6} \cmidrule(r){8-13}
		$[1, 7]$ & ABT   & 3.05\cdot10^{-1} & 2.65\cdot10^{-1} & 2.85\cdot10^{-1} & 2.70\cdot10^{-1} &       & $[1, 5]$ & ABT   & 3.05\cdot10^{-1} & 2.65\cdot10^{-1} & 2.85\cdot10^{-1} & 2.72\cdot10^{-1} \\*
          & ASB   & 4.78\cdot10^{-1} & 1.33\cdot10^{-1} & 3.05\cdot10^{-1} & 1.76\cdot10^{-1} &       &       & ASB   & 4.34\cdot10^{-1} & 1.53\cdot10^{-1} & 2.93\cdot10^{-1} & 1.99\cdot10^{-1} \\*
          & ADC   & 0.000    & 3.33\cdot10^{-1} & 1.67\cdot10^{-1} & 2.92\cdot10^{-1} &       &       & ADC   & 0.000    & 3.33\cdot10^{-1} & 1.67\cdot10^{-1} & 2.78\cdot10^{-1} \\*
          & CB0   & 9.84\cdot10^{-1} & 0.000    & 4.92\cdot10^{-1} & 1.23\cdot10^{-1} &       &       & CB0   & 9.68\cdot10^{-1} & 0.000    & 4.84\cdot10^{-1} & 1.61\cdot10^{-1} \\*
          & CB1   & 1.000    & 0.000    & 5.00\cdot10^{-1} & 1.25\cdot10^{-1} &       &       & CB1   & 1.000    & 0.000    & 5.00\cdot10^{-1} & 1.67\cdot10^{-1} \\*
          & CB2   & 1.000    & 0.000    & 5.00\cdot10^{-1} & 1.25\cdot10^{-1} &       &       & CB2   & 1.000    & 0.000    & 5.00\cdot10^{-1} & 1.67\cdot10^{-1} \\*
          & AC1   & 5.88\cdot10^{-1} & 7.43\cdot10^{-2} & 3.31\cdot10^{-1} & 1.39\cdot10^{-1} &       &       & AC1   & 5.62\cdot10^{-1} & 8.84\cdot10^{-2} & 3.25\cdot10^{-1} & 1.67\cdot10^{-1} \\*
          & AC2   & 9.36\cdot10^{-1} & 4.02\cdot10^{-3} & 4.70\cdot10^{-1} & 1.20\cdot10^{-1} &       &       & AC2   & 9.78\cdot10^{-1} & 0.000    & 4.89\cdot10^{-1} & 1.63\cdot10^{-1} \\*
          & AC3   & 9.82\cdot10^{-1} & 0.000    & 4.91\cdot10^{-1} & 1.23\cdot10^{-1} &       &       & AC3   & 9.74\cdot10^{-1} & 0.000    & 4.87\cdot10^{-1} & 1.62\cdot10^{-1} \\*
          & CSA   & 3.35\cdot10^{-1} & 2.73\cdot10^{-1} & 3.04\cdot10^{-1} & 2.81\cdot10^{-1} &       &       & CSA   & 3.07\cdot10^{-1} & 2.71\cdot10^{-1} & 2.89\cdot10^{-1} & 2.77\cdot10^{-1} \\*
          & CGA   & 4.72\cdot10^{-1} & 1.31\cdot10^{-1} & 3.01\cdot10^{-1} & 1.73\cdot10^{-1} &       &       & CGA   & 4.42\cdot10^{-1} & 1.53\cdot10^{-1} & 2.97\cdot10^{-1} & 2.01\cdot10^{-1} \\*
      \cmidrule(r){1-6} \cmidrule(r){8-13}
		$[1, 3]$ & ABT   & 3.05\cdot10^{-1} & 2.65\cdot10^{-1} & 2.85\cdot10^{-1} & 2.75\cdot10^{-1} &       & $[1, 2]$ & ABT   & 2.97\cdot10^{-1} & 2.69\cdot10^{-1} & 2.83\cdot10^{-1} & 2.78\cdot10^{-1} \\*
          & ASB   & 4.02\cdot10^{-1} & 1.97\cdot10^{-1} & 2.99\cdot10^{-1} & 2.48\cdot10^{-1} &       &       & ASB   & 3.61\cdot10^{-1} & 2.23\cdot10^{-1} & 2.92\cdot10^{-1} & 2.69\cdot10^{-1} \\*
          & ADC   & 8.43\cdot10^{-2} & 9.68\cdot10^{-1} & 5.26\cdot10^{-1} & 7.47\cdot10^{-1} &       &       & ADC   & 9.64\cdot10^{-2} & 9.56\cdot10^{-1} & 5.26\cdot10^{-1} & 6.69\cdot10^{-1} \\*
          & CB0   & 9.12\cdot10^{-1} & 6.02\cdot10^{-3} & 4.59\cdot10^{-1} & 2.32\cdot10^{-1} &       &       & CB0   & 9.68\cdot10^{-1} & 0.000    & 4.84\cdot10^{-1} & 3.23\cdot10^{-1} \\*
          & CB1   & 1.000    & 0.000    & 5.00\cdot10^{-1} & 2.50\cdot10^{-1} &       &       & CB1   & 1.000    & 0.000    & 5.00\cdot10^{-1} & 3.33\cdot10^{-1} \\*
          & CB2   & 1.000    & 0.000    & 5.00\cdot10^{-1} & 2.50\cdot10^{-1} &       &       & CB2   & 1.000    & 0.000    & 5.00\cdot10^{-1} & 3.33\cdot10^{-1} \\*
          & AC1   & 4.58\cdot10^{-1} & 1.75\cdot10^{-1} & 3.16\cdot10^{-1} & 2.45\cdot10^{-1} &       &       & AC1   & 4.02\cdot10^{-1} & 2.23\cdot10^{-1} & 3.12\cdot10^{-1} & 2.82\cdot10^{-1} \\*
          & AC2   & 9.58\cdot10^{-1} & 4.02\cdot10^{-3} & 4.81\cdot10^{-1} & 2.42\cdot10^{-1} &       &       & AC2   & 9.62\cdot10^{-1} & 0.000    & 4.81\cdot10^{-1} & 3.21\cdot10^{-1} \\*
          & AC3   & 9.68\cdot10^{-1} & 0.000    & 4.84\cdot10^{-1} & 2.42\cdot10^{-1} &       &       & AC3   & 1.000    & 0.000    & 5.00\cdot10^{-1} & 3.33\cdot10^{-1} \\*
          & CSA   & 2.99\cdot10^{-1} & 2.71\cdot10^{-1} & 2.85\cdot10^{-1} & 2.78\cdot10^{-1} &       &       & CSA   & 2.97\cdot10^{-1} & 2.75\cdot10^{-1} & 2.86\cdot10^{-1} & 2.82\cdot10^{-1} \\*
          & CGA   & 3.96\cdot10^{-1} & 1.99\cdot10^{-1} & 2.97\cdot10^{-1} & 2.48\cdot10^{-1} &       &       & CGA   & 3.59\cdot10^{-1} & 2.21\cdot10^{-1} & 2.90\cdot10^{-1} & 2.67\cdot10^{-1} \\*
      \cmidrule(r){1-6} \cmidrule(r){8-13}
		$[2, 3]$ & ABT   & 2.97\cdot10^{-1} & 2.69\cdot10^{-1} & 2.83\cdot10^{-1} & 2.80\cdot10^{-1} &       & $[1, 1]$ & ABT   & 2.77\cdot10^{-1} & 2.81\cdot10^{-1} & 2.79\cdot10^{-1} & 2.79\cdot10^{-1} \\*
          & ASB   & 3.29\cdot10^{-1} & 2.39\cdot10^{-1} & 2.84\cdot10^{-1} & 2.75\cdot10^{-1} &       &       & ASB   & 2.95\cdot10^{-1} & 2.69\cdot10^{-1} & 2.82\cdot10^{-1} & 2.82\cdot10^{-1} \\*
          & ADC   & 9.64\cdot10^{-2} & 9.56\cdot10^{-1} & 5.26\cdot10^{-1} & 6.12\cdot10^{-1} &       &       & ADC   & 4.80\cdot10^{-1} & 6.27\cdot10^{-1} & 5.53\cdot10^{-1} & 5.53\cdot10^{-1} \\*
          & CB0   & 9.12\cdot10^{-1} & 4.02\cdot10^{-3} & 4.58\cdot10^{-1} & 3.67\cdot10^{-1} &       &       & CB0   & 5.20\cdot10^{-1} & 3.73\cdot10^{-1} & 4.47\cdot10^{-1} & 4.47\cdot10^{-1} \\*
          & CB1   & 1.000    & 0.000    & 5.00\cdot10^{-1} & 4.00\cdot10^{-1} &       &       & CB1   & 4.10\cdot10^{-1} & 5.60\cdot10^{-1} & 4.85\cdot10^{-1} & 4.85\cdot10^{-1} \\*
          & CB2   & 1.000    & 0.000    & 5.00\cdot10^{-1} & 4.00\cdot10^{-1} &       &       & CB2   & 2.95\cdot10^{-1} & 2.69\cdot10^{-1} & 2.82\cdot10^{-1} & 2.82\cdot10^{-1} \\*
          & AC1   & 3.47\cdot10^{-1} & 2.57\cdot10^{-1} & 3.02\cdot10^{-1} & 2.93\cdot10^{-1} &       &       & AC1   & 2.93\cdot10^{-1} & 2.95\cdot10^{-1} & 2.94\cdot10^{-1} & 2.94\cdot10^{-1} \\*
          & AC2   & 1.000    & 0.000    & 5.00\cdot10^{-1} & 4.00\cdot10^{-1} &       &       & AC2   & 2.95\cdot10^{-1} & 2.69\cdot10^{-1} & 2.82\cdot10^{-1} & 2.82\cdot10^{-1} \\*
          & AC3   & 9.88\cdot10^{-1} & 0.000    & 4.94\cdot10^{-1} & 3.95\cdot10^{-1} &       &       & AC3   & 2.93\cdot10^{-1} & 2.95\cdot10^{-1} & 2.94\cdot10^{-1} & 2.94\cdot10^{-1} \\*
          & CSA   & 2.85\cdot10^{-1} & 2.81\cdot10^{-1} & 2.83\cdot10^{-1} & 2.83\cdot10^{-1} &       &       & CSA   & 2.95\cdot10^{-1} & 2.69\cdot10^{-1} & 2.82\cdot10^{-1} & 2.82\cdot10^{-1} \\*
          & CGA   & 3.39\cdot10^{-1} & 2.43\cdot10^{-1} & 2.91\cdot10^{-1} & 2.82\cdot10^{-1} &       &       & CGA   & 2.95\cdot10^{-1} & 2.69\cdot10^{-1} & 2.82\cdot10^{-1} & 2.82\cdot10^{-1} \\*
      \cmidrule(r){1-6} \cmidrule(r){8-13}
			
		$[3, 2]$ & ABT   & 2.77\cdot10^{-1} & 2.81\cdot10^{-1} & 2.79\cdot10^{-1} & 2.79\cdot10^{-1} &       & $[2, 1]$ & ABT   & 2.77\cdot10^{-1} & 2.81\cdot10^{-1} & 2.79\cdot10^{-1} & 2.78\cdot10^{-1} \\*
          & ASB   & 2.61\cdot10^{-1} & 3.05\cdot10^{-1} & 2.83\cdot10^{-1} & 2.79\cdot10^{-1} &       &       & ASB   & 2.43\cdot10^{-1} & 3.33\cdot10^{-1} & 2.88\cdot10^{-1} & 2.73\cdot10^{-1} \\*
          & ADC   & 9.78\cdot10^{-1} & 6.83\cdot10^{-2} & 5.23\cdot10^{-1} & 6.14\cdot10^{-1} &       &       & ADC   & 9.78\cdot10^{-1} & 6.83\cdot10^{-2} & 5.23\cdot10^{-1} & 6.75\cdot10^{-1} \\*
          & CB0   & 2.01\cdot10^{-3} & 9.70\cdot10^{-1} & 4.86\cdot10^{-1} & 3.89\cdot10^{-1} &       &       & CB0   & 6.02\cdot10^{-3} & 9.48\cdot10^{-1} & 4.77\cdot10^{-1} & 3.20\cdot10^{-1} \\*
          & CB1   & 2.01\cdot10^{-3} & 9.82\cdot10^{-1} & 4.92\cdot10^{-1} & 3.94\cdot10^{-1} &       &       & CB1   & 0.000    & 1.000    & 5.00\cdot10^{-1} & 3.33\cdot10^{-1} \\*
          & CB2   & 0.000    & 1.000    & 5.00\cdot10^{-1} & 4.00\cdot10^{-1} &       &       & CB2   & 0.000    & 9.92\cdot10^{-1} & 4.96\cdot10^{-1} & 3.31\cdot10^{-1} \\*
          & AC1   & 2.53\cdot10^{-1} & 3.49\cdot10^{-1} & 3.01\cdot10^{-1} & 2.92\cdot10^{-1} &       &       & AC1   & 2.33\cdot10^{-1} & 3.78\cdot10^{-1} & 3.05\cdot10^{-1} & 2.81\cdot10^{-1} \\*
          & AC2   & 0.000    & 1.000    & 5.00\cdot10^{-1} & 4.00\cdot10^{-1} &       &       & AC2   & 0.000    & 9.92\cdot10^{-1} & 4.96\cdot10^{-1} & 3.31\cdot10^{-1} \\*
          & AC3   & 0.000    & 9.92\cdot10^{-1} & 4.96\cdot10^{-1} & 3.97\cdot10^{-1} &       &       & AC3   & 0.000    & 9.92\cdot10^{-1} & 4.96\cdot10^{-1} & 3.31\cdot10^{-1} \\*
          & CSA   & 2.81\cdot10^{-1} & 2.67\cdot10^{-1} & 2.74\cdot10^{-1} & 2.76\cdot10^{-1} &       &       & CSA   & 2.99\cdot10^{-1} & 2.81\cdot10^{-1} & 2.90\cdot10^{-1} & 2.93\cdot10^{-1} \\*
          & CGA   & 2.71\cdot10^{-1} & 2.95\cdot10^{-1} & 2.83\cdot10^{-1} & 2.81\cdot10^{-1} &       &       & CGA   & 2.49\cdot10^{-1} & 3.35\cdot10^{-1} & 2.92\cdot10^{-1} & 2.78\cdot10^{-1} \\*
      \cmidrule(r){1-6} \cmidrule(r){8-13}
		$[3, 1]$ & ABT   & 2.77\cdot10^{-1} & 2.81\cdot10^{-1} & 2.79\cdot10^{-1} & 2.78\cdot10^{-1} &       & $[5, 1]$ & ABT   & 2.75\cdot10^{-1} & 2.87\cdot10^{-1} & 2.81\cdot10^{-1} & 2.77\cdot10^{-1} \\*
          & ASB   & 2.23\cdot10^{-1} & 3.69\cdot10^{-1} & 2.96\cdot10^{-1} & 2.60\cdot10^{-1} &       &       & ASB   & 1.79\cdot10^{-1} & 4.08\cdot10^{-1} & 2.93\cdot10^{-1} & 2.17\cdot10^{-1} \\*
          & ADC   & 9.78\cdot10^{-1} & 6.83\cdot10^{-2} & 5.23\cdot10^{-1} & 7.51\cdot10^{-1} &       &       & ADC   & 9.98\cdot10^{-1} & 5.22\cdot10^{-2} & 5.25\cdot10^{-1} & 8.40\cdot10^{-1} \\*
          & CB0   & 2.01\cdot10^{-3} & 9.78\cdot10^{-1} & 4.90\cdot10^{-1} & 2.46\cdot10^{-1} &       &       & CB0   & 2.01\cdot10^{-3} & 9.78\cdot10^{-1} & 4.90\cdot10^{-1} & 1.65\cdot10^{-1} \\*
          & CB1   & 0.000    & 1.000    & 5.00\cdot10^{-1} & 2.50\cdot10^{-1} &       &       & CB1   & 0.000    & 1.000    & 5.00\cdot10^{-1} & 1.67\cdot10^{-1} \\*
          & CB2   & 0.000    & 9.98\cdot10^{-1} & 4.99\cdot10^{-1} & 2.49\cdot10^{-1} &       &       & CB2   & 0.000    & 1.000    & 5.00\cdot10^{-1} & 1.67\cdot10^{-1} \\*
          & AC1   & 1.79\cdot10^{-1} & 4.62\cdot10^{-1} & 3.20\cdot10^{-1} & 2.49\cdot10^{-1} &       &       & AC1   & 1.33\cdot10^{-1} & 5.18\cdot10^{-1} & 3.25\cdot10^{-1} & 1.97\cdot10^{-1} \\*
          & AC2   & 0.000    & 1.000    & 5.00\cdot10^{-1} & 2.50\cdot10^{-1} &       &       & AC2   & 0.000    & 9.92\cdot10^{-1} & 4.96\cdot10^{-1} & 1.65\cdot10^{-1} \\*
          & AC3   & 0.000    & 9.92\cdot10^{-1} & 4.96\cdot10^{-1} & 2.48\cdot10^{-1} &       &       & AC3   & 0.000    & 9.92\cdot10^{-1} & 4.96\cdot10^{-1} & 1.65\cdot10^{-1} \\*
          & CSA   & 2.95\cdot10^{-1} & 2.79\cdot10^{-1} & 2.87\cdot10^{-1} & 2.91\cdot10^{-1} &       &       & CSA   & 2.83\cdot10^{-1} & 2.97\cdot10^{-1} & 2.90\cdot10^{-1} & 2.85\cdot10^{-1} \\*
          & CGA   & 2.13\cdot10^{-1} & 3.67\cdot10^{-1} & 2.90\cdot10^{-1} & 2.52\cdot10^{-1} &       &       & CGA   & 1.81\cdot10^{-1} & 4.20\cdot10^{-1} & 3.00\cdot10^{-1} & 2.21\cdot10^{-1} \\*
      \cmidrule(r){1-6} \cmidrule(r){8-13}
		$[7, 1]$ & ABT   & 2.75\cdot10^{-1} & 2.87\cdot10^{-1} & 2.81\cdot10^{-1} & 2.77\cdot10^{-1} &       & $[10, 1]$ & ABT   & 2.75\cdot10^{-1} & 2.87\cdot10^{-1} & 2.81\cdot10^{-1} & 2.76\cdot10^{-1} \\*
          & ASB   & 1.57\cdot10^{-1} & 4.42\cdot10^{-1} & 2.99\cdot10^{-1} & 1.92\cdot10^{-1} &       &       & ASB   & 1.35\cdot10^{-1} & 4.66\cdot10^{-1} & 3.00\cdot10^{-1} & 1.65\cdot10^{-1} \\*
          & ADC   & 9.90\cdot10^{-1} & 5.22\cdot10^{-2} & 5.21\cdot10^{-1} & 8.73\cdot10^{-1} &       &       & ADC   & 9.90\cdot10^{-1} & 5.22\cdot10^{-2} & 5.21\cdot10^{-1} & 9.05\cdot10^{-1} \\*
          & CB0   & 2.01\cdot10^{-3} & 9.78\cdot10^{-1} & 4.90\cdot10^{-1} & 1.24\cdot10^{-1} &       &       & CB0   & 2.01\cdot10^{-3} & 9.78\cdot10^{-1} & 4.90\cdot10^{-1} & 9.07\cdot10^{-2} \\*
          & CB1   & 0.000    & 1.000    & 5.00\cdot10^{-1} & 1.25\cdot10^{-1} &       &       & CB1   & 0.000    & 1.000    & 5.00\cdot10^{-1} & 9.09\cdot10^{-2} \\*
          & CB2   & 2.01\cdot10^{-3} & 9.98\cdot10^{-1} & 5.00\cdot10^{-1} & 1.27\cdot10^{-1} &       &       & CB2   & 0.000    & 9.92\cdot10^{-1} & 4.96\cdot10^{-1} & 9.02\cdot10^{-2} \\*
          & AC1   & 6.63\cdot10^{-2} & 5.94\cdot10^{-1} & 3.30\cdot10^{-1} & 1.32\cdot10^{-1} &       &       & AC1   & 2.81\cdot10^{-2} & 6.49\cdot10^{-1} & 3.38\cdot10^{-1} & 8.45\cdot10^{-2} \\*
          & AC2   & 0.000    & 9.92\cdot10^{-1} & 4.96\cdot10^{-1} & 1.24\cdot10^{-1} &       &       & AC2   & 0.000    & 9.92\cdot10^{-1} & 4.96\cdot10^{-1} & 9.02\cdot10^{-2} \\*
          & AC3   & 0.000    & 9.92\cdot10^{-1} & 4.96\cdot10^{-1} & 1.24\cdot10^{-1} &       &       & AC3   & 0.000    & 9.92\cdot10^{-1} & 4.96\cdot10^{-1} & 9.02\cdot10^{-2} \\*
          & CSA   & 2.77\cdot10^{-1} & 3.03\cdot10^{-1} & 2.90\cdot10^{-1} & 2.80\cdot10^{-1} &       &       & CSA   & 2.73\cdot10^{-1} & 3.03\cdot10^{-1} & 2.88\cdot10^{-1} & 2.76\cdot10^{-1} \\*
          & CGA   & 1.51\cdot10^{-1} & 4.42\cdot10^{-1} & 2.96\cdot10^{-1} & 1.87\cdot10^{-1} &       &       & CGA   & 1.29\cdot10^{-1} & 4.82\cdot10^{-1} & 3.05\cdot10^{-1} & 1.61\cdot10^{-1} \\*
    \cmidrule(r){1-6} \cmidrule(r){8-13} 
	$[25, 1]$ & ABT   & 2.55\cdot10^{-1} & 3.13\cdot10^{-1} & 2.84\cdot10^{-1} & 2.57\cdot10^{-1} &       & $[50, 1]$ & ABT   & 2.39\cdot10^{-1} & 3.39\cdot10^{-1} & 2.89\cdot10^{-1} & 2.41\cdot10^{-1} \\*
          & ASB   & 1.06\cdot10^{-1} & 5.12\cdot10^{-1} & 3.09\cdot10^{-1} & 1.22\cdot10^{-1} &       &       & ASB   & 9.24\cdot10^{-2} & 5.38\cdot10^{-1} & 3.15\cdot10^{-1} & 1.01\cdot10^{-1} \\*
          & ADC   & 3.27\cdot10^{-1} & 6.75\cdot10^{-1} & 5.01\cdot10^{-1} & 3.41\cdot10^{-1} &       &       & ADC   & 1.00\cdot10^{-2} & 9.48\cdot10^{-1} & 4.79\cdot10^{-1} & 2.84\cdot10^{-2} \\*
          & CB0   & 1.00\cdot10^{-2} & 9.48\cdot10^{-1} & 4.79\cdot10^{-1} & 4.61\cdot10^{-2} &       &       & CB0   & 1.00\cdot10^{-2} & 9.48\cdot10^{-1} & 4.79\cdot10^{-1} & 2.84\cdot10^{-2} \\*
          & CB1   & 0.000    & 1.000    & 5.00\cdot10^{-1} & 3.85\cdot10^{-2} &       &       & CB1   & 0.000    & 1.000    & 5.00\cdot10^{-1} & 1.96\cdot10^{-2} \\*
          & CB2   & 0.000    & 9.94\cdot10^{-1} & 4.97\cdot10^{-1} & 3.82\cdot10^{-2} &       &       & CB2   & 0.000    & 9.94\cdot10^{-1} & 4.97\cdot10^{-1} & 1.95\cdot10^{-2} \\*
          & AC1   & 1.00\cdot10^{-2} & 9.48\cdot10^{-1} & 4.79\cdot10^{-1} & 4.61\cdot10^{-2} &       &       & AC1   & 2.01\cdot10^{-3} & 9.48\cdot10^{-1} & 4.75\cdot10^{-1} & 2.06\cdot10^{-2} \\*
          & AC2   & 0.000    & 9.92\cdot10^{-1} & 4.96\cdot10^{-1} & 3.82\cdot10^{-2} &       &       & AC2   & 0.000    & 9.92\cdot10^{-1} & 4.96\cdot10^{-1} & 1.95\cdot10^{-2} \\*
          & AC3   & 2.01\cdot10^{-3} & 9.70\cdot10^{-1} & 4.86\cdot10^{-1} & 3.92\cdot10^{-2} &       &       & AC3   & 2.01\cdot10^{-3} & 9.70\cdot10^{-1} & 4.86\cdot10^{-1} & 2.10\cdot10^{-2} \\*
          & CSA   & 2.53\cdot10^{-1} & 3.53\cdot10^{-1} & 3.03\cdot10^{-1} & 2.57\cdot10^{-1} &       &       & CSA   & 2.15\cdot10^{-1} & 4.18\cdot10^{-1} & 3.16\cdot10^{-1} & 2.19\cdot10^{-1} \\*
          & CGA   & 8.03\cdot10^{-2} & 5.86\cdot10^{-1} & 3.33\cdot10^{-1} & 9.98\cdot10^{-2} &       &       & CGA   & 3.21\cdot10^{-2} & 6.71\cdot10^{-1} & 3.51\cdot10^{-1} & 4.46\cdot10^{-2} \\*
      \cmidrule(r){1-6} \cmidrule(r){8-13}
		$[100, 1]$ & ABT   & 2.29\cdot10^{-1} & 3.59\cdot10^{-1} & 2.94\cdot10^{-1} & 2.30\cdot10^{-1} &       &       &       &       &       &       &  \\*
          & ASB   & 9.64\cdot10^{-2} & 5.12\cdot10^{-1} & 3.04\cdot10^{-1} & 1.01\cdot10^{-1} &       &       &       &       &       &       &  \\*
          & ADC   & 1.00\cdot10^{-2} & 9.48\cdot10^{-1} & 4.79\cdot10^{-1} & 1.93\cdot10^{-2} &       &       &       &       &       &       &  \\*
          & CB0   & 1.00\cdot10^{-2} & 9.48\cdot10^{-1} & 4.79\cdot10^{-1} & 1.93\cdot10^{-2} &       &       &       &       &       &       &  \\*
          & CB1   & 1.00\cdot10^{-2} & 9.48\cdot10^{-1} & 4.79\cdot10^{-1} & 1.93\cdot10^{-2} &       &       &       &       &       &       &  \\*
          & CB2   & 1.00\cdot10^{-2} & 9.48\cdot10^{-1} & 4.79\cdot10^{-1} & 1.93\cdot10^{-2} &       &       &       &       &       &       &  \\*
          & AC1   & 2.01\cdot10^{-3} & 9.48\cdot10^{-1} & 4.75\cdot10^{-1} & 1.14\cdot10^{-2} &       &       &       &       &       &       &  \\*
          & AC2   & 2.01\cdot10^{-3} & 9.48\cdot10^{-1} & 4.75\cdot10^{-1} & 1.14\cdot10^{-2} &       &       &       &       &       &       &  \\*
          & AC3   & 1.00\cdot10^{-2} & 9.48\cdot10^{-1} & 4.79\cdot10^{-1} & 1.93\cdot10^{-2} &       &       &       &       &       &       &  \\*
          & CSA   & 1.31\cdot10^{-1} & 5.22\cdot10^{-1} & 3.26\cdot10^{-1} & 1.34\cdot10^{-1} &       &       &       &       &       &       &  \\*
          & CGA   & 1.00\cdot10^{-2} & 8.03\cdot10^{-1} & 4.07\cdot10^{-1} & 1.79\cdot10^{-2} &       &       &       &       &       &       &  \\*

\end{longtabu}
}

\newpage

%%%%%%%%%%%%%%%%%%
%%% TWO CLOUDS %%%
%%%%%%%%%%%%%%%%%%

\begin{landscape}
% GRAPHICS
\begin{figure}[p]
\centering
\includegraphics[width=.6\paperheight]{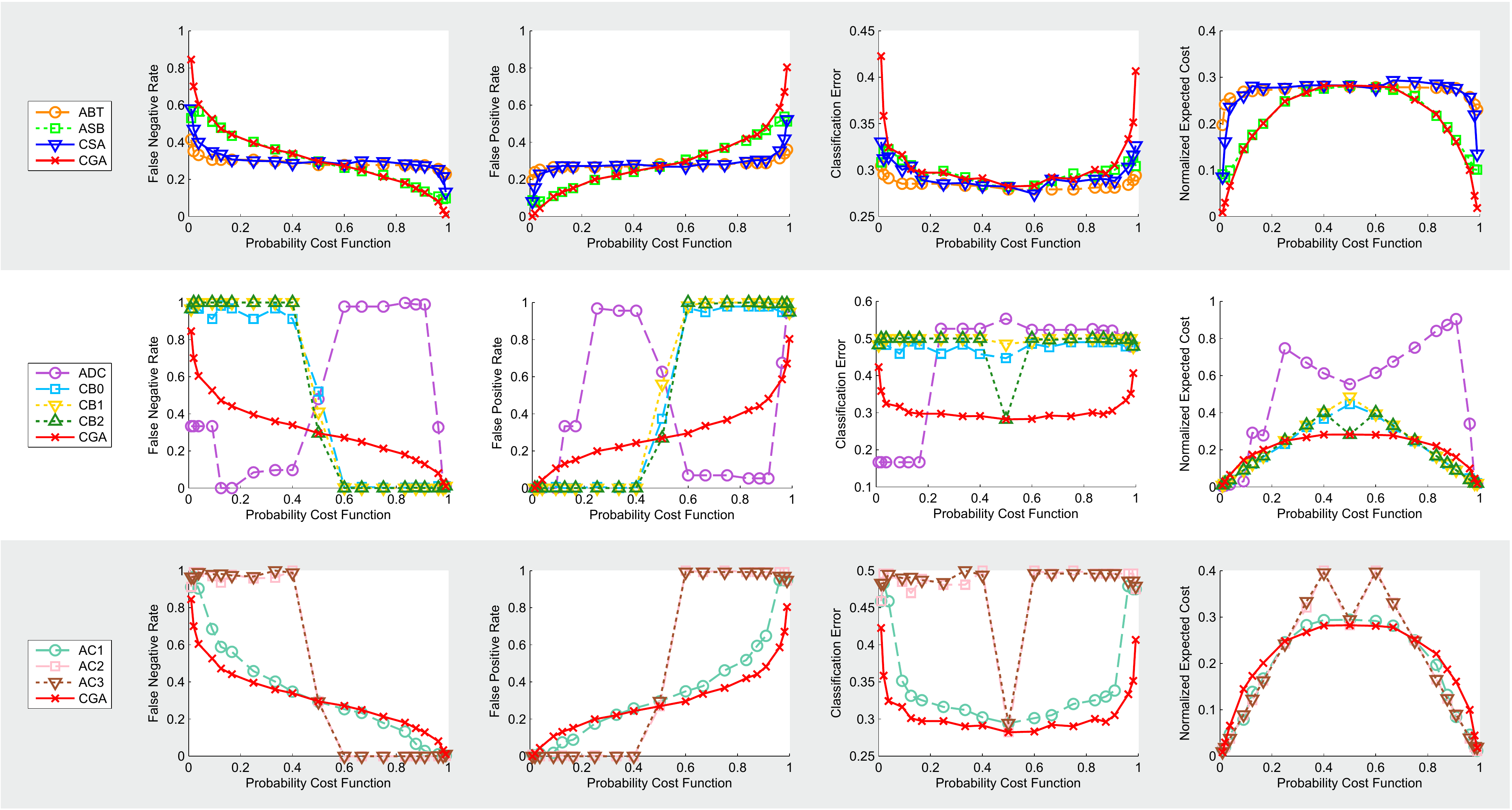}
\caption[Results obtained for the Two Clouds Dataset.] {Results obtained for the Two Clouds Dataset. First column of the illustration corresponds to False Negative Rate, second column to False Positive Rate, third column to Classification Error and the fourth one corresponds to Normalized Expected Cost. For a clearer visualization, algorithms have been divided into three groups so each row of the illustration corresponds to a different group. Cost Generalized AdaBoost is plotted in all the graphs to have a common reference across the representations.}
\label{TwoClouds_perform_fig} % caption for the whole figure
\end{figure}
\end{landscape}

\newpage

%%%%%%%%%%%%%%%%%%
%%% UCI BREAST %%%
%%%%%%%%%%%%%%%%%%

% TABLE

{
\tiny
\begin{longtabu} to \textwidth {llX[$l]@{}@{}X[$l]@{}@{}X[$l]@{}@{}X[$l]@{}@{}l@{}llX[$l]@{}@{}X[$l]@{}@{}X[$l]@{}@{}X[$l]@{}}

\toprule
{\textbf{Cost}}  & {\textbf{Alg}}   & {\textbf{FNR}}    & {\textbf{FPR}}    & {\textbf{CE}}    & {\textbf{NEC}}   && {\textbf{Cost}}  & {\textbf{Alg}}   & {\textbf{FNR}}    & {\textbf{FPR}}    & {\textbf{CE}} & {\textbf{NEC}}\\* 
\cmidrule(r){1-6} \cmidrule(r){8-13}
\endfirsthead

%\multicolumn{13}{l}% {\tiny{Continued from previous page}} \\*
\toprule
{\textbf{Cost}}  & {\textbf{Alg}}   & {\textbf{FNR}}    & {\textbf{FPR}}    & {\textbf{CE}}    & {\textbf{NEC}}   && {\textbf{Cost}}  & {\textbf{Alg}}   & {\textbf{FNR}}    & {\textbf{FPR}}    & {\textbf{CE}} & {\textbf{NEC}}\\* 
\cmidrule(r){1-6} \cmidrule(r){8-13}
\endhead

\multicolumn{13}{r}{{\tiny{\tablename\ \thetable{} - Continued on next page $\rightarrow$}}} \\* 
\endfoot

\bottomrule
  \caption{Results obtained for the UCI Breast Cancer Dataset.}
	\label{tab:breast_perform}%
\endlastfoot

    $[1, 100]$ & ABT   & 2.11\cdot10^{-1} & 2.95\cdot10^{-2} & 1.20\cdot10^{-1} & 3.13\cdot10^{-2} &       & $[1, 50]$ & ABT   & 1.90\cdot10^{-1} & 3.80\cdot10^{-2} & 1.14\cdot10^{-1} & 4.10\cdot10^{-2} \\*
          & ASB   & 1.56\cdot10^{-1} & 4.22\cdot10^{-2} & 9.92\cdot10^{-2} & 4.33\cdot10^{-2} &       &       & ASB   & 1.52\cdot10^{-1} & 3.80\cdot10^{-2} & 9.49\cdot10^{-2} & 4.02\cdot10^{-2} \\*
          & ADC   & 5.78\cdot10^{-1} & 8.44\cdot10^{-3} & 2.93\cdot10^{-1} & 1.41\cdot10^{-2} &       &       & ADC   & 4.94\cdot10^{-1} & 1.27\cdot10^{-2} & 2.53\cdot10^{-1} & 2.21\cdot10^{-2} \\*
          & CB0   & 5.78\cdot10^{-1} & 8.44\cdot10^{-3} & 2.93\cdot10^{-1} & 1.41\cdot10^{-2} &       &       & CB0   & 4.94\cdot10^{-1} & 1.27\cdot10^{-2} & 2.53\cdot10^{-1} & 2.21\cdot10^{-2} \\*
          & CB1   & 9.32\cdot10^{-1} & 0.000    & 4.66\cdot10^{-1} & 9.23\cdot10^{-3} &       &       & CB1   & 9.70\cdot10^{-1} & 0.000    & 4.85\cdot10^{-1} & 1.90\cdot10^{-2} \\*
          & CB2   & 9.28\cdot10^{-1} & 0.000    & 4.64\cdot10^{-1} & 9.19\cdot10^{-3} &       &       & CB2   & 9.79\cdot10^{-1} & 0.000    & 4.89\cdot10^{-1} & 1.92\cdot10^{-2} \\*
          & AC1   & 4.09\cdot10^{-1} & 1.27\cdot10^{-2} & 2.11\cdot10^{-1} & 1.66\cdot10^{-2} &       &       & AC1   & 2.19\cdot10^{-1} & 1.27\cdot10^{-2} & 1.16\cdot10^{-1} & 1.67\cdot10^{-2} \\*
          & AC2   & 8.40\cdot10^{-1} & 0.000    & 4.20\cdot10^{-1} & 8.31\cdot10^{-3} &       &       & AC2   & 8.31\cdot10^{-1} & 0.000    & 4.16\cdot10^{-1} & 1.63\cdot10^{-2} \\*
          & AC3   & 9.07\cdot10^{-1} & 0.000    & 4.54\cdot10^{-1} & 8.98\cdot10^{-3} &       &       & AC3   & 9.07\cdot10^{-1} & 0.000    & 4.54\cdot10^{-1} & 1.78\cdot10^{-2} \\*
          & CSA   & 1.56\cdot10^{-1} & 4.64\cdot10^{-2} & 1.01\cdot10^{-1} & 4.75\cdot10^{-2} &       &       & CSA   & 1.43\cdot10^{-1} & 3.80\cdot10^{-2} & 9.07\cdot10^{-2} & 4.00\cdot10^{-2} \\*
          & CGA   & 2.36\cdot10^{-1} & 2.53\cdot10^{-2} & 1.31\cdot10^{-1} & 2.74\cdot10^{-2} &       &       & CGA   & 2.07\cdot10^{-1} & 3.38\cdot10^{-2} & 1.20\cdot10^{-1} & 3.71\cdot10^{-2} \\*
     \cmidrule(r){1-6} \cmidrule(r){8-13} 
		$[1, 25]$ & ABT   & 1.98\cdot10^{-1} & 3.80\cdot10^{-2} & 1.18\cdot10^{-1} & 4.41\cdot10^{-2} &       & $[1, 10]$ & ABT   & 1.77\cdot10^{-1} & 3.80\cdot10^{-2} & 1.08\cdot10^{-1} & 5.06\cdot10^{-2} \\*
          & ASB   & 1.65\cdot10^{-1} & 3.80\cdot10^{-2} & 1.01\cdot10^{-1} & 4.28\cdot10^{-2} &       &       & ASB   & 1.73\cdot10^{-1} & 3.80\cdot10^{-2} & 1.05\cdot10^{-1} & 5.02\cdot10^{-2} \\*
          & ADC   & 4.77\cdot10^{-1} & 4.22\cdot10^{-3} & 2.41\cdot10^{-1} & 2.24\cdot10^{-2} &       &       & ADC   & 3.42\cdot10^{-1} & 8.44\cdot10^{-3} & 1.75\cdot10^{-1} & 3.87\cdot10^{-2} \\*
          & CB0   & 6.08\cdot10^{-1} & 8.44\cdot10^{-3} & 3.08\cdot10^{-1} & 3.15\cdot10^{-2} &       &       & CB0   & 5.91\cdot10^{-1} & 8.44\cdot10^{-3} & 3.00\cdot10^{-1} & 6.14\cdot10^{-2} \\*
          & CB1   & 1.000    & 0.000    & 5.00\cdot10^{-1} & 3.85\cdot10^{-2} &       &       & CB1   & 1.000    & 0.000    & 5.00\cdot10^{-1} & 9.09\cdot10^{-2} \\*
          & CB2   & 9.83\cdot10^{-1} & 0.000    & 4.92\cdot10^{-1} & 3.78\cdot10^{-2} &       &       & CB2   & 1.000    & 0.000    & 5.00\cdot10^{-1} & 9.09\cdot10^{-2} \\*
          & AC1   & 1.90\cdot10^{-1} & 2.11\cdot10^{-2} & 1.05\cdot10^{-1} & 2.76\cdot10^{-2} &       &       & AC1   & 1.52\cdot10^{-1} & 3.80\cdot10^{-2} & 9.49\cdot10^{-2} & 4.83\cdot10^{-2} \\*
          & AC2   & 9.37\cdot10^{-1} & 0.000    & 4.68\cdot10^{-1} & 3.60\cdot10^{-2} &       &       & AC2   & 8.95\cdot10^{-1} & 0.000    & 4.47\cdot10^{-1} & 8.13\cdot10^{-2} \\*
          & AC3   & 9.28\cdot10^{-1} & 0.000    & 4.64\cdot10^{-1} & 3.57\cdot10^{-2} &       &       & AC3   & 9.79\cdot10^{-1} & 0.000    & 4.89\cdot10^{-1} & 8.90\cdot10^{-2} \\*
          & CSA   & 9.70\cdot10^{-2} & 4.22\cdot10^{-2} & 6.96\cdot10^{-2} & 4.43\cdot10^{-2} &       &       & CSA   & 1.01\cdot10^{-1} & 4.22\cdot10^{-2} & 7.17\cdot10^{-2} & 4.76\cdot10^{-2} \\*
          & CGA   & 2.19\cdot10^{-1} & 3.38\cdot10^{-2} & 1.27\cdot10^{-1} & 4.09\cdot10^{-2} &       &       & CGA   & 2.03\cdot10^{-1} & 3.38\cdot10^{-2} & 1.18\cdot10^{-1} & 4.91\cdot10^{-2} \\*
     \cmidrule(r){1-6} \cmidrule(r){8-13} 
		$[1, 7]$ & ABT   & 1.69\cdot10^{-1} & 3.80\cdot10^{-2} & 1.03\cdot10^{-1} & 5.43\cdot10^{-2} &       & $[1, 5]$ & ABT   & 1.60\cdot10^{-1} & 3.80\cdot10^{-2} & 9.92\cdot10^{-2} & 5.84\cdot10^{-2} \\*
          & ASB   & 1.86\cdot10^{-1} & 3.80\cdot10^{-2} & 1.12\cdot10^{-1} & 5.64\cdot10^{-2} &       &       & ASB   & 1.56\cdot10^{-1} & 3.80\cdot10^{-2} & 9.70\cdot10^{-2} & 5.77\cdot10^{-2} \\*
          & ADC   & 2.62\cdot10^{-1} & 8.44\cdot10^{-3} & 1.35\cdot10^{-1} & 4.01\cdot10^{-2} &       &       & ADC   & 2.07\cdot10^{-1} & 8.44\cdot10^{-3} & 1.08\cdot10^{-1} & 4.15\cdot10^{-2} \\*
          & CB0   & 5.82\cdot10^{-1} & 8.44\cdot10^{-3} & 2.95\cdot10^{-1} & 8.02\cdot10^{-2} &       &       & CB0   & 5.82\cdot10^{-1} & 8.44\cdot10^{-3} & 2.95\cdot10^{-1} & 1.04\cdot10^{-1} \\*
          & CB1   & 9.87\cdot10^{-1} & 0.000    & 4.94\cdot10^{-1} & 1.23\cdot10^{-1} &       &       & CB1   & 9.96\cdot10^{-1} & 0.000    & 4.98\cdot10^{-1} & 1.66\cdot10^{-1} \\*
          & CB2   & 1.000    & 0.000    & 5.00\cdot10^{-1} & 1.25\cdot10^{-1} &       &       & CB2   & 9.92\cdot10^{-1} & 0.000    & 4.96\cdot10^{-1} & 1.65\cdot10^{-1} \\*
          & AC1   & 1.22\cdot10^{-1} & 4.22\cdot10^{-2} & 8.23\cdot10^{-2} & 5.22\cdot10^{-2} &       &       & AC1   & 1.14\cdot10^{-1} & 4.22\cdot10^{-2} & 7.81\cdot10^{-2} & 5.41\cdot10^{-2} \\*
          & AC2   & 9.92\cdot10^{-1} & 0.000    & 4.96\cdot10^{-1} & 1.24\cdot10^{-1} &       &       & AC2   & 9.83\cdot10^{-1} & 0.000    & 4.92\cdot10^{-1} & 1.64\cdot10^{-1} \\*
          & AC3   & 1.000    & 2.11\cdot10^{-2} & 5.11\cdot10^{-1} & 1.43\cdot10^{-1} &       &       & AC3   & 1.000    & 2.11\cdot10^{-2} & 5.11\cdot10^{-1} & 1.84\cdot10^{-1} \\*
          & CSA   & 1.10\cdot10^{-1} & 4.22\cdot10^{-2} & 7.59\cdot10^{-2} & 5.06\cdot10^{-2} &       &       & CSA   & 1.10\cdot10^{-1} & 3.80\cdot10^{-2} & 7.38\cdot10^{-2} & 4.99\cdot10^{-2} \\*
          & CGA   & 1.98\cdot10^{-1} & 3.38\cdot10^{-2} & 1.16\cdot10^{-1} & 5.43\cdot10^{-2} &       &       & CGA   & 1.77\cdot10^{-1} & 3.80\cdot10^{-2} & 1.08\cdot10^{-1} & 6.12\cdot10^{-2} \\*
     \cmidrule(r){1-6} \cmidrule(r){8-13} 
		$[1, 3]$ & ABT   & 1.69\cdot10^{-1} & 3.80\cdot10^{-2} & 1.03\cdot10^{-1} & 7.07\cdot10^{-2} &       & $[1, 2]$ & ABT   & 1.69\cdot10^{-1} & 3.80\cdot10^{-2} & 1.03\cdot10^{-1} & 8.16\cdot10^{-2} \\*
          & ASB   & 1.60\cdot10^{-1} & 3.80\cdot10^{-2} & 9.92\cdot10^{-2} & 6.86\cdot10^{-2} &       &       & ASB   & 1.69\cdot10^{-1} & 3.80\cdot10^{-2} & 1.03\cdot10^{-1} & 8.16\cdot10^{-2} \\*
          & ADC   & 1.60\cdot10^{-1} & 1.27\cdot10^{-2} & 8.65\cdot10^{-2} & 4.96\cdot10^{-2} &       &       & ADC   & 1.27\cdot10^{-1} & 3.38\cdot10^{-2} & 8.02\cdot10^{-2} & 6.47\cdot10^{-2} \\*
          & CB0   & 5.82\cdot10^{-1} & 8.44\cdot10^{-3} & 2.95\cdot10^{-1} & 1.52\cdot10^{-1} &       &       & CB0   & 5.82\cdot10^{-1} & 8.44\cdot10^{-3} & 2.95\cdot10^{-1} & 2.00\cdot10^{-1} \\*
          & CB1   & 1.000    & 0.000    & 5.00\cdot10^{-1} & 2.50\cdot10^{-1} &       &       & CB1   & 9.75\cdot10^{-1} & 0.000    & 4.87\cdot10^{-1} & 3.25\cdot10^{-1} \\*
          & CB2   & 9.66\cdot10^{-1} & 0.000    & 4.83\cdot10^{-1} & 2.42\cdot10^{-1} &       &       & CB2   & 9.20\cdot10^{-1} & 0.000    & 4.60\cdot10^{-1} & 3.07\cdot10^{-1} \\*
          & AC1   & 1.18\cdot10^{-1} & 4.22\cdot10^{-2} & 8.02\cdot10^{-2} & 6.12\cdot10^{-2} &       &       & AC1   & 1.18\cdot10^{-1} & 4.22\cdot10^{-2} & 8.02\cdot10^{-2} & 6.75\cdot10^{-2} \\*
          & AC2   & 8.35\cdot10^{-1} & 0.000    & 4.18\cdot10^{-1} & 2.09\cdot10^{-1} &       &       & AC2   & 9.24\cdot10^{-1} & 0.000    & 4.62\cdot10^{-1} & 3.08\cdot10^{-1} \\*
          & AC3   & 1.000    & 0.000    & 5.00\cdot10^{-1} & 2.50\cdot10^{-1} &       &       & AC3   & 9.28\cdot10^{-1} & 0.000    & 4.64\cdot10^{-1} & 3.09\cdot10^{-1} \\*
          & CSA   & 1.22\cdot10^{-1} & 3.80\cdot10^{-2} & 8.02\cdot10^{-2} & 5.91\cdot10^{-2} &       &       & CSA   & 1.48\cdot10^{-1} & 4.22\cdot10^{-2} & 9.49\cdot10^{-2} & 7.74\cdot10^{-2} \\*
          & CGA   & 1.90\cdot10^{-1} & 3.38\cdot10^{-2} & 1.12\cdot10^{-1} & 7.28\cdot10^{-2} &       &       & CGA   & 1.69\cdot10^{-1} & 3.80\cdot10^{-2} & 1.03\cdot10^{-1} & 8.16\cdot10^{-2} \\*
     \cmidrule(r){1-6} \cmidrule(r){8-13} 
		$[2, 3]$ & ABT   & 1.69\cdot10^{-1} & 3.80\cdot10^{-2} & 1.03\cdot10^{-1} & 9.03\cdot10^{-2} &       & $[1, 1]$ & ABT   & 1.60\cdot10^{-1} & 3.80\cdot10^{-2} & 9.92\cdot10^{-2} & 9.92\cdot10^{-2} \\*
          & ASB   & 1.60\cdot10^{-1} & 3.80\cdot10^{-2} & 9.92\cdot10^{-2} & 8.69\cdot10^{-2} &       &       & ASB   & 1.60\cdot10^{-1} & 3.80\cdot10^{-2} & 9.92\cdot10^{-2} & 9.92\cdot10^{-2} \\*
          & ADC   & 1.10\cdot10^{-1} & 4.22\cdot10^{-2} & 7.59\cdot10^{-2} & 6.92\cdot10^{-2} &       &       & ADC   & 4.64\cdot10^{-2} & 6.33\cdot10^{-2} & 5.49\cdot10^{-2} & 5.49\cdot10^{-2} \\*
          & CB0   & 5.78\cdot10^{-1} & 8.44\cdot10^{-3} & 2.93\cdot10^{-1} & 2.36\cdot10^{-1} &       &       & CB0   & 1.27\cdot10^{-1} & 9.28\cdot10^{-2} & 1.10\cdot10^{-1} & 1.10\cdot10^{-1} \\*
          & CB1   & 9.11\cdot10^{-1} & 0.000    & 4.56\cdot10^{-1} & 3.65\cdot10^{-1} &       &       & CB1   & 1.56\cdot10^{-1} & 1.18\cdot10^{-1} & 1.37\cdot10^{-1} & 1.37\cdot10^{-1} \\*
          & CB2   & 7.72\cdot10^{-1} & 4.22\cdot10^{-3} & 3.88\cdot10^{-1} & 3.11\cdot10^{-1} &       &       & CB2   & 1.60\cdot10^{-1} & 3.80\cdot10^{-2} & 9.92\cdot10^{-2} & 9.92\cdot10^{-2} \\*
          & AC1   & 1.22\cdot10^{-1} & 4.22\cdot10^{-2} & 8.23\cdot10^{-2} & 7.43\cdot10^{-2} &       &       & AC1   & 1.52\cdot10^{-1} & 4.64\cdot10^{-2} & 9.92\cdot10^{-2} & 9.92\cdot10^{-2} \\*
          & AC2   & 6.79\cdot10^{-1} & 4.22\cdot10^{-3} & 3.42\cdot10^{-1} & 2.74\cdot10^{-1} &       &       & AC2   & 1.60\cdot10^{-1} & 3.80\cdot10^{-2} & 9.92\cdot10^{-2} & 9.92\cdot10^{-2} \\*
          & AC3   & 8.27\cdot10^{-1} & 0.000    & 4.14\cdot10^{-1} & 3.31\cdot10^{-1} &       &       & AC3   & 1.52\cdot10^{-1} & 4.64\cdot10^{-2} & 9.92\cdot10^{-2} & 9.92\cdot10^{-2} \\*
          & CSA   & 1.52\cdot10^{-1} & 4.64\cdot10^{-2} & 9.92\cdot10^{-2} & 8.86\cdot10^{-2} &       &       & CSA   & 1.60\cdot10^{-1} & 3.80\cdot10^{-2} & 9.92\cdot10^{-2} & 9.92\cdot10^{-2} \\*
          & CGA   & 1.56\cdot10^{-1} & 4.22\cdot10^{-2} & 9.92\cdot10^{-2} & 8.78\cdot10^{-2} &       &       & CGA   & 1.60\cdot10^{-1} & 3.80\cdot10^{-2} & 9.92\cdot10^{-2} & 9.92\cdot10^{-2} \\*
     \cmidrule(r){1-6} \cmidrule(r){8-13}
		
		$[3, 2]$ & ABT   & 1.65\cdot10^{-1} & 3.80\cdot10^{-2} & 1.01\cdot10^{-1} & 1.14\cdot10^{-1} &       & $[2, 1]$ & ABT   & 1.65\cdot10^{-1} & 3.80\cdot10^{-2} & 1.01\cdot10^{-1} & 1.22\cdot10^{-1} \\*
          & ASB   & 1.56\cdot10^{-1} & 3.80\cdot10^{-2} & 9.70\cdot10^{-2} & 1.09\cdot10^{-1} &       &       & ASB   & 1.65\cdot10^{-1} & 3.80\cdot10^{-2} & 1.01\cdot10^{-1} & 1.22\cdot10^{-1} \\*
          & ADC   & 1.69\cdot10^{-2} & 8.86\cdot10^{-2} & 5.27\cdot10^{-2} & 4.56\cdot10^{-2} &       &       & ADC   & 1.69\cdot10^{-2} & 1.10\cdot10^{-1} & 6.33\cdot10^{-2} & 4.78\cdot10^{-2} \\*
          & CB0   & 1.69\cdot10^{-2} & 7.00\cdot10^{-1} & 3.59\cdot10^{-1} & 2.90\cdot10^{-1} &       &       & CB0   & 1.69\cdot10^{-2} & 7.00\cdot10^{-1} & 3.59\cdot10^{-1} & 2.45\cdot10^{-1} \\*
          & CB1   & 0.000    & 7.55\cdot10^{-1} & 3.78\cdot10^{-1} & 3.02\cdot10^{-1} &       &       & CB1   & 0.000    & 9.37\cdot10^{-1} & 4.68\cdot10^{-1} & 3.12\cdot10^{-1} \\*
          & CB2   & 0.000    & 7.05\cdot10^{-1} & 3.52\cdot10^{-1} & 2.82\cdot10^{-1} &       &       & CB2   & 0.000    & 8.73\cdot10^{-1} & 4.37\cdot10^{-1} & 2.91\cdot10^{-1} \\*
          & AC1   & 1.60\cdot10^{-1} & 4.64\cdot10^{-2} & 1.03\cdot10^{-1} & 1.15\cdot10^{-1} &       &       & AC1   & 1.69\cdot10^{-1} & 4.64\cdot10^{-2} & 1.08\cdot10^{-1} & 1.28\cdot10^{-1} \\*
          & AC2   & 4.22\cdot10^{-3} & 9.58\cdot10^{-1} & 4.81\cdot10^{-1} & 3.86\cdot10^{-1} &       &       & AC2   & 0.000    & 1.000    & 5.00\cdot10^{-1} & 3.33\cdot10^{-1} \\*
          & AC3   & 1.52\cdot10^{-1} & 7.05\cdot10^{-1} & 4.28\cdot10^{-1} & 3.73\cdot10^{-1} &       &       & AC3   & 1.52\cdot10^{-1} & 7.05\cdot10^{-1} & 4.28\cdot10^{-1} & 3.36\cdot10^{-1} \\*
          & CSA   & 1.90\cdot10^{-1} & 3.80\cdot10^{-2} & 1.14\cdot10^{-1} & 1.29\cdot10^{-1} &       &       & CSA   & 1.48\cdot10^{-1} & 3.80\cdot10^{-2} & 9.28\cdot10^{-2} & 1.11\cdot10^{-1} \\*
          & CGA   & 1.73\cdot10^{-1} & 3.80\cdot10^{-2} & 1.05\cdot10^{-1} & 1.19\cdot10^{-1} &       &       & CGA   & 1.48\cdot10^{-1} & 3.80\cdot10^{-2} & 9.28\cdot10^{-2} & 1.11\cdot10^{-1} \\*
     \cmidrule(r){1-6} \cmidrule(r){8-13} 
		$[3, 1]$ & ABT   & 1.65\cdot10^{-1} & 3.80\cdot10^{-2} & 1.01\cdot10^{-1} & 1.33\cdot10^{-1} &       & $[5, 1]$ & ABT   & 1.56\cdot10^{-1} & 3.80\cdot10^{-2} & 9.70\cdot10^{-2} & 1.36\cdot10^{-1} \\*
          & ASB   & 1.65\cdot10^{-1} & 3.80\cdot10^{-2} & 1.01\cdot10^{-1} & 1.33\cdot10^{-1} &       &       & ASB   & 1.77\cdot10^{-1} & 4.22\cdot10^{-2} & 1.10\cdot10^{-1} & 1.55\cdot10^{-1} \\*
          & ADC   & 1.69\cdot10^{-2} & 1.69\cdot10^{-1} & 9.28\cdot10^{-2} & 5.49\cdot10^{-2} &       &       & ADC   & 1.69\cdot10^{-2} & 2.03\cdot10^{-1} & 1.10\cdot10^{-1} & 4.78\cdot10^{-2} \\*
          & CB0   & 1.69\cdot10^{-2} & 7.00\cdot10^{-1} & 3.59\cdot10^{-1} & 1.88\cdot10^{-1} &       &       & CB0   & 1.69\cdot10^{-2} & 7.05\cdot10^{-1} & 3.61\cdot10^{-1} & 1.32\cdot10^{-1} \\*
          & CB1   & 0.000    & 1.000    & 5.00\cdot10^{-1} & 2.50\cdot10^{-1} &       &       & CB1   & 0.000    & 1.000    & 5.00\cdot10^{-1} & 1.67\cdot10^{-1} \\*
          & CB2   & 0.000    & 1.000    & 5.00\cdot10^{-1} & 2.50\cdot10^{-1} &       &       & CB2   & 0.000    & 1.000    & 5.00\cdot10^{-1} & 1.67\cdot10^{-1} \\*
          & AC1   & 1.60\cdot10^{-1} & 3.80\cdot10^{-2} & 9.92\cdot10^{-2} & 1.30\cdot10^{-1} &       &       & AC1   & 1.31\cdot10^{-1} & 4.22\cdot10^{-2} & 8.65\cdot10^{-2} & 1.16\cdot10^{-1} \\*
          & AC2   & 1.43\cdot10^{-1} & 9.83\cdot10^{-1} & 5.63\cdot10^{-1} & 3.53\cdot10^{-1} &       &       & AC2   & 1.43\cdot10^{-1} & 9.83\cdot10^{-1} & 5.63\cdot10^{-1} & 2.83\cdot10^{-1} \\*
          & AC3   & 1.43\cdot10^{-1} & 9.83\cdot10^{-1} & 5.63\cdot10^{-1} & 3.53\cdot10^{-1} &       &       & AC3   & 1.43\cdot10^{-1} & 9.83\cdot10^{-1} & 5.63\cdot10^{-1} & 2.83\cdot10^{-1} \\*
          & CSA   & 1.98\cdot10^{-1} & 3.80\cdot10^{-2} & 1.18\cdot10^{-1} & 1.58\cdot10^{-1} &       &       & CSA   & 2.03\cdot10^{-1} & 3.80\cdot10^{-2} & 1.20\cdot10^{-1} & 1.75\cdot10^{-1} \\*
          & CGA   & 1.60\cdot10^{-1} & 4.22\cdot10^{-2} & 1.01\cdot10^{-1} & 1.31\cdot10^{-1} &       &       & CGA   & 1.27\cdot10^{-1} & 4.22\cdot10^{-2} & 8.44\cdot10^{-2} & 1.13\cdot10^{-1} \\*
     \cmidrule(r){1-6} \cmidrule(r){8-13} 
		$[7, 1]$ & ABT   & 1.56\cdot10^{-1} & 3.80\cdot10^{-2} & 9.70\cdot10^{-2} & 1.41\cdot10^{-1} &       & $[10, 1]$ & ABT   & 1.56\cdot10^{-1} & 3.80\cdot10^{-2} & 9.70\cdot10^{-2} & 1.45\cdot10^{-1} \\*
          & ASB   & 1.69\cdot10^{-1} & 3.80\cdot10^{-2} & 1.03\cdot10^{-1} & 1.52\cdot10^{-1} &       &       & ASB   & 1.73\cdot10^{-1} & 3.80\cdot10^{-2} & 1.05\cdot10^{-1} & 1.61\cdot10^{-1} \\*
          & ADC   & 1.69\cdot10^{-2} & 2.24\cdot10^{-1} & 1.20\cdot10^{-1} & 4.27\cdot10^{-2} &       &       & ADC   & 1.69\cdot10^{-2} & 2.36\cdot10^{-1} & 1.27\cdot10^{-1} & 3.68\cdot10^{-2} \\*
          & CB0   & 1.69\cdot10^{-2} & 7.05\cdot10^{-1} & 3.61\cdot10^{-1} & 1.03\cdot10^{-1} &       &       & CB0   & 1.69\cdot10^{-2} & 7.09\cdot10^{-1} & 3.63\cdot10^{-1} & 7.98\cdot10^{-2} \\*
          & CB1   & 0.000    & 1.000    & 5.00\cdot10^{-1} & 1.25\cdot10^{-1} &       &       & CB1   & 0.000    & 1.000    & 5.00\cdot10^{-1} & 9.09\cdot10^{-2} \\*
          & CB2   & 0.000    & 1.000    & 5.00\cdot10^{-1} & 1.25\cdot10^{-1} &       &       & CB2   & 0.000    & 1.000    & 5.00\cdot10^{-1} & 9.09\cdot10^{-2} \\*
          & AC1   & 1.10\cdot10^{-1} & 4.22\cdot10^{-2} & 7.59\cdot10^{-2} & 1.01\cdot10^{-1} &       &       & AC1   & 9.28\cdot10^{-2} & 4.22\cdot10^{-2} & 6.75\cdot10^{-2} & 8.82\cdot10^{-2} \\*
          & AC2   & 0.000    & 1.000    & 5.00\cdot10^{-1} & 1.25\cdot10^{-1} &       &       & AC2   & 1.27\cdot10^{-2} & 7.22\cdot10^{-1} & 3.67\cdot10^{-1} & 7.71\cdot10^{-2} \\*
          & AC3   & 1.43\cdot10^{-1} & 9.83\cdot10^{-1} & 5.63\cdot10^{-1} & 2.48\cdot10^{-1} &       &       & AC3   & 1.43\cdot10^{-1} & 9.83\cdot10^{-1} & 5.63\cdot10^{-1} & 2.20\cdot10^{-1} \\*
          & CSA   & 1.98\cdot10^{-1} & 3.38\cdot10^{-2} & 1.16\cdot10^{-1} & 1.78\cdot10^{-1} &       &       & CSA   & 1.98\cdot10^{-1} & 2.11\cdot10^{-2} & 1.10\cdot10^{-1} & 1.82\cdot10^{-1} \\*
          & CGA   & 1.27\cdot10^{-1} & 4.64\cdot10^{-2} & 8.65\cdot10^{-2} & 1.17\cdot10^{-1} &       &       & CGA   & 1.27\cdot10^{-1} & 4.22\cdot10^{-2} & 8.44\cdot10^{-2} & 1.19\cdot10^{-1} \\*
     \cmidrule(r){1-6} \cmidrule(r){8-13} 
		$[25, 1]$ & ABT   & 1.18\cdot10^{-1} & 4.64\cdot10^{-2} & 8.23\cdot10^{-2} & 1.15\cdot10^{-1} &       & $[50, 1]$ & ABT   & 6.33\cdot10^{-2} & 5.06\cdot10^{-2} & 5.70\cdot10^{-2} & 6.30\cdot10^{-2} \\*
          & ASB   & 1.56\cdot10^{-1} & 3.80\cdot10^{-2} & 9.70\cdot10^{-2} & 1.52\cdot10^{-1} &       &       & ASB   & 1.52\cdot10^{-1} & 4.22\cdot10^{-2} & 9.70\cdot10^{-2} & 1.50\cdot10^{-1} \\*
          & ADC   & 1.69\cdot10^{-2} & 2.36\cdot10^{-1} & 1.27\cdot10^{-1} & 2.53\cdot10^{-2} &       &       & ADC   & 1.69\cdot10^{-2} & 2.36\cdot10^{-1} & 1.27\cdot10^{-1} & 2.12\cdot10^{-2} \\*
          & CB0   & 1.69\cdot10^{-2} & 7.05\cdot10^{-1} & 3.61\cdot10^{-1} & 4.33\cdot10^{-2} &       &       & CB0   & 1.69\cdot10^{-2} & 7.05\cdot10^{-1} & 3.61\cdot10^{-1} & 3.04\cdot10^{-2} \\*
          & CB1   & 0.000    & 1.000    & 5.00\cdot10^{-1} & 3.85\cdot10^{-2} &       &       & CB1   & 0.000    & 1.000    & 5.00\cdot10^{-1} & 1.96\cdot10^{-2} \\*
          & CB2   & 0.000    & 1.000    & 5.00\cdot10^{-1} & 3.85\cdot10^{-2} &       &       & CB2   & 0.000    & 1.000    & 5.00\cdot10^{-1} & 1.96\cdot10^{-2} \\*
          & AC1   & 2.11\cdot10^{-2} & 9.28\cdot10^{-2} & 5.70\cdot10^{-2} & 2.39\cdot10^{-2} &       &       & AC1   & 1.69\cdot10^{-2} & 1.05\cdot10^{-1} & 6.12\cdot10^{-2} & 1.86\cdot10^{-2} \\*
          & AC2   & 1.48\cdot10^{-1} & 9.41\cdot10^{-1} & 5.44\cdot10^{-1} & 1.78\cdot10^{-1} &       &       & AC2   & 1.48\cdot10^{-1} & 9.41\cdot10^{-1} & 5.44\cdot10^{-1} & 1.63\cdot10^{-1} \\*
          & AC3   & 1.43\cdot10^{-1} & 9.83\cdot10^{-1} & 5.63\cdot10^{-1} & 1.76\cdot10^{-1} &       &       & AC3   & 1.43\cdot10^{-1} & 9.83\cdot10^{-1} & 5.63\cdot10^{-1} & 1.60\cdot10^{-1} \\*
          & CSA   & 6.75\cdot10^{-2} & 3.38\cdot10^{-2} & 5.06\cdot10^{-2} & 6.62\cdot10^{-2} &       &       & CSA   & 8.02\cdot10^{-2} & 8.44\cdot10^{-2} & 8.23\cdot10^{-2} & 8.03\cdot10^{-2} \\*
          & CGA   & 9.28\cdot10^{-2} & 6.33\cdot10^{-2} & 7.81\cdot10^{-2} & 9.17\cdot10^{-2} &       &       & CGA   & 5.49\cdot10^{-2} & 8.86\cdot10^{-2} & 7.17\cdot10^{-2} & 5.55\cdot10^{-2} \\*
     \cmidrule(r){1-6} \cmidrule(r){8-13} 
		$[100, 1]$ & ABT   & 2.53\cdot10^{-2} & 9.70\cdot10^{-2} & 6.12\cdot10^{-2} & 2.60\cdot10^{-2} &       &       &       &       &       &       &  \\*
          & ASB   & 1.35\cdot10^{-1} & 4.22\cdot10^{-2} & 8.86\cdot10^{-2} & 1.34\cdot10^{-1} &       &       &       &       &       &       &  \\*
          & ADC   & 4.94\cdot10^{-1} & 3.59\cdot10^{-1} & 4.26\cdot10^{-1} & 4.92\cdot10^{-1} &       &       &       &       &       &       &  \\*
          & CB0   & 1.60\cdot10^{-1} & 6.92\cdot10^{-1} & 4.26\cdot10^{-1} & 1.66\cdot10^{-1} &       &       &       &       &       &       &  \\*
          & CB1   & 0.000    & 9.41\cdot10^{-1} & 4.70\cdot10^{-1} & 9.32\cdot10^{-3} &       &       &       &       &       &       &  \\*
          & CB2   & 0.000    & 9.37\cdot10^{-1} & 4.68\cdot10^{-1} & 9.27\cdot10^{-3} &       &       &       &       &       &       &  \\*
          & AC1   & 1.69\cdot10^{-2} & 2.66\cdot10^{-1} & 1.41\cdot10^{-1} & 1.93\cdot10^{-2} &       &       &       &       &       &       &  \\*
          & AC2   & 1.48\cdot10^{-1} & 9.41\cdot10^{-1} & 5.44\cdot10^{-1} & 1.56\cdot10^{-1} &       &       &       &       &       &       &  \\*
          & AC3   & 1.43\cdot10^{-1} & 9.83\cdot10^{-1} & 5.63\cdot10^{-1} & 1.52\cdot10^{-1} &       &       &       &       &       &       &  \\*
          & CSA   & 3.80\cdot10^{-2} & 1.05\cdot10^{-1} & 7.17\cdot10^{-2} & 3.86\cdot10^{-2} &       &       &       &       &       &       &  \\*
          & CGA   & 4.64\cdot10^{-2} & 9.28\cdot10^{-2} & 6.96\cdot10^{-2} & 4.69\cdot10^{-2} &       &       &       &       &       &       &  \\*

\end{longtabu}
}

\newpage

%%%%%%%%%%%%%%%%%%
%%% UCI BREAST %%%
%%%%%%%%%%%%%%%%%%

\begin{landscape}
% GRAPHICS
\begin{figure}[p]
\centering
\includegraphics[width=.6\paperheight]{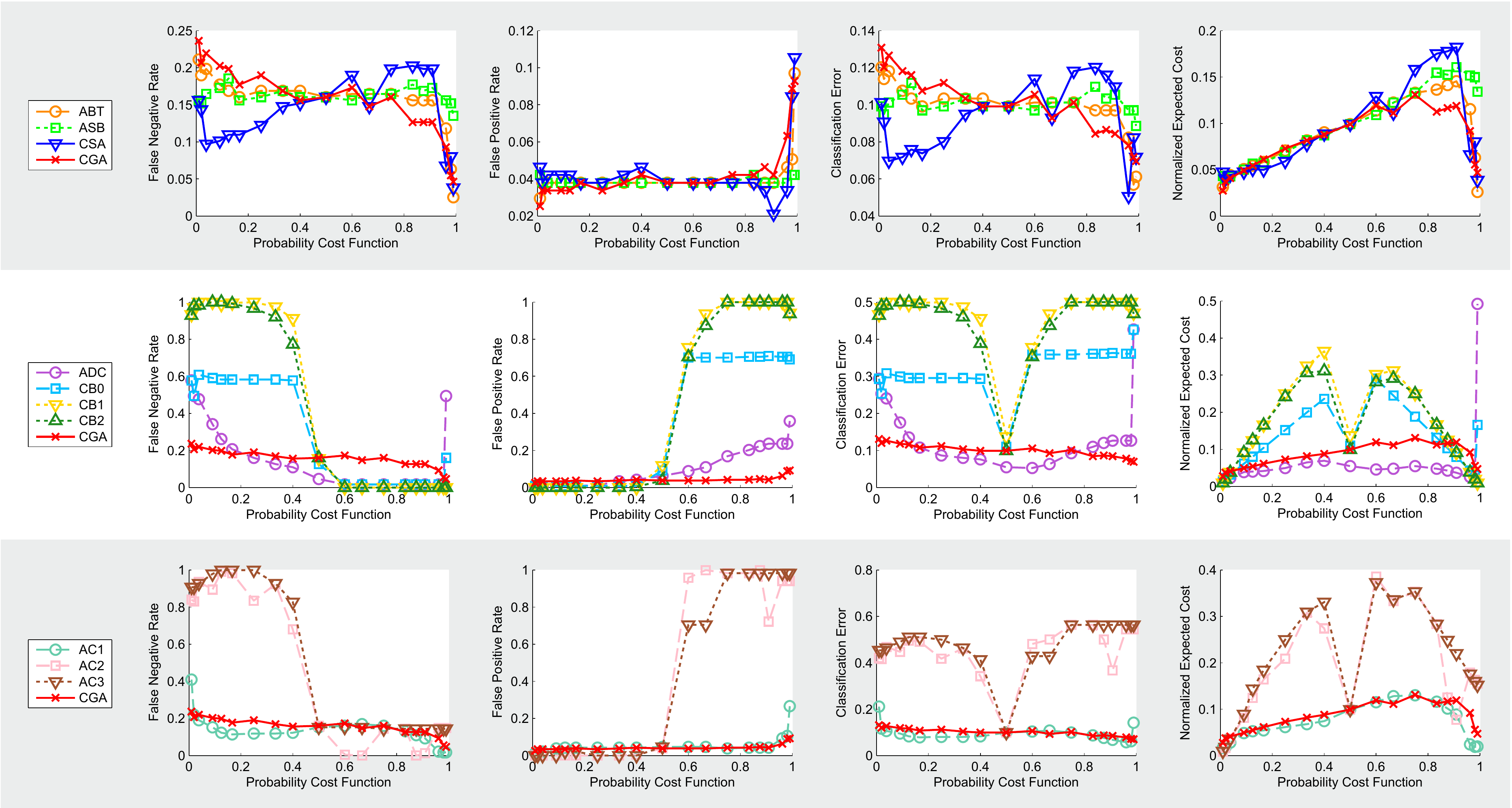}
\caption[Results obtained for the UCI Breast Cancer Dataset.]{Results obtained for the UCI Breast Cancer Dataset. First column of the illustration corresponds to False Negative Rate, second column to False Positive Rate, third column to Classification Error and the fourth one corresponds to Normalized Expected Cost. For a clearer visualization, algorithms have been divided into three groups so each row of the illustration corresponds to a different group. Cost Generalized AdaBoost is plotted in all the graphs to have a common reference across the representations.}
\label{fig:twoclouds_perform} % caption for the whole figure
\end{figure}
\end{landscape}
\newpage

%%%%%%%%%%%%%%%%%%
%%% UCI CREDIT %%%
%%%%%%%%%%%%%%%%%%

% TABLE

{
\tiny
\begin{longtabu} to \textwidth {llX[$l]@{}@{}X[$l]@{}@{}X[$l]@{}@{}X[$l]@{}@{}l@{}llX[$l]@{}@{}X[$l]@{}@{}X[$l]@{}@{}X[$l]@{}}

\toprule
{\textbf{Cost}}  & {\textbf{Alg}}   & {\textbf{FNR}}    & {\textbf{FPR}}    & {\textbf{CE}}    & {\textbf{NEC}}   && {\textbf{Cost}}  & {\textbf{Alg}}   & {\textbf{FNR}}    & {\textbf{FPR}}    & {\textbf{CE}} & {\textbf{NEC}}\\* 
\cmidrule(r){1-6} \cmidrule(r){8-13}
\endfirsthead

%\multicolumn{13}{l}% {\tiny{Continued from previous page}} \\*
\toprule
{\textbf{Cost}}  & {\textbf{Alg}}   & {\textbf{FNR}}    & {\textbf{FPR}}    & {\textbf{CE}}    & {\textbf{NEC}}   && {\textbf{Cost}}  & {\textbf{Alg}}   & {\textbf{FNR}}    & {\textbf{FPR}}    & {\textbf{CE}} & {\textbf{NEC}}\\* 
\cmidrule(r){1-6} \cmidrule(r){8-13}
\endhead

\multicolumn{13}{r}{{\tiny{\tablename\ \thetable{} - Continued on next page $\rightarrow$}}} \\* 
\endfoot

\bottomrule
  \caption{Results obtained for the UCI Credit Dataset.}
	\label{tab:credit_perform}%
\endlastfoot

    $[1, 100]$ & ABT   & 8.17\cdot10^{-1} & 2.67\cdot10^{-2} & 4.22\cdot10^{-1} & 3.45\cdot10^{-2} &       & $[1, 50]$ & ABT   & 7.30\cdot10^{-1} & 7.33\cdot10^{-2} & 4.02\cdot10^{-1} & 8.62\cdot10^{-2} \\*
          & ASB   & 7.83\cdot10^{-1} & 4.67\cdot10^{-2} & 4.15\cdot10^{-1} & 5.40\cdot10^{-2} &       &       & ASB   & 8.23\cdot10^{-1} & 3.33\cdot10^{-2} & 4.28\cdot10^{-1} & 4.88\cdot10^{-2} \\*
          & ADC   & 9.67\cdot10^{-1} & 3.33\cdot10^{-3} & 4.85\cdot10^{-1} & 1.29\cdot10^{-2} &       &       & ADC   & 9.67\cdot10^{-1} & 3.33\cdot10^{-3} & 4.85\cdot10^{-1} & 2.22\cdot10^{-2} \\*
          & CB0   & 9.67\cdot10^{-1} & 3.33\cdot10^{-3} & 4.85\cdot10^{-1} & 1.29\cdot10^{-2} &       &       & CB0   & 9.67\cdot10^{-1} & 3.33\cdot10^{-3} & 4.85\cdot10^{-1} & 2.22\cdot10^{-2} \\*
          & CB1   & 9.67\cdot10^{-1} & 3.33\cdot10^{-3} & 4.85\cdot10^{-1} & 1.29\cdot10^{-2} &       &       & CB1   & 9.67\cdot10^{-1} & 3.33\cdot10^{-3} & 4.85\cdot10^{-1} & 2.22\cdot10^{-2} \\*
          & CB2   & 9.67\cdot10^{-1} & 3.33\cdot10^{-3} & 4.85\cdot10^{-1} & 1.29\cdot10^{-2} &       &       & CB2   & 9.67\cdot10^{-1} & 3.33\cdot10^{-3} & 4.85\cdot10^{-1} & 2.22\cdot10^{-2} \\*
          & AC1   & 9.67\cdot10^{-1} & 3.33\cdot10^{-3} & 4.85\cdot10^{-1} & 1.29\cdot10^{-2} &       &       & AC1   & 9.67\cdot10^{-1} & 3.33\cdot10^{-3} & 4.85\cdot10^{-1} & 2.22\cdot10^{-2} \\*
          & AC2   & 9.67\cdot10^{-1} & 3.33\cdot10^{-3} & 4.85\cdot10^{-1} & 1.29\cdot10^{-2} &       &       & AC2   & 9.67\cdot10^{-1} & 3.33\cdot10^{-3} & 4.85\cdot10^{-1} & 2.22\cdot10^{-2} \\*
          & AC3   & 9.67\cdot10^{-1} & 3.33\cdot10^{-3} & 4.85\cdot10^{-1} & 1.29\cdot10^{-2} &       &       & AC3   & 9.67\cdot10^{-1} & 3.33\cdot10^{-3} & 4.85\cdot10^{-1} & 2.22\cdot10^{-2} \\*
          & CSA   & 7.90\cdot10^{-1} & 6.67\cdot10^{-2} & 4.28\cdot10^{-1} & 7.38\cdot10^{-2} &       &       & CSA   & 6.77\cdot10^{-1} & 1.17\cdot10^{-1} & 3.97\cdot10^{-1} & 1.28\cdot10^{-1} \\*
          & CGA   & 9.67\cdot10^{-1} & 3.33\cdot10^{-3} & 4.85\cdot10^{-1} & 1.29\cdot10^{-2} &       &       & CGA   & 9.50\cdot10^{-1} & 3.33\cdot10^{-3} & 4.77\cdot10^{-1} & 2.19\cdot10^{-2} \\*
     \cmidrule(r){1-6} \cmidrule(r){8-13} 
		$[1, 25]$ & ABT   & 6.43\cdot10^{-1} & 9.67\cdot10^{-2} & 3.70\cdot10^{-1} & 1.18\cdot10^{-1} &       & $[1, 10]$ & ABT   & 6.13\cdot10^{-1} & 1.33\cdot10^{-1} & 3.73\cdot10^{-1} & 1.77\cdot10^{-1} \\*
          & ASB   & 7.80\cdot10^{-1} & 4.00\cdot10^{-2} & 4.10\cdot10^{-1} & 6.85\cdot10^{-2} &       &       & ASB   & 6.83\cdot10^{-1} & 7.00\cdot10^{-2} & 3.77\cdot10^{-1} & 1.26\cdot10^{-1} \\*
          & ADC   & 3.33\cdot10^{-1} & 6.70\cdot10^{-1} & 5.02\cdot10^{-1} & 6.57\cdot10^{-1} &       &       & ADC   & 3.33\cdot10^{-2} & 9.97\cdot10^{-1} & 5.15\cdot10^{-1} & 9.09\cdot10^{-1} \\*
          & CB0   & 9.67\cdot10^{-1} & 3.33\cdot10^{-3} & 4.85\cdot10^{-1} & 4.04\cdot10^{-2} &       &       & CB0   & 9.67\cdot10^{-1} & 3.33\cdot10^{-3} & 4.85\cdot10^{-1} & 9.09\cdot10^{-2} \\*
          & CB1   & 9.73\cdot10^{-1} & 3.33\cdot10^{-3} & 4.88\cdot10^{-1} & 4.06\cdot10^{-2} &       &       & CB1   & 1.000    & 0.000    & 5.00\cdot10^{-1} & 9.09\cdot10^{-2} \\*
          & CB2   & 9.83\cdot10^{-1} & 0.000    & 4.92\cdot10^{-1} & 3.78\cdot10^{-2} &       &       & CB2   & 9.97\cdot10^{-1} & 0.000    & 4.98\cdot10^{-1} & 9.06\cdot10^{-2} \\*
          & AC1   & 9.67\cdot10^{-1} & 3.33\cdot10^{-3} & 4.85\cdot10^{-1} & 4.04\cdot10^{-2} &       &       & AC1   & 8.77\cdot10^{-1} & 2.67\cdot10^{-2} & 4.52\cdot10^{-1} & 1.04\cdot10^{-1} \\*
          & AC2   & 9.93\cdot10^{-1} & 0.000    & 4.97\cdot10^{-1} & 3.82\cdot10^{-2} &       &       & AC2   & 1.000    & 0.000    & 5.00\cdot10^{-1} & 9.09\cdot10^{-2} \\*
          & AC3   & 9.83\cdot10^{-1} & 3.33\cdot10^{-3} & 4.93\cdot10^{-1} & 4.10\cdot10^{-2} &       &       & AC3   & 1.000    & 0.000    & 5.00\cdot10^{-1} & 9.09\cdot10^{-2} \\*
          & CSA   & 5.80\cdot10^{-1} & 1.60\cdot10^{-1} & 3.70\cdot10^{-1} & 1.76\cdot10^{-1} &       &       & CSA   & 5.00\cdot10^{-1} & 2.03\cdot10^{-1} & 3.52\cdot10^{-1} & 2.30\cdot10^{-1} \\*
          & CGA   & 8.67\cdot10^{-1} & 2.67\cdot10^{-2} & 4.47\cdot10^{-1} & 5.90\cdot10^{-2} &       &       & CGA   & 6.90\cdot10^{-1} & 7.00\cdot10^{-2} & 3.80\cdot10^{-1} & 1.26\cdot10^{-1} \\*
     \cmidrule(r){1-6} \cmidrule(r){8-13} 
		$[1, 7]$ & ABT   & 6.13\cdot10^{-1} & 1.33\cdot10^{-1} & 3.73\cdot10^{-1} & 1.93\cdot10^{-1} &       & $[1, 5]$ & ABT   & 5.83\cdot10^{-1} & 1.47\cdot10^{-1} & 3.65\cdot10^{-1} & 2.19\cdot10^{-1} \\*
          & ASB   & 6.47\cdot10^{-1} & 1.00\cdot10^{-1} & 3.73\cdot10^{-1} & 1.68\cdot10^{-1} &       &       & ASB   & 5.83\cdot10^{-1} & 1.43\cdot10^{-1} & 3.63\cdot10^{-1} & 2.17\cdot10^{-1} \\*
          & ADC   & 3.33\cdot10^{-2} & 9.97\cdot10^{-1} & 5.15\cdot10^{-1} & 8.76\cdot10^{-1} &       &       & ADC   & 4.33\cdot10^{-2} & 9.93\cdot10^{-1} & 5.18\cdot10^{-1} & 8.35\cdot10^{-1} \\*
          & CB0   & 9.67\cdot10^{-1} & 3.33\cdot10^{-3} & 4.85\cdot10^{-1} & 1.24\cdot10^{-1} &       &       & CB0   & 9.57\cdot10^{-1} & 6.67\cdot10^{-3} & 4.82\cdot10^{-1} & 1.65\cdot10^{-1} \\*
          & CB1   & 1.000    & 0.000    & 5.00\cdot10^{-1} & 1.25\cdot10^{-1} &       &       & CB1   & 9.97\cdot10^{-1} & 3.33\cdot10^{-3} & 5.00\cdot10^{-1} & 1.69\cdot10^{-1} \\*
          & CB2   & 1.000    & 0.000    & 5.00\cdot10^{-1} & 1.25\cdot10^{-1} &       &       & CB2   & 1.000    & 0.000    & 5.00\cdot10^{-1} & 1.67\cdot10^{-1} \\*
          & AC1   & 7.70\cdot10^{-1} & 5.67\cdot10^{-2} & 4.13\cdot10^{-1} & 1.46\cdot10^{-1} &       &       & AC1   & 6.80\cdot10^{-1} & 8.33\cdot10^{-2} & 3.82\cdot10^{-1} & 1.83\cdot10^{-1} \\*
          & AC2   & 1.000    & 0.000    & 5.00\cdot10^{-1} & 1.25\cdot10^{-1} &       &       & AC2   & 9.97\cdot10^{-1} & 0.000    & 4.98\cdot10^{-1} & 1.66\cdot10^{-1} \\*
          & AC3   & 1.000    & 0.000    & 5.00\cdot10^{-1} & 1.25\cdot10^{-1} &       &       & AC3   & 1.000    & 0.000    & 5.00\cdot10^{-1} & 1.67\cdot10^{-1} \\*
          & CSA   & 4.70\cdot10^{-1} & 2.07\cdot10^{-1} & 3.38\cdot10^{-1} & 2.40\cdot10^{-1} &       &       & CSA   & 4.50\cdot10^{-1} & 2.20\cdot10^{-1} & 3.35\cdot10^{-1} & 2.58\cdot10^{-1} \\*
          & CGA   & 6.47\cdot10^{-1} & 1.07\cdot10^{-1} & 3.77\cdot10^{-1} & 1.74\cdot10^{-1} &       &       & CGA   & 5.63\cdot10^{-1} & 1.50\cdot10^{-1} & 3.57\cdot10^{-1} & 2.19\cdot10^{-1} \\*
     \cmidrule(r){1-6} \cmidrule(r){8-13} 
		$[1, 3]$ & ABT   & 5.17\cdot10^{-1} & 1.87\cdot10^{-1} & 3.52\cdot10^{-1} & 2.69\cdot10^{-1} &       & $[1, 2]$ & ABT   & 5.10\cdot10^{-1} & 1.90\cdot10^{-1} & 3.50\cdot10^{-1} & 2.97\cdot10^{-1} \\*
          & ASB   & 4.90\cdot10^{-1} & 1.83\cdot10^{-1} & 3.37\cdot10^{-1} & 2.60\cdot10^{-1} &       &       & ASB   & 4.40\cdot10^{-1} & 2.50\cdot10^{-1} & 3.45\cdot10^{-1} & 3.13\cdot10^{-1} \\*
          & ADC   & 1.43\cdot10^{-1} & 9.50\cdot10^{-1} & 5.47\cdot10^{-1} & 7.48\cdot10^{-1} &       &       & ADC   & 1.43\cdot10^{-1} & 9.50\cdot10^{-1} & 5.47\cdot10^{-1} & 6.81\cdot10^{-1} \\*
          & CB0   & 9.67\cdot10^{-1} & 3.33\cdot10^{-3} & 4.85\cdot10^{-1} & 2.44\cdot10^{-1} &       &       & CB0   & 9.67\cdot10^{-1} & 3.33\cdot10^{-3} & 4.85\cdot10^{-1} & 3.24\cdot10^{-1} \\*
          & CB1   & 9.93\cdot10^{-1} & 3.33\cdot10^{-3} & 4.98\cdot10^{-1} & 2.51\cdot10^{-1} &       &       & CB1   & 1.000    & 0.000    & 5.00\cdot10^{-1} & 3.33\cdot10^{-1} \\*
          & CB2   & 1.000    & 0.000    & 5.00\cdot10^{-1} & 2.50\cdot10^{-1} &       &       & CB2   & 1.000    & 0.000    & 5.00\cdot10^{-1} & 3.33\cdot10^{-1} \\*
          & AC1   & 5.27\cdot10^{-1} & 1.77\cdot10^{-1} & 3.52\cdot10^{-1} & 2.64\cdot10^{-1} &       &       & AC1   & 4.03\cdot10^{-1} & 2.27\cdot10^{-1} & 3.15\cdot10^{-1} & 2.86\cdot10^{-1} \\*
          & AC2   & 1.000    & 0.000    & 5.00\cdot10^{-1} & 2.50\cdot10^{-1} &       &       & AC2   & 1.000    & 0.000    & 5.00\cdot10^{-1} & 3.33\cdot10^{-1} \\*
          & AC3   & 1.000    & 0.000    & 5.00\cdot10^{-1} & 2.50\cdot10^{-1} &       &       & AC3   & 1.000    & 0.000    & 5.00\cdot10^{-1} & 3.33\cdot10^{-1} \\*
          & CSA   & 4.07\cdot10^{-1} & 2.70\cdot10^{-1} & 3.38\cdot10^{-1} & 3.04\cdot10^{-1} &       &       & CSA   & 3.77\cdot10^{-1} & 3.03\cdot10^{-1} & 3.40\cdot10^{-1} & 3.28\cdot10^{-1} \\*
          & CGA   & 4.83\cdot10^{-1} & 1.90\cdot10^{-1} & 3.37\cdot10^{-1} & 2.63\cdot10^{-1} &       &       & CGA   & 4.33\cdot10^{-1} & 2.63\cdot10^{-1} & 3.48\cdot10^{-1} & 3.20\cdot10^{-1} \\*
     \cmidrule(r){1-6} \cmidrule(r){8-13} 
		$[2, 3]$ & ABT   & 3.13\cdot10^{-1} & 3.47\cdot10^{-1} & 3.30\cdot10^{-1} & 3.33\cdot10^{-1} &       & $[1, 1]$ & ABT   & 2.77\cdot10^{-1} & 3.80\cdot10^{-1} & 3.28\cdot10^{-1} & 3.28\cdot10^{-1} \\*
          & ASB   & 3.60\cdot10^{-1} & 3.03\cdot10^{-1} & 3.32\cdot10^{-1} & 3.26\cdot10^{-1} &       &       & ASB   & 2.93\cdot10^{-1} & 3.60\cdot10^{-1} & 3.27\cdot10^{-1} & 3.27\cdot10^{-1} \\*
          & ADC   & 5.37\cdot10^{-1} & 7.17\cdot10^{-1} & 6.27\cdot10^{-1} & 6.45\cdot10^{-1} &       &       & ADC   & 8.00\cdot10^{-1} & 5.50\cdot10^{-1} & 6.75\cdot10^{-1} & 6.75\cdot10^{-1} \\*
          & CB0   & 9.67\cdot10^{-1} & 6.67\cdot10^{-3} & 4.87\cdot10^{-1} & 3.91\cdot10^{-1} &       &       & CB0   & 2.00\cdot10^{-1} & 4.50\cdot10^{-1} & 3.25\cdot10^{-1} & 3.25\cdot10^{-1} \\*
          & CB1   & 9.93\cdot10^{-1} & 3.33\cdot10^{-3} & 4.98\cdot10^{-1} & 3.99\cdot10^{-1} &       &       & CB1   & 8.00\cdot10^{-1} & 5.50\cdot10^{-1} & 6.75\cdot10^{-1} & 6.75\cdot10^{-1} \\*
          & CB2   & 9.97\cdot10^{-1} & 0.000    & 4.98\cdot10^{-1} & 3.99\cdot10^{-1} &       &       & CB2   & 2.93\cdot10^{-1} & 3.60\cdot10^{-1} & 3.27\cdot10^{-1} & 3.27\cdot10^{-1} \\*
          & AC1   & 3.47\cdot10^{-1} & 2.73\cdot10^{-1} & 3.10\cdot10^{-1} & 3.03\cdot10^{-1} &       &       & AC1   & 2.83\cdot10^{-1} & 3.50\cdot10^{-1} & 3.17\cdot10^{-1} & 3.17\cdot10^{-1} \\*
          & AC2   & 1.000    & 0.000    & 5.00\cdot10^{-1} & 4.00\cdot10^{-1} &       &       & AC2   & 2.93\cdot10^{-1} & 3.60\cdot10^{-1} & 3.27\cdot10^{-1} & 3.27\cdot10^{-1} \\*
          & AC3   & 9.83\cdot10^{-1} & 3.33\cdot10^{-3} & 4.93\cdot10^{-1} & 3.95\cdot10^{-1} &       &       & AC3   & 2.83\cdot10^{-1} & 3.50\cdot10^{-1} & 3.17\cdot10^{-1} & 3.17\cdot10^{-1} \\*
          & CSA   & 3.30\cdot10^{-1} & 3.27\cdot10^{-1} & 3.28\cdot10^{-1} & 3.28\cdot10^{-1} &       &       & CSA   & 2.93\cdot10^{-1} & 3.60\cdot10^{-1} & 3.27\cdot10^{-1} & 3.27\cdot10^{-1} \\*
          & CGA   & 3.77\cdot10^{-1} & 3.13\cdot10^{-1} & 3.45\cdot10^{-1} & 3.39\cdot10^{-1} &       &       & CGA   & 2.93\cdot10^{-1} & 3.60\cdot10^{-1} & 3.27\cdot10^{-1} & 3.27\cdot10^{-1} \\*
     \cmidrule(r){1-6} \cmidrule(r){8-13} 
		
		$[3, 2]$ & ABT   & 2.40\cdot10^{-1} & 4.53\cdot10^{-1} & 3.47\cdot10^{-1} & 3.25\cdot10^{-1} &       & $[2, 1]$ & ABT   & 2.33\cdot10^{-1} & 4.67\cdot10^{-1} & 3.50\cdot10^{-1} & 3.11\cdot10^{-1} \\*
          & ASB   & 2.57\cdot10^{-1} & 4.37\cdot10^{-1} & 3.47\cdot10^{-1} & 3.29\cdot10^{-1} &       &       & ASB   & 2.27\cdot10^{-1} & 4.80\cdot10^{-1} & 3.53\cdot10^{-1} & 3.11\cdot10^{-1} \\*
          & ADC   & 8.13\cdot10^{-1} & 5.20\cdot10^{-1} & 6.67\cdot10^{-1} & 6.96\cdot10^{-1} &       &       & ADC   & 8.27\cdot10^{-1} & 4.97\cdot10^{-1} & 6.62\cdot10^{-1} & 7.17\cdot10^{-1} \\*
          & CB0   & 0.000    & 9.93\cdot10^{-1} & 4.97\cdot10^{-1} & 3.97\cdot10^{-1} &       &       & CB0   & 0.000    & 9.97\cdot10^{-1} & 4.98\cdot10^{-1} & 3.32\cdot10^{-1} \\*
          & CB1   & 0.000    & 9.97\cdot10^{-1} & 4.98\cdot10^{-1} & 3.99\cdot10^{-1} &       &       & CB1   & 0.000    & 1.000    & 5.00\cdot10^{-1} & 3.33\cdot10^{-1} \\*
          & CB2   & 0.000    & 9.97\cdot10^{-1} & 4.98\cdot10^{-1} & 3.99\cdot10^{-1} &       &       & CB2   & 0.000    & 1.000    & 5.00\cdot10^{-1} & 3.33\cdot10^{-1} \\*
          & AC1   & 2.20\cdot10^{-1} & 3.90\cdot10^{-1} & 3.05\cdot10^{-1} & 2.88\cdot10^{-1} &       &       & AC1   & 1.67\cdot10^{-1} & 4.60\cdot10^{-1} & 3.13\cdot10^{-1} & 2.64\cdot10^{-1} \\*
          & AC2   & 0.000    & 9.93\cdot10^{-1} & 4.97\cdot10^{-1} & 3.97\cdot10^{-1} &       &       & AC2   & 0.000    & 1.000    & 5.00\cdot10^{-1} & 3.33\cdot10^{-1} \\*
          & AC3   & 0.000    & 9.97\cdot10^{-1} & 4.98\cdot10^{-1} & 3.99\cdot10^{-1} &       &       & AC3   & 0.000    & 9.97\cdot10^{-1} & 4.98\cdot10^{-1} & 3.32\cdot10^{-1} \\*
          & CSA   & 2.67\cdot10^{-1} & 4.13\cdot10^{-1} & 3.40\cdot10^{-1} & 3.25\cdot10^{-1} &       &       & CSA   & 2.57\cdot10^{-1} & 4.40\cdot10^{-1} & 3.48\cdot10^{-1} & 3.18\cdot10^{-1} \\*
          & CGA   & 2.57\cdot10^{-1} & 4.33\cdot10^{-1} & 3.45\cdot10^{-1} & 3.27\cdot10^{-1} &       &       & CGA   & 2.20\cdot10^{-1} & 4.77\cdot10^{-1} & 3.48\cdot10^{-1} & 3.06\cdot10^{-1} \\*
     \cmidrule(r){1-6} \cmidrule(r){8-13} 
		$[3, 1]$ & ABT   & 2.30\cdot10^{-1} & 4.67\cdot10^{-1} & 3.48\cdot10^{-1} & 2.89\cdot10^{-1} &       & $[5, 1]$ & ABT   & 1.30\cdot10^{-1} & 5.87\cdot10^{-1} & 3.58\cdot10^{-1} & 2.06\cdot10^{-1} \\*
          & ASB   & 1.70\cdot10^{-1} & 5.30\cdot10^{-1} & 3.50\cdot10^{-1} & 2.60\cdot10^{-1} &       &       & ASB   & 1.20\cdot10^{-1} & 5.87\cdot10^{-1} & 3.53\cdot10^{-1} & 1.98\cdot10^{-1} \\*
          & ADC   & 9.43\cdot10^{-1} & 1.53\cdot10^{-1} & 5.48\cdot10^{-1} & 7.46\cdot10^{-1} &       &       & ADC   & 1.000    & 6.67\cdot10^{-3} & 5.03\cdot10^{-1} & 8.34\cdot10^{-1} \\*
          & CB0   & 0.000    & 9.93\cdot10^{-1} & 4.97\cdot10^{-1} & 2.48\cdot10^{-1} &       &       & CB0   & 0.000    & 9.93\cdot10^{-1} & 4.97\cdot10^{-1} & 1.66\cdot10^{-1} \\*
          & CB1   & 0.000    & 1.000    & 5.00\cdot10^{-1} & 2.50\cdot10^{-1} &       &       & CB1   & 0.000    & 9.97\cdot10^{-1} & 4.98\cdot10^{-1} & 1.66\cdot10^{-1} \\*
          & CB2   & 0.000    & 1.000    & 5.00\cdot10^{-1} & 2.50\cdot10^{-1} &       &       & CB2   & 0.000    & 9.97\cdot10^{-1} & 4.98\cdot10^{-1} & 1.66\cdot10^{-1} \\*
          & AC1   & 1.23\cdot10^{-1} & 5.17\cdot10^{-1} & 3.20\cdot10^{-1} & 2.22\cdot10^{-1} &       &       & AC1   & 8.33\cdot10^{-2} & 6.20\cdot10^{-1} & 3.52\cdot10^{-1} & 1.73\cdot10^{-1} \\*
          & AC2   & 0.000    & 1.000    & 5.00\cdot10^{-1} & 2.50\cdot10^{-1} &       &       & AC2   & 0.000    & 9.97\cdot10^{-1} & 4.98\cdot10^{-1} & 1.66\cdot10^{-1} \\*
          & AC3   & 0.000    & 9.97\cdot10^{-1} & 4.98\cdot10^{-1} & 2.49\cdot10^{-1} &       &       & AC3   & 0.000    & 9.97\cdot10^{-1} & 4.98\cdot10^{-1} & 1.66\cdot10^{-1} \\*
          & CSA   & 2.37\cdot10^{-1} & 4.70\cdot10^{-1} & 3.53\cdot10^{-1} & 2.95\cdot10^{-1} &       &       & CSA   & 2.03\cdot10^{-1} & 4.97\cdot10^{-1} & 3.50\cdot10^{-1} & 2.52\cdot10^{-1} \\*
          & CGA   & 1.70\cdot10^{-1} & 5.20\cdot10^{-1} & 3.45\cdot10^{-1} & 2.58\cdot10^{-1} &       &       & CGA   & 1.13\cdot10^{-1} & 5.87\cdot10^{-1} & 3.50\cdot10^{-1} & 1.92\cdot10^{-1} \\*
     \cmidrule(r){1-6} \cmidrule(r){8-13} 
		$[7, 1]$ & ABT   & 1.03\cdot10^{-1} & 6.33\cdot10^{-1} & 3.68\cdot10^{-1} & 1.70\cdot10^{-1} &       & $[10, 1]$ & ABT   & 8.00\cdot10^{-2} & 6.93\cdot10^{-1} & 3.87\cdot10^{-1} & 1.36\cdot10^{-1} \\*
          & ASB   & 1.00\cdot10^{-1} & 6.33\cdot10^{-1} & 3.67\cdot10^{-1} & 1.67\cdot10^{-1} &       &       & ASB   & 8.33\cdot10^{-2} & 6.97\cdot10^{-1} & 3.90\cdot10^{-1} & 1.39\cdot10^{-1} \\*
          & ADC   & 1.000    & 6.67\cdot10^{-3} & 5.03\cdot10^{-1} & 8.76\cdot10^{-1} &       &       & ADC   & 1.000    & 6.67\cdot10^{-3} & 5.03\cdot10^{-1} & 9.10\cdot10^{-1} \\*
          & CB0   & 0.000    & 9.93\cdot10^{-1} & 4.97\cdot10^{-1} & 1.24\cdot10^{-1} &       &       & CB0   & 0.000    & 9.93\cdot10^{-1} & 4.97\cdot10^{-1} & 9.03\cdot10^{-2} \\*
          & CB1   & 0.000    & 9.97\cdot10^{-1} & 4.98\cdot10^{-1} & 1.25\cdot10^{-1} &       &       & CB1   & 0.000    & 9.93\cdot10^{-1} & 4.97\cdot10^{-1} & 9.03\cdot10^{-2} \\*
          & CB2   & 0.000    & 9.97\cdot10^{-1} & 4.98\cdot10^{-1} & 1.25\cdot10^{-1} &       &       & CB2   & 0.000    & 9.93\cdot10^{-1} & 4.97\cdot10^{-1} & 9.03\cdot10^{-2} \\*
          & AC1   & 5.00\cdot10^{-2} & 7.03\cdot10^{-1} & 3.77\cdot10^{-1} & 1.32\cdot10^{-1} &       &       & AC1   & 2.33\cdot10^{-2} & 7.80\cdot10^{-1} & 4.02\cdot10^{-1} & 9.21\cdot10^{-2} \\*
          & AC2   & 0.000    & 9.93\cdot10^{-1} & 4.97\cdot10^{-1} & 1.24\cdot10^{-1} &       &       & AC2   & 0.000    & 9.97\cdot10^{-1} & 4.98\cdot10^{-1} & 9.06\cdot10^{-2} \\*
          & AC3   & 0.000    & 9.97\cdot10^{-1} & 4.98\cdot10^{-1} & 1.25\cdot10^{-1} &       &       & AC3   & 0.000    & 9.97\cdot10^{-1} & 4.98\cdot10^{-1} & 9.06\cdot10^{-2} \\*
          & CSA   & 1.97\cdot10^{-1} & 4.97\cdot10^{-1} & 3.47\cdot10^{-1} & 2.34\cdot10^{-1} &       &       & CSA   & 1.77\cdot10^{-1} & 5.23\cdot10^{-1} & 3.50\cdot10^{-1} & 2.08\cdot10^{-1} \\*
          & CGA   & 1.07\cdot10^{-1} & 6.37\cdot10^{-1} & 3.72\cdot10^{-1} & 1.73\cdot10^{-1} &       &       & CGA   & 7.67\cdot10^{-2} & 6.73\cdot10^{-1} & 3.75\cdot10^{-1} & 1.31\cdot10^{-1} \\*
     \cmidrule(r){1-6} \cmidrule(r){8-13} 
		$[25, 1]$ & ABT   & 6.67\cdot10^{-2} & 6.93\cdot10^{-1} & 3.80\cdot10^{-1} & 9.08\cdot10^{-2} &       & $[50, 1]$ & ABT   & 4.00\cdot10^{-2} & 7.50\cdot10^{-1} & 3.95\cdot10^{-1} & 5.39\cdot10^{-2} \\*
          & ASB   & 4.00\cdot10^{-2} & 7.63\cdot10^{-1} & 4.02\cdot10^{-1} & 6.78\cdot10^{-2} &       &       & ASB   & 3.33\cdot10^{-2} & 7.97\cdot10^{-1} & 4.15\cdot10^{-1} & 4.83\cdot10^{-2} \\*
          & ADC   & 1.000    & 6.67\cdot10^{-3} & 5.03\cdot10^{-1} & 9.62\cdot10^{-1} &       &       & ADC   & 1.000    & 6.67\cdot10^{-3} & 5.03\cdot10^{-1} & 9.81\cdot10^{-1} \\*
          & CB0   & 0.000    & 9.93\cdot10^{-1} & 4.97\cdot10^{-1} & 3.82\cdot10^{-2} &       &       & CB0   & 0.000    & 9.93\cdot10^{-1} & 4.97\cdot10^{-1} & 1.95\cdot10^{-2} \\*
          & CB1   & 0.000    & 9.93\cdot10^{-1} & 4.97\cdot10^{-1} & 3.82\cdot10^{-2} &       &       & CB1   & 0.000    & 9.93\cdot10^{-1} & 4.97\cdot10^{-1} & 1.95\cdot10^{-2} \\*
          & CB2   & 0.000    & 9.93\cdot10^{-1} & 4.97\cdot10^{-1} & 3.82\cdot10^{-2} &       &       & CB2   & 0.000    & 9.93\cdot10^{-1} & 4.97\cdot10^{-1} & 1.95\cdot10^{-2} \\*
          & AC1   & 0.000    & 9.57\cdot10^{-1} & 4.78\cdot10^{-1} & 3.68\cdot10^{-2} &       &       & AC1   & 0.000    & 9.97\cdot10^{-1} & 4.98\cdot10^{-1} & 1.95\cdot10^{-2} \\*
          & AC2   & 0.000    & 9.97\cdot10^{-1} & 4.98\cdot10^{-1} & 3.83\cdot10^{-2} &       &       & AC2   & 0.000    & 9.97\cdot10^{-1} & 4.98\cdot10^{-1} & 1.95\cdot10^{-2} \\*
          & AC3   & 0.000    & 9.93\cdot10^{-1} & 4.97\cdot10^{-1} & 3.82\cdot10^{-2} &       &       & AC3   & 0.000    & 9.93\cdot10^{-1} & 4.97\cdot10^{-1} & 1.95\cdot10^{-2} \\*
          & CSA   & 1.20\cdot10^{-1} & 5.63\cdot10^{-1} & 3.42\cdot10^{-1} & 1.37\cdot10^{-1} &       &       & CSA   & 9.33\cdot10^{-2} & 6.20\cdot10^{-1} & 3.57\cdot10^{-1} & 1.04\cdot10^{-1} \\*
          & CGA   & 2.00\cdot10^{-2} & 8.53\cdot10^{-1} & 4.37\cdot10^{-1} & 5.21\cdot10^{-2} &       &       & CGA   & 0.000    & 9.63\cdot10^{-1} & 4.82\cdot10^{-1} & 1.89\cdot10^{-2} \\*
     \cmidrule(r){1-6} \cmidrule(r){8-13} 
		$[100, 1]$ & ABT   & 3.00\cdot10^{-2} & 8.13\cdot10^{-1} & 4.22\cdot10^{-1} & 3.78\cdot10^{-2} &       &       &       &       &       &       &  \\*
          & ASB   & 4.00\cdot10^{-2} & 7.60\cdot10^{-1} & 4.00\cdot10^{-1} & 4.71\cdot10^{-2} &       &       &       &       &       &       &  \\*
          & ADC   & 3.33\cdot10^{-1} & 6.67\cdot10^{-1} & 5.00\cdot10^{-1} & 3.37\cdot10^{-1} &       &       &       &       &       &       &  \\*
          & CB0   & 0.000    & 9.93\cdot10^{-1} & 4.97\cdot10^{-1} & 9.83\cdot10^{-3} &       &       &       &       &       &       &  \\*
          & CB1   & 0.000    & 9.93\cdot10^{-1} & 4.97\cdot10^{-1} & 9.83\cdot10^{-3} &       &       &       &       &       &       &  \\*
          & CB2   & 0.000    & 9.93\cdot10^{-1} & 4.97\cdot10^{-1} & 9.83\cdot10^{-3} &       &       &       &       &       &       &  \\*
          & AC1   & 0.000    & 9.93\cdot10^{-1} & 4.97\cdot10^{-1} & 9.83\cdot10^{-3} &       &       &       &       &       &       &  \\*
          & AC2   & 0.000    & 9.93\cdot10^{-1} & 4.97\cdot10^{-1} & 9.83\cdot10^{-3} &       &       &       &       &       &       &  \\*
          & AC3   & 0.000    & 9.93\cdot10^{-1} & 4.97\cdot10^{-1} & 9.83\cdot10^{-3} &       &       &       &       &       &       &  \\*
          & CSA   & 5.67\cdot10^{-2} & 7.60\cdot10^{-1} & 4.08\cdot10^{-1} & 6.36\cdot10^{-2} &       &       &       &       &       &       &  \\*
          & CGA   & 0.000    & 9.93\cdot10^{-1} & 4.97\cdot10^{-1} & 9.83\cdot10^{-3} &       &       &       &       &       &       &  \\*

\end{longtabu}
}

\newpage

%%%%%%%%%%%%%%%%%%
%%% UCI CREDIT %%%
%%%%%%%%%%%%%%%%%%

\begin{landscape}
% GRAPHICS
\begin{figure}[p]
\centering
\includegraphics[width=.6\paperheight]{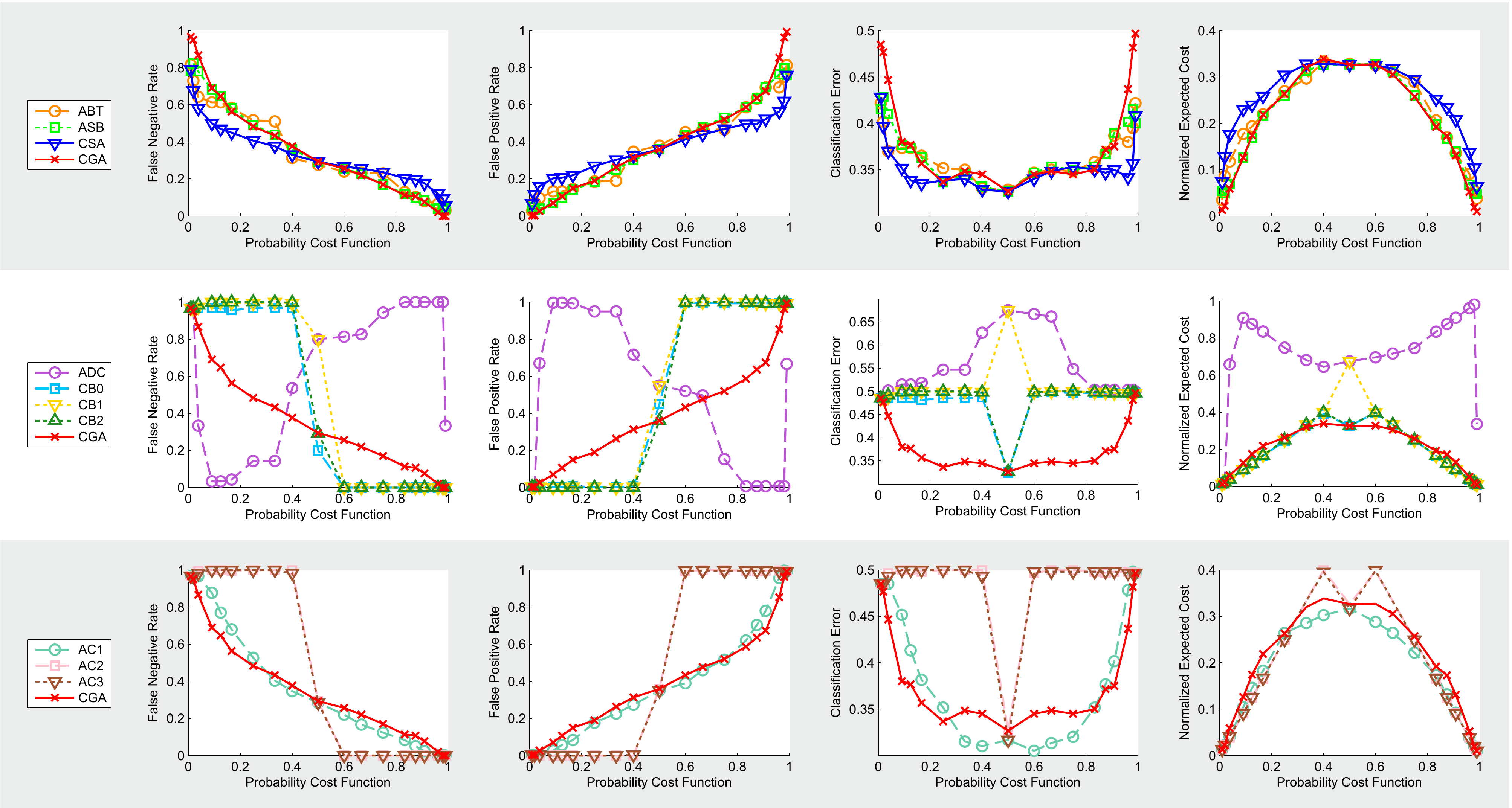}
\caption[Results obtained for the UCI Credit Dataset.]{Results obtained for the UCI Credit Dataset. First column of the illustration corresponds to False Negative Rate, second column to False Positive Rate, third column to Classification Error and the fourth one corresponds to Normalized Expected Cost. For a clearer visualization, algorithms have been divided into three groups so each row of the illustration corresponds to a different group. Cost Generalized AdaBoost is plotted in all the graphs to have a common reference across the representations.}
\label{fig:credit_perform} % caption for the whole figure
\end{figure}
\end{landscape}
\newpage

%%%%%%%%%%%%%%%%%%%%
%%% UCI DIABETES %%%
%%%%%%%%%%%%%%%%%%%%

% TABLE

{
\tiny
\begin{longtabu} to \textwidth {llX[$l]@{}@{}X[$l]@{}@{}X[$l]@{}@{}X[$l]@{}@{}l@{}llX[$l]@{}@{}X[$l]@{}@{}X[$l]@{}@{}X[$l]@{}}

\toprule
{\textbf{Cost}}  & {\textbf{Alg}}   & {\textbf{FNR}}    & {\textbf{FPR}}    & {\textbf{CE}}    & {\textbf{NEC}}   && {\textbf{Cost}}  & {\textbf{Alg}}   & {\textbf{FNR}}    & {\textbf{FPR}}    & {\textbf{CE}} & {\textbf{NEC}}\\* 
\cmidrule(r){1-6} \cmidrule(r){8-13}
\endfirsthead

%\multicolumn{13}{l}% {\tiny{Continued from previous page}} \\*
\toprule
{\textbf{Cost}}  & {\textbf{Alg}}   & {\textbf{FNR}}    & {\textbf{FPR}}    & {\textbf{CE}}    & {\textbf{NEC}}   && {\textbf{Cost}}  & {\textbf{Alg}}   & {\textbf{FNR}}    & {\textbf{FPR}}    & {\textbf{CE}} & {\textbf{NEC}}\\* 
\cmidrule(r){1-6} \cmidrule(r){8-13}
\endhead

\multicolumn{13}{r}{{\tiny{\tablename\ \thetable{} - Continued on next page $\rightarrow$}}} \\* 
\endfoot

\bottomrule
  \caption{Results obtained for the UCI Diabetes Dataset.}
	\label{tab:diabetes_perform}%
\endlastfoot

    $[1, 100]$ & ABT   & 5.54\cdot10^{-1} & 1.16\cdot10^{-1} & 3.35\cdot10^{-1} & 1.20\cdot10^{-1} &       & $[1, 50]$ & ABT   & 4.98\cdot10^{-1} & 1.57\cdot10^{-1} & 3.28\cdot10^{-1} & 1.64\cdot10^{-1} \\*
          & ASB   & 5.92\cdot10^{-1} & 9.74\cdot10^{-2} & 3.45\cdot10^{-1} & 1.02\cdot10^{-1} &       &       & ASB   & 6.14\cdot10^{-1} & 9.36\cdot10^{-2} & 3.54\cdot10^{-1} & 1.04\cdot10^{-1} \\*
          & ADC   & 9.55\cdot10^{-1} & 1.50\cdot10^{-2} & 4.85\cdot10^{-1} & 2.43\cdot10^{-2} &       &       & ADC   & 6.29\cdot10^{-1} & 3.41\cdot10^{-1} & 4.85\cdot10^{-1} & 3.46\cdot10^{-1} \\*
          & CB0   & 9.55\cdot10^{-1} & 1.50\cdot10^{-2} & 4.85\cdot10^{-1} & 2.43\cdot10^{-2} &       &       & CB0   & 9.55\cdot10^{-1} & 1.50\cdot10^{-2} & 4.85\cdot10^{-1} & 3.34\cdot10^{-2} \\*
          & CB1   & 9.55\cdot10^{-1} & 1.50\cdot10^{-2} & 4.85\cdot10^{-1} & 2.43\cdot10^{-2} &       &       & CB1   & 9.59\cdot10^{-1} & 1.12\cdot10^{-2} & 4.85\cdot10^{-1} & 2.98\cdot10^{-2} \\*
          & CB2   & 9.55\cdot10^{-1} & 1.50\cdot10^{-2} & 4.85\cdot10^{-1} & 2.43\cdot10^{-2} &       &       & CB2   & 1.000    & 0.000    & 5.00\cdot10^{-1} & 1.96\cdot10^{-2} \\*
          & AC1   & 9.55\cdot10^{-1} & 1.50\cdot10^{-2} & 4.85\cdot10^{-1} & 2.43\cdot10^{-2} &       &       & AC1   & 9.55\cdot10^{-1} & 1.50\cdot10^{-2} & 4.85\cdot10^{-1} & 3.34\cdot10^{-2} \\*
          & AC2   & 9.55\cdot10^{-1} & 1.50\cdot10^{-2} & 4.85\cdot10^{-1} & 2.43\cdot10^{-2} &       &       & AC2   & 9.55\cdot10^{-1} & 1.50\cdot10^{-2} & 4.85\cdot10^{-1} & 3.34\cdot10^{-2} \\*
          & AC3   & 9.55\cdot10^{-1} & 1.50\cdot10^{-2} & 4.85\cdot10^{-1} & 2.43\cdot10^{-2} &       &       & AC3   & 9.55\cdot10^{-1} & 1.50\cdot10^{-2} & 4.85\cdot10^{-1} & 3.34\cdot10^{-2} \\*
          & CSA   & 6.03\cdot10^{-1} & 8.99\cdot10^{-2} & 3.46\cdot10^{-1} & 9.50\cdot10^{-2} &       &       & CSA   & 4.64\cdot10^{-1} & 1.54\cdot10^{-1} & 3.09\cdot10^{-1} & 1.60\cdot10^{-1} \\*
          & CGA   & 9.03\cdot10^{-1} & 1.50\cdot10^{-2} & 4.59\cdot10^{-1} & 2.38\cdot10^{-2} &       &       & CGA   & 7.53\cdot10^{-1} & 4.12\cdot10^{-2} & 3.97\cdot10^{-1} & 5.52\cdot10^{-2} \\*
     \cmidrule(r){1-6} \cmidrule(r){8-13} 
		$[1, 25]$ & ABT   & 4.31\cdot10^{-1} & 1.95\cdot10^{-1} & 3.13\cdot10^{-1} & 2.04\cdot10^{-1} &       & $[1, 10]$ & ABT   & 3.97\cdot10^{-1} & 2.21\cdot10^{-1} & 3.09\cdot10^{-1} & 2.37\cdot10^{-1} \\*
          & ASB   & 6.03\cdot10^{-1} & 1.05\cdot10^{-1} & 3.54\cdot10^{-1} & 1.24\cdot10^{-1} &       &       & ASB   & 5.47\cdot10^{-1} & 1.46\cdot10^{-1} & 3.46\cdot10^{-1} & 1.82\cdot10^{-1} \\*
          & ADC   & 2.96\cdot10^{-1} & 6.74\cdot10^{-1} & 4.85\cdot10^{-1} & 6.60\cdot10^{-1} &       &       & ADC   & 2.51\cdot10^{-1} & 6.78\cdot10^{-1} & 4.64\cdot10^{-1} & 6.39\cdot10^{-1} \\*
          & CB0   & 9.55\cdot10^{-1} & 1.50\cdot10^{-2} & 4.85\cdot10^{-1} & 5.11\cdot10^{-2} &       &       & CB0   & 9.10\cdot10^{-1} & 1.87\cdot10^{-2} & 4.64\cdot10^{-1} & 9.98\cdot10^{-2} \\*
          & CB1   & 9.78\cdot10^{-1} & 0.000    & 4.89\cdot10^{-1} & 3.76\cdot10^{-2} &       &       & CB1   & 1.000    & 0.000    & 5.00\cdot10^{-1} & 9.09\cdot10^{-2} \\*
          & CB2   & 1.000    & 0.000    & 5.00\cdot10^{-1} & 3.85\cdot10^{-2} &       &       & CB2   & 1.000    & 0.000    & 5.00\cdot10^{-1} & 9.09\cdot10^{-2} \\*
          & AC1   & 8.35\cdot10^{-1} & 2.25\cdot10^{-2} & 4.29\cdot10^{-1} & 5.37\cdot10^{-2} &       &       & AC1   & 6.40\cdot10^{-1} & 5.24\cdot10^{-2} & 3.46\cdot10^{-1} & 1.06\cdot10^{-1} \\*
          & AC2   & 1.000    & 0.000    & 5.00\cdot10^{-1} & 3.85\cdot10^{-2} &       &       & AC2   & 1.000    & 3.75\cdot10^{-3} & 5.02\cdot10^{-1} & 9.43\cdot10^{-2} \\*
          & AC3   & 1.000    & 0.000    & 5.00\cdot10^{-1} & 3.85\cdot10^{-2} &       &       & AC3   & 1.000    & 0.000    & 5.00\cdot10^{-1} & 9.09\cdot10^{-2} \\*
          & CSA   & 4.08\cdot10^{-1} & 1.69\cdot10^{-1} & 2.88\cdot10^{-1} & 1.78\cdot10^{-1} &       &       & CSA   & 3.71\cdot10^{-1} & 2.21\cdot10^{-1} & 2.96\cdot10^{-1} & 2.35\cdot10^{-1} \\*
          & CGA   & 6.48\cdot10^{-1} & 7.87\cdot10^{-2} & 3.63\cdot10^{-1} & 1.01\cdot10^{-1} &       &       & CGA   & 5.24\cdot10^{-1} & 1.42\cdot10^{-1} & 3.33\cdot10^{-1} & 1.77\cdot10^{-1} \\*
     \cmidrule(r){1-6} \cmidrule(r){8-13} 
		$[1, 7]$ & ABT   & 3.82\cdot10^{-1} & 2.21\cdot10^{-1} & 3.01\cdot10^{-1} & 2.41\cdot10^{-1} &       & $[1, 5]$ & ABT   & 3.67\cdot10^{-1} & 2.36\cdot10^{-1} & 3.01\cdot10^{-1} & 2.58\cdot10^{-1} \\*
          & ASB   & 5.13\cdot10^{-1} & 1.76\cdot10^{-1} & 3.45\cdot10^{-1} & 2.18\cdot10^{-1} &       &       & ASB   & 4.68\cdot10^{-1} & 1.72\cdot10^{-1} & 3.20\cdot10^{-1} & 2.22\cdot10^{-1} \\*
          & ADC   & 4.42\cdot10^{-1} & 6.59\cdot10^{-1} & 5.51\cdot10^{-1} & 6.32\cdot10^{-1} &       &       & ADC   & 4.42\cdot10^{-1} & 6.59\cdot10^{-1} & 5.51\cdot10^{-1} & 6.23\cdot10^{-1} \\*
          & CB0   & 9.55\cdot10^{-1} & 1.50\cdot10^{-2} & 4.85\cdot10^{-1} & 1.32\cdot10^{-1} &       &       & CB0   & 9.55\cdot10^{-1} & 1.50\cdot10^{-2} & 4.85\cdot10^{-1} & 1.72\cdot10^{-1} \\*
          & CB1   & 1.000    & 0.000    & 5.00\cdot10^{-1} & 1.25\cdot10^{-1} &       &       & CB1   & 1.000    & 0.000    & 5.00\cdot10^{-1} & 1.67\cdot10^{-1} \\*
          & CB2   & 9.93\cdot10^{-1} & 0.000    & 4.96\cdot10^{-1} & 1.24\cdot10^{-1} &       &       & CB2   & 9.85\cdot10^{-1} & 0.000    & 4.93\cdot10^{-1} & 1.64\cdot10^{-1} \\*
          & AC1   & 5.54\cdot10^{-1} & 8.99\cdot10^{-2} & 3.22\cdot10^{-1} & 1.48\cdot10^{-1} &       &       & AC1   & 5.02\cdot10^{-1} & 1.20\cdot10^{-1} & 3.11\cdot10^{-1} & 1.84\cdot10^{-1} \\*
          & AC2   & 1.000    & 3.75\cdot10^{-3} & 5.02\cdot10^{-1} & 1.28\cdot10^{-1} &       &       & AC2   & 9.96\cdot10^{-1} & 7.49\cdot10^{-3} & 5.02\cdot10^{-1} & 1.72\cdot10^{-1} \\*
          & AC3   & 1.000    & 0.000    & 5.00\cdot10^{-1} & 1.25\cdot10^{-1} &       &       & AC3   & 9.93\cdot10^{-1} & 0.000    & 4.96\cdot10^{-1} & 1.65\cdot10^{-1} \\*
          & CSA   & 3.56\cdot10^{-1} & 2.43\cdot10^{-1} & 3.00\cdot10^{-1} & 2.57\cdot10^{-1} &       &       & CSA   & 3.48\cdot10^{-1} & 2.43\cdot10^{-1} & 2.96\cdot10^{-1} & 2.61\cdot10^{-1} \\*
          & CGA   & 4.91\cdot10^{-1} & 1.69\cdot10^{-1} & 3.30\cdot10^{-1} & 2.09\cdot10^{-1} &       &       & CGA   & 4.64\cdot10^{-1} & 1.87\cdot10^{-1} & 3.26\cdot10^{-1} & 2.33\cdot10^{-1} \\*
     \cmidrule(r){1-6} \cmidrule(r){8-13} 
		$[1, 3]$ & ABT   & 3.63\cdot10^{-1} & 2.58\cdot10^{-1} & 3.11\cdot10^{-1} & 2.85\cdot10^{-1} &       & $[1, 2]$ & ABT   & 3.37\cdot10^{-1} & 2.73\cdot10^{-1} & 3.05\cdot10^{-1} & 2.95\cdot10^{-1} \\*
          & ASB   & 4.04\cdot10^{-1} & 2.10\cdot10^{-1} & 3.07\cdot10^{-1} & 2.58\cdot10^{-1} &       &       & ASB   & 3.67\cdot10^{-1} & 2.28\cdot10^{-1} & 2.98\cdot10^{-1} & 2.75\cdot10^{-1} \\*
          & ADC   & 4.61\cdot10^{-1} & 6.59\cdot10^{-1} & 5.60\cdot10^{-1} & 6.10\cdot10^{-1} &       &       & ADC   & 3.97\cdot10^{-1} & 9.14\cdot10^{-1} & 6.55\cdot10^{-1} & 7.42\cdot10^{-1} \\*
          & CB0   & 9.59\cdot10^{-1} & 1.12\cdot10^{-2} & 4.85\cdot10^{-1} & 2.48\cdot10^{-1} &       &       & CB0   & 9.55\cdot10^{-1} & 1.12\cdot10^{-2} & 4.83\cdot10^{-1} & 3.26\cdot10^{-1} \\*
          & CB1   & 1.000    & 0.000    & 5.00\cdot10^{-1} & 2.50\cdot10^{-1} &       &       & CB1   & 1.000    & 0.000    & 5.00\cdot10^{-1} & 3.33\cdot10^{-1} \\*
          & CB2   & 9.85\cdot10^{-1} & 0.000    & 4.93\cdot10^{-1} & 2.46\cdot10^{-1} &       &       & CB2   & 1.000    & 0.000    & 5.00\cdot10^{-1} & 3.33\cdot10^{-1} \\*
          & AC1   & 4.42\cdot10^{-1} & 1.69\cdot10^{-1} & 3.05\cdot10^{-1} & 2.37\cdot10^{-1} &       &       & AC1   & 3.78\cdot10^{-1} & 2.06\cdot10^{-1} & 2.92\cdot10^{-1} & 2.63\cdot10^{-1} \\*
          & AC2   & 1.000    & 3.75\cdot10^{-3} & 5.02\cdot10^{-1} & 2.53\cdot10^{-1} &       &       & AC2   & 1.000    & 0.000    & 5.00\cdot10^{-1} & 3.33\cdot10^{-1} \\*
          & AC3   & 9.93\cdot10^{-1} & 0.000    & 4.96\cdot10^{-1} & 2.48\cdot10^{-1} &       &       & AC3   & 1.000    & 3.75\cdot10^{-3} & 5.02\cdot10^{-1} & 3.36\cdot10^{-1} \\*
          & CSA   & 3.26\cdot10^{-1} & 2.66\cdot10^{-1} & 2.96\cdot10^{-1} & 2.81\cdot10^{-1} &       &       & CSA   & 3.07\cdot10^{-1} & 2.81\cdot10^{-1} & 2.94\cdot10^{-1} & 2.90\cdot10^{-1} \\*
          & CGA   & 4.08\cdot10^{-1} & 2.17\cdot10^{-1} & 3.13\cdot10^{-1} & 2.65\cdot10^{-1} &       &       & CGA   & 3.78\cdot10^{-1} & 2.43\cdot10^{-1} & 3.11\cdot10^{-1} & 2.88\cdot10^{-1} \\*
     \cmidrule(r){1-6} \cmidrule(r){8-13} 
		$[2, 3]$ & ABT   & 3.37\cdot10^{-1} & 2.73\cdot10^{-1} & 3.05\cdot10^{-1} & 2.99\cdot10^{-1} &       & $[1, 1]$ & ABT   & 3.22\cdot10^{-1} & 3.07\cdot10^{-1} & 3.15\cdot10^{-1} & 3.15\cdot10^{-1} \\*
          & ASB   & 3.41\cdot10^{-1} & 2.58\cdot10^{-1} & 3.00\cdot10^{-1} & 2.91\cdot10^{-1} &       &       & ASB   & 3.03\cdot10^{-1} & 3.11\cdot10^{-1} & 3.07\cdot10^{-1} & 3.07\cdot10^{-1} \\*
          & ADC   & 4.08\cdot10^{-1} & 8.99\cdot10^{-1} & 6.54\cdot10^{-1} & 7.03\cdot10^{-1} &       &       & ADC   & 7.53\cdot10^{-1} & 6.52\cdot10^{-1} & 7.02\cdot10^{-1} & 7.02\cdot10^{-1} \\*
          & CB0   & 9.55\cdot10^{-1} & 1.12\cdot10^{-2} & 4.83\cdot10^{-1} & 3.89\cdot10^{-1} &       &       & CB0   & 2.47\cdot10^{-1} & 3.48\cdot10^{-1} & 2.98\cdot10^{-1} & 2.98\cdot10^{-1} \\*
          & CB1   & 1.000    & 0.000    & 5.00\cdot10^{-1} & 4.00\cdot10^{-1} &       &       & CB1   & 7.45\cdot10^{-1} & 6.59\cdot10^{-1} & 7.02\cdot10^{-1} & 7.02\cdot10^{-1} \\*
          & CB2   & 1.000    & 0.000    & 5.00\cdot10^{-1} & 4.00\cdot10^{-1} &       &       & CB2   & 3.03\cdot10^{-1} & 3.11\cdot10^{-1} & 3.07\cdot10^{-1} & 3.07\cdot10^{-1} \\*
          & AC1   & 3.41\cdot10^{-1} & 2.40\cdot10^{-1} & 2.90\cdot10^{-1} & 2.80\cdot10^{-1} &       &       & AC1   & 2.88\cdot10^{-1} & 3.00\cdot10^{-1} & 2.94\cdot10^{-1} & 2.94\cdot10^{-1} \\*
          & AC2   & 1.000    & 3.75\cdot10^{-3} & 5.02\cdot10^{-1} & 4.02\cdot10^{-1} &       &       & AC2   & 3.03\cdot10^{-1} & 3.11\cdot10^{-1} & 3.07\cdot10^{-1} & 3.07\cdot10^{-1} \\*
          & AC3   & 9.96\cdot10^{-1} & 3.75\cdot10^{-3} & 5.00\cdot10^{-1} & 4.01\cdot10^{-1} &       &       & AC3   & 2.88\cdot10^{-1} & 3.00\cdot10^{-1} & 2.94\cdot10^{-1} & 2.94\cdot10^{-1} \\*
          & CSA   & 3.03\cdot10^{-1} & 3.07\cdot10^{-1} & 3.05\cdot10^{-1} & 3.06\cdot10^{-1} &       &       & CSA   & 3.03\cdot10^{-1} & 3.11\cdot10^{-1} & 3.07\cdot10^{-1} & 3.07\cdot10^{-1} \\*
          & CGA   & 3.45\cdot10^{-1} & 2.73\cdot10^{-1} & 3.09\cdot10^{-1} & 3.02\cdot10^{-1} &       &       & CGA   & 3.03\cdot10^{-1} & 3.11\cdot10^{-1} & 3.07\cdot10^{-1} & 3.07\cdot10^{-1} \\*
     \cmidrule(r){1-6} \cmidrule(r){8-13} 
		
		$[3, 2]$ & ABT   & 2.96\cdot10^{-1} & 3.33\cdot10^{-1} & 3.15\cdot10^{-1} & 3.11\cdot10^{-1} &       & $[2, 1]$ & ABT   & 2.70\cdot10^{-1} & 3.52\cdot10^{-1} & 3.11\cdot10^{-1} & 2.97\cdot10^{-1} \\*
          & ASB   & 2.66\cdot10^{-1} & 3.37\cdot10^{-1} & 3.01\cdot10^{-1} & 2.94\cdot10^{-1} &       &       & ASB   & 2.32\cdot10^{-1} & 3.60\cdot10^{-1} & 2.96\cdot10^{-1} & 2.75\cdot10^{-1} \\*
          & ADC   & 8.31\cdot10^{-1} & 5.36\cdot10^{-1} & 6.84\cdot10^{-1} & 7.13\cdot10^{-1} &       &       & ADC   & 8.43\cdot10^{-1} & 4.42\cdot10^{-1} & 6.42\cdot10^{-1} & 7.09\cdot10^{-1} \\*
          & CB0   & 0.000    & 9.96\cdot10^{-1} & 4.98\cdot10^{-1} & 3.99\cdot10^{-1} &       &       & CB0   & 3.75\cdot10^{-3} & 9.96\cdot10^{-1} & 5.00\cdot10^{-1} & 3.35\cdot10^{-1} \\*
          & CB1   & 0.000    & 1.000    & 5.00\cdot10^{-1} & 4.00\cdot10^{-1} &       &       & CB1   & 0.000    & 1.000    & 5.00\cdot10^{-1} & 3.33\cdot10^{-1} \\*
          & CB2   & 0.000    & 1.000    & 5.00\cdot10^{-1} & 4.00\cdot10^{-1} &       &       & CB2   & 0.000    & 9.96\cdot10^{-1} & 4.98\cdot10^{-1} & 3.32\cdot10^{-1} \\*
          & AC1   & 2.36\cdot10^{-1} & 3.52\cdot10^{-1} & 2.94\cdot10^{-1} & 2.82\cdot10^{-1} &       &       & AC1   & 2.13\cdot10^{-1} & 3.63\cdot10^{-1} & 2.88\cdot10^{-1} & 2.63\cdot10^{-1} \\*
          & AC2   & 0.000    & 1.000    & 5.00\cdot10^{-1} & 4.00\cdot10^{-1} &       &       & AC2   & 0.000    & 1.000    & 5.00\cdot10^{-1} & 3.33\cdot10^{-1} \\*
          & AC3   & 0.000    & 1.000    & 5.00\cdot10^{-1} & 4.00\cdot10^{-1} &       &       & AC3   & 0.000    & 1.000    & 5.00\cdot10^{-1} & 3.33\cdot10^{-1} \\*
          & CSA   & 2.88\cdot10^{-1} & 3.03\cdot10^{-1} & 2.96\cdot10^{-1} & 2.94\cdot10^{-1} &       &       & CSA   & 2.77\cdot10^{-1} & 3.30\cdot10^{-1} & 3.03\cdot10^{-1} & 2.95\cdot10^{-1} \\*
          & CGA   & 2.40\cdot10^{-1} & 3.30\cdot10^{-1} & 2.85\cdot10^{-1} & 2.76\cdot10^{-1} &       &       & CGA   & 2.25\cdot10^{-1} & 3.82\cdot10^{-1} & 3.03\cdot10^{-1} & 2.77\cdot10^{-1} \\*
     \cmidrule(r){1-6} \cmidrule(r){8-13} 
		$[3, 1]$ & ABT   & 2.43\cdot10^{-1} & 3.60\cdot10^{-1} & 3.01\cdot10^{-1} & 2.72\cdot10^{-1} &       & $[5, 1]$ & ABT   & 2.06\cdot10^{-1} & 3.97\cdot10^{-1} & 3.01\cdot10^{-1} & 2.38\cdot10^{-1} \\*
          & ASB   & 1.84\cdot10^{-1} & 4.12\cdot10^{-1} & 2.98\cdot10^{-1} & 2.41\cdot10^{-1} &       &       & ASB   & 1.65\cdot10^{-1} & 4.72\cdot10^{-1} & 3.18\cdot10^{-1} & 2.16\cdot10^{-1} \\*
          & ADC   & 9.21\cdot10^{-1} & 3.15\cdot10^{-1} & 6.18\cdot10^{-1} & 7.70\cdot10^{-1} &       &       & ADC   & 9.36\cdot10^{-1} & 2.58\cdot10^{-1} & 5.97\cdot10^{-1} & 8.23\cdot10^{-1} \\*
          & CB0   & 3.75\cdot10^{-3} & 9.96\cdot10^{-1} & 5.00\cdot10^{-1} & 2.52\cdot10^{-1} &       &       & CB0   & 3.75\cdot10^{-3} & 9.96\cdot10^{-1} & 5.00\cdot10^{-1} & 1.69\cdot10^{-1} \\*
          & CB1   & 0.000    & 1.000    & 5.00\cdot10^{-1} & 2.50\cdot10^{-1} &       &       & CB1   & 0.000    & 1.000    & 5.00\cdot10^{-1} & 1.67\cdot10^{-1} \\*
          & CB2   & 0.000    & 9.96\cdot10^{-1} & 4.98\cdot10^{-1} & 2.49\cdot10^{-1} &       &       & CB2   & 0.000    & 9.81\cdot10^{-1} & 4.91\cdot10^{-1} & 1.64\cdot10^{-1} \\*
          & AC1   & 1.72\cdot10^{-1} & 4.01\cdot10^{-1} & 2.87\cdot10^{-1} & 2.29\cdot10^{-1} &       &       & AC1   & 1.24\cdot10^{-1} & 4.53\cdot10^{-1} & 2.88\cdot10^{-1} & 1.79\cdot10^{-1} \\*
          & AC2   & 0.000    & 1.000    & 5.00\cdot10^{-1} & 2.50\cdot10^{-1} &       &       & AC2   & 0.000    & 1.000    & 5.00\cdot10^{-1} & 1.67\cdot10^{-1} \\*
          & AC3   & 0.000    & 1.000    & 5.00\cdot10^{-1} & 2.50\cdot10^{-1} &       &       & AC3   & 0.000    & 1.000    & 5.00\cdot10^{-1} & 1.67\cdot10^{-1} \\*
          & CSA   & 2.70\cdot10^{-1} & 3.41\cdot10^{-1} & 3.05\cdot10^{-1} & 2.87\cdot10^{-1} &       &       & CSA   & 2.58\cdot10^{-1} & 3.45\cdot10^{-1} & 3.01\cdot10^{-1} & 2.73\cdot10^{-1} \\*
          & CGA   & 1.76\cdot10^{-1} & 4.04\cdot10^{-1} & 2.90\cdot10^{-1} & 2.33\cdot10^{-1} &       &       & CGA   & 1.57\cdot10^{-1} & 4.53\cdot10^{-1} & 3.05\cdot10^{-1} & 2.07\cdot10^{-1} \\*
     \cmidrule(r){1-6} \cmidrule(r){8-13} 
		$[7, 1]$ & ABT   & 1.80\cdot10^{-1} & 4.19\cdot10^{-1} & 3.00\cdot10^{-1} & 2.10\cdot10^{-1} &       & $[10, 1]$ & ABT   & 1.80\cdot10^{-1} & 4.19\cdot10^{-1} & 3.00\cdot10^{-1} & 2.02\cdot10^{-1} \\*
          & ASB   & 1.42\cdot10^{-1} & 4.83\cdot10^{-1} & 3.13\cdot10^{-1} & 1.85\cdot10^{-1} &       &       & ASB   & 1.12\cdot10^{-1} & 5.13\cdot10^{-1} & 3.13\cdot10^{-1} & 1.49\cdot10^{-1} \\*
          & ADC   & 9.81\cdot10^{-1} & 1.35\cdot10^{-1} & 5.58\cdot10^{-1} & 8.75\cdot10^{-1} &       &       & ADC   & 9.89\cdot10^{-1} & 1.42\cdot10^{-1} & 5.66\cdot10^{-1} & 9.12\cdot10^{-1} \\*
          & CB0   & 3.75\cdot10^{-3} & 9.96\cdot10^{-1} & 5.00\cdot10^{-1} & 1.28\cdot10^{-1} &       &       & CB0   & 3.75\cdot10^{-3} & 9.96\cdot10^{-1} & 5.00\cdot10^{-1} & 9.40\cdot10^{-2} \\*
          & CB1   & 0.000    & 1.000    & 5.00\cdot10^{-1} & 1.25\cdot10^{-1} &       &       & CB1   & 0.000    & 1.000    & 5.00\cdot10^{-1} & 9.09\cdot10^{-2} \\*
          & CB2   & 0.000    & 1.000    & 5.00\cdot10^{-1} & 1.25\cdot10^{-1} &       &       & CB2   & 0.000    & 1.000    & 5.00\cdot10^{-1} & 9.09\cdot10^{-2} \\*
          & AC1   & 1.09\cdot10^{-1} & 4.94\cdot10^{-1} & 3.01\cdot10^{-1} & 1.57\cdot10^{-1} &       &       & AC1   & 8.24\cdot10^{-2} & 5.21\cdot10^{-1} & 3.01\cdot10^{-1} & 1.22\cdot10^{-1} \\*
          & AC2   & 0.000    & 9.96\cdot10^{-1} & 4.98\cdot10^{-1} & 1.25\cdot10^{-1} &       &       & AC2   & 0.000    & 1.000    & 5.00\cdot10^{-1} & 9.09\cdot10^{-2} \\*
          & AC3   & 0.000    & 1.000    & 5.00\cdot10^{-1} & 1.25\cdot10^{-1} &       &       & AC3   & 0.000    & 1.000    & 5.00\cdot10^{-1} & 9.09\cdot10^{-2} \\*
          & CSA   & 2.62\cdot10^{-1} & 3.67\cdot10^{-1} & 3.15\cdot10^{-1} & 2.75\cdot10^{-1} &       &       & CSA   & 2.51\cdot10^{-1} & 3.71\cdot10^{-1} & 3.11\cdot10^{-1} & 2.62\cdot10^{-1} \\*
          & CGA   & 1.46\cdot10^{-1} & 4.94\cdot10^{-1} & 3.20\cdot10^{-1} & 1.90\cdot10^{-1} &       &       & CGA   & 1.24\cdot10^{-1} & 5.09\cdot10^{-1} & 3.16\cdot10^{-1} & 1.59\cdot10^{-1} \\*
     \cmidrule(r){1-6} \cmidrule(r){8-13} 
		$[25, 1]$ & ABT   & 1.76\cdot10^{-1} & 4.31\cdot10^{-1} & 3.03\cdot10^{-1} & 1.86\cdot10^{-1} &       & $[50, 1]$ & ABT   & 1.72\cdot10^{-1} & 4.53\cdot10^{-1} & 3.13\cdot10^{-1} & 1.78\cdot10^{-1} \\*
          & ASB   & 7.87\cdot10^{-2} & 5.69\cdot10^{-1} & 3.24\cdot10^{-1} & 9.75\cdot10^{-2} &       &       & ASB   & 8.61\cdot10^{-2} & 5.84\cdot10^{-1} & 3.35\cdot10^{-1} & 9.59\cdot10^{-2} \\*
          & ADC   & 6.70\cdot10^{-1} & 3.30\cdot10^{-1} & 5.00\cdot10^{-1} & 6.57\cdot10^{-1} &       &       & ADC   & 6.70\cdot10^{-1} & 3.30\cdot10^{-1} & 5.00\cdot10^{-1} & 6.64\cdot10^{-1} \\*
          & CB0   & 3.75\cdot10^{-3} & 9.96\cdot10^{-1} & 5.00\cdot10^{-1} & 4.19\cdot10^{-2} &       &       & CB0   & 3.75\cdot10^{-3} & 9.96\cdot10^{-1} & 5.00\cdot10^{-1} & 2.32\cdot10^{-2} \\*
          & CB1   & 3.75\cdot10^{-3} & 9.96\cdot10^{-1} & 5.00\cdot10^{-1} & 4.19\cdot10^{-2} &       &       & CB1   & 3.75\cdot10^{-3} & 9.96\cdot10^{-1} & 5.00\cdot10^{-1} & 2.32\cdot10^{-2} \\*
          & CB2   & 3.75\cdot10^{-3} & 9.96\cdot10^{-1} & 5.00\cdot10^{-1} & 4.19\cdot10^{-2} &       &       & CB2   & 3.75\cdot10^{-3} & 9.96\cdot10^{-1} & 5.00\cdot10^{-1} & 2.32\cdot10^{-2} \\*
          & AC1   & 1.87\cdot10^{-2} & 6.89\cdot10^{-1} & 3.54\cdot10^{-1} & 4.45\cdot10^{-2} &       &       & AC1   & 3.75\cdot10^{-3} & 8.61\cdot10^{-1} & 4.33\cdot10^{-1} & 2.06\cdot10^{-2} \\*
          & AC2   & 3.75\cdot10^{-3} & 9.96\cdot10^{-1} & 5.00\cdot10^{-1} & 4.19\cdot10^{-2} &       &       & AC2   & 3.75\cdot10^{-3} & 9.96\cdot10^{-1} & 5.00\cdot10^{-1} & 2.32\cdot10^{-2} \\*
          & AC3   & 3.75\cdot10^{-3} & 9.96\cdot10^{-1} & 5.00\cdot10^{-1} & 4.19\cdot10^{-2} &       &       & AC3   & 3.75\cdot10^{-3} & 9.96\cdot10^{-1} & 5.00\cdot10^{-1} & 2.32\cdot10^{-2} \\*
          & CSA   & 1.99\cdot10^{-1} & 3.90\cdot10^{-1} & 2.94\cdot10^{-1} & 2.06\cdot10^{-1} &       &       & CSA   & 1.27\cdot10^{-1} & 4.64\cdot10^{-1} & 2.96\cdot10^{-1} & 1.34\cdot10^{-1} \\*
          & CGA   & 5.24\cdot10^{-2} & 6.67\cdot10^{-1} & 3.60\cdot10^{-1} & 7.61\cdot10^{-2} &       &       & CGA   & 3.37\cdot10^{-2} & 7.27\cdot10^{-1} & 3.80\cdot10^{-1} & 4.73\cdot10^{-2} \\*
     \cmidrule(r){1-6} \cmidrule(r){8-13} 
		$[100, 1]$ & ABT   & 1.39\cdot10^{-1} & 4.72\cdot10^{-1} & 3.05\cdot10^{-1} & 1.42\cdot10^{-1} &       &       &       &       &       &       &  \\*
          & ASB   & 8.24\cdot10^{-2} & 5.36\cdot10^{-1} & 3.09\cdot10^{-1} & 8.69\cdot10^{-2} &       &       &       &       &       &       &  \\*
          & ADC   & 3.37\cdot10^{-1} & 6.63\cdot10^{-1} & 5.00\cdot10^{-1} & 3.40\cdot10^{-1} &       &       &       &       &       &       &  \\*
          & CB0   & 3.75\cdot10^{-3} & 9.96\cdot10^{-1} & 5.00\cdot10^{-1} & 1.36\cdot10^{-2} &       &       &       &       &       &       &  \\*
          & CB1   & 3.75\cdot10^{-3} & 9.96\cdot10^{-1} & 5.00\cdot10^{-1} & 1.36\cdot10^{-2} &       &       &       &       &       &       &  \\*
          & CB2   & 3.75\cdot10^{-3} & 9.96\cdot10^{-1} & 5.00\cdot10^{-1} & 1.36\cdot10^{-2} &       &       &       &       &       &       &  \\*
          & AC1   & 3.75\cdot10^{-3} & 9.96\cdot10^{-1} & 5.00\cdot10^{-1} & 1.36\cdot10^{-2} &       &       &       &       &       &       &  \\*
          & AC2   & 3.75\cdot10^{-3} & 9.96\cdot10^{-1} & 5.00\cdot10^{-1} & 1.36\cdot10^{-2} &       &       &       &       &       &       &  \\*
          & AC3   & 3.75\cdot10^{-3} & 9.96\cdot10^{-1} & 5.00\cdot10^{-1} & 1.36\cdot10^{-2} &       &       &       &       &       &       &  \\*
          & CSA   & 7.12\cdot10^{-2} & 5.88\cdot10^{-1} & 3.30\cdot10^{-1} & 7.63\cdot10^{-2} &       &       &       &       &       &       &  \\*
          & CGA   & 7.49\cdot10^{-3} & 8.76\cdot10^{-1} & 4.42\cdot10^{-1} & 1.61\cdot10^{-2} &       &       &       &       &       &       &  \\*

\end{longtabu}
}

\newpage

%%%%%%%%%%%%%%%%%%%%
%%% UCI DIABETES %%%
%%%%%%%%%%%%%%%%%%%%

\begin{landscape}
% GRAPHICS
\begin{figure}[p]
\centering
\includegraphics[width=.6\paperheight]{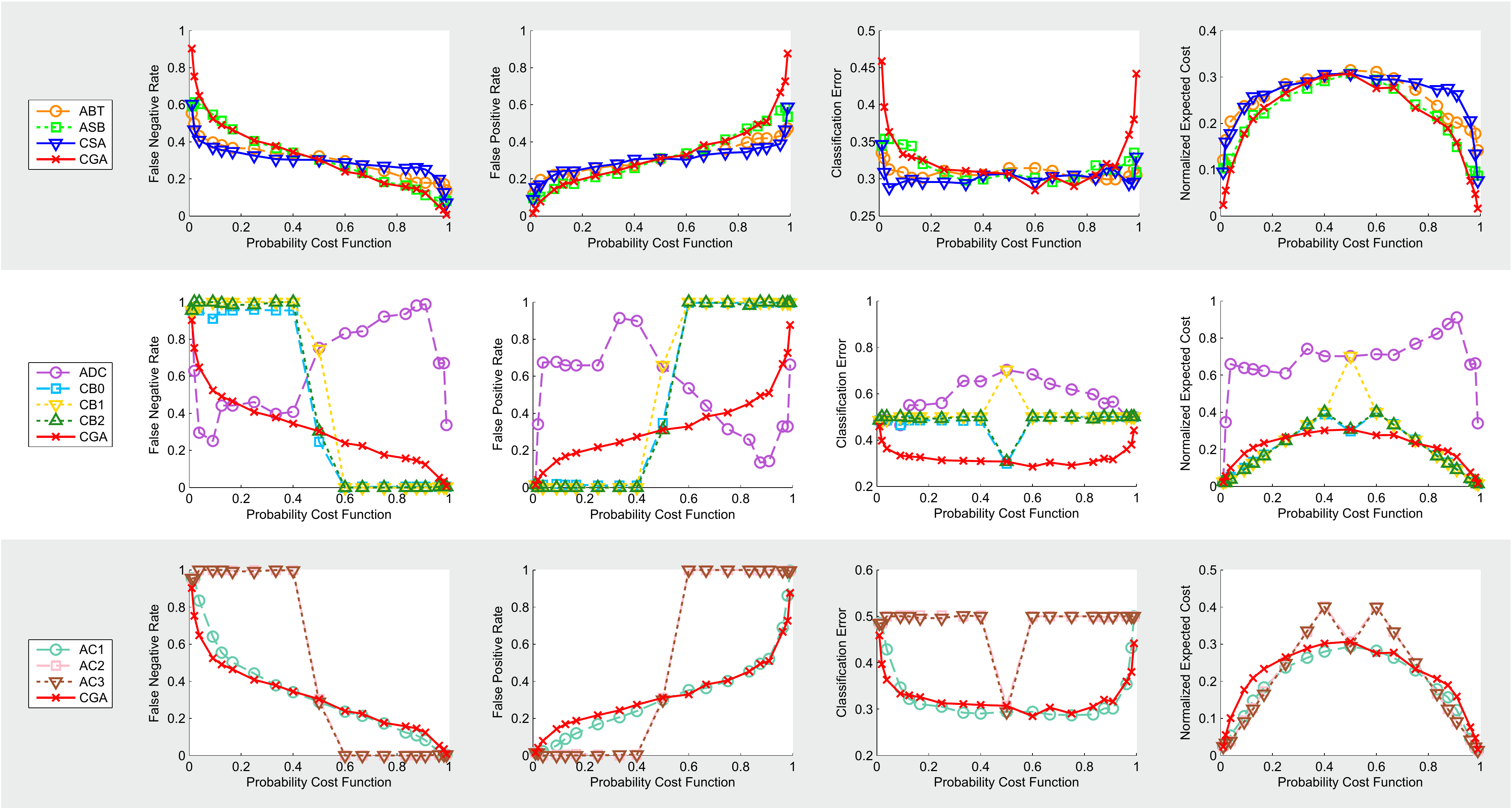}
\caption[Results obtained for the UCI Diabetes.]{Results obtained for the UCI Diabetes. First column of the illustration corresponds to False Negative Rate, second column to False Positive Rate, third column to Classification Error and the fourth one corresponds to Normalized Expected Cost. For a clearer visualization, algorithms have been divided into three groups so each row of the illustration corresponds to a different group. Cost Generalized AdaBoost is plotted in all the graphs to have a common reference across the representations.}
\label{fig:diabetes_perform} % caption for the whole figure
\end{figure}
\end{landscape}
\newpage

%%%%%%%%%%%%%%%%%%%%%%
%%% UCI IONOSPHERE %%%
%%%%%%%%%%%%%%%%%%%%%%

% TABLE

{
\tiny
\begin{longtabu} to \textwidth {llX[$l]@{}@{}X[$l]@{}@{}X[$l]@{}@{}X[$l]@{}@{}l@{}llX[$l]@{}@{}X[$l]@{}@{}X[$l]@{}@{}X[$l]@{}}

\toprule
{\textbf{Cost}}  & {\textbf{Alg}}   & {\textbf{FNR}}    & {\textbf{FPR}}    & {\textbf{CE}}    & {\textbf{NEC}}   && {\textbf{Cost}}  & {\textbf{Alg}}   & {\textbf{FNR}}    & {\textbf{FPR}}    & {\textbf{CE}} & {\textbf{NEC}}\\* 
\cmidrule(r){1-6} \cmidrule(r){8-13}
\endfirsthead

%\multicolumn{13}{l}% {\tiny{Continued from previous page}} \\*
\toprule
{\textbf{Cost}}  & {\textbf{Alg}}   & {\textbf{FNR}}    & {\textbf{FPR}}    & {\textbf{CE}}    & {\textbf{NEC}}   && {\textbf{Cost}}  & {\textbf{Alg}}   & {\textbf{FNR}}    & {\textbf{FPR}}    & {\textbf{CE}} & {\textbf{NEC}}\\* 
\cmidrule(r){1-6} \cmidrule(r){8-13}
\endhead

\multicolumn{13}{r}{{\tiny{\tablename\ \thetable{} - Continued on next page $\rightarrow$}}} \\* 
\endfoot

\bottomrule
  \caption{Results obtained for the UCI Ionosphere Dataset.}
	\label{tab:ionosphere_perform}%
\endlastfoot

    $[1, 100]$ & ABT   & 1.98\cdot10^{-1} & 1.11\cdot10^{-1} & 1.55\cdot10^{-1} & 1.12\cdot10^{-1} &       & $[1, 50]$ & ABT   & 1.59\cdot10^{-1} & 1.43\cdot10^{-1} & 1.51\cdot10^{-1} & 1.43\cdot10^{-1} \\*
          & ASB   & 1.11\cdot10^{-1} & 1.98\cdot10^{-1} & 1.55\cdot10^{-1} & 1.98\cdot10^{-1} &       &       & ASB   & 7.94\cdot10^{-2} & 1.98\cdot10^{-1} & 1.39\cdot10^{-1} & 1.96\cdot10^{-1} \\*
          & ADC   & 0.000    & 1.000    & 5.00\cdot10^{-1} & 9.90\cdot10^{-1} &       &       & ADC   & 0.000    & 1.000    & 5.00\cdot10^{-1} & 9.80\cdot10^{-1} \\*
          & CB0   & 1.000    & 0.000    & 5.00\cdot10^{-1} & 9.90\cdot10^{-3} &       &       & CB0   & 1.000    & 0.000    & 5.00\cdot10^{-1} & 1.96\cdot10^{-2} \\*
          & CB1   & 1.000    & 0.000    & 5.00\cdot10^{-1} & 9.90\cdot10^{-3} &       &       & CB1   & 1.000    & 0.000    & 5.00\cdot10^{-1} & 1.96\cdot10^{-2} \\*
          & CB2   & 1.000    & 0.000    & 5.00\cdot10^{-1} & 9.90\cdot10^{-3} &       &       & CB2   & 1.000    & 0.000    & 5.00\cdot10^{-1} & 1.96\cdot10^{-2} \\*
          & AC1   & 1.000    & 0.000    & 5.00\cdot10^{-1} & 9.90\cdot10^{-3} &       &       & AC1   & 1.000    & 0.000    & 5.00\cdot10^{-1} & 1.96\cdot10^{-2} \\*
          & AC2   & 1.000    & 0.000    & 5.00\cdot10^{-1} & 9.90\cdot10^{-3} &       &       & AC2   & 1.000    & 0.000    & 5.00\cdot10^{-1} & 1.96\cdot10^{-2} \\*
          & AC3   & 1.000    & 0.000    & 5.00\cdot10^{-1} & 9.90\cdot10^{-3} &       &       & AC3   & 1.000    & 0.000    & 5.00\cdot10^{-1} & 1.96\cdot10^{-2} \\*
          & CSA   & 1.51\cdot10^{-1} & 1.75\cdot10^{-1} & 1.63\cdot10^{-1} & 1.74\cdot10^{-1} &       &       & CSA   & 7.94\cdot10^{-2} & 2.94\cdot10^{-1} & 1.87\cdot10^{-1} & 2.89\cdot10^{-1} \\*
          & CGA   & 3.41\cdot10^{-1} & 7.94\cdot10^{-2} & 2.10\cdot10^{-1} & 8.20\cdot10^{-2} &       &       & CGA   & 2.30\cdot10^{-1} & 9.52\cdot10^{-2} & 1.63\cdot10^{-1} & 9.79\cdot10^{-2} \\*
     \cmidrule(r){1-6} \cmidrule(r){8-13} 
		$[1, 25]$ & ABT   & 1.19\cdot10^{-1} & 2.22\cdot10^{-1} & 1.71\cdot10^{-1} & 2.18\cdot10^{-1} &       & $[1, 10]$ & ABT   & 9.52\cdot10^{-2} & 2.06\cdot10^{-1} & 1.51\cdot10^{-1} & 1.96\cdot10^{-1} \\*
          & ASB   & 7.94\cdot10^{-2} & 1.98\cdot10^{-1} & 1.39\cdot10^{-1} & 1.94\cdot10^{-1} &       &       & ASB   & 8.73\cdot10^{-2} & 2.30\cdot10^{-1} & 1.59\cdot10^{-1} & 2.17\cdot10^{-1} \\*
          & ADC   & 0.000    & 1.000    & 5.00\cdot10^{-1} & 9.62\cdot10^{-1} &       &       & ADC   & 0.000    & 1.000    & 5.00\cdot10^{-1} & 9.09\cdot10^{-1} \\*
          & CB0   & 1.000    & 0.000    & 5.00\cdot10^{-1} & 3.85\cdot10^{-2} &       &       & CB0   & 1.000    & 0.000    & 5.00\cdot10^{-1} & 9.09\cdot10^{-2} \\*
          & CB1   & 1.000    & 0.000    & 5.00\cdot10^{-1} & 3.85\cdot10^{-2} &       &       & CB1   & 1.000    & 0.000    & 5.00\cdot10^{-1} & 9.09\cdot10^{-2} \\*
          & CB2   & 1.000    & 0.000    & 5.00\cdot10^{-1} & 3.85\cdot10^{-2} &       &       & CB2   & 1.000    & 0.000    & 5.00\cdot10^{-1} & 9.09\cdot10^{-2} \\*
          & AC1   & 4.29\cdot10^{-1} & 3.17\cdot10^{-2} & 2.30\cdot10^{-1} & 4.70\cdot10^{-2} &       &       & AC1   & 1.59\cdot10^{-1} & 1.59\cdot10^{-1} & 1.59\cdot10^{-1} & 1.59\cdot10^{-1} \\*
          & AC2   & 1.000    & 0.000    & 5.00\cdot10^{-1} & 3.85\cdot10^{-2} &       &       & AC2   & 1.000    & 0.000    & 5.00\cdot10^{-1} & 9.09\cdot10^{-2} \\*
          & AC3   & 1.000    & 0.000    & 5.00\cdot10^{-1} & 3.85\cdot10^{-2} &       &       & AC3   & 1.000    & 0.000    & 5.00\cdot10^{-1} & 9.09\cdot10^{-2} \\*
          & CSA   & 2.38\cdot10^{-2} & 2.54\cdot10^{-1} & 1.39\cdot10^{-1} & 2.45\cdot10^{-1} &       &       & CSA   & 3.17\cdot10^{-2} & 3.25\cdot10^{-1} & 1.79\cdot10^{-1} & 2.99\cdot10^{-1} \\*
          & CGA   & 2.30\cdot10^{-1} & 9.52\cdot10^{-2} & 1.63\cdot10^{-1} & 1.00\cdot10^{-1} &       &       & CGA   & 1.51\cdot10^{-1} & 1.11\cdot10^{-1} & 1.31\cdot10^{-1} & 1.15\cdot10^{-1} \\*
     \cmidrule(r){1-6} \cmidrule(r){8-13} 
		$[1, 7]$ & ABT   & 9.52\cdot10^{-2} & 2.06\cdot10^{-1} & 1.51\cdot10^{-1} & 1.92\cdot10^{-1} &       & $[1, 5]$ & ABT   & 1.03\cdot10^{-1} & 1.75\cdot10^{-1} & 1.39\cdot10^{-1} & 1.63\cdot10^{-1} \\*
          & ASB   & 1.03\cdot10^{-1} & 1.75\cdot10^{-1} & 1.39\cdot10^{-1} & 1.66\cdot10^{-1} &       &       & ASB   & 7.14\cdot10^{-2} & 2.22\cdot10^{-1} & 1.47\cdot10^{-1} & 1.97\cdot10^{-1} \\*
          & ADC   & 0.000    & 1.000    & 5.00\cdot10^{-1} & 8.75\cdot10^{-1} &       &       & ADC   & 0.000    & 1.000    & 5.00\cdot10^{-1} & 8.33\cdot10^{-1} \\*
          & CB0   & 1.000    & 0.000    & 5.00\cdot10^{-1} & 1.25\cdot10^{-1} &       &       & CB0   & 1.000    & 0.000    & 5.00\cdot10^{-1} & 1.67\cdot10^{-1} \\*
          & CB1   & 1.000    & 0.000    & 5.00\cdot10^{-1} & 1.25\cdot10^{-1} &       &       & CB1   & 1.000    & 0.000    & 5.00\cdot10^{-1} & 1.67\cdot10^{-1} \\*
          & CB2   & 1.000    & 0.000    & 5.00\cdot10^{-1} & 1.25\cdot10^{-1} &       &       & CB2   & 1.000    & 0.000    & 5.00\cdot10^{-1} & 1.67\cdot10^{-1} \\*
          & AC1   & 7.94\cdot10^{-2} & 1.98\cdot10^{-1} & 1.39\cdot10^{-1} & 1.84\cdot10^{-1} &       &       & AC1   & 6.35\cdot10^{-2} & 2.54\cdot10^{-1} & 1.59\cdot10^{-1} & 2.22\cdot10^{-1} \\*
          & AC2   & 1.000    & 0.000    & 5.00\cdot10^{-1} & 1.25\cdot10^{-1} &       &       & AC2   & 1.000    & 0.000    & 5.00\cdot10^{-1} & 1.67\cdot10^{-1} \\*
          & AC3   & 1.000    & 0.000    & 5.00\cdot10^{-1} & 1.25\cdot10^{-1} &       &       & AC3   & 1.000    & 0.000    & 5.00\cdot10^{-1} & 1.67\cdot10^{-1} \\*
          & CSA   & 3.97\cdot10^{-2} & 3.02\cdot10^{-1} & 1.71\cdot10^{-1} & 2.69\cdot10^{-1} &       &       & CSA   & 3.17\cdot10^{-2} & 3.17\cdot10^{-1} & 1.75\cdot10^{-1} & 2.70\cdot10^{-1} \\*
          & CGA   & 1.90\cdot10^{-1} & 1.27\cdot10^{-1} & 1.59\cdot10^{-1} & 1.35\cdot10^{-1} &       &       & CGA   & 1.35\cdot10^{-1} & 1.43\cdot10^{-1} & 1.39\cdot10^{-1} & 1.42\cdot10^{-1} \\*
     \cmidrule(r){1-6} \cmidrule(r){8-13} 
		$[1, 3]$ & ABT   & 1.03\cdot10^{-1} & 1.75\cdot10^{-1} & 1.39\cdot10^{-1} & 1.57\cdot10^{-1} &       & $[1, 2]$ & ABT   & 7.14\cdot10^{-2} & 2.38\cdot10^{-1} & 1.55\cdot10^{-1} & 1.83\cdot10^{-1} \\*
          & ASB   & 4.76\cdot10^{-2} & 2.38\cdot10^{-1} & 1.43\cdot10^{-1} & 1.90\cdot10^{-1} &       &       & ASB   & 7.14\cdot10^{-2} & 1.98\cdot10^{-1} & 1.35\cdot10^{-1} & 1.56\cdot10^{-1} \\*
          & ADC   & 2.06\cdot10^{-1} & 8.49\cdot10^{-1} & 5.28\cdot10^{-1} & 6.88\cdot10^{-1} &       &       & ADC   & 8.65\cdot10^{-1} & 5.00\cdot10^{-1} & 6.83\cdot10^{-1} & 6.22\cdot10^{-1} \\*
          & CB0   & 1.000    & 0.000    & 5.00\cdot10^{-1} & 2.50\cdot10^{-1} &       &       & CB0   & 1.000    & 0.000    & 5.00\cdot10^{-1} & 3.33\cdot10^{-1} \\*
          & CB1   & 9.92\cdot10^{-1} & 0.000    & 4.96\cdot10^{-1} & 2.48\cdot10^{-1} &       &       & CB1   & 1.000    & 0.000    & 5.00\cdot10^{-1} & 3.33\cdot10^{-1} \\*
          & CB2   & 1.000    & 0.000    & 5.00\cdot10^{-1} & 2.50\cdot10^{-1} &       &       & CB2   & 1.000    & 0.000    & 5.00\cdot10^{-1} & 3.33\cdot10^{-1} \\*
          & AC1   & 4.76\cdot10^{-2} & 2.54\cdot10^{-1} & 1.51\cdot10^{-1} & 2.02\cdot10^{-1} &       &       & AC1   & 3.97\cdot10^{-2} & 2.38\cdot10^{-1} & 1.39\cdot10^{-1} & 1.72\cdot10^{-1} \\*
          & AC2   & 1.000    & 0.000    & 5.00\cdot10^{-1} & 2.50\cdot10^{-1} &       &       & AC2   & 1.000    & 0.000    & 5.00\cdot10^{-1} & 3.33\cdot10^{-1} \\*
          & AC3   & 1.000    & 0.000    & 5.00\cdot10^{-1} & 2.50\cdot10^{-1} &       &       & AC3   & 1.000    & 0.000    & 5.00\cdot10^{-1} & 3.33\cdot10^{-1} \\*
          & CSA   & 2.38\cdot10^{-2} & 2.78\cdot10^{-1} & 1.51\cdot10^{-1} & 2.14\cdot10^{-1} &       &       & CSA   & 7.14\cdot10^{-2} & 2.38\cdot10^{-1} & 1.55\cdot10^{-1} & 1.83\cdot10^{-1} \\*
          & CGA   & 1.43\cdot10^{-1} & 1.59\cdot10^{-1} & 1.51\cdot10^{-1} & 1.55\cdot10^{-1} &       &       & CGA   & 1.19\cdot10^{-1} & 1.90\cdot10^{-1} & 1.55\cdot10^{-1} & 1.67\cdot10^{-1} \\*
     \cmidrule(r){1-6} \cmidrule(r){8-13} 
		$[2, 3]$ & ABT   & 7.14\cdot10^{-2} & 2.38\cdot10^{-1} & 1.55\cdot10^{-1} & 1.71\cdot10^{-1} &       & $[1, 1]$ & ABT   & 7.14\cdot10^{-2} & 2.38\cdot10^{-1} & 1.55\cdot10^{-1} & 1.55\cdot10^{-1} \\*
          & ASB   & 6.35\cdot10^{-2} & 2.06\cdot10^{-1} & 1.35\cdot10^{-1} & 1.49\cdot10^{-1} &       &       & ASB   & 1.03\cdot10^{-1} & 2.14\cdot10^{-1} & 1.59\cdot10^{-1} & 1.59\cdot10^{-1} \\*
          & ADC   & 6.51\cdot10^{-1} & 5.63\cdot10^{-1} & 6.07\cdot10^{-1} & 5.98\cdot10^{-1} &       &       & ADC   & 3.17\cdot10^{-2} & 4.37\cdot10^{-1} & 2.34\cdot10^{-1} & 2.34\cdot10^{-1} \\*
          & CB0   & 1.000    & 0.000    & 5.00\cdot10^{-1} & 4.00\cdot10^{-1} &       &       & CB0   & 1.59\cdot10^{-2} & 5.00\cdot10^{-1} & 2.58\cdot10^{-1} & 2.58\cdot10^{-1} \\*
          & CB1   & 1.000    & 0.000    & 5.00\cdot10^{-1} & 4.00\cdot10^{-1} &       &       & CB1   & 1.27\cdot10^{-1} & 2.54\cdot10^{-1} & 1.90\cdot10^{-1} & 1.90\cdot10^{-1} \\*
          & CB2   & 1.000    & 0.000    & 5.00\cdot10^{-1} & 4.00\cdot10^{-1} &       &       & CB2   & 1.03\cdot10^{-1} & 2.14\cdot10^{-1} & 1.59\cdot10^{-1} & 1.59\cdot10^{-1} \\*
          & AC1   & 6.35\cdot10^{-2} & 2.30\cdot10^{-1} & 1.47\cdot10^{-1} & 1.63\cdot10^{-1} &       &       & AC1   & 5.56\cdot10^{-2} & 1.98\cdot10^{-1} & 1.27\cdot10^{-1} & 1.27\cdot10^{-1} \\*
          & AC2   & 1.000    & 0.000    & 5.00\cdot10^{-1} & 4.00\cdot10^{-1} &       &       & AC2   & 1.03\cdot10^{-1} & 2.14\cdot10^{-1} & 1.59\cdot10^{-1} & 1.59\cdot10^{-1} \\*
          & AC3   & 1.000    & 0.000    & 5.00\cdot10^{-1} & 4.00\cdot10^{-1} &       &       & AC3   & 5.56\cdot10^{-2} & 1.98\cdot10^{-1} & 1.27\cdot10^{-1} & 1.27\cdot10^{-1} \\*
          & CSA   & 7.14\cdot10^{-2} & 2.38\cdot10^{-1} & 1.55\cdot10^{-1} & 1.71\cdot10^{-1} &       &       & CSA   & 1.03\cdot10^{-1} & 2.14\cdot10^{-1} & 1.59\cdot10^{-1} & 1.59\cdot10^{-1} \\*
          & CGA   & 1.11\cdot10^{-1} & 1.83\cdot10^{-1} & 1.47\cdot10^{-1} & 1.54\cdot10^{-1} &       &       & CGA   & 1.03\cdot10^{-1} & 2.14\cdot10^{-1} & 1.59\cdot10^{-1} & 1.59\cdot10^{-1} \\*
     \cmidrule(r){1-6} \cmidrule(r){8-13} 
		
		$[3, 2]$ & ABT   & 7.14\cdot10^{-2} & 2.38\cdot10^{-1} & 1.55\cdot10^{-1} & 1.38\cdot10^{-1} &       & $[2, 1]$ & ABT   & 7.14\cdot10^{-2} & 2.38\cdot10^{-1} & 1.55\cdot10^{-1} & 1.27\cdot10^{-1} \\*
          & ASB   & 1.03\cdot10^{-1} & 1.83\cdot10^{-1} & 1.43\cdot10^{-1} & 1.35\cdot10^{-1} &       &       & ASB   & 1.03\cdot10^{-1} & 2.30\cdot10^{-1} & 1.67\cdot10^{-1} & 1.46\cdot10^{-1} \\*
          & ADC   & 2.38\cdot10^{-2} & 4.29\cdot10^{-1} & 2.26\cdot10^{-1} & 1.86\cdot10^{-1} &       &       & ADC   & 1.59\cdot10^{-2} & 4.37\cdot10^{-1} & 2.26\cdot10^{-1} & 1.56\cdot10^{-1} \\*
          & CB0   & 3.97\cdot10^{-2} & 4.13\cdot10^{-1} & 2.26\cdot10^{-1} & 1.89\cdot10^{-1} &       &       & CB0   & 3.97\cdot10^{-2} & 4.13\cdot10^{-1} & 2.26\cdot10^{-1} & 1.64\cdot10^{-1} \\*
          & CB1   & 0.000    & 9.60\cdot10^{-1} & 4.80\cdot10^{-1} & 3.84\cdot10^{-1} &       &       & CB1   & 0.000    & 1.000    & 5.00\cdot10^{-1} & 3.33\cdot10^{-1} \\*
          & CB2   & 0.000    & 8.97\cdot10^{-1} & 4.48\cdot10^{-1} & 3.59\cdot10^{-1} &       &       & CB2   & 0.000    & 9.92\cdot10^{-1} & 4.96\cdot10^{-1} & 3.31\cdot10^{-1} \\*
          & AC1   & 8.73\cdot10^{-2} & 1.75\cdot10^{-1} & 1.31\cdot10^{-1} & 1.22\cdot10^{-1} &       &       & AC1   & 6.35\cdot10^{-2} & 1.90\cdot10^{-1} & 1.27\cdot10^{-1} & 1.06\cdot10^{-1} \\*
          & AC2   & 0.000    & 9.05\cdot10^{-1} & 4.52\cdot10^{-1} & 3.62\cdot10^{-1} &       &       & AC2   & 0.000    & 9.37\cdot10^{-1} & 4.68\cdot10^{-1} & 3.12\cdot10^{-1} \\*
          & AC3   & 0.000    & 9.29\cdot10^{-1} & 4.64\cdot10^{-1} & 3.71\cdot10^{-1} &       &       & AC3   & 0.000    & 9.21\cdot10^{-1} & 4.60\cdot10^{-1} & 3.07\cdot10^{-1} \\*
          & CSA   & 7.14\cdot10^{-2} & 1.98\cdot10^{-1} & 1.35\cdot10^{-1} & 1.22\cdot10^{-1} &       &       & CSA   & 1.27\cdot10^{-1} & 1.83\cdot10^{-1} & 1.55\cdot10^{-1} & 1.46\cdot10^{-1} \\*
          & CGA   & 1.03\cdot10^{-1} & 2.30\cdot10^{-1} & 1.67\cdot10^{-1} & 1.54\cdot10^{-1} &       &       & CGA   & 6.35\cdot10^{-2} & 2.54\cdot10^{-1} & 1.59\cdot10^{-1} & 1.27\cdot10^{-1} \\*
     \cmidrule(r){1-6} \cmidrule(r){8-13} 
		$[3, 1]$ & ABT   & 8.73\cdot10^{-2} & 1.75\cdot10^{-1} & 1.31\cdot10^{-1} & 1.09\cdot10^{-1} &       & $[5, 1]$ & ABT   & 8.73\cdot10^{-2} & 1.75\cdot10^{-1} & 1.31\cdot10^{-1} & 1.02\cdot10^{-1} \\*
          & ASB   & 7.94\cdot10^{-2} & 2.38\cdot10^{-1} & 1.59\cdot10^{-1} & 1.19\cdot10^{-1} &       &       & ASB   & 7.14\cdot10^{-2} & 2.46\cdot10^{-1} & 1.59\cdot10^{-1} & 1.01\cdot10^{-1} \\*
          & ADC   & 1.59\cdot10^{-2} & 4.60\cdot10^{-1} & 2.38\cdot10^{-1} & 1.27\cdot10^{-1} &       &       & ADC   & 1.59\cdot10^{-2} & 4.60\cdot10^{-1} & 2.38\cdot10^{-1} & 8.99\cdot10^{-2} \\*
          & CB0   & 3.97\cdot10^{-2} & 4.13\cdot10^{-1} & 2.26\cdot10^{-1} & 1.33\cdot10^{-1} &       &       & CB0   & 3.97\cdot10^{-2} & 4.13\cdot10^{-1} & 2.26\cdot10^{-1} & 1.02\cdot10^{-1} \\*
          & CB1   & 0.000    & 1.000    & 5.00\cdot10^{-1} & 2.50\cdot10^{-1} &       &       & CB1   & 0.000    & 1.000    & 5.00\cdot10^{-1} & 1.67\cdot10^{-1} \\*
          & CB2   & 0.000    & 1.000    & 5.00\cdot10^{-1} & 2.50\cdot10^{-1} &       &       & CB2   & 0.000    & 1.000    & 5.00\cdot10^{-1} & 1.67\cdot10^{-1} \\*
          & AC1   & 9.52\cdot10^{-2} & 1.90\cdot10^{-1} & 1.43\cdot10^{-1} & 1.19\cdot10^{-1} &       &       & AC1   & 8.73\cdot10^{-2} & 2.14\cdot10^{-1} & 1.51\cdot10^{-1} & 1.08\cdot10^{-1} \\*
          & AC2   & 0.000    & 9.60\cdot10^{-1} & 4.80\cdot10^{-1} & 2.40\cdot10^{-1} &       &       & AC2   & 0.000    & 9.92\cdot10^{-1} & 4.96\cdot10^{-1} & 1.65\cdot10^{-1} \\*
          & AC3   & 0.000    & 9.68\cdot10^{-1} & 4.84\cdot10^{-1} & 2.42\cdot10^{-1} &       &       & AC3   & 0.000    & 1.000    & 5.00\cdot10^{-1} & 1.67\cdot10^{-1} \\*
          & CSA   & 1.35\cdot10^{-1} & 1.67\cdot10^{-1} & 1.51\cdot10^{-1} & 1.43\cdot10^{-1} &       &       & CSA   & 1.67\cdot10^{-1} & 1.35\cdot10^{-1} & 1.51\cdot10^{-1} & 1.61\cdot10^{-1} \\*
          & CGA   & 2.38\cdot10^{-2} & 2.62\cdot10^{-1} & 1.43\cdot10^{-1} & 8.33\cdot10^{-2} &       &       & CGA   & 4.76\cdot10^{-2} & 2.70\cdot10^{-1} & 1.59\cdot10^{-1} & 8.47\cdot10^{-2} \\*
     \cmidrule(r){1-6} \cmidrule(r){8-13} 
		$[7, 1]$ & ABT   & 7.14\cdot10^{-2} & 2.14\cdot10^{-1} & 1.43\cdot10^{-1} & 8.93\cdot10^{-2} &       & $[10, 1]$ & ABT   & 7.14\cdot10^{-2} & 2.14\cdot10^{-1} & 1.43\cdot10^{-1} & 8.44\cdot10^{-2} \\*
          & ASB   & 4.76\cdot10^{-2} & 2.94\cdot10^{-1} & 1.71\cdot10^{-1} & 7.84\cdot10^{-2} &       &       & ASB   & 8.73\cdot10^{-2} & 2.86\cdot10^{-1} & 1.87\cdot10^{-1} & 1.05\cdot10^{-1} \\*
          & ADC   & 1.59\cdot10^{-2} & 4.60\cdot10^{-1} & 2.38\cdot10^{-1} & 7.14\cdot10^{-2} &       &       & ADC   & 1.59\cdot10^{-2} & 4.60\cdot10^{-1} & 2.38\cdot10^{-1} & 5.63\cdot10^{-2} \\*
          & CB0   & 3.97\cdot10^{-2} & 4.13\cdot10^{-1} & 2.26\cdot10^{-1} & 8.63\cdot10^{-2} &       &       & CB0   & 3.97\cdot10^{-2} & 4.13\cdot10^{-1} & 2.26\cdot10^{-1} & 7.36\cdot10^{-2} \\*
          & CB1   & 0.000    & 9.76\cdot10^{-1} & 4.88\cdot10^{-1} & 1.22\cdot10^{-1} &       &       & CB1   & 0.000    & 9.92\cdot10^{-1} & 4.96\cdot10^{-1} & 9.02\cdot10^{-2} \\*
          & CB2   & 0.000    & 9.84\cdot10^{-1} & 4.92\cdot10^{-1} & 1.23\cdot10^{-1} &       &       & CB2   & 0.000    & 1.000    & 5.00\cdot10^{-1} & 9.09\cdot10^{-2} \\*
          & AC1   & 7.94\cdot10^{-2} & 2.54\cdot10^{-1} & 1.67\cdot10^{-1} & 1.01\cdot10^{-1} &       &       & AC1   & 7.14\cdot10^{-2} & 3.17\cdot10^{-1} & 1.94\cdot10^{-1} & 9.38\cdot10^{-2} \\*
          & AC2   & 0.000    & 9.13\cdot10^{-1} & 4.56\cdot10^{-1} & 1.14\cdot10^{-1} &       &       & AC2   & 0.000    & 1.000    & 5.00\cdot10^{-1} & 9.09\cdot10^{-2} \\*
          & AC3   & 0.000    & 9.92\cdot10^{-1} & 4.96\cdot10^{-1} & 1.24\cdot10^{-1} &       &       & AC3   & 0.000    & 1.000    & 5.00\cdot10^{-1} & 9.09\cdot10^{-2} \\*
          & CSA   & 1.75\cdot10^{-1} & 1.59\cdot10^{-1} & 1.67\cdot10^{-1} & 1.73\cdot10^{-1} &       &       & CSA   & 1.83\cdot10^{-1} & 1.51\cdot10^{-1} & 1.67\cdot10^{-1} & 1.80\cdot10^{-1} \\*
          & CGA   & 5.56\cdot10^{-2} & 2.86\cdot10^{-1} & 1.71\cdot10^{-1} & 8.43\cdot10^{-2} &       &       & CGA   & 4.76\cdot10^{-2} & 3.33\cdot10^{-1} & 1.90\cdot10^{-1} & 7.36\cdot10^{-2} \\*
     \cmidrule(r){1-6} \cmidrule(r){8-13} 
		$[25, 1]$ & ABT   & 5.56\cdot10^{-2} & 2.62\cdot10^{-1} & 1.59\cdot10^{-1} & 6.35\cdot10^{-2} &       & $[50, 1]$ & ABT   & 8.73\cdot10^{-2} & 2.22\cdot10^{-1} & 1.55\cdot10^{-1} & 8.99\cdot10^{-2} \\*
          & ASB   & 7.94\cdot10^{-2} & 2.70\cdot10^{-1} & 1.75\cdot10^{-1} & 8.67\cdot10^{-2} &       &       & ASB   & 3.97\cdot10^{-2} & 2.86\cdot10^{-1} & 1.63\cdot10^{-1} & 4.45\cdot10^{-2} \\*
          & ADC   & 3.97\cdot10^{-2} & 4.13\cdot10^{-1} & 2.26\cdot10^{-1} & 5.40\cdot10^{-2} &       &       & ADC   & 3.97\cdot10^{-2} & 4.13\cdot10^{-1} & 2.26\cdot10^{-1} & 4.70\cdot10^{-2} \\*
          & CB0   & 3.97\cdot10^{-2} & 4.13\cdot10^{-1} & 2.26\cdot10^{-1} & 5.40\cdot10^{-2} &       &       & CB0   & 3.97\cdot10^{-2} & 4.13\cdot10^{-1} & 2.26\cdot10^{-1} & 4.70\cdot10^{-2} \\*
          & CB1   & 0.000    & 9.84\cdot10^{-1} & 4.92\cdot10^{-1} & 3.79\cdot10^{-2} &       &       & CB1   & 0.000    & 1.000    & 5.00\cdot10^{-1} & 1.96\cdot10^{-2} \\*
          & CB2   & 0.000    & 9.60\cdot10^{-1} & 4.80\cdot10^{-1} & 3.69\cdot10^{-2} &       &       & CB2   & 0.000    & 9.60\cdot10^{-1} & 4.80\cdot10^{-1} & 1.88\cdot10^{-2} \\*
          & AC1   & 3.97\cdot10^{-2} & 4.13\cdot10^{-1} & 2.26\cdot10^{-1} & 5.40\cdot10^{-2} &       &       & AC1   & 3.97\cdot10^{-2} & 4.13\cdot10^{-1} & 2.26\cdot10^{-1} & 4.70\cdot10^{-2} \\*
          & AC2   & 0.000    & 9.84\cdot10^{-1} & 4.92\cdot10^{-1} & 3.79\cdot10^{-2} &       &       & AC2   & 0.000    & 9.68\cdot10^{-1} & 4.84\cdot10^{-1} & 1.90\cdot10^{-2} \\*
          & AC3   & 0.000    & 1.000    & 5.00\cdot10^{-1} & 3.85\cdot10^{-2} &       &       & AC3   & 0.000    & 1.000    & 5.00\cdot10^{-1} & 1.96\cdot10^{-2} \\*
          & CSA   & 1.67\cdot10^{-1} & 1.43\cdot10^{-1} & 1.55\cdot10^{-1} & 1.66\cdot10^{-1} &       &       & CSA   & 2.46\cdot10^{-1} & 1.75\cdot10^{-1} & 2.10\cdot10^{-1} & 2.45\cdot10^{-1} \\*
          & CGA   & 2.38\cdot10^{-2} & 3.57\cdot10^{-1} & 1.90\cdot10^{-1} & 3.66\cdot10^{-2} &       &       & CGA   & 2.38\cdot10^{-2} & 3.33\cdot10^{-1} & 1.79\cdot10^{-1} & 2.99\cdot10^{-2} \\*
     \cmidrule(r){1-6} \cmidrule(r){8-13} 
		$[100, 1]$ & ABT   & 7.14\cdot10^{-2} & 2.70\cdot10^{-1} & 1.71\cdot10^{-1} & 7.34\cdot10^{-2} &       &       &       &       &       &       &  \\*
          & ASB   & 3.97\cdot10^{-2} & 2.78\cdot10^{-1} & 1.59\cdot10^{-1} & 4.20\cdot10^{-2} &       &       &       &       &       &       &  \\*
          & ADC   & 3.97\cdot10^{-2} & 4.13\cdot10^{-1} & 2.26\cdot10^{-1} & 4.34\cdot10^{-2} &       &       &       &       &       &       &  \\*
          & CB0   & 3.97\cdot10^{-2} & 4.13\cdot10^{-1} & 2.26\cdot10^{-1} & 4.34\cdot10^{-2} &       &       &       &       &       &       &  \\*
          & CB1   & 0.000    & 9.60\cdot10^{-1} & 4.80\cdot10^{-1} & 9.51\cdot10^{-3} &       &       &       &       &       &       &  \\*
          & CB2   & 0.000    & 9.60\cdot10^{-1} & 4.80\cdot10^{-1} & 9.51\cdot10^{-3} &       &       &       &       &       &       &  \\*
          & AC1   & 3.97\cdot10^{-2} & 4.13\cdot10^{-1} & 2.26\cdot10^{-1} & 4.34\cdot10^{-2} &       &       &       &       &       &       &  \\*
          & AC2   & 0.000    & 9.21\cdot10^{-1} & 4.60\cdot10^{-1} & 9.12\cdot10^{-3} &       &       &       &       &       &       &  \\*
          & AC3   & 0.000    & 9.21\cdot10^{-1} & 4.60\cdot10^{-1} & 9.12\cdot10^{-3} &       &       &       &       &       &       &  \\*
          & CSA   & 2.14\cdot10^{-1} & 2.86\cdot10^{-1} & 2.50\cdot10^{-1} & 2.15\cdot10^{-1} &       &       &       &       &       &       &  \\*
          & CGA   & 2.38\cdot10^{-2} & 3.73\cdot10^{-1} & 1.98\cdot10^{-1} & 2.73\cdot10^{-2} &       &       &       &       &       &       &  \\*

\end{longtabu}
}

\newpage

%%%%%%%%%%%%%%%%%%%%%%
%%% UCI IONOSPHERE %%%
%%%%%%%%%%%%%%%%%%%%%%

\begin{landscape}
% GRAPHICS
\begin{figure}[p]
\centering
\includegraphics[width=.6\paperheight]{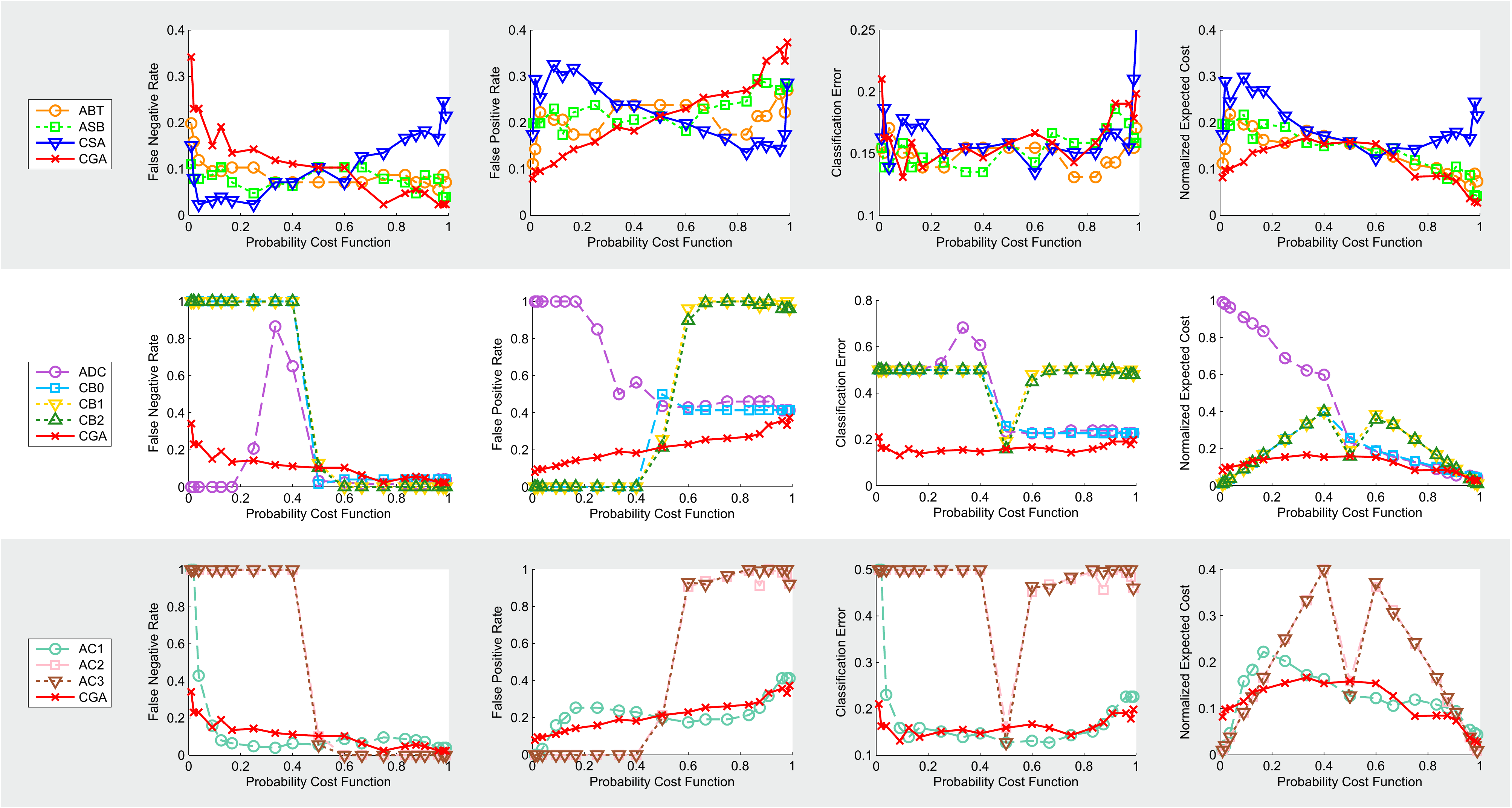}
\caption[Results obtained for the UCI Ionosphere Dataset.]{Results obtained for the UCI Ionosphere Dataset. First column of the illustration corresponds to False Negative Rate, second column to False Positive Rate, third column to Classification Error and the fourth one corresponds to Normalized Expected Cost. For a clearer visualization, algorithms have been divided into three groups so each row of the illustration corresponds to a different group. Cost Generalized AdaBoost is plotted in all the graphs to have a common reference across the representations.}
\label{fig:ionosphere_perform} % caption for the whole figure
\end{figure}
\end{landscape}
\newpage

%%%%%%%%%%%%%%%%
%%% UCI SPAM %%%
%%%%%%%%%%%%%%%%

% TABLE

{
\tiny
\begin{longtabu} to \textwidth {llX[$l]@{}@{}X[$l]@{}@{}X[$l]@{}@{}X[$l]@{}@{}l@{}llX[$l]@{}@{}X[$l]@{}@{}X[$l]@{}@{}X[$l]@{}}

\toprule
{\textbf{Cost}}  & {\textbf{Alg}}   & {\textbf{FNR}}    & {\textbf{FPR}}    & {\textbf{CE}}    & {\textbf{NEC}}   && {\textbf{Cost}}  & {\textbf{Alg}}   & {\textbf{FNR}}    & {\textbf{FPR}}    & {\textbf{CE}} & {\textbf{NEC}}\\* 
\cmidrule(r){1-6} \cmidrule(r){8-13}
\endfirsthead

%\multicolumn{13}{l}% {\tiny{Continued from previous page}} \\*
\toprule
{\textbf{Cost}}  & {\textbf{Alg}}   & {\textbf{FNR}}    & {\textbf{FPR}}    & {\textbf{CE}}    & {\textbf{NEC}}   && {\textbf{Cost}}  & {\textbf{Alg}}   & {\textbf{FNR}}    & {\textbf{FPR}}    & {\textbf{CE}} & {\textbf{NEC}}\\* 
\cmidrule(r){1-6} \cmidrule(r){8-13}
\endhead

\multicolumn{13}{r}{{\tiny{\tablename\ \thetable{} - Continued on next page $\rightarrow$}}} \\* 
\endfoot

\bottomrule
  \caption{Results obtained for the UCI Spam Dataset.}
  \label{tab:spam_perform}%
\endlastfoot

    $[1, 100]$ & ABT   & 1.87\cdot10^{-1} & 2.26\cdot10^{-2} & 1.05\cdot10^{-1} & 2.42\cdot10^{-2} &       & $[1, 50]$ & ABT   & 1.01\cdot10^{-1} & 4.47\cdot10^{-2} & 7.28\cdot10^{-2} & 4.58\cdot10^{-2} \\*
          & ASB   & 8.61\cdot10^{-2} & 4.53\cdot10^{-2} & 6.57\cdot10^{-2} & 4.57\cdot10^{-2} &       &       & ASB   & 8.28\cdot10^{-2} & 4.97\cdot10^{-2} & 6.62\cdot10^{-2} & 5.03\cdot10^{-2} \\*
          & ADC   & 7.15\cdot10^{-1} & 1.10\cdot10^{-3} & 3.58\cdot10^{-1} & 8.17\cdot10^{-3} &       &       & ADC   & 7.15\cdot10^{-1} & 1.10\cdot10^{-3} & 3.58\cdot10^{-1} & 1.51\cdot10^{-2} \\*
          & CB0   & 7.15\cdot10^{-1} & 1.10\cdot10^{-3} & 3.58\cdot10^{-1} & 8.17\cdot10^{-3} &       &       & CB0   & 7.15\cdot10^{-1} & 1.10\cdot10^{-3} & 3.58\cdot10^{-1} & 1.51\cdot10^{-2} \\*
          & CB1   & 9.75\cdot10^{-1} & 0.000    & 4.87\cdot10^{-1} & 9.65\cdot10^{-3} &       &       & CB1   & 9.96\cdot10^{-1} & 0.000    & 4.98\cdot10^{-1} & 1.95\cdot10^{-2} \\*
          & CB2   & 9.47\cdot10^{-1} & 0.000    & 4.74\cdot10^{-1} & 9.38\cdot10^{-3} &       &       & CB2   & 9.47\cdot10^{-1} & 0.000    & 4.74\cdot10^{-1} & 1.86\cdot10^{-2} \\*
          & AC1   & 6.21\cdot10^{-1} & 1.10\cdot10^{-3} & 3.11\cdot10^{-1} & 7.24\cdot10^{-3} &       &       & AC1   & 4.33\cdot10^{-1} & 4.42\cdot10^{-3} & 2.19\cdot10^{-1} & 1.28\cdot10^{-2} \\*
          & AC2   & 8.98\cdot10^{-1} & 0.000    & 4.49\cdot10^{-1} & 8.90\cdot10^{-3} &       &       & AC2   & 9.94\cdot10^{-1} & 0.000    & 4.97\cdot10^{-1} & 1.95\cdot10^{-2} \\*
          & AC3   & 9.39\cdot10^{-1} & 0.000    & 4.70\cdot10^{-1} & 9.30\cdot10^{-3} &       &       & AC3   & 9.66\cdot10^{-1} & 0.000    & 4.83\cdot10^{-1} & 1.89\cdot10^{-2} \\*
          & CSA   & 1.56\cdot10^{-1} & 4.42\cdot10^{-2} & 1.00\cdot10^{-1} & 4.53\cdot10^{-2} &       &       & CSA   & 8.66\cdot10^{-2} & 7.34\cdot10^{-2} & 8.00\cdot10^{-2} & 7.37\cdot10^{-2} \\*
          & CGA   & 2.94\cdot10^{-1} & 1.32\cdot10^{-2} & 1.53\cdot10^{-1} & 1.60\cdot10^{-2} &       &       & CGA   & 1.77\cdot10^{-1} & 2.59\cdot10^{-2} & 1.02\cdot10^{-1} & 2.89\cdot10^{-2} \\*
     \cmidrule(r){1-6} \cmidrule(r){8-13} 
		$[1, 25]$ & ABT   & 8.33\cdot10^{-2} & 5.30\cdot10^{-2} & 6.82\cdot10^{-2} & 5.41\cdot10^{-2} &       & $[1, 10]$ & ABT   & 6.73\cdot10^{-2} & 6.68\cdot10^{-2} & 6.71\cdot10^{-2} & 6.68\cdot10^{-2} \\*
          & ASB   & 7.84\cdot10^{-2} & 5.68\cdot10^{-2} & 6.76\cdot10^{-2} & 5.77\cdot10^{-2} &       &       & ASB   & 7.51\cdot10^{-2} & 5.63\cdot10^{-2} & 6.57\cdot10^{-2} & 5.80\cdot10^{-2} \\*
          & ADC   & 7.15\cdot10^{-1} & 1.10\cdot10^{-3} & 3.58\cdot10^{-1} & 2.85\cdot10^{-2} &       &       & ADC   & 6.31\cdot10^{-1} & 1.43\cdot10^{-2} & 3.23\cdot10^{-1} & 7.04\cdot10^{-2} \\*
          & CB0   & 8.23\cdot10^{-1} & 5.52\cdot10^{-4} & 4.12\cdot10^{-1} & 3.22\cdot10^{-2} &       &       & CB0   & 8.18\cdot10^{-1} & 5.52\cdot10^{-4} & 4.09\cdot10^{-1} & 7.49\cdot10^{-2} \\*
          & CB1   & 9.87\cdot10^{-1} & 0.000    & 4.93\cdot10^{-1} & 3.80\cdot10^{-2} &       &       & CB1   & 9.95\cdot10^{-1} & 0.000    & 4.98\cdot10^{-1} & 9.05\cdot10^{-2} \\*
          & CB2   & 1.000    & 0.000    & 5.00\cdot10^{-1} & 3.85\cdot10^{-2} &       &       & CB2   & 9.98\cdot10^{-1} & 0.000    & 4.99\cdot10^{-1} & 9.07\cdot10^{-2} \\*
          & AC1   & 2.35\cdot10^{-1} & 2.15\cdot10^{-2} & 1.28\cdot10^{-1} & 2.97\cdot10^{-2} &       &       & AC1   & 1.11\cdot10^{-1} & 4.19\cdot10^{-2} & 7.67\cdot10^{-2} & 4.83\cdot10^{-2} \\*
          & AC2   & 9.91\cdot10^{-1} & 0.000    & 4.95\cdot10^{-1} & 3.81\cdot10^{-2} &       &       & AC2   & 9.98\cdot10^{-1} & 0.000    & 4.99\cdot10^{-1} & 9.07\cdot10^{-2} \\*
          & AC3   & 9.74\cdot10^{-1} & 0.000    & 4.87\cdot10^{-1} & 3.74\cdot10^{-2} &       &       & AC3   & 9.87\cdot10^{-1} & 0.000    & 4.93\cdot10^{-1} & 8.97\cdot10^{-2} \\*
          & CSA   & 6.40\cdot10^{-2} & 7.95\cdot10^{-2} & 7.17\cdot10^{-2} & 7.89\cdot10^{-2} &       &       & CSA   & 5.08\cdot10^{-2} & 9.38\cdot10^{-2} & 7.23\cdot10^{-2} & 8.99\cdot10^{-2} \\*
          & CGA   & 1.24\cdot10^{-1} & 4.08\cdot10^{-2} & 8.25\cdot10^{-2} & 4.40\cdot10^{-2} &       &       & CGA   & 9.11\cdot10^{-2} & 4.86\cdot10^{-2} & 6.98\cdot10^{-2} & 5.24\cdot10^{-2} \\*
     \cmidrule(r){1-6} \cmidrule(r){8-13} 
		$[1, 7]$ & ABT   & 6.46\cdot10^{-2} & 7.23\cdot10^{-2} & 6.84\cdot10^{-2} & 7.13\cdot10^{-2} &       & $[1, 5]$ & ABT   & 6.51\cdot10^{-2} & 7.23\cdot10^{-2} & 6.87\cdot10^{-2} & 7.11\cdot10^{-2} \\*
          & ASB   & 7.28\cdot10^{-2} & 5.91\cdot10^{-2} & 6.59\cdot10^{-2} & 6.08\cdot10^{-2} &       &       & ASB   & 7.06\cdot10^{-2} & 6.24\cdot10^{-2} & 6.65\cdot10^{-2} & 6.37\cdot10^{-2} \\*
          & ADC   & 6.31\cdot10^{-1} & 1.43\cdot10^{-2} & 3.23\cdot10^{-1} & 9.14\cdot10^{-2} &       &       & ADC   & 5.70\cdot10^{-1} & 1.82\cdot10^{-2} & 2.94\cdot10^{-1} & 1.10\cdot10^{-1} \\*
          & CB0   & 8.13\cdot10^{-1} & 3.31\cdot10^{-3} & 4.08\cdot10^{-1} & 1.05\cdot10^{-1} &       &       & CB0   & 8.13\cdot10^{-1} & 3.31\cdot10^{-3} & 4.08\cdot10^{-1} & 1.38\cdot10^{-1} \\*
          & CB1   & 9.98\cdot10^{-1} & 0.000    & 4.99\cdot10^{-1} & 1.25\cdot10^{-1} &       &       & CB1   & 9.93\cdot10^{-1} & 0.000    & 4.97\cdot10^{-1} & 1.66\cdot10^{-1} \\*
          & CB2   & 9.99\cdot10^{-1} & 0.000    & 5.00\cdot10^{-1} & 1.25\cdot10^{-1} &       &       & CB2   & 9.98\cdot10^{-1} & 0.000    & 4.99\cdot10^{-1} & 1.66\cdot10^{-1} \\*
          & AC1   & 9.71\cdot10^{-2} & 5.30\cdot10^{-2} & 7.51\cdot10^{-2} & 5.85\cdot10^{-2} &       &       & AC1   & 7.40\cdot10^{-2} & 6.02\cdot10^{-2} & 6.71\cdot10^{-2} & 6.25\cdot10^{-2} \\*
          & AC2   & 9.94\cdot10^{-1} & 0.000    & 4.97\cdot10^{-1} & 1.24\cdot10^{-1} &       &       & AC2   & 9.84\cdot10^{-1} & 0.000    & 4.92\cdot10^{-1} & 1.64\cdot10^{-1} \\*
          & AC3   & 9.92\cdot10^{-1} & 0.000    & 4.96\cdot10^{-1} & 1.24\cdot10^{-1} &       &       & AC3   & 9.97\cdot10^{-1} & 0.000    & 4.99\cdot10^{-1} & 1.66\cdot10^{-1} \\*
          & CSA   & 5.02\cdot10^{-2} & 9.55\cdot10^{-2} & 7.28\cdot10^{-2} & 8.98\cdot10^{-2} &       &       & CSA   & 4.75\cdot10^{-2} & 9.82\cdot10^{-2} & 7.28\cdot10^{-2} & 8.98\cdot10^{-2} \\*
          & CGA   & 8.17\cdot10^{-2} & 5.19\cdot10^{-2} & 6.68\cdot10^{-2} & 5.56\cdot10^{-2} &       &       & CGA   & 7.73\cdot10^{-2} & 5.79\cdot10^{-2} & 6.76\cdot10^{-2} & 6.12\cdot10^{-2} \\*
     \cmidrule(r){1-6} \cmidrule(r){8-13} 
		$[1, 3]$ & ABT   & 6.24\cdot10^{-2} & 7.45\cdot10^{-2} & 6.84\cdot10^{-2} & 7.15\cdot10^{-2} &       & $[1, 2]$ & ABT   & 5.96\cdot10^{-2} & 7.89\cdot10^{-2} & 6.93\cdot10^{-2} & 7.25\cdot10^{-2} \\*
          & ASB   & 6.73\cdot10^{-2} & 6.51\cdot10^{-2} & 6.62\cdot10^{-2} & 6.57\cdot10^{-2} &       &       & ASB   & 6.46\cdot10^{-2} & 6.95\cdot10^{-2} & 6.71\cdot10^{-2} & 6.79\cdot10^{-2} \\*
          & ADC   & 5.60\cdot10^{-1} & 1.71\cdot10^{-2} & 2.88\cdot10^{-1} & 1.53\cdot10^{-1} &       &       & ADC   & 5.03\cdot10^{-1} & 3.15\cdot10^{-2} & 2.67\cdot10^{-1} & 1.89\cdot10^{-1} \\*
          & CB0   & 8.13\cdot10^{-1} & 3.31\cdot10^{-3} & 4.08\cdot10^{-1} & 2.06\cdot10^{-1} &       &       & CB0   & 8.14\cdot10^{-1} & 3.31\cdot10^{-3} & 4.09\cdot10^{-1} & 2.74\cdot10^{-1} \\*
          & CB1   & 9.99\cdot10^{-1} & 0.000    & 5.00\cdot10^{-1} & 2.50\cdot10^{-1} &       &       & CB1   & 1.000    & 0.000    & 5.00\cdot10^{-1} & 3.33\cdot10^{-1} \\*
          & CB2   & 1.000    & 0.000    & 5.00\cdot10^{-1} & 2.50\cdot10^{-1} &       &       & CB2   & 9.98\cdot10^{-1} & 0.000    & 4.99\cdot10^{-1} & 3.33\cdot10^{-1} \\*
          & AC1   & 5.68\cdot10^{-2} & 6.18\cdot10^{-2} & 5.93\cdot10^{-2} & 6.06\cdot10^{-2} &       &       & AC1   & 5.02\cdot10^{-2} & 6.46\cdot10^{-2} & 5.74\cdot10^{-2} & 5.98\cdot10^{-2} \\*
          & AC2   & 9.99\cdot10^{-1} & 0.000    & 4.99\cdot10^{-1} & 2.50\cdot10^{-1} &       &       & AC2   & 9.98\cdot10^{-1} & 0.000    & 4.99\cdot10^{-1} & 3.33\cdot10^{-1} \\*
          & AC3   & 9.97\cdot10^{-1} & 0.000    & 4.99\cdot10^{-1} & 2.49\cdot10^{-1} &       &       & AC3   & 9.98\cdot10^{-1} & 0.000    & 4.99\cdot10^{-1} & 3.33\cdot10^{-1} \\*
          & CSA   & 4.75\cdot10^{-2} & 9.44\cdot10^{-2} & 7.09\cdot10^{-2} & 8.26\cdot10^{-2} &       &       & CSA   & 4.75\cdot10^{-2} & 8.89\cdot10^{-2} & 6.82\cdot10^{-2} & 7.51\cdot10^{-2} \\*
          & CGA   & 7.06\cdot10^{-2} & 6.24\cdot10^{-2} & 6.65\cdot10^{-2} & 6.44\cdot10^{-2} &       &       & CGA   & 6.35\cdot10^{-2} & 6.73\cdot10^{-2} & 6.54\cdot10^{-2} & 6.60\cdot10^{-2} \\*
     \cmidrule(r){1-6} \cmidrule(r){8-13} 
		$[2, 3]$ & ABT   & 6.02\cdot10^{-2} & 7.89\cdot10^{-2} & 6.95\cdot10^{-2} & 7.14\cdot10^{-2} &       & $[1, 1]$ & ABT   & 5.96\cdot10^{-2} & 7.84\cdot10^{-2} & 6.90\cdot10^{-2} & 6.90\cdot10^{-2} \\*
          & ASB   & 6.13\cdot10^{-2} & 7.23\cdot10^{-2} & 6.68\cdot10^{-2} & 6.79\cdot10^{-2} &       &       & ASB   & 5.85\cdot10^{-2} & 7.89\cdot10^{-2} & 6.87\cdot10^{-2} & 6.87\cdot10^{-2} \\*
          & ADC   & 2.29\cdot10^{-1} & 1.67\cdot10^{-1} & 1.98\cdot10^{-1} & 1.92\cdot10^{-1} &       &       & ADC   & 1.91\cdot10^{-1} & 2.18\cdot10^{-1} & 2.04\cdot10^{-1} & 2.04\cdot10^{-1} \\*
          & CB0   & 8.14\cdot10^{-1} & 2.76\cdot10^{-3} & 4.08\cdot10^{-1} & 3.27\cdot10^{-1} &       &       & CB0   & 1.91\cdot10^{-1} & 2.18\cdot10^{-1} & 2.04\cdot10^{-1} & 2.04\cdot10^{-1} \\*
          & CB1   & 1.000    & 0.000    & 5.00\cdot10^{-1} & 4.00\cdot10^{-1} &       &       & CB1   & 2.72\cdot10^{-1} & 4.06\cdot10^{-1} & 3.39\cdot10^{-1} & 3.39\cdot10^{-1} \\*
          & CB2   & 9.99\cdot10^{-1} & 0.000    & 4.99\cdot10^{-1} & 4.00\cdot10^{-1} &       &       & CB2   & 5.85\cdot10^{-2} & 7.89\cdot10^{-2} & 6.87\cdot10^{-2} & 6.87\cdot10^{-2} \\*
          & AC1   & 4.97\cdot10^{-2} & 6.29\cdot10^{-2} & 5.63\cdot10^{-2} & 5.76\cdot10^{-2} &       &       & AC1   & 5.24\cdot10^{-2} & 6.79\cdot10^{-2} & 6.02\cdot10^{-2} & 6.02\cdot10^{-2} \\*
          & AC2   & 9.98\cdot10^{-1} & 0.000    & 4.99\cdot10^{-1} & 3.99\cdot10^{-1} &       &       & AC2   & 5.85\cdot10^{-2} & 7.89\cdot10^{-2} & 6.87\cdot10^{-2} & 6.87\cdot10^{-2} \\*
          & AC3   & 9.97\cdot10^{-1} & 0.000    & 4.99\cdot10^{-1} & 3.99\cdot10^{-1} &       &       & AC3   & 5.24\cdot10^{-2} & 6.79\cdot10^{-2} & 6.02\cdot10^{-2} & 6.02\cdot10^{-2} \\*
          & CSA   & 5.30\cdot10^{-2} & 8.39\cdot10^{-2} & 6.84\cdot10^{-2} & 7.15\cdot10^{-2} &       &       & CSA   & 5.85\cdot10^{-2} & 7.89\cdot10^{-2} & 6.87\cdot10^{-2} & 6.87\cdot10^{-2} \\*
          & CGA   & 6.24\cdot10^{-2} & 7.45\cdot10^{-2} & 6.84\cdot10^{-2} & 6.96\cdot10^{-2} &       &       & CGA   & 5.85\cdot10^{-2} & 7.89\cdot10^{-2} & 6.87\cdot10^{-2} & 6.87\cdot10^{-2} \\*
     \cmidrule(r){1-6} \cmidrule(r){8-13} 
		
		$[3, 2]$ & ABT   & 5.85\cdot10^{-2} & 7.89\cdot10^{-2} & 6.87\cdot10^{-2} & 6.67\cdot10^{-2} &       & $[2, 1]$ & ABT   & 5.46\cdot10^{-2} & 8.00\cdot10^{-2} & 6.73\cdot10^{-2} & 6.31\cdot10^{-2} \\*
          & ASB   & 5.63\cdot10^{-2} & 7.95\cdot10^{-2} & 6.79\cdot10^{-2} & 6.56\cdot10^{-2} &       &       & ASB   & 5.35\cdot10^{-2} & 8.33\cdot10^{-2} & 6.84\cdot10^{-2} & 6.35\cdot10^{-2} \\*
          & ADC   & 1.56\cdot10^{-1} & 2.31\cdot10^{-1} & 1.93\cdot10^{-1} & 1.86\cdot10^{-1} &       &       & ADC   & 2.32\cdot10^{-2} & 4.06\cdot10^{-1} & 2.14\cdot10^{-1} & 1.51\cdot10^{-1} \\*
          & CB0   & 6.07\cdot10^{-3} & 9.29\cdot10^{-1} & 4.67\cdot10^{-1} & 3.75\cdot10^{-1} &       &       & CB0   & 6.07\cdot10^{-3} & 9.28\cdot10^{-1} & 4.67\cdot10^{-1} & 3.13\cdot10^{-1} \\*
          & CB1   & 0.000    & 1.000    & 5.00\cdot10^{-1} & 4.00\cdot10^{-1} &       &       & CB1   & 0.000    & 9.99\cdot10^{-1} & 4.99\cdot10^{-1} & 3.33\cdot10^{-1} \\*
          & CB2   & 0.000    & 9.97\cdot10^{-1} & 4.99\cdot10^{-1} & 3.99\cdot10^{-1} &       &       & CB2   & 0.000    & 9.99\cdot10^{-1} & 4.99\cdot10^{-1} & 3.33\cdot10^{-1} \\*
          & AC1   & 5.19\cdot10^{-2} & 6.90\cdot10^{-2} & 6.04\cdot10^{-2} & 5.87\cdot10^{-2} &       &       & AC1   & 5.02\cdot10^{-2} & 7.01\cdot10^{-2} & 6.02\cdot10^{-2} & 5.68\cdot10^{-2} \\*
          & AC2   & 0.000    & 9.99\cdot10^{-1} & 4.99\cdot10^{-1} & 4.00\cdot10^{-1} &       &       & AC2   & 5.52\cdot10^{-4} & 9.97\cdot10^{-1} & 4.99\cdot10^{-1} & 3.33\cdot10^{-1} \\*
          & AC3   & 0.000    & 1.000    & 5.00\cdot10^{-1} & 4.00\cdot10^{-1} &       &       & AC3   & 0.000    & 1.000    & 5.00\cdot10^{-1} & 3.33\cdot10^{-1} \\*
          & CSA   & 6.24\cdot10^{-2} & 7.12\cdot10^{-2} & 6.68\cdot10^{-2} & 6.59\cdot10^{-2} &       &       & CSA   & 6.46\cdot10^{-2} & 6.29\cdot10^{-2} & 6.37\cdot10^{-2} & 6.40\cdot10^{-2} \\*
          & CGA   & 5.35\cdot10^{-2} & 8.11\cdot10^{-2} & 6.73\cdot10^{-2} & 6.46\cdot10^{-2} &       &       & CGA   & 5.24\cdot10^{-2} & 8.28\cdot10^{-2} & 6.76\cdot10^{-2} & 6.25\cdot10^{-2} \\*
     \cmidrule(r){1-6} \cmidrule(r){8-13} 
		$[3, 1]$ & ABT   & 5.46\cdot10^{-2} & 8.06\cdot10^{-2} & 6.76\cdot10^{-2} & 6.11\cdot10^{-2} &       & $[5, 1]$ & ABT   & 5.24\cdot10^{-2} & 8.06\cdot10^{-2} & 6.65\cdot10^{-2} & 5.71\cdot10^{-2} \\*
          & ASB   & 5.08\cdot10^{-2} & 8.33\cdot10^{-2} & 6.71\cdot10^{-2} & 5.89\cdot10^{-2} &       &       & ASB   & 4.97\cdot10^{-2} & 8.55\cdot10^{-2} & 6.76\cdot10^{-2} & 5.56\cdot10^{-2} \\*
          & ADC   & 1.32\cdot10^{-2} & 4.56\cdot10^{-1} & 2.35\cdot10^{-1} & 1.24\cdot10^{-1} &       &       & ADC   & 1.27\cdot10^{-2} & 5.64\cdot10^{-1} & 2.88\cdot10^{-1} & 1.05\cdot10^{-1} \\*
          & CB0   & 6.07\cdot10^{-3} & 9.29\cdot10^{-1} & 4.67\cdot10^{-1} & 2.37\cdot10^{-1} &       &       & CB0   & 6.07\cdot10^{-3} & 9.29\cdot10^{-1} & 4.68\cdot10^{-1} & 1.60\cdot10^{-1} \\*
          & CB1   & 0.000    & 9.99\cdot10^{-1} & 5.00\cdot10^{-1} & 2.50\cdot10^{-1} &       &       & CB1   & 0.000    & 1.000    & 5.00\cdot10^{-1} & 1.67\cdot10^{-1} \\*
          & CB2   & 0.000    & 1.000    & 5.00\cdot10^{-1} & 2.50\cdot10^{-1} &       &       & CB2   & 0.000    & 9.97\cdot10^{-1} & 4.98\cdot10^{-1} & 1.66\cdot10^{-1} \\*
          & AC1   & 4.97\cdot10^{-2} & 7.17\cdot10^{-2} & 6.07\cdot10^{-2} & 5.52\cdot10^{-2} &       &       & AC1   & 4.80\cdot10^{-2} & 8.66\cdot10^{-2} & 6.73\cdot10^{-2} & 5.45\cdot10^{-2} \\*
          & AC2   & 0.000    & 9.93\cdot10^{-1} & 4.96\cdot10^{-1} & 2.48\cdot10^{-1} &       &       & AC2   & 0.000    & 9.90\cdot10^{-1} & 4.95\cdot10^{-1} & 1.65\cdot10^{-1} \\*
          & AC3   & 0.000    & 9.99\cdot10^{-1} & 4.99\cdot10^{-1} & 2.50\cdot10^{-1} &       &       & AC3   & 0.000    & 1.000    & 5.00\cdot10^{-1} & 1.67\cdot10^{-1} \\*
          & CSA   & 7.23\cdot10^{-2} & 5.57\cdot10^{-2} & 6.40\cdot10^{-2} & 6.82\cdot10^{-2} &       &       & CSA   & 7.56\cdot10^{-2} & 5.68\cdot10^{-2} & 6.62\cdot10^{-2} & 7.25\cdot10^{-2} \\*
          & CGA   & 4.97\cdot10^{-2} & 8.50\cdot10^{-2} & 6.73\cdot10^{-2} & 5.85\cdot10^{-2} &       &       & CGA   & 4.25\cdot10^{-2} & 9.55\cdot10^{-2} & 6.90\cdot10^{-2} & 5.13\cdot10^{-2} \\*
     \cmidrule(r){1-6} \cmidrule(r){8-13} 
		$[7, 1]$ & ABT   & 5.13\cdot10^{-2} & 8.11\cdot10^{-2} & 6.62\cdot10^{-2} & 5.50\cdot10^{-2} &       & $[10, 1]$ & ABT   & 4.80\cdot10^{-2} & 8.28\cdot10^{-2} & 6.54\cdot10^{-2} & 5.12\cdot10^{-2} \\*
          & ASB   & 4.53\cdot10^{-2} & 8.61\cdot10^{-2} & 6.57\cdot10^{-2} & 5.04\cdot10^{-2} &       &       & ASB   & 4.42\cdot10^{-2} & 8.61\cdot10^{-2} & 6.51\cdot10^{-2} & 4.80\cdot10^{-2} \\*
          & ADC   & 1.27\cdot10^{-2} & 5.86\cdot10^{-1} & 2.99\cdot10^{-1} & 8.43\cdot10^{-2} &       &       & ADC   & 1.05\cdot10^{-2} & 6.36\cdot10^{-1} & 3.23\cdot10^{-1} & 6.73\cdot10^{-2} \\*
          & CB0   & 6.07\cdot10^{-3} & 9.29\cdot10^{-1} & 4.68\cdot10^{-1} & 1.21\cdot10^{-1} &       &       & CB0   & 6.07\cdot10^{-3} & 9.36\cdot10^{-1} & 4.71\cdot10^{-1} & 9.06\cdot10^{-2} \\*
          & CB1   & 0.000    & 1.000    & 5.00\cdot10^{-1} & 1.25\cdot10^{-1} &       &       & CB1   & 0.000    & 9.94\cdot10^{-1} & 4.97\cdot10^{-1} & 9.04\cdot10^{-2} \\*
          & CB2   & 5.52\cdot10^{-4} & 9.93\cdot10^{-1} & 4.97\cdot10^{-1} & 1.25\cdot10^{-1} &       &       & CB2   & 0.000    & 9.93\cdot10^{-1} & 4.97\cdot10^{-1} & 9.03\cdot10^{-2} \\*
          & AC1   & 4.53\cdot10^{-2} & 9.66\cdot10^{-2} & 7.09\cdot10^{-2} & 5.17\cdot10^{-2} &       &       & AC1   & 3.64\cdot10^{-2} & 1.14\cdot10^{-1} & 7.51\cdot10^{-2} & 4.34\cdot10^{-2} \\*
          & AC2   & 0.000    & 9.98\cdot10^{-1} & 4.99\cdot10^{-1} & 1.25\cdot10^{-1} &       &       & AC2   & 0.000    & 9.98\cdot10^{-1} & 4.99\cdot10^{-1} & 9.08\cdot10^{-2} \\*
          & AC3   & 0.000    & 1.000    & 5.00\cdot10^{-1} & 1.25\cdot10^{-1} &       &       & AC3   & 0.000    & 1.000    & 5.00\cdot10^{-1} & 9.09\cdot10^{-2} \\*
          & CSA   & 7.23\cdot10^{-2} & 5.52\cdot10^{-2} & 6.37\cdot10^{-2} & 7.02\cdot10^{-2} &       &       & CSA   & 7.45\cdot10^{-2} & 6.24\cdot10^{-2} & 6.84\cdot10^{-2} & 7.34\cdot10^{-2} \\*
          & CGA   & 3.97\cdot10^{-2} & 9.66\cdot10^{-2} & 6.82\cdot10^{-2} & 4.68\cdot10^{-2} &       &       & CGA   & 3.59\cdot10^{-2} & 1.03\cdot10^{-1} & 6.93\cdot10^{-2} & 4.19\cdot10^{-2} \\*
     \cmidrule(r){1-6} \cmidrule(r){8-13} 
		$[25, 1]$ & ABT   & 4.03\cdot10^{-2} & 9.38\cdot10^{-2} & 6.71\cdot10^{-2} & 4.23\cdot10^{-2} &       & $[50, 1]$ & ABT   & 2.98\cdot10^{-2} & 1.20\cdot10^{-1} & 7.51\cdot10^{-2} & 3.16\cdot10^{-2} \\*
          & ASB   & 4.03\cdot10^{-2} & 8.83\cdot10^{-2} & 6.43\cdot10^{-2} & 4.21\cdot10^{-2} &       &       & ASB   & 3.86\cdot10^{-2} & 9.33\cdot10^{-2} & 6.59\cdot10^{-2} & 3.97\cdot10^{-2} \\*
          & ADC   & 1.43\cdot10^{-2} & 6.95\cdot10^{-1} & 3.55\cdot10^{-1} & 4.05\cdot10^{-2} &       &       & ADC   & 8.28\cdot10^{-3} & 8.82\cdot10^{-1} & 4.45\cdot10^{-1} & 2.54\cdot10^{-2} \\*
          & CB0   & 4.42\cdot10^{-3} & 9.45\cdot10^{-1} & 4.75\cdot10^{-1} & 4.06\cdot10^{-2} &       &       & CB0   & 8.28\cdot10^{-3} & 8.82\cdot10^{-1} & 4.45\cdot10^{-1} & 2.54\cdot10^{-2} \\*
          & CB1   & 0.000    & 9.70\cdot10^{-1} & 4.85\cdot10^{-1} & 3.73\cdot10^{-2} &       &       & CB1   & 0.000    & 9.59\cdot10^{-1} & 4.79\cdot10^{-1} & 1.88\cdot10^{-2} \\*
          & CB2   & 5.52\cdot10^{-4} & 9.81\cdot10^{-1} & 4.91\cdot10^{-1} & 3.83\cdot10^{-2} &       &       & CB2   & 0.000    & 9.79\cdot10^{-1} & 4.90\cdot10^{-1} & 1.92\cdot10^{-2} \\*
          & AC1   & 1.55\cdot10^{-2} & 2.43\cdot10^{-1} & 1.29\cdot10^{-1} & 2.42\cdot10^{-2} &       &       & AC1   & 1.10\cdot10^{-2} & 3.31\cdot10^{-1} & 1.71\cdot10^{-1} & 1.73\cdot10^{-2} \\*
          & AC2   & 0.000    & 9.83\cdot10^{-1} & 4.92\cdot10^{-1} & 3.78\cdot10^{-2} &       &       & AC2   & 0.000    & 9.93\cdot10^{-1} & 4.97\cdot10^{-1} & 1.95\cdot10^{-2} \\*
          & AC3   & 0.000    & 9.99\cdot10^{-1} & 4.99\cdot10^{-1} & 3.84\cdot10^{-2} &       &       & AC3   & 0.000    & 9.99\cdot10^{-1} & 5.00\cdot10^{-1} & 1.96\cdot10^{-2} \\*
          & CSA   & 6.62\cdot10^{-2} & 7.17\cdot10^{-2} & 6.90\cdot10^{-2} & 6.64\cdot10^{-2} &       &       & CSA   & 5.52\cdot10^{-2} & 9.55\cdot10^{-2} & 7.53\cdot10^{-2} & 5.60\cdot10^{-2} \\*
          & CGA   & 3.04\cdot10^{-2} & 1.31\cdot10^{-1} & 8.06\cdot10^{-2} & 3.42\cdot10^{-2} &       &       & CGA   & 2.10\cdot10^{-2} & 1.68\cdot10^{-1} & 9.46\cdot10^{-2} & 2.39\cdot10^{-2} \\*
     \cmidrule(r){1-6} \cmidrule(r){8-13} 
		$[100, 1]$ & ABT   & 2.10\cdot10^{-2} & 1.55\cdot10^{-1} & 8.80\cdot10^{-2} & 2.23\cdot10^{-2} &       &       &       &       &       &       &  \\*
          & ASB   & 3.75\cdot10^{-2} & 9.49\cdot10^{-2} & 6.62\cdot10^{-2} & 3.81\cdot10^{-2} &       &       &       &       &       &       &  \\*
          & ADC   & 6.62\cdot10^{-3} & 9.28\cdot10^{-1} & 4.67\cdot10^{-1} & 1.57\cdot10^{-2} &       &       &       &       &       &       &  \\*
          & CB0   & 6.62\cdot10^{-3} & 9.28\cdot10^{-1} & 4.67\cdot10^{-1} & 1.57\cdot10^{-2} &       &       &       &       &       &       &  \\*
          & CB1   & 6.62\cdot10^{-3} & 9.28\cdot10^{-1} & 4.67\cdot10^{-1} & 1.57\cdot10^{-2} &       &       &       &       &       &       &  \\*
          & CB2   & 6.62\cdot10^{-3} & 9.28\cdot10^{-1} & 4.67\cdot10^{-1} & 1.57\cdot10^{-2} &       &       &       &       &       &       &  \\*
          & AC1   & 7.17\cdot10^{-3} & 4.95\cdot10^{-1} & 2.51\cdot10^{-1} & 1.20\cdot10^{-2} &       &       &       &       &       &       &  \\*
          & AC2   & 6.62\cdot10^{-3} & 9.28\cdot10^{-1} & 4.67\cdot10^{-1} & 1.57\cdot10^{-2} &       &       &       &       &       &       &  \\*
          & AC3   & 6.62\cdot10^{-3} & 9.28\cdot10^{-1} & 4.67\cdot10^{-1} & 1.57\cdot10^{-2} &       &       &       &       &       &       &  \\*
          & CSA   & 3.48\cdot10^{-2} & 1.52\cdot10^{-1} & 9.33\cdot10^{-2} & 3.59\cdot10^{-2} &       &       &       &       &       &       &  \\*
          & CGA   & 1.60\cdot10^{-2} & 2.36\cdot10^{-1} & 1.26\cdot10^{-1} & 1.82\cdot10^{-2} &       &       &       &       &       &       &  \\*

\end{longtabu}
}

\newpage

%%%%%%%%%%%%%%%%
%%% UCI SPAM %%%
%%%%%%%%%%%%%%%%

\begin{landscape}
% GRAPHICS
\begin{figure}[p]
\centering
\includegraphics[width=.6\paperheight]{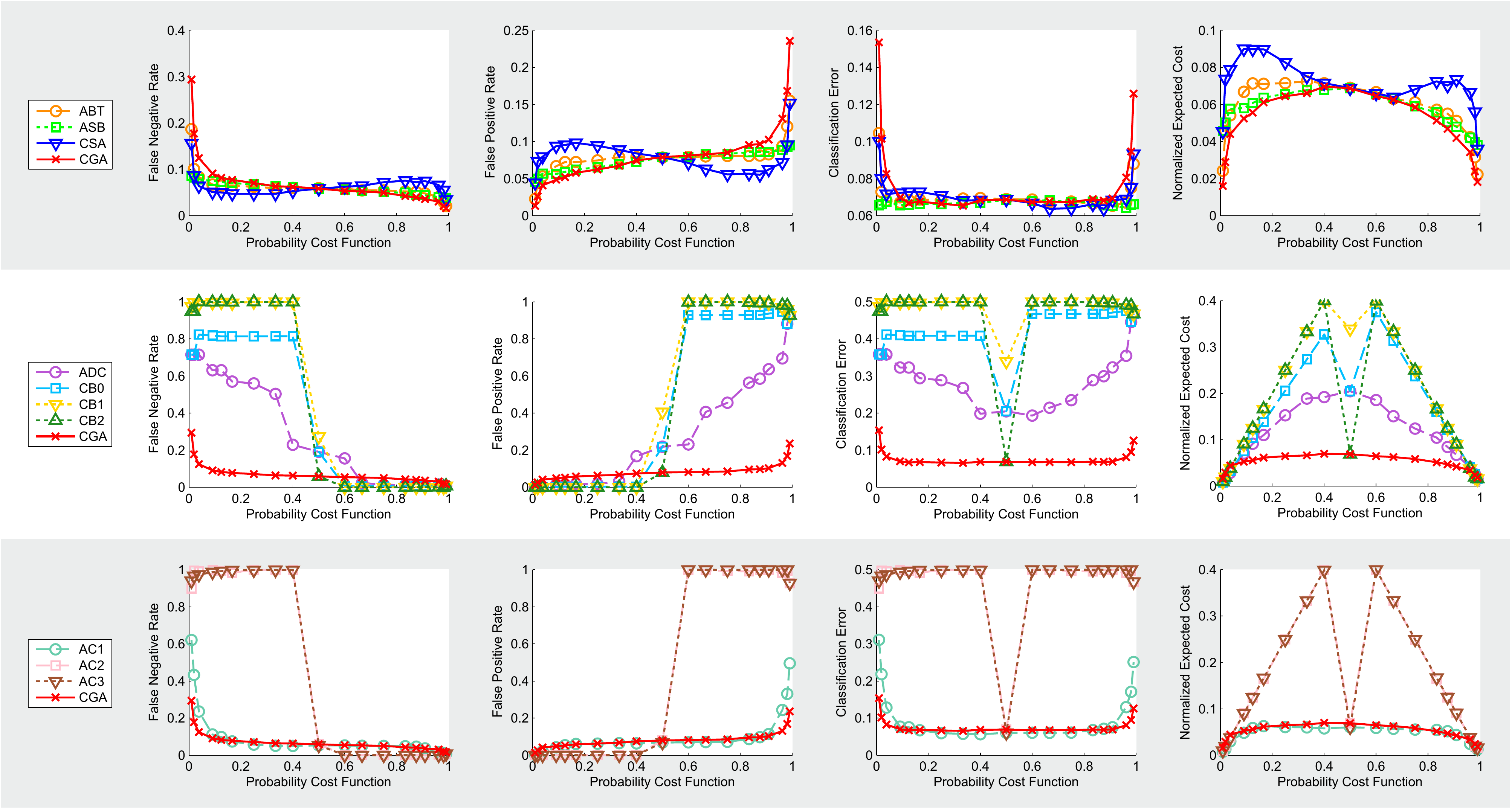}
\caption[Results obtained for the UCI Spam Dataset.]{Results obtained for the UCI Spam Dataset. First column of the illustration corresponds to False Negative Rate, second column to False Positive Rate, third column to Classification Error and the fourth one corresponds to Normalized Expected Cost. For a clearer visualization, algorithms have been divided into three groups so each row of the illustration corresponds to a different group. Cost Generalized AdaBoost is plotted in all the graphs to have a common reference across the representations.}
\label{fig:spam_perform} % caption for the whole figure
\end{figure}
\end{landscape}

%%%%%%%%%%%%
%%% CBCL %%%
%%%%%%%%%%%%

% TABLE

{
	
\tiny
\begin{longtabu} to \textwidth {llX[$l]@{}@{}X[$l]@{}@{}X[$l]@{}@{}X[$l]@{}@{}l@{}llX[$l]@{}@{}X[$l]@{}@{}X[$l]@{}@{}X[$l]@{}}

\toprule
{\textbf{Cost}}  & {\textbf{Alg}}   & {\textbf{FNR}}    & {\textbf{FPR}}    & {\textbf{CE}}    & {\textbf{NEC}}   && {\textbf{Cost}}  & {\textbf{Alg}}   & {\textbf{FNR}}    & {\textbf{FPR}}    & {\textbf{CE}} & {\textbf{NEC}}\\* 
\cmidrule(r){1-6} \cmidrule(r){8-13}
\endfirsthead

%\multicolumn{13}{l}% {\tiny{Continued from previous page}} \\*
\toprule
{\textbf{Cost}}  & {\textbf{Alg}}   & {\textbf{FNR}}    & {\textbf{FPR}}    & {\textbf{CE}}    & {\textbf{NEC}}   && {\textbf{Cost}}  & {\textbf{Alg}}   & {\textbf{FNR}}    & {\textbf{FPR}}    & {\textbf{CE}} & {\textbf{NEC}}\\* 
\cmidrule(r){1-6} \cmidrule(r){8-13}
\endhead

\multicolumn{13}{r}{{\tiny{\tablename\ \thetable{} - Continued on next page $\rightarrow$}}} \\* 
\endfoot

\bottomrule
    \caption{Results obtained for the CBCL Dataset.}
    \label{tab:cbcl_perform}%
\endlastfoot

    $[1, 100]$ & ABT   & 2.05\cdot10^{-1} & 3.10\cdot10^{-2} & 1.18\cdot10^{-1} & 3.28\cdot10^{-2} &       & $[1, 50]$ & ABT   & 1.75\cdot10^{-1} & 4.50\cdot10^{-2} & 1.10\cdot10^{-1} & 4.76\cdot10^{-2} \\*
          & ASB   & 1.00\cdot10^{-1} & 6.51\cdot10^{-2} & 8.26\cdot10^{-2} & 6.54\cdot10^{-2} &       &       & ASB   & 1.00\cdot10^{-1} & 6.51\cdot10^{-2} & 8.26\cdot10^{-2} & 6.58\cdot10^{-2} \\*
          & ADC   & 8.36\cdot10^{-1} & 6.01\cdot10^{-3} & 4.21\cdot10^{-1} & 1.42\cdot10^{-2} &       &       & ADC   & 8.36\cdot10^{-1} & 6.01\cdot10^{-3} & 4.21\cdot10^{-1} & 2.23\cdot10^{-2} \\*
          & CB0   & 8.36\cdot10^{-1} & 6.01\cdot10^{-3} & 4.21\cdot10^{-1} & 1.42\cdot10^{-2} &       &       & CB0   & 8.36\cdot10^{-1} & 6.01\cdot10^{-3} & 4.21\cdot10^{-1} & 2.23\cdot10^{-2} \\*
          & CB1   & 9.64\cdot10^{-1} & 0.000    & 4.82\cdot10^{-1} & 9.54\cdot10^{-3} &       &       & CB1   & 9.87\cdot10^{-1} & 0.000    & 4.93\cdot10^{-1} & 1.94\cdot10^{-2} \\*
          & CB2   & 9.48\cdot10^{-1} & 0.000    & 4.74\cdot10^{-1} & 9.39\cdot10^{-3} &       &       & CB2   & 9.87\cdot10^{-1} & 0.000    & 4.93\cdot10^{-1} & 1.94\cdot10^{-2} \\*
          & AC1   & 3.21\cdot10^{-1} & 3.00\cdot10^{-3} & 1.62\cdot10^{-1} & 6.15\cdot10^{-3} &       &       & AC1   & 1.61\cdot10^{-1} & 1.10\cdot10^{-2} & 8.61\cdot10^{-2} & 1.40\cdot10^{-2} \\*
          & AC2   & 9.30\cdot10^{-1} & 0.000    & 4.65\cdot10^{-1} & 9.21\cdot10^{-3} &       &       & AC2   & 9.71\cdot10^{-1} & 0.000    & 4.85\cdot10^{-1} & 1.90\cdot10^{-2} \\*
          & AC3   & 9.33\cdot10^{-1} & 0.000    & 4.66\cdot10^{-1} & 9.24\cdot10^{-3} &       &       & AC3   & 9.77\cdot10^{-1} & 0.000    & 4.88\cdot10^{-1} & 1.92\cdot10^{-2} \\*
          & CSA   & 4.00\cdot10^{-2} & 2.47\cdot10^{-1} & 1.44\cdot10^{-1} & 2.45\cdot10^{-1} &       &       & CSA   & 1.43\cdot10^{-1} & 7.41\cdot10^{-2} & 1.09\cdot10^{-1} & 7.54\cdot10^{-2} \\*
          & CGA   & 2.36\cdot10^{-1} & 1.50\cdot10^{-2} & 1.26\cdot10^{-1} & 1.72\cdot10^{-2} &       &       & CGA   & 2.06\cdot10^{-1} & 1.40\cdot10^{-2} & 1.10\cdot10^{-1} & 1.78\cdot10^{-2} \\*
     \cmidrule(r){1-6} \cmidrule(r){8-13} 
		$[1, 25]$ & ABT   & 1.41\cdot10^{-1} & 6.51\cdot10^{-2} & 1.03\cdot10^{-1} & 6.80\cdot10^{-2} &       & $[1, 10]$ & ABT   & 1.06\cdot10^{-1} & 6.71\cdot10^{-2} & 8.66\cdot10^{-2} & 7.06\cdot10^{-2} \\*
          & ASB   & 9.41\cdot10^{-2} & 5.71\cdot10^{-2} & 7.56\cdot10^{-2} & 5.85\cdot10^{-2} &       &       & ASB   & 9.41\cdot10^{-2} & 5.71\cdot10^{-2} & 7.56\cdot10^{-2} & 6.04\cdot10^{-2} \\*
          & ADC   & 8.23\cdot10^{-1} & 1.00\cdot10^{-3} & 4.12\cdot10^{-1} & 3.26\cdot10^{-2} &       &       & ADC   & 7.14\cdot10^{-1} & 3.00\cdot10^{-3} & 3.58\cdot10^{-1} & 6.76\cdot10^{-2} \\*
          & CB0   & 8.36\cdot10^{-1} & 6.01\cdot10^{-3} & 4.21\cdot10^{-1} & 3.79\cdot10^{-2} &       &       & CB0   & 8.38\cdot10^{-1} & 6.01\cdot10^{-3} & 4.22\cdot10^{-1} & 8.16\cdot10^{-2} \\*
          & CB1   & 9.94\cdot10^{-1} & 0.000    & 4.97\cdot10^{-1} & 3.82\cdot10^{-2} &       &       & CB1   & 9.95\cdot10^{-1} & 0.000    & 4.97\cdot10^{-1} & 9.05\cdot10^{-2} \\*
          & CB2   & 9.96\cdot10^{-1} & 0.000    & 4.98\cdot10^{-1} & 3.83\cdot10^{-2} &       &       & CB2   & 1.000    & 0.000    & 5.00\cdot10^{-1} & 9.09\cdot10^{-2} \\*
          & AC1   & 8.31\cdot10^{-2} & 3.90\cdot10^{-2} & 6.11\cdot10^{-2} & 4.07\cdot10^{-2} &       &       & AC1   & 5.91\cdot10^{-2} & 8.61\cdot10^{-2} & 7.26\cdot10^{-2} & 8.36\cdot10^{-2} \\*
          & AC2   & 9.82\cdot10^{-1} & 0.000    & 4.91\cdot10^{-1} & 3.78\cdot10^{-2} &       &       & AC2   & 9.98\cdot10^{-1} & 0.000    & 4.99\cdot10^{-1} & 9.07\cdot10^{-2} \\*
          & AC3   & 9.76\cdot10^{-1} & 0.000    & 4.88\cdot10^{-1} & 3.75\cdot10^{-2} &       &       & AC3   & 9.96\cdot10^{-1} & 0.000    & 4.98\cdot10^{-1} & 9.05\cdot10^{-2} \\*
          & CSA   & 6.51\cdot10^{-2} & 1.09\cdot10^{-1} & 8.71\cdot10^{-2} & 1.07\cdot10^{-1} &       &       & CSA   & 6.91\cdot10^{-2} & 1.01\cdot10^{-1} & 8.51\cdot10^{-2} & 9.82\cdot10^{-2} \\*
          & CGA   & 1.77\cdot10^{-1} & 2.10\cdot10^{-2} & 9.91\cdot10^{-2} & 2.70\cdot10^{-2} &       &       & CGA   & 1.36\cdot10^{-1} & 2.60\cdot10^{-2} & 8.11\cdot10^{-2} & 3.60\cdot10^{-2} \\*
     \cmidrule(r){1-6} \cmidrule(r){8-13} 
		$[1, 7]$ & ABT   & 1.06\cdot10^{-1} & 6.71\cdot10^{-2} & 8.66\cdot10^{-2} & 7.19\cdot10^{-2} &       & $[1, 5]$ & ABT   & 1.10\cdot10^{-1} & 6.81\cdot10^{-2} & 8.91\cdot10^{-2} & 7.51\cdot10^{-2} \\*
          & ASB   & 9.41\cdot10^{-2} & 5.71\cdot10^{-2} & 7.56\cdot10^{-2} & 6.17\cdot10^{-2} &       &       & ASB   & 9.41\cdot10^{-2} & 5.71\cdot10^{-2} & 7.56\cdot10^{-2} & 6.32\cdot10^{-2} \\*
          & ADC   & 7.45\cdot10^{-1} & 1.00\cdot10^{-3} & 3.73\cdot10^{-1} & 9.40\cdot10^{-2} &       &       & ADC   & 6.33\cdot10^{-1} & 1.90\cdot10^{-2} & 3.26\cdot10^{-1} & 1.21\cdot10^{-1} \\*
          & CB0   & 8.40\cdot10^{-1} & 6.01\cdot10^{-3} & 4.23\cdot10^{-1} & 1.10\cdot10^{-1} &       &       & CB0   & 8.36\cdot10^{-1} & 5.01\cdot10^{-3} & 4.20\cdot10^{-1} & 1.43\cdot10^{-1} \\*
          & CB1   & 9.98\cdot10^{-1} & 0.000    & 4.99\cdot10^{-1} & 1.25\cdot10^{-1} &       &       & CB1   & 9.98\cdot10^{-1} & 0.000    & 4.99\cdot10^{-1} & 1.66\cdot10^{-1} \\*
          & CB2   & 1.000    & 0.000    & 5.00\cdot10^{-1} & 1.25\cdot10^{-1} &       &       & CB2   & 1.000    & 0.000    & 5.00\cdot10^{-1} & 1.67\cdot10^{-1} \\*
          & AC1   & 5.51\cdot10^{-2} & 9.31\cdot10^{-2} & 7.41\cdot10^{-2} & 8.83\cdot10^{-2} &       &       & AC1   & 6.11\cdot10^{-2} & 9.31\cdot10^{-2} & 7.71\cdot10^{-2} & 8.78\cdot10^{-2} \\*
          & AC2   & 9.96\cdot10^{-1} & 0.000    & 4.98\cdot10^{-1} & 1.24\cdot10^{-1} &       &       & AC2   & 9.98\cdot10^{-1} & 0.000    & 4.99\cdot10^{-1} & 1.66\cdot10^{-1} \\*
          & AC3   & 9.96\cdot10^{-1} & 0.000    & 4.98\cdot10^{-1} & 1.24\cdot10^{-1} &       &       & AC3   & 9.95\cdot10^{-1} & 0.000    & 4.97\cdot10^{-1} & 1.66\cdot10^{-1} \\*
          & CSA   & 6.21\cdot10^{-2} & 1.13\cdot10^{-1} & 8.76\cdot10^{-2} & 1.07\cdot10^{-1} &       &       & CSA   & 7.41\cdot10^{-2} & 1.05\cdot10^{-1} & 8.96\cdot10^{-2} & 9.99\cdot10^{-2} \\*
          & CGA   & 1.66\cdot10^{-1} & 4.60\cdot10^{-2} & 1.06\cdot10^{-1} & 6.11\cdot10^{-2} &       &       & CGA   & 1.46\cdot10^{-1} & 4.50\cdot10^{-2} & 9.56\cdot10^{-2} & 6.19\cdot10^{-2} \\*
     \cmidrule(r){1-6} \cmidrule(r){8-13} 
		$[1, 3]$ & ABT   & 1.08\cdot10^{-1} & 5.41\cdot10^{-2} & 8.11\cdot10^{-2} & 6.76\cdot10^{-2} &       & $[1, 2]$ & ABT   & 1.07\cdot10^{-1} & 5.71\cdot10^{-2} & 8.21\cdot10^{-2} & 7.37\cdot10^{-2} \\*
          & ASB   & 8.81\cdot10^{-2} & 6.41\cdot10^{-2} & 7.61\cdot10^{-2} & 7.01\cdot10^{-2} &       &       & ASB   & 9.01\cdot10^{-2} & 6.31\cdot10^{-2} & 7.66\cdot10^{-2} & 7.21\cdot10^{-2} \\*
          & ADC   & 4.55\cdot10^{-1} & 5.41\cdot10^{-2} & 2.55\cdot10^{-1} & 1.54\cdot10^{-1} &       &       & ADC   & 3.37\cdot10^{-1} & 7.01\cdot10^{-2} & 2.04\cdot10^{-1} & 1.59\cdot10^{-1} \\*
          & CB0   & 8.37\cdot10^{-1} & 2.00\cdot10^{-3} & 4.19\cdot10^{-1} & 2.11\cdot10^{-1} &       &       & CB0   & 8.37\cdot10^{-1} & 1.00\cdot10^{-3} & 4.19\cdot10^{-1} & 2.80\cdot10^{-1} \\*
          & CB1   & 9.99\cdot10^{-1} & 0.000    & 4.99\cdot10^{-1} & 2.50\cdot10^{-1} &       &       & CB1   & 9.95\cdot10^{-1} & 0.000    & 4.97\cdot10^{-1} & 3.32\cdot10^{-1} \\*
          & CB2   & 9.96\cdot10^{-1} & 0.000    & 4.98\cdot10^{-1} & 2.49\cdot10^{-1} &       &       & CB2   & 9.93\cdot10^{-1} & 0.000    & 4.96\cdot10^{-1} & 3.31\cdot10^{-1} \\*
          & AC1   & 6.41\cdot10^{-2} & 9.21\cdot10^{-2} & 7.81\cdot10^{-2} & 8.51\cdot10^{-2} &       &       & AC1   & 7.11\cdot10^{-2} & 7.71\cdot10^{-2} & 7.41\cdot10^{-2} & 7.51\cdot10^{-2} \\*
          & AC2   & 9.95\cdot10^{-1} & 0.000    & 4.97\cdot10^{-1} & 2.49\cdot10^{-1} &       &       & AC2   & 9.95\cdot10^{-1} & 0.000    & 4.97\cdot10^{-1} & 3.32\cdot10^{-1} \\*
          & AC3   & 9.97\cdot10^{-1} & 0.000    & 4.98\cdot10^{-1} & 2.49\cdot10^{-1} &       &       & AC3   & 9.97\cdot10^{-1} & 0.000    & 4.98\cdot10^{-1} & 3.32\cdot10^{-1} \\*
          & CSA   & 7.01\cdot10^{-2} & 7.61\cdot10^{-2} & 7.31\cdot10^{-2} & 7.46\cdot10^{-2} &       &       & CSA   & 1.00\cdot10^{-1} & 7.01\cdot10^{-2} & 8.51\cdot10^{-2} & 8.01\cdot10^{-2} \\*
          & CGA   & 1.23\cdot10^{-1} & 4.30\cdot10^{-2} & 8.31\cdot10^{-2} & 6.31\cdot10^{-2} &       &       & CGA   & 8.71\cdot10^{-2} & 5.71\cdot10^{-2} & 7.21\cdot10^{-2} & 6.71\cdot10^{-2} \\*
     \cmidrule(r){1-6} \cmidrule(r){8-13} 
		$[2, 3]$ & ABT   & 1.07\cdot10^{-1} & 5.71\cdot10^{-2} & 8.21\cdot10^{-2} & 7.71\cdot10^{-2} &       & $[1, 1]$ & ABT   & 1.06\cdot10^{-1} & 6.21\cdot10^{-2} & 8.41\cdot10^{-2} & 8.41\cdot10^{-2} \\*
          & ASB   & 9.01\cdot10^{-2} & 6.31\cdot10^{-2} & 7.66\cdot10^{-2} & 7.39\cdot10^{-2} &       &       & ASB   & 1.04\cdot10^{-1} & 6.21\cdot10^{-2} & 8.31\cdot10^{-2} & 8.31\cdot10^{-2} \\*
          & ADC   & 2.38\cdot10^{-1} & 9.81\cdot10^{-2} & 1.68\cdot10^{-1} & 1.54\cdot10^{-1} &       &       & ADC   & 1.33\cdot10^{-1} & 1.39\cdot10^{-1} & 1.36\cdot10^{-1} & 1.36\cdot10^{-1} \\*
          & CB0   & 8.37\cdot10^{-1} & 1.00\cdot10^{-3} & 4.19\cdot10^{-1} & 3.35\cdot10^{-1} &       &       & CB0   & 2.56\cdot10^{-1} & 2.02\cdot10^{-1} & 2.29\cdot10^{-1} & 2.29\cdot10^{-1} \\*
          & CB1   & 7.29\cdot10^{-1} & 0.000    & 3.64\cdot10^{-1} & 2.91\cdot10^{-1} &       &       & CB1   & 1.13\cdot10^{-1} & 6.81\cdot10^{-2} & 9.06\cdot10^{-2} & 9.06\cdot10^{-2} \\*
          & CB2   & 7.10\cdot10^{-1} & 0.000    & 3.55\cdot10^{-1} & 2.84\cdot10^{-1} &       &       & CB2   & 1.04\cdot10^{-1} & 6.21\cdot10^{-2} & 8.31\cdot10^{-2} & 8.31\cdot10^{-2} \\*
          & AC1   & 8.61\cdot10^{-2} & 6.31\cdot10^{-2} & 7.46\cdot10^{-2} & 7.23\cdot10^{-2} &       &       & AC1   & 9.41\cdot10^{-2} & 6.01\cdot10^{-2} & 7.71\cdot10^{-2} & 7.71\cdot10^{-2} \\*
          & AC2   & 8.80\cdot10^{-1} & 0.000    & 4.40\cdot10^{-1} & 3.52\cdot10^{-1} &       &       & AC2   & 1.04\cdot10^{-1} & 6.21\cdot10^{-2} & 8.31\cdot10^{-2} & 8.31\cdot10^{-2} \\*
          & AC3   & 9.98\cdot10^{-1} & 0.000    & 4.99\cdot10^{-1} & 3.99\cdot10^{-1} &       &       & AC3   & 9.41\cdot10^{-2} & 6.01\cdot10^{-2} & 7.71\cdot10^{-2} & 7.71\cdot10^{-2} \\*
          & CSA   & 1.04\cdot10^{-1} & 8.01\cdot10^{-2} & 9.21\cdot10^{-2} & 8.97\cdot10^{-2} &       &       & CSA   & 9.01\cdot10^{-2} & 6.11\cdot10^{-2} & 7.56\cdot10^{-2} & 7.56\cdot10^{-2} \\*
          & CGA   & 1.09\cdot10^{-1} & 5.61\cdot10^{-2} & 8.26\cdot10^{-2} & 7.73\cdot10^{-2} &       &       & CGA   & 1.04\cdot10^{-1} & 6.21\cdot10^{-2} & 8.31\cdot10^{-2} & 8.31\cdot10^{-2} \\*
     \cmidrule(r){1-6} \cmidrule(r){8-13} 
		
		$[3, 2]$ & ABT   & 1.06\cdot10^{-1} & 6.21\cdot10^{-2} & 8.41\cdot10^{-2} & 8.85\cdot10^{-2} &       & $[2, 1]$ & ABT   & 1.06\cdot10^{-1} & 6.21\cdot10^{-2} & 8.41\cdot10^{-2} & 9.14\cdot10^{-2} \\*
          & ASB   & 8.91\cdot10^{-2} & 6.11\cdot10^{-2} & 7.51\cdot10^{-2} & 7.79\cdot10^{-2} &       &       & ASB   & 8.91\cdot10^{-2} & 6.11\cdot10^{-2} & 7.51\cdot10^{-2} & 7.97\cdot10^{-2} \\*
          & ADC   & 8.71\cdot10^{-2} & 2.32\cdot10^{-1} & 1.60\cdot10^{-1} & 1.45\cdot10^{-1} &       &       & ADC   & 6.71\cdot10^{-2} & 3.03\cdot10^{-1} & 1.85\cdot10^{-1} & 1.46\cdot10^{-1} \\*
          & CB0   & 2.20\cdot10^{-2} & 6.51\cdot10^{-1} & 3.36\cdot10^{-1} & 2.73\cdot10^{-1} &       &       & CB0   & 2.20\cdot10^{-2} & 6.59\cdot10^{-1} & 3.40\cdot10^{-1} & 2.34\cdot10^{-1} \\*
          & CB1   & 4.00\cdot10^{-3} & 6.53\cdot10^{-1} & 3.28\cdot10^{-1} & 2.63\cdot10^{-1} &       &       & CB1   & 0.000    & 9.79\cdot10^{-1} & 4.89\cdot10^{-1} & 3.26\cdot10^{-1} \\*
          & CB2   & 6.01\cdot10^{-3} & 6.48\cdot10^{-1} & 3.27\cdot10^{-1} & 2.63\cdot10^{-1} &       &       & CB2   & 0.000    & 9.62\cdot10^{-1} & 4.81\cdot10^{-1} & 3.21\cdot10^{-1} \\*
          & AC1   & 1.00\cdot10^{-1} & 4.40\cdot10^{-2} & 7.21\cdot10^{-2} & 7.77\cdot10^{-2} &       &       & AC1   & 1.09\cdot10^{-1} & 3.20\cdot10^{-2} & 7.06\cdot10^{-2} & 8.34\cdot10^{-2} \\*
          & AC2   & 3.00\cdot10^{-3} & 7.34\cdot10^{-1} & 3.68\cdot10^{-1} & 2.95\cdot10^{-1} &       &       & AC2   & 0.000    & 9.74\cdot10^{-1} & 4.87\cdot10^{-1} & 3.25\cdot10^{-1} \\*
          & AC3   & 0.000    & 9.98\cdot10^{-1} & 4.99\cdot10^{-1} & 3.99\cdot10^{-1} &       &       & AC3   & 0.000    & 1.000    & 5.00\cdot10^{-1} & 3.33\cdot10^{-1} \\*
          & CSA   & 1.06\cdot10^{-1} & 6.91\cdot10^{-2} & 8.76\cdot10^{-2} & 9.13\cdot10^{-2} &       &       & CSA   & 1.09\cdot10^{-1} & 4.00\cdot10^{-2} & 7.46\cdot10^{-2} & 8.61\cdot10^{-2} \\*
          & CGA   & 8.91\cdot10^{-2} & 6.01\cdot10^{-2} & 7.46\cdot10^{-2} & 7.75\cdot10^{-2} &       &       & CGA   & 1.01\cdot10^{-1} & 8.41\cdot10^{-2} & 9.26\cdot10^{-2} & 9.54\cdot10^{-2} \\*
     \cmidrule(r){1-6} \cmidrule(r){8-13} 
		$[3, 1]$ & ABT   & 1.06\cdot10^{-1} & 6.21\cdot10^{-2} & 8.41\cdot10^{-2} & 9.51\cdot10^{-2} &       & $[5, 1]$ & ABT   & 1.04\cdot10^{-1} & 6.91\cdot10^{-2} & 8.66\cdot10^{-2} & 9.83\cdot10^{-2} \\*
          & ASB   & 8.91\cdot10^{-2} & 6.11\cdot10^{-2} & 7.51\cdot10^{-2} & 8.21\cdot10^{-2} &       &       & ASB   & 9.51\cdot10^{-2} & 6.71\cdot10^{-2} & 8.11\cdot10^{-2} & 9.04\cdot10^{-2} \\*
          & ADC   & 4.20\cdot10^{-2} & 4.20\cdot10^{-1} & 2.31\cdot10^{-1} & 1.37\cdot10^{-1} &       &       & ADC   & 1.90\cdot10^{-2} & 5.13\cdot10^{-1} & 2.66\cdot10^{-1} & 1.01\cdot10^{-1} \\*
          & CB0   & 2.20\cdot10^{-2} & 6.61\cdot10^{-1} & 3.41\cdot10^{-1} & 1.82\cdot10^{-1} &       &       & CB0   & 2.20\cdot10^{-2} & 6.55\cdot10^{-1} & 3.38\cdot10^{-1} & 1.27\cdot10^{-1} \\*
          & CB1   & 0.000    & 1.000    & 5.00\cdot10^{-1} & 2.50\cdot10^{-1} &       &       & CB1   & 0.000    & 9.94\cdot10^{-1} & 4.97\cdot10^{-1} & 1.66\cdot10^{-1} \\*
          & CB2   & 0.000    & 9.95\cdot10^{-1} & 4.97\cdot10^{-1} & 2.49\cdot10^{-1} &       &       & CB2   & 0.000    & 9.99\cdot10^{-1} & 4.99\cdot10^{-1} & 1.66\cdot10^{-1} \\*
          & AC1   & 1.17\cdot10^{-1} & 2.60\cdot10^{-2} & 7.16\cdot10^{-2} & 9.43\cdot10^{-2} &       &       & AC1   & 1.21\cdot10^{-1} & 4.10\cdot10^{-2} & 8.11\cdot10^{-2} & 1.08\cdot10^{-1} \\*
          & AC2   & 0.000    & 9.95\cdot10^{-1} & 4.97\cdot10^{-1} & 2.49\cdot10^{-1} &       &       & AC2   & 0.000    & 9.94\cdot10^{-1} & 4.97\cdot10^{-1} & 1.66\cdot10^{-1} \\*
          & AC3   & 0.000    & 9.99\cdot10^{-1} & 4.99\cdot10^{-1} & 2.50\cdot10^{-1} &       &       & AC3   & 0.000    & 9.99\cdot10^{-1} & 4.99\cdot10^{-1} & 1.66\cdot10^{-1} \\*
          & CSA   & 1.30\cdot10^{-1} & 5.41\cdot10^{-2} & 9.21\cdot10^{-2} & 1.11\cdot10^{-1} &       &       & CSA   & 1.32\cdot10^{-1} & 5.11\cdot10^{-2} & 9.16\cdot10^{-2} & 1.19\cdot10^{-1} \\*
          & CGA   & 7.91\cdot10^{-2} & 7.71\cdot10^{-2} & 7.81\cdot10^{-2} & 7.86\cdot10^{-2} &       &       & CGA   & 7.71\cdot10^{-2} & 9.91\cdot10^{-2} & 8.81\cdot10^{-2} & 8.07\cdot10^{-2} \\*
     \cmidrule(r){1-6} \cmidrule(r){8-13} 
		$[7, 1]$ & ABT   & 9.01\cdot10^{-2} & 9.61\cdot10^{-2} & 9.31\cdot10^{-2} & 9.08\cdot10^{-2} &       & $[10, 1]$ & ABT   & 9.01\cdot10^{-2} & 9.61\cdot10^{-2} & 9.31\cdot10^{-2} & 9.06\cdot10^{-2} \\*
          & ASB   & 9.51\cdot10^{-2} & 6.71\cdot10^{-2} & 8.11\cdot10^{-2} & 9.16\cdot10^{-2} &       &       & ASB   & 9.61\cdot10^{-2} & 5.81\cdot10^{-2} & 7.71\cdot10^{-2} & 9.26\cdot10^{-2} \\*
          & ADC   & 2.00\cdot10^{-2} & 5.53\cdot10^{-1} & 2.86\cdot10^{-1} & 8.66\cdot10^{-2} &       &       & ADC   & 2.10\cdot10^{-2} & 6.07\cdot10^{-1} & 3.14\cdot10^{-1} & 7.43\cdot10^{-2} \\*
          & CB0   & 2.20\cdot10^{-2} & 6.60\cdot10^{-1} & 3.41\cdot10^{-1} & 1.02\cdot10^{-1} &       &       & CB0   & 2.20\cdot10^{-2} & 6.60\cdot10^{-1} & 3.41\cdot10^{-1} & 8.00\cdot10^{-2} \\*
          & CB1   & 0.000    & 9.91\cdot10^{-1} & 4.95\cdot10^{-1} & 1.24\cdot10^{-1} &       &       & CB1   & 1.00\cdot10^{-3} & 9.91\cdot10^{-1} & 4.96\cdot10^{-1} & 9.10\cdot10^{-2} \\*
          & CB2   & 0.000    & 9.96\cdot10^{-1} & 4.98\cdot10^{-1} & 1.24\cdot10^{-1} &       &       & CB2   & 0.000    & 9.98\cdot10^{-1} & 4.99\cdot10^{-1} & 9.07\cdot10^{-2} \\*
          & AC1   & 1.24\cdot10^{-1} & 4.30\cdot10^{-2} & 8.36\cdot10^{-2} & 1.14\cdot10^{-1} &       &       & AC1   & 1.18\cdot10^{-1} & 5.61\cdot10^{-2} & 8.71\cdot10^{-2} & 1.12\cdot10^{-1} \\*
          & AC2   & 0.000    & 9.95\cdot10^{-1} & 4.97\cdot10^{-1} & 1.24\cdot10^{-1} &       &       & AC2   & 0.000    & 9.96\cdot10^{-1} & 4.98\cdot10^{-1} & 9.05\cdot10^{-2} \\*
          & AC3   & 0.000    & 9.99\cdot10^{-1} & 4.99\cdot10^{-1} & 1.25\cdot10^{-1} &       &       & AC3   & 0.000    & 1.000    & 5.00\cdot10^{-1} & 9.09\cdot10^{-2} \\*
          & CSA   & 1.75\cdot10^{-1} & 5.91\cdot10^{-2} & 1.17\cdot10^{-1} & 1.61\cdot10^{-1} &       &       & CSA   & 1.43\cdot10^{-1} & 4.60\cdot10^{-2} & 9.46\cdot10^{-2} & 1.34\cdot10^{-1} \\*
          & CGA   & 8.01\cdot10^{-2} & 1.16\cdot10^{-1} & 9.81\cdot10^{-2} & 8.46\cdot10^{-2} &       &       & CGA   & 7.01\cdot10^{-2} & 1.07\cdot10^{-1} & 8.86\cdot10^{-2} & 7.34\cdot10^{-2} \\*
     \cmidrule(r){1-6} \cmidrule(r){8-13} 
		$[25, 1]$ & ABT   & 7.11\cdot10^{-2} & 1.20\cdot10^{-1} & 9.56\cdot10^{-2} & 7.30\cdot10^{-2} &       & $[50, 1]$ & ABT   & 7.41\cdot10^{-2} & 1.51\cdot10^{-1} & 1.13\cdot10^{-1} & 7.56\cdot10^{-2} \\*
          & ASB   & 9.61\cdot10^{-2} & 5.81\cdot10^{-2} & 7.71\cdot10^{-2} & 9.46\cdot10^{-2} &       &       & ASB   & 9.61\cdot10^{-2} & 5.81\cdot10^{-2} & 7.71\cdot10^{-2} & 9.54\cdot10^{-2} \\*
          & ADC   & 2.60\cdot10^{-2} & 6.50\cdot10^{-1} & 3.38\cdot10^{-1} & 5.00\cdot10^{-2} &       &       & ADC   & 2.20\cdot10^{-2} & 6.55\cdot10^{-1} & 3.38\cdot10^{-1} & 3.44\cdot10^{-2} \\*
          & CB0   & 2.40\cdot10^{-2} & 6.40\cdot10^{-1} & 3.32\cdot10^{-1} & 4.77\cdot10^{-2} &       &       & CB0   & 2.20\cdot10^{-2} & 6.55\cdot10^{-1} & 3.38\cdot10^{-1} & 3.44\cdot10^{-2} \\*
          & CB1   & 0.000    & 9.94\cdot10^{-1} & 4.97\cdot10^{-1} & 3.82\cdot10^{-2} &       &       & CB1   & 0.000    & 9.86\cdot10^{-1} & 4.93\cdot10^{-1} & 1.93\cdot10^{-2} \\*
          & CB2   & 0.000    & 9.83\cdot10^{-1} & 4.91\cdot10^{-1} & 3.78\cdot10^{-2} &       &       & CB2   & 0.000    & 9.93\cdot10^{-1} & 4.96\cdot10^{-1} & 1.95\cdot10^{-2} \\*
          & AC1   & 1.00\cdot10^{-1} & 9.71\cdot10^{-2} & 9.86\cdot10^{-2} & 1.00\cdot10^{-1} &       &       & AC1   & 8.01\cdot10^{-2} & 1.97\cdot10^{-1} & 1.39\cdot10^{-1} & 8.24\cdot10^{-2} \\*
          & AC2   & 0.000    & 9.90\cdot10^{-1} & 4.95\cdot10^{-1} & 3.81\cdot10^{-2} &       &       & AC2   & 2.00\cdot10^{-3} & 9.95\cdot10^{-1} & 4.98\cdot10^{-1} & 2.15\cdot10^{-2} \\*
          & AC3   & 0.000    & 9.98\cdot10^{-1} & 4.99\cdot10^{-1} & 3.84\cdot10^{-2} &       &       & AC3   & 1.00\cdot10^{-3} & 9.90\cdot10^{-1} & 4.95\cdot10^{-1} & 2.04\cdot10^{-2} \\*
          & CSA   & 1.40\cdot10^{-1} & 8.91\cdot10^{-2} & 1.15\cdot10^{-1} & 1.38\cdot10^{-1} &       &       & CSA   & 2.85\cdot10^{-1} & 1.00\cdot10^{-3} & 1.43\cdot10^{-1} & 2.80\cdot10^{-1} \\*
          & CGA   & 6.01\cdot10^{-2} & 1.18\cdot10^{-1} & 8.91\cdot10^{-2} & 6.23\cdot10^{-2} &       &       & CGA   & 5.61\cdot10^{-2} & 1.45\cdot10^{-1} & 1.01\cdot10^{-1} & 5.78\cdot10^{-2} \\*
     \cmidrule(r){1-6} \cmidrule(r){8-13} 
		$[100, 1]$ & ABT   & 6.91\cdot10^{-2} & 2.05\cdot10^{-1} & 1.37\cdot10^{-1} & 7.04\cdot10^{-2} &       &       &       &       &       &       &  \\*
          & ASB   & 9.61\cdot10^{-2} & 5.81\cdot10^{-2} & 7.71\cdot10^{-2} & 9.57\cdot10^{-2} &       &       &       &       &       &       &  \\*
          & ADC   & 2.20\cdot10^{-2} & 6.55\cdot10^{-1} & 3.38\cdot10^{-1} & 2.83\cdot10^{-2} &       &       &       &       &       &       &  \\*
          & CB0   & 2.20\cdot10^{-2} & 6.55\cdot10^{-1} & 3.38\cdot10^{-1} & 2.83\cdot10^{-2} &       &       &       &       &       &       &  \\*
          & CB1   & 2.00\cdot10^{-3} & 9.69\cdot10^{-1} & 4.85\cdot10^{-1} & 1.16\cdot10^{-2} &       &       &       &       &       &       &  \\*
          & CB2   & 0.000    & 9.73\cdot10^{-1} & 4.86\cdot10^{-1} & 9.63\cdot10^{-3} &       &       &       &       &       &       &  \\*
          & AC1   & 5.61\cdot10^{-2} & 3.22\cdot10^{-1} & 1.89\cdot10^{-1} & 5.87\cdot10^{-2} &       &       &       &       &       &       &  \\*
          & AC2   & 4.00\cdot10^{-3} & 9.51\cdot10^{-1} & 4.77\cdot10^{-1} & 1.34\cdot10^{-2} &       &       &       &       &       &       &  \\*
          & AC3   & 0.000    & 9.95\cdot10^{-1} & 4.97\cdot10^{-1} & 9.85\cdot10^{-3} &       &       &       &       &       &       &  \\*
          & CSA   & 2.51\cdot10^{-1} & 1.10\cdot10^{-1} & 1.81\cdot10^{-1} & 2.50\cdot10^{-1} &       &       &       &       &       &       &  \\*
          & CGA   & 4.90\cdot10^{-2} & 2.24\cdot10^{-1} & 1.37\cdot10^{-1} & 5.08\cdot10^{-2} &       &       &       &       &       &       &  \\*

\end{longtabu}
}

\newpage

%%%%%%%%%%%%
%%% CBCL %%%
%%%%%%%%%%%%

\begin{landscape}
% GRAPHICS
\begin{figure}[p]
\centering
\includegraphics[width=.6\paperheight]{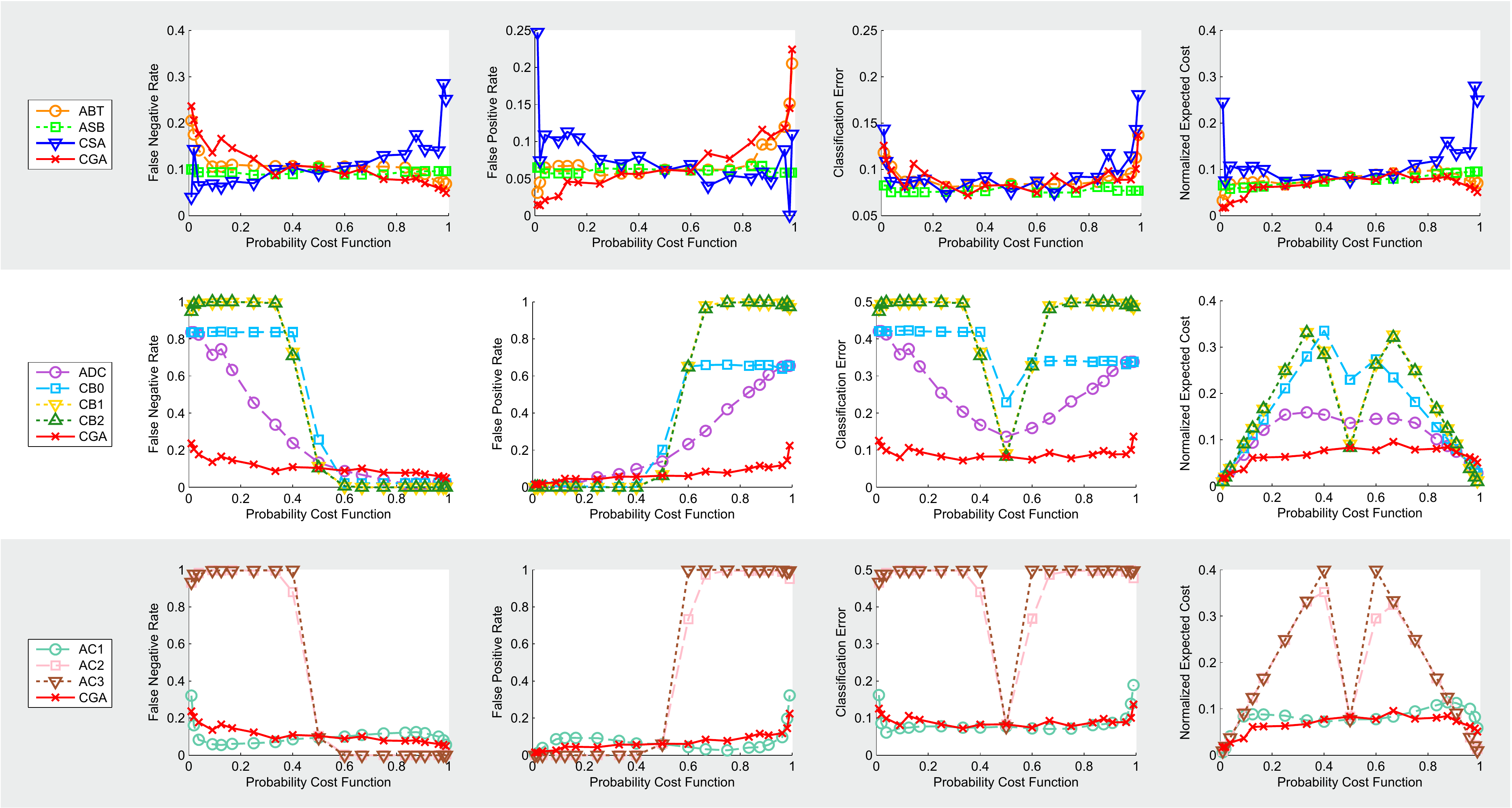}
\caption[Results obtained for the CBCL Dataset.] {Results obtained for the CBCL Dataset. First column of the illustration corresponds to False Negative Rate, second column to False Positive Rate, third column to Classification Error and the fourth one corresponds to Normalized Expected Cost. For a clearer visualization, algorithms have been divided into three groups so each row of the illustration corresponds to a different group. Cost Generalized AdaBoost is plotted in all the graphs to have a common reference across the representations.}
\label{fig:cbcl_perform} % caption for the whole figure
\end{figure}

\end{landscape}

%\renewcommand{\thesection}{\arabic{section}}
%
%\newcounter{partcount3}
%\renewcommand{\thepartcount3}{\alph{partcount3}}
%
%\makeatletter 
%\renewcommand{\thefigure}{\@arabic\c@figure}
%\makeatother
%
%\makeatletter 
%\renewcommand{\thetable}{\@arabic\c@table}
%\makeatother

\end{appendices}

%\section*{References} \pdfbookmark{References}{refs}
\bibliography{Revisiting}

\end{document}